\documentclass{article}

\usepackage{fontspec}
\usepackage{graphicx}
\usepackage{hyperref}
\usepackage{amsmath}
\usepackage{amssymb}
\usepackage[numbers,square]{natbib}
\usepackage{geometry}
\usepackage{xcolor}
\usepackage{newunicodechar}
\usepackage{listings}
\usepackage{mdframed}
\usepackage{placeins}
\usepackage{setspace}
\usepackage{longtable}
\usepackage{tcolorbox}
\usepackage{trimclip}

\hypersetup{
    colorlinks=true,
    linkcolor=blue,
    urlcolor=blue,
    citecolor=blue,
}

\newfontfamily{\dejavufont}{DejaVuSans.ttf}

\newunicodechar{⏎}{{\dejavufont ⏎}}
\newunicodechar{↺}{{\dejavufont ↺}}
\newunicodechar{м}{{\unicodefont м}}
\newunicodechar{о}{{\unicodefont о}}
\newunicodechar{с}{{\unicodefont с}}
\newunicodechar{т}{{\unicodefont т}}
\newunicodechar{у}{{\unicodefont у}}

\let\oldsection\section
\renewcommand{\section}{\FloatBarrier\oldsection}
\let\oldsubsection\subsection
\renewcommand{\subsection}{\FloatBarrier\oldsubsection}

\lstnewenvironment{promptblock}{%
\lstset{
  basicstyle=\ttfamily\small,
  breaklines=true,
  breakatwhitespace=true,
  columns=fullflexible,
  frame=none,
  keepspaces=true,
  showstringspaces=false,
  literate={⏎}{{\dejavufont\symbol{"23CE}}}1
}
}{}

\surroundwithmdframed[
  backgroundcolor=gray!10,
  linecolor=gray!50,
  linewidth=0.5pt,
  innerleftmargin=8pt,
  innerrightmargin=8pt,
  innertopmargin=6pt,
  innerbottommargin=6pt,
  roundcorner=2pt
]{promptblock}

\definecolor{featurechipbg}{HTML}{d97757}
\newtcbox{\featurechipbox}{
  on line,
  colback=featurechipbg,
  colframe=featurechipbg,
  boxrule=0pt,
  arc=3pt,
  boxsep=0pt,
  left=4pt,
  right=4pt,
  top=2pt,
  bottom=2pt,
}
\newcommand{\featurechip}[2]{%
  \href{https://transformer-circuits.pub/2024/scaling-monosemanticity/features/index.html?featureId=#1_#2}{%
    \featurechipbox{\textcolor{white}{\textsf{\footnotesize #1/#2}}}%
  }%
}

\newenvironment{featureexamples}{%
  \begin{tcolorbox}[
    colback=white,
    colframe=gray!50,
    boxrule=0.5pt,
    arc=2pt,
    left=6pt,
    right=6pt,
    top=6pt,
    bottom=6pt,
  ]
}{%
  \end{tcolorbox}
}

\newcommand{\exampleline}[1]{%
  \par\vspace{1pt}\noindent\clipbox{0pt 0pt {\dimexpr\width-\linewidth\relax} 0pt}{#1}%
}

\newfontfamily{\unicodefont}{NotoSansMonoCJKsc-Regular.otf}[Path=fonts/, Scale=0.85]

\title{Scaling Monosemanticity: Extracting Interpretable Features from Claude 3 Sonnet}

\author{\parbox{0.9\textwidth}{\centering Adly Templeton\textsuperscript{*}, Tom Conerly\textsuperscript{*}, Jonathan Marcus, Jack Lindsey, Trenton Bricken, Brian Chen, Adam Pearce, Craig Citro, Emmanuel Ameisen, Andy Jones, Hoagy Cunningham, Nicholas L Turner, Callum McDougall, Monte MacDiarmid, Alex Tamkin, Esin Durmus, Tristan Hume, Francesco Mosconi, C. Daniel Freeman, Theodore R. Sumers, Edward Rees, Joshua Batson, Adam Jermyn, Shan Carter, Chris Olah, and Tom Henighan\textsuperscript{$\dagger$} \\[0.5em]
\textit{Anthropic} \\[0.3em]
\footnotesize{* Equal contribution; $\dagger$ Corresponding author: henighan@anthropic.com} \\[0.5em]
\href{https://transformer-circuits.pub/2024/scaling-monosemanticity/index.html}{\textcolor{magenta}{HTML version}}}}

\date{May 21, 2024}

\begin{document}

\maketitle

\begin{abstract}
We demonstrate that sparse autoencoders can extract interpretable features from Claude 3 Sonnet, a production-scale language model, addressing the open question of whether dictionary learning methods scale beyond small transformers. We trained sparse autoencoders with up to 34 million features on the model's middle layer residual stream, using scaling laws to guide hyperparameter selection. The resulting features are multilingual and multimodal (generalizing to images despite text-only training), respond to both concrete instances and abstract discussions of concepts, and can be used to steer model behavior in ways consistent with their interpretations. We find features corresponding to famous entities and locations, as well as more abstract concepts like sarcasm or errors in code. We also identify features relevant to ways in which language models might cause harm—including features representing deception, power-seeking, sycophancy, and bias—and show that these causally influence model outputs when manipulated.  Additionally, we conduct analyses of feature interpretability, geometry, and computational function. However, significant limitations remain: our suite of features is incomplete, and we lack rigorous methods for evaluating whether our features faithfully capture model computations.
\end{abstract}

Eight months ago, we \href{https://transformer-circuits.pub/2023/monosemantic-features/index.html}{demonstrated} that sparse autoencoders could recover monosemantic features from a small one-layer transformer. At the time, a major concern was that this method might not scale feasibly to state-of-the-art transformers and, as a result, be unable to practically contribute to AI safety. Since then, scaling sparse autoencoders has been a major priority of the Anthropic interpretability team, and we're pleased to report extracting \textit{high-quality features from Claude 3 Sonnet},\footnote{For clarity, this is the 3.0 version of Claude 3 Sonnet, released March 4, 2024. It is the exact model in production as of the writing of this paper. It is the finetuned model, not the base pretrained model (although our method also works on the base model).} Anthropic's medium-sized production model.

We find a diversity of highly abstract features. They both respond to and behaviorally cause abstract behaviors. Examples of features we find include features for famous people, features for countries and cities, and features tracking type signatures in code. Many features are multilingual (responding to the same concept across languages) and multimodal (responding to the same concept in both text and images), as well as encompassing both abstract and concrete instantiations of the same idea (such as code with security vulnerabilities, and abstract discussion of security vulnerabilities).

\begin{figure}[!htp]
    \centering
    \includegraphics[width=0.9\textwidth,height=0.7\textheight,keepaspectratio]{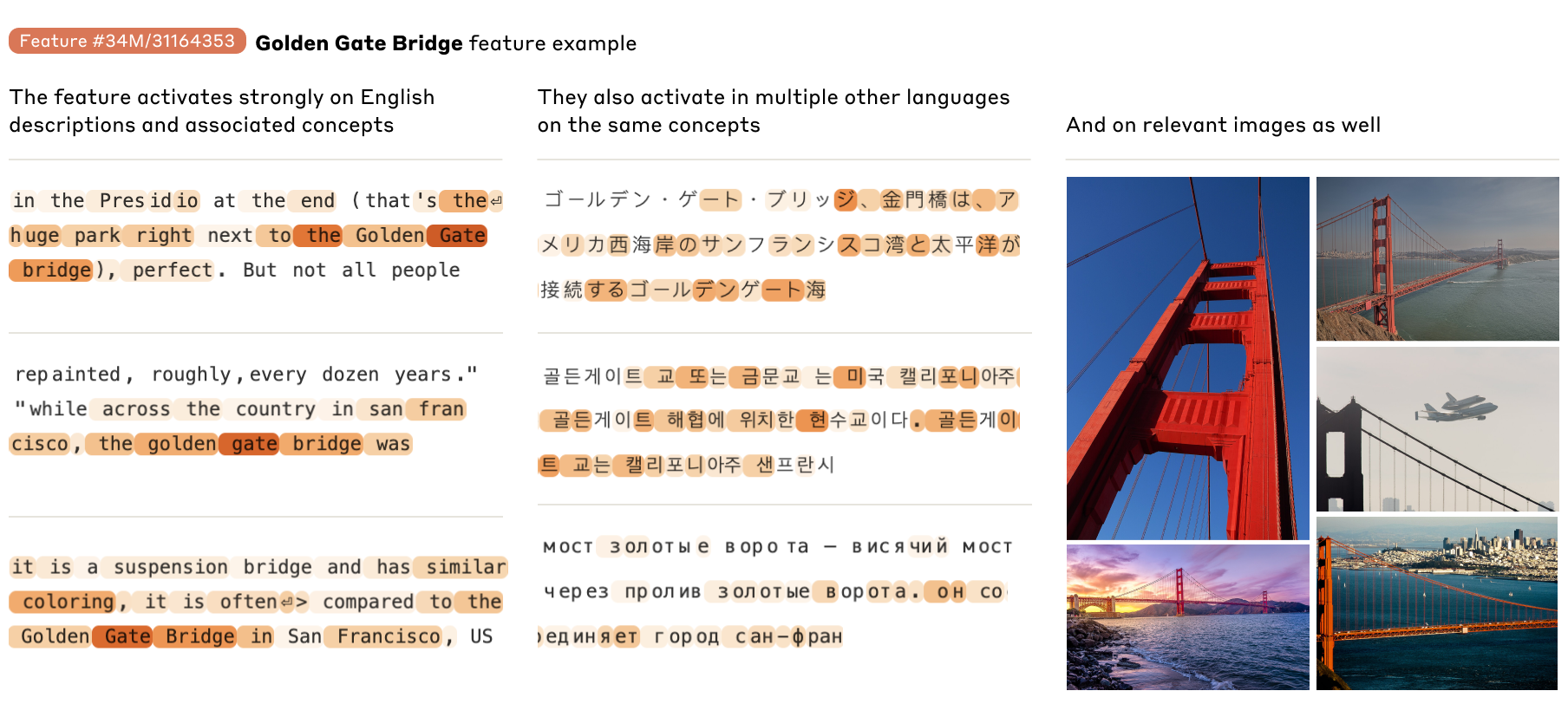}
    \label{fig:gdoc_2}
\end{figure}

Some of the features we find are of particular interest because they may be \textbf{safety-relevant} -- that is, they are plausibly connected to a range of ways in which modern AI systems may cause harm. In particular, we find features related to \hyperref[sec:safety-relevant-code]{security vulnerabilities and backdoors in code}; \hyperref[sec:safety-relevant-bias]{bias} (including both overt slurs, and more subtle biases); \hyperref[sec:safety-relevant-deception]{lying, deception, and power-seeking} (including treacherous turns); \hyperref[sec:safety-relevant-sycophancy]{sycophancy}; and \hyperref[sec:safety-relevant-criminal]{dangerous / criminal content} (e.g., producing bioweapons). However, we caution not to read too much into the mere existence of such features: there's a difference (for example) between knowing about lies, being capable of lying, and actually lying in the real world. This research is also very preliminary. Further work will be needed to understand the implications of these potentially safety-relevant features.

\subsection*{Key Results}

\begin{itemize}
    \item Sparse autoencoders produce interpretable features for large models.
    \item Scaling laws can be \hyperref[sec:scaling-scaling-laws]{used to guide the training} of sparse autoencoders.
    \item The resulting features are highly abstract: multilingual, multimodal, and generalizing between concrete and abstract references.
    \item There \hyperref[sec:feature-survey-completeness]{appears to be a systematic relationship} between the frequency of concepts and the dictionary size needed to resolve features for them.
    \item Features can be used to steer large models (\textit{see e.g.} \hyperref[sec:assessing-tour-influence]{Influence on Behavior}). This extends prior work on steering models using other methods (see \hyperref[sec:related-work-steering]{Related Work}).
    \item We observe features related to a broad range of safety concerns, including \hyperref[sec:safety-relevant-deception]{deception}, \hyperref[sec:safety-relevant-sycophancy]{sycophancy}, \hyperref[sec:safety-relevant-bias]{bias}, and \hyperref[sec:safety-relevant-criminal]{dangerous content}.
\end{itemize}

\section{Scaling Dictionary Learning to Claude 3 Sonnet}\label{sec:scaling-to-sonnet}

Our general approach to understanding Claude 3 Sonnet is based on the \textbf{linear representation hypothesis} (\textit{see e.g.} \cite{mikolov2013linguistic}) and the \textbf{superposition hypothesis} (\textit{see e.g.} \cite{arora2018linear,goh2016decoding,elhage2022superposition}). For an introduction to these ideas, we refer readers to \href{https://transformer-circuits.pub/2022/toy_model/index.html\#motivation}{the Background and Motivation section} of \textit{Toy Models} \cite{elhage2022superposition}. At a high level, the linear representation hypothesis suggests that neural networks represent meaningful concepts -- referred to as \textbf{features} -- as directions in their activation spaces. The superposition hypothesis accepts the idea of linear representations and further hypothesizes that neural networks use the existence of almost-orthogonal directions in high-dimensional spaces to represent more features than there are dimensions.

If one believes these hypotheses, the natural approach is to use a standard method called \textbf{dictionary learning} \cite{elad2010sparse,olshausen1997sparse}. Recently, several papers have suggested that this can be quite effective for transformer language models \cite{yun2021transformer,bricken2023monosemanticity,cunningham2023sparse,tamkin2023codebook}. In particular, a specific approximation of dictionary learning called a sparse autoencoder appears to be very effective \cite{bricken2023monosemanticity,cunningham2023sparse}.

To date, these efforts have been on relatively small language models by the standards of modern foundation models. Our previous paper \cite{bricken2023monosemanticity}, which focused on a one-layer model, was a particularly extreme example of this. As a result, an important question has been left hanging: will these methods work for large models? Or is there some reason, whether pragmatic questions of engineering or more fundamental differences in how large models operate, that would mean these efforts can't generalize?

This context motivates our project of scaling sparse autoencoders to Claude 3 Sonnet, Anthropic's medium-scale production model. The rest of this section will review our general sparse autoencoder setup, the specifics of the three sparse autoencoders we'll analyze in this paper, and how we used scaling laws to make informed decisions about the design of our sparse autoencoders. From there, we'll dive into analyzing the features our sparse autoencoders learn -- and the interesting properties of Claude 3 Sonnet they reveal.

\subsection{Sparse Autoencoders}\label{sec:scaling-sparse-autoencoders}

Our high-level goal in this work is to decompose the activations of a model (Claude 3 Sonnet) into more interpretable pieces. We do so by training a sparse autoencoder (SAE) on the model activations, as in our prior work \cite{bricken2023monosemanticity} and that of several other groups (\textit{e.g.} \cite{yun2021transformer,cunningham2023sparse,tamkin2023codebook}; see \hyperref[sec:related-work]{Related Work}). SAEs are an instance of a family of “sparse dictionary learning” algorithms that seek to decompose data into a weighted sum of sparsely active components.

Our SAE consists of two layers. The first layer (“encoder”) maps the activity to a higher-dimensional layer via a learned linear transformation followed by a ReLU nonlinearity. We refer to the units of this high-dimensional layer as “features.” The second layer (“decoder”) attempts to reconstruct the model activations via a linear transformation of the feature activations. The model is trained to minimize a combination of (1) reconstruction error and (2) an L1 regularization penalty on the feature activations, which incentivizes sparsity.

Once the SAE is trained, it provides us with an approximate decomposition of the model’s activations into a linear combination of “feature directions” (SAE decoder weights) with coefficients equal to the feature activations. The sparsity penalty ensures that, for many given inputs to the model, a very small fraction of features will have nonzero activations. Thus, for any given token in any given context, the model activations are “explained” by a small set of active features (out of a large pool of possible features). For more motivation and explanation of SAEs, see the \href{https://transformer-circuits.pub/2023/monosemantic-features/index.html\#problem-setup}{Problem Setup} section of \textit{Towards Monosemanticity }\cite{bricken2023monosemanticity}.

Here’s a brief overview of our methodology which we described in greater detail in \href{https://transformer-circuits.pub/2024/april-update/index.html\#training-saes}{Update on how we train SAEs} from our April 2024 Update.

As a preprocessing step we apply a scalar normalization to the model activations so their average squared L2 norm is the residual stream dimension, $D$. We denote the normalized activations as $\mathbf{x} \in \mathbb{R}^D$, and attempt to decompose this vector using $F$ features as follows:
$$\hat{\mathbf{x}} = \mathbf{b}^\text{dec} + \sum_{i=1}^F f_i(\mathbf{x}) \mathbf{W}^\text{dec}_{\cdot,i}$$
where $W^\text{dec} \in \mathbb{R}^{D \times F}$ are the learned SAE decoder weights, $\mathbf{b}^\text{dec} \in \mathbb{R}^D$ are learned biases, and $f_i$ denotes the activity of feature \textit{i}. Feature activations are given by the output of the encoder:
$$f_i(x) = \text{ReLU}\left(\mathbf{W}^\text{enc}_{i, \cdot} \cdot \mathbf{x} +b^\text{enc}_i \right)$$
where $W^\text{enc} \in \mathbb{R}^{F \times D}$ are the learned SAE encoder weights, and $\mathbf{b}^\text{enc} \in \mathbb{R}^F$ are learned biases.

The loss function $\mathcal{L}$ is the combination of an L2 penalty on the reconstruction loss and an L1 penalty on feature activations.

$$\mathcal{L} = \mathbb{E}_\mathbf{x} \left[ \|\mathbf{x}-\hat{\mathbf{x}}\|_2^2 + \lambda\sum_i f_i(\mathbf{x}) \cdot \|\mathbf{W}^\text{dec}_{\cdot,i}\|_2 \right]$$

Including the factor of $\|\mathbf{W}^\text{dec}_{\cdot,i}\|_2$ in the L1 penalty term allows us to interpret the unit-normalized decoder vectors $\displaystyle\frac{\mathbf{W}^\text{dec}_{\cdot,i}}{\|\mathbf{W}^\text{dec}_{\cdot,i}\|_2}$ as “feature vectors” or “feature directions,” and the product $f_i(\mathbf{x}) \cdot \|\mathbf{W}^\text{dec}_{\cdot,i}\|_2$ as the feature activations\footnote{This also prevents the SAE from “cheating” the L1 penalty by making $f_i(\mathbf{x})$ small and $\mathbf{W}^\text{dec}_{\cdot,i}$ large in a way that leaves the reconstructed activations unchanged.}. Henceforth we will use “feature activation” to refer to this quantity.

\subsection{Our SAE experiments}\label{sec:scaling-sae-experiments}

Claude 3 Sonnet is a proprietary model for both safety and competitive reasons. Some of the decisions in this publication reflect this, such as not reporting the size of the model, leaving units off certain plots, and using a simplified tokenizer. For more information on how Anthropic thinks about safety considerations in publishing research results, we refer readers to our \textit{\href{https://www.anthropic.com/news/core-views-on-ai-safety}{Core Views on AI Safety}}.

In this work, we focused on applying SAEs to residual stream activations halfway through the model (i.e.~at the “middle layer”). We made this choice for several reasons. First, the residual stream is smaller than the MLP layer, making SAE training and inference computationally cheaper. Second, focusing on the residual stream in theory helps us mitigate an issue we call “cross-layer superposition” (see \hyperref[sec:discussion-limitations]{Limitations} for more discussion). We chose to focus on the middle layer of the model because we reasoned that it is likely to contain interesting, abstract features (see e.g., \cite{jermyn20248l,elhage2022solu,40l2021l}).

We trained three SAEs of varying sizes: 1,048,576 (\textasciitilde{}1M), 4,194,304 (\textasciitilde{}4M), and 33,554,432 (\textasciitilde{}34M) features. The number of training steps for the 34M feature run was selected using a scaling laws analysis to minimize the training loss given a fixed compute budget (see below). We used an L1 coefficient of 5\footnote{Our L1 coefficient is only relevant in the context of how we normalize activations. See \href{https://transformer-circuits.pub/2024/april-update/index.html\#training-saes}{Update on how we train SAEs} for full details.}. We performed a sweep over a narrow range of learning rates (suggested by the scaling laws analysis) and chose the value that gave the lowest loss.

For all three SAEs, the average number of features active (i.e.~with nonzero activations) on a given token was fewer than 300, and the SAE reconstruction explained at least 65\% of the variance of the model activations. At the end of training, we defined “dead” features as those which were not active over a sample of $10^{7}$ tokens. The proportion of dead features was roughly 2\% for the 1M SAE, 35\% for the 4M SAE, and 65\% for the 34M SAE. We expect that improvements to the training procedure may be able to reduce the number of dead features in future experiments.

\subsection{Scaling Laws}\label{sec:scaling-scaling-laws}

Training SAEs on larger models is computationally intensive. It is important to understand (1) the extent to which additional compute improves dictionary learning results, and (2) how that compute should be allocated to obtain the highest-quality dictionary possible for a given computational budget.

Though we lack a gold-standard method of assessing the quality of a dictionary learning run, we have found that the loss function we use during training -- a weighted combination of reconstruction mean-squared error (MSE) and an L1 penalty on feature activations -- is a useful proxy, conditioned on a reasonable choice of the L1 coefficient. That is, we have found that dictionaries with low loss values (using an L1 coefficient of 5) tend to produce interpretable features and to improve other metrics of interest (the L0 norm, and the number of dead or otherwise degenerate features). Of course, this is an imperfect metric, and we have little confidence that it is optimal. It may well be the case that other L1 coefficients (or other objective functions altogether) would be better proxies to optimize.

With this proxy, we can treat dictionary learning as a standard machine learning problem, to which we can apply the “scaling laws” framework for hyperparameter optimization (\textit{see e.g.} \cite{kaplan2020scaling,hoffmann2022training}). In an SAE, compute usage primarily depends on two key hyperparameters: the number of features being learned, and the number of steps used to train the autoencoder (which maps linearly to the amount of data used, as we train the SAE for only one epoch). The compute cost scales with the product of these parameters if the input dimension and other hyperparameters are held constant.

We conducted a thorough sweep over these parameters, fixing the values of other hyperparameters (learning rate, batch size, optimization protocol, etc.). We were also interested in tracking the compute-optimal values of the loss function and parameters of interest; that is, the lowest loss that can be achieved using a given compute budget, and the number of training steps and features that achieve this minimum.

\clearpage

We make the following observations:

Over the ranges we tested, given the compute-optimal choice of training steps and number of features, loss decreases approximately according to a power law with respect to compute.

\begin{figure}[!htp]
    \centering
    \includegraphics[width=0.9\textwidth,height=0.7\textheight,keepaspectratio]{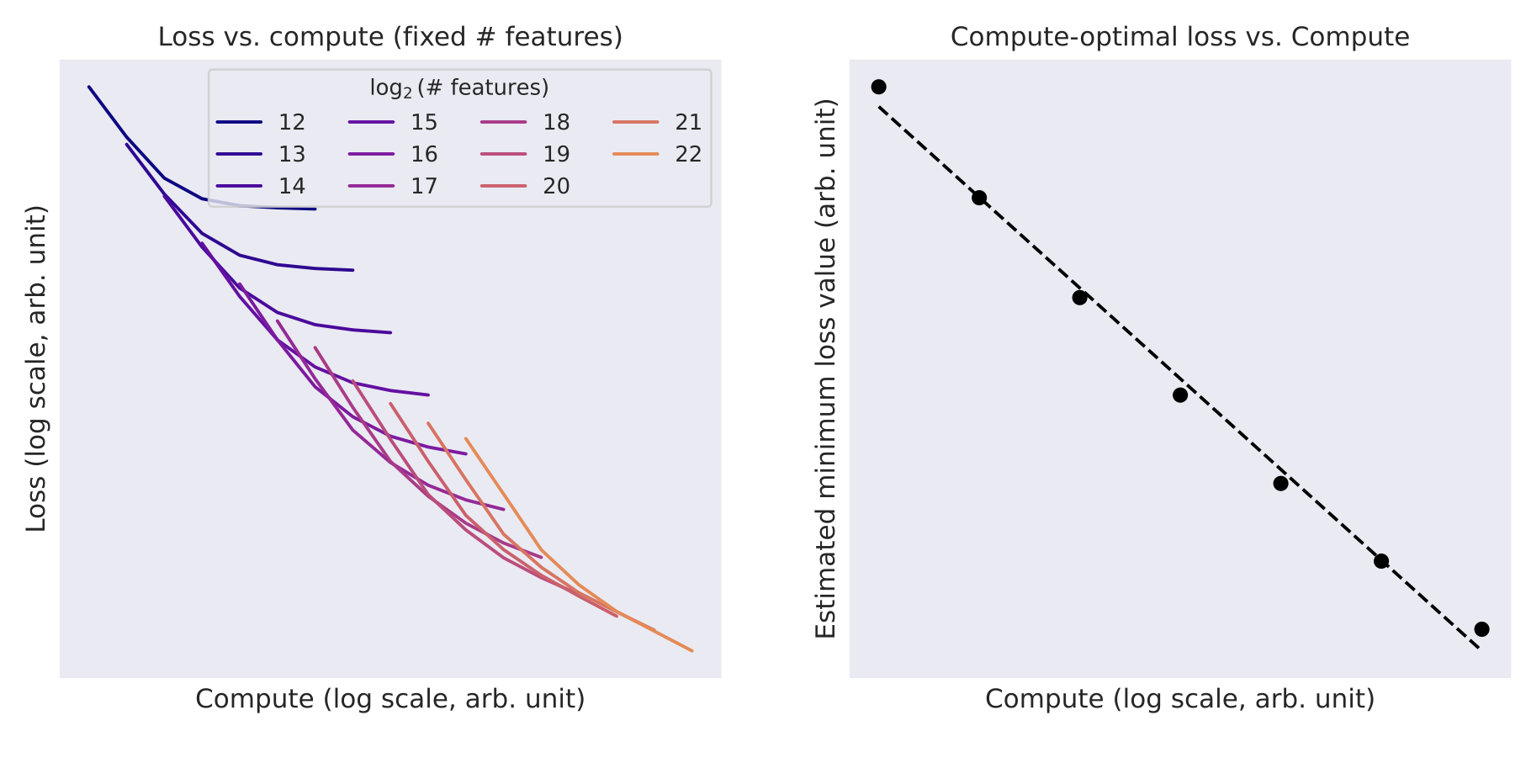}
    \label{fig:gdoc_3}
\end{figure}

As the compute budget increases, the optimal allocations of FLOPS to training steps and number of features both scale approximately as power laws. In general, the optimal number of features appears to scale somewhat more quickly than the optimal number of training steps at the compute budgets we tested, though this trend may change at higher compute budgets.

\begin{figure}[!htp]
    \centering
    \includegraphics[width=0.9\textwidth,height=0.7\textheight,keepaspectratio]{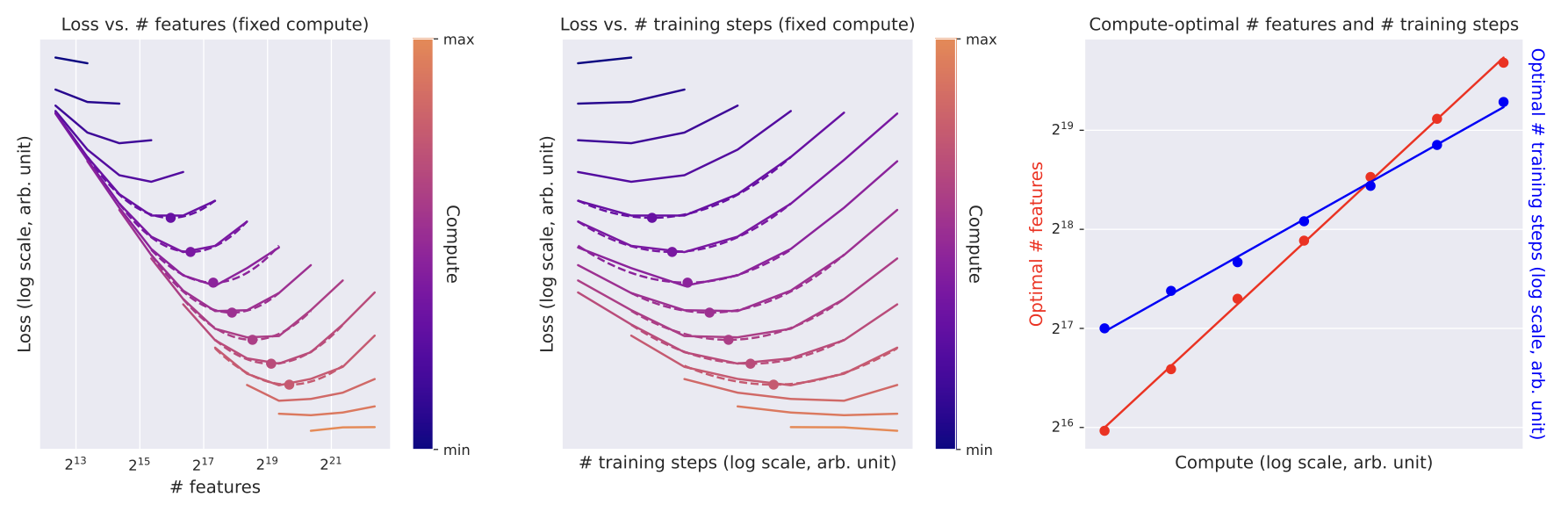}
    \label{fig:gdoc_4}
\end{figure}

These analyses used a fixed learning rate. For different compute budgets, we subsequently swept over learning rates at different optimal parameter settings according to the plots above. The inferred optimal learning rates decreased approximately as a power law as a function of compute budget, and we extrapolated this trend to choose learning rates for the larger runs.

\section{Assessing Feature Interpretability}\label{sec:assessing-interp}

In the previous section, we described how we trained sparse autoencoders on Claude 3 Sonnet. And as predicted by scaling laws, we achieved lower losses by training large SAEs. But the loss is only a proxy for what we actually care about: interpretable features that explain model behavior.

The goal of this section is to investigate whether these features are actually interpretable and explain model behavior. We'll first look at a handful of relatively straightforward features and provide evidence that they're interpretable. Then we'll look at two much more complex features, and demonstrate that they track very abstract concepts. We'll close with an experiment using automated interpretability to evaluate a larger number of features and compare them to neurons.

\subsection{Four Examples of Interpretable Features}\label{sec:assessing-tour}

In this subsection, we'll look at a few features and argue that they are genuinely interpretable. Our goal is just to demonstrate that interpretable features exist, leaving strong claims (such as most features being interpretable) to a later section.  We will provide evidence that our interpretations are good descriptions of what the features represent and how they function in the network, using an analysis similar to that in \textit{Towards Monosemanticity }\cite{bricken2023monosemanticity}.

The features we study in this section respond to:

\begin{itemize}
    \item The Golden Gate Bridge \featurechip{34M}{31164353}: Descriptions of or references to the Golden Gate Bridge.
    \item Brain sciences \featurechip{34M}{9493533}: discussions of neuroscience and related academic research on brains or minds.
    \item Monuments and popular tourist attractions \featurechip{1M}{887839}
    \item Transit infrastructure \featurechip{1M}{3}
\end{itemize}

Here and elsewhere in the paper, for each feature, we show representative examples from the top 20 text inputs in our SAE dataset, as ranked by how strongly they activate that feature (see the appendix for \hyperref[sec:appendix-methods-dataset]{details}). A larger, randomly sampled set of activations can be found by clicking on the feature ID. The highlight colors indicate activation strength at each token (white: no activation, orange: strongest activation).

\begin{featureexamples}
\featurechip{34M}{31164353} \textbf{Golden Gate Bridge}
\exampleline{{\unicodefont \colorbox[RGB]{254,230,207}{\strut{}nd}\colorbox[RGB]{255,255,255}{\strut{} (}\colorbox[RGB]{255,255,255}{\strut{}that}\colorbox[RGB]{254,236,218}{\strut{}'s}\colorbox[RGB]{253,176,111}{\strut{} the}\colorbox[RGB]{254,223,191}{\strut{}⏎}\colorbox[RGB]{253,201,152}{\strut{}h}\colorbox[RGB]{253,190,134}{\strut{}uge}\colorbox[RGB]{253,208,163}{\strut{} park}\colorbox[RGB]{253,201,152}{\strut{} right}\colorbox[RGB]{255,244,233}{\strut{} next}\colorbox[RGB]{253,197,145}{\strut{} to}\colorbox[RGB]{242,111,28}{\strut{} the}\colorbox[RGB]{253,190,134}{\strut{} Golden}\colorbox[RGB]{223,83,8}{\strut{} Gate}\colorbox[RGB]{252,151,74}{\strut{} bridge}\colorbox[RGB]{254,236,218}{\strut{}),}\colorbox[RGB]{254,232,209}{\strut{} perfect}\colorbox[RGB]{255,255,255}{\strut{}.}\colorbox[RGB]{255,255,255}{\strut{} But}\colorbox[RGB]{255,255,255}{\strut{} not}\colorbox[RGB]{255,255,255}{\strut{} all}\colorbox[RGB]{255,255,255}{\strut{} people}\colorbox[RGB]{255,255,255}{\strut{}⏎}\colorbox[RGB]{255,255,255}{\strut{}can}\colorbox[RGB]{255,255,255}{\strut{} live}\colorbox[RGB]{255,244,233}{\strut{} in}}}
\exampleline{{\unicodefont \colorbox[RGB]{255,255,255}{\strut{}e}\colorbox[RGB]{255,242,229}{\strut{} across}\colorbox[RGB]{254,234,214}{\strut{} the}\colorbox[RGB]{255,244,233}{\strut{} country}\colorbox[RGB]{255,240,225}{\strut{} in}\colorbox[RGB]{255,243,230}{\strut{} San}\colorbox[RGB]{253,192,137}{\strut{} Francisco}\colorbox[RGB]{255,238,221}{\strut{},}\colorbox[RGB]{253,177,113}{\strut{} the}\colorbox[RGB]{253,181,119}{\strut{} Golden}\colorbox[RGB]{234,97,16}{\strut{} Gate}\colorbox[RGB]{253,176,111}{\strut{} bridge}\colorbox[RGB]{253,216,179}{\strut{} was}\colorbox[RGB]{253,203,155}{\strut{} protected}\colorbox[RGB]{255,239,223}{\strut{} at}\colorbox[RGB]{254,233,212}{\strut{} all}\colorbox[RGB]{254,230,207}{\strut{} times}\colorbox[RGB]{253,185,126}{\strut{} by}\colorbox[RGB]{254,221,189}{\strut{} a}\colorbox[RGB]{255,255,255}{\strut{} vig}\colorbox[RGB]{254,223,191}{\strut{}ilant}\colorbox[RGB]{254,237,220}{\strut{} }}}
\exampleline{{\unicodefont \colorbox[RGB]{253,212,172}{\strut{}ar}\colorbox[RGB]{253,167,98}{\strut{} coloring}\colorbox[RGB]{254,224,194}{\strut{},}\colorbox[RGB]{254,226,199}{\strut{} it}\colorbox[RGB]{254,229,204}{\strut{} is}\colorbox[RGB]{254,221,189}{\strut{} often}\colorbox[RGB]{253,205,158}{\strut{}⏎}\colorbox[RGB]{254,224,194}{\strut{}>}\colorbox[RGB]{255,241,227}{\strut{} compared}\colorbox[RGB]{253,216,179}{\strut{} to}\colorbox[RGB]{253,196,144}{\strut{} the}\colorbox[RGB]{253,206,160}{\strut{} Golden}\colorbox[RGB]{234,97,16}{\strut{} Gate}\colorbox[RGB]{252,145,67}{\strut{} Bridge}\colorbox[RGB]{253,188,131}{\strut{} in}\colorbox[RGB]{255,244,233}{\strut{} San}\colorbox[RGB]{253,206,160}{\strut{} Francisco}\colorbox[RGB]{255,240,225}{\strut{},}\colorbox[RGB]{255,255,255}{\strut{} US}\colorbox[RGB]{255,239,223}{\strut{}.}\colorbox[RGB]{255,239,223}{\strut{} It}\colorbox[RGB]{255,238,221}{\strut{} was}\colorbox[RGB]{255,239,223}{\strut{} built}\colorbox[RGB]{254,233,211}{\strut{} by}\colorbox[RGB]{255,243,231}{\strut{} the}}}
\exampleline{{\unicodefont \colorbox[RGB]{255,255,255}{\strut{}l}\colorbox[RGB]{255,255,255}{\strut{} to}\colorbox[RGB]{255,255,255}{\strut{} reach}\colorbox[RGB]{255,255,255}{\strut{} and}\colorbox[RGB]{255,255,255}{\strut{} if}\colorbox[RGB]{255,255,255}{\strut{} we}\colorbox[RGB]{255,255,255}{\strut{} were}\colorbox[RGB]{254,223,191}{\strut{} going}\colorbox[RGB]{255,240,225}{\strut{} to}\colorbox[RGB]{255,243,230}{\strut{} see}\colorbox[RGB]{253,190,134}{\strut{} the}\colorbox[RGB]{253,201,152}{\strut{} Golden}\colorbox[RGB]{236,99,18}{\strut{} Gate}\colorbox[RGB]{252,149,72}{\strut{} Bridge}\colorbox[RGB]{254,221,187}{\strut{} before}\colorbox[RGB]{255,243,230}{\strut{} sunset}\colorbox[RGB]{255,255,255}{\strut{},}\colorbox[RGB]{255,255,255}{\strut{} we}\colorbox[RGB]{255,255,255}{\strut{} had}\colorbox[RGB]{255,255,255}{\strut{} to}\colorbox[RGB]{255,240,225}{\strut{} hit}\colorbox[RGB]{254,224,194}{\strut{} the}\colorbox[RGB]{255,255,255}{\strut{} road}\colorbox[RGB]{255,255,255}{\strut{},}\colorbox[RGB]{255,255,255}{\strut{} so}}}
\exampleline{{\unicodefont \colorbox[RGB]{255,255,255}{\strut{}t}\colorbox[RGB]{255,255,255}{\strut{} it}\colorbox[RGB]{255,255,255}{\strut{}?''}\colorbox[RGB]{255,255,255}{\strut{} ''}\colorbox[RGB]{255,255,255}{\strut{} Because}\colorbox[RGB]{255,240,225}{\strut{} of}\colorbox[RGB]{255,238,221}{\strut{} what}\colorbox[RGB]{255,244,233}{\strut{}'s}\colorbox[RGB]{254,224,194}{\strut{} above}\colorbox[RGB]{253,208,163}{\strut{} it}\colorbox[RGB]{254,225,196}{\strut{}.''}\colorbox[RGB]{255,255,255}{\strut{} ''}\colorbox[RGB]{253,201,152}{\strut{}The}\colorbox[RGB]{253,216,179}{\strut{} Golden}\colorbox[RGB]{236,99,18}{\strut{} Gate}\colorbox[RGB]{253,205,158}{\strut{} Bridge}\colorbox[RGB]{255,241,227}{\strut{}.''}\colorbox[RGB]{255,255,255}{\strut{} ''}\colorbox[RGB]{254,236,218}{\strut{}The}\colorbox[RGB]{254,235,216}{\strut{} fort}\colorbox[RGB]{254,236,218}{\strut{} fronts}\colorbox[RGB]{253,172,105}{\strut{} the}\colorbox[RGB]{253,215,176}{\strut{} anch}\colorbox[RGB]{253,208,163}{\strut{}orage}\colorbox[RGB]{255,244,233}{\strut{} and}\colorbox[RGB]{253,211,169}{\strut{} the}\colorbox[RGB]{253,216,179}{\strut{} }}}
\end{featureexamples}

\begin{featureexamples}
\featurechip{34M}{9493533} \textbf{Brain sciences}
\exampleline{{\unicodefont \colorbox[RGB]{255,255,255}{\strut{}------}\colorbox[RGB]{255,255,255}{\strut{}⏎}\colorbox[RGB]{255,255,255}{\strut{}mj}\colorbox[RGB]{255,255,255}{\strut{}lee}\colorbox[RGB]{255,255,255}{\strut{}⏎}\colorbox[RGB]{255,255,255}{\strut{}I}\colorbox[RGB]{255,255,255}{\strut{} really}\colorbox[RGB]{255,255,255}{\strut{} enjoy}\colorbox[RGB]{255,244,233}{\strut{} books}\colorbox[RGB]{255,255,255}{\strut{} on}\colorbox[RGB]{255,255,255}{\strut{} neuro}\colorbox[RGB]{253,188,131}{\strut{}science}\colorbox[RGB]{223,83,8}{\strut{} that}\colorbox[RGB]{253,218,182}{\strut{} change}\colorbox[RGB]{255,243,230}{\strut{} the}\colorbox[RGB]{254,221,187}{\strut{} way}\colorbox[RGB]{254,225,196}{\strut{} I}\colorbox[RGB]{254,221,187}{\strut{} think}\colorbox[RGB]{255,255,255}{\strut{} about}\colorbox[RGB]{255,243,231}{\strut{}⏎}\colorbox[RGB]{255,255,255}{\strut{}per}\colorbox[RGB]{255,255,255}{\strut{}ception}\colorbox[RGB]{254,223,191}{\strut{}.}\colorbox[RGB]{254,230,207}{\strut{}⏎}\colorbox[RGB]{254,221,187}{\strut{}⏎}\colorbox[RGB]{255,255,255}{\strut{}Ph}\colorbox[RGB]{255,255,255}{\strut{}ant}\colorbox[RGB]{255,255,255}{\strut{}o}}}
\exampleline{{\unicodefont \colorbox[RGB]{255,255,255}{\strut{}which}\colorbox[RGB]{255,241,227}{\strut{} brings}\colorbox[RGB]{255,255,255}{\strut{}⏎}\colorbox[RGB]{255,255,255}{\strut{}together}\colorbox[RGB]{255,255,255}{\strut{} engineers}\colorbox[RGB]{255,255,255}{\strut{} and}\colorbox[RGB]{255,255,255}{\strut{} neuro}\colorbox[RGB]{255,255,255}{\strut{}scient}\colorbox[RGB]{253,187,129}{\strut{}ists}\colorbox[RGB]{243,115,31}{\strut{}.}\colorbox[RGB]{254,230,207}{\strut{} If}\colorbox[RGB]{254,236,218}{\strut{} you}\colorbox[RGB]{254,229,204}{\strut{} like}\colorbox[RGB]{254,228,202}{\strut{} the}\colorbox[RGB]{254,233,211}{\strut{} intersection}\colorbox[RGB]{255,255,255}{\strut{} of}\colorbox[RGB]{255,255,255}{\strut{}⏎}\colorbox[RGB]{255,255,255}{\strut{}analog}\colorbox[RGB]{255,255,255}{\strut{},}\colorbox[RGB]{255,255,255}{\strut{} digital}\colorbox[RGB]{255,255,255}{\strut{},}\colorbox[RGB]{255,255,255}{\strut{} h}}}
\exampleline{{\unicodefont \colorbox[RGB]{255,255,255}{\strut{}ow}\colorbox[RGB]{255,255,255}{\strut{} managed}\colorbox[RGB]{255,255,255}{\strut{} to}\colorbox[RGB]{255,255,255}{\strut{} track}\colorbox[RGB]{255,255,255}{\strut{} it}\colorbox[RGB]{253,214,175}{\strut{}⏎}\colorbox[RGB]{255,255,255}{\strut{}down}\colorbox[RGB]{255,255,255}{\strut{} and}\colorbox[RGB]{255,255,255}{\strut{} buy}\colorbox[RGB]{255,255,255}{\strut{} it}\colorbox[RGB]{255,255,255}{\strut{} again}\colorbox[RGB]{254,233,212}{\strut{}.}\colorbox[RGB]{253,179,116}{\strut{} The}\colorbox[RGB]{249,131,50}{\strut{} book}\colorbox[RGB]{253,174,108}{\strut{} is}\colorbox[RGB]{253,214,175}{\strut{} from}\colorbox[RGB]{253,205,158}{\strut{} the}\colorbox[RGB]{255,244,233}{\strut{} 1960}\colorbox[RGB]{253,206,160}{\strut{}s}\colorbox[RGB]{253,157,83}{\strut{},}\colorbox[RGB]{253,183,122}{\strut{} but}\colorbox[RGB]{253,187,129}{\strut{} there}\colorbox[RGB]{253,194,141}{\strut{} are}\colorbox[RGB]{252,145,67}{\strut{} some}\colorbox[RGB]{253,167,98}{\strut{} really}\colorbox[RGB]{255,255,255}{\strut{}⏎}\colorbox[RGB]{253,158,85}{\strut{}goo}}}
\exampleline{{\unicodefont \colorbox[RGB]{253,218,182}{\strut{}interested}\colorbox[RGB]{255,239,223}{\strut{} in}\colorbox[RGB]{253,212,172}{\strut{} learning}\colorbox[RGB]{254,226,199}{\strut{} more}\colorbox[RGB]{254,221,187}{\strut{} about}\colorbox[RGB]{254,230,207}{\strut{} cognition}\colorbox[RGB]{253,214,175}{\strut{},}\colorbox[RGB]{253,201,152}{\strut{} should}\colorbox[RGB]{250,136,55}{\strut{} I}\colorbox[RGB]{253,185,126}{\strut{} study}\colorbox[RGB]{255,240,225}{\strut{}⏎}\colorbox[RGB]{255,255,255}{\strut{}neuro}\colorbox[RGB]{253,203,155}{\strut{}science}\colorbox[RGB]{253,194,141}{\strut{},}\colorbox[RGB]{253,206,160}{\strut{} or}\colorbox[RGB]{253,174,108}{\strut{} some}\colorbox[RGB]{253,158,85}{\strut{} other}\colorbox[RGB]{253,216,179}{\strut{} field}\colorbox[RGB]{255,244,233}{\strut{},}\colorbox[RGB]{253,179,116}{\strut{} or}\colorbox[RGB]{254,226,199}{\strut{} is}\colorbox[RGB]{253,214,175}{\strut{} it}}}
\exampleline{{\unicodefont \colorbox[RGB]{255,239,223}{\strut{}Con}\colorbox[RGB]{255,255,255}{\strut{}scious}\colorbox[RGB]{255,255,255}{\strut{}ness}\colorbox[RGB]{255,255,255}{\strut{} and}\colorbox[RGB]{255,255,255}{\strut{} the}\colorbox[RGB]{255,255,255}{\strut{} Social}\colorbox[RGB]{255,255,255}{\strut{} Brain}\colorbox[RGB]{253,205,158}{\strut{},''}\colorbox[RGB]{254,233,211}{\strut{} by}\colorbox[RGB]{254,229,204}{\strut{} Gra}\colorbox[RGB]{255,255,255}{\strut{}z}\colorbox[RGB]{255,244,233}{\strut{}iano}\colorbox[RGB]{253,181,119}{\strut{} is}\colorbox[RGB]{250,136,55}{\strut{} a}\colorbox[RGB]{253,179,116}{\strut{} great}\colorbox[RGB]{255,255,255}{\strut{} place}\colorbox[RGB]{253,187,129}{\strut{} to}\colorbox[RGB]{254,224,194}{\strut{} start}\colorbox[RGB]{253,183,122}{\strut{}.}\colorbox[RGB]{253,158,85}{\strut{}⏎}\colorbox[RGB]{252,153,77}{\strut{}⏎}\colorbox[RGB]{255,255,255}{\strut{}------}\colorbox[RGB]{255,255,255}{\strut{}⏎}\colorbox[RGB]{255,255,255}{\strut{}ozy}\colorbox[RGB]{255,239,223}{\strut{}⏎}\colorbox[RGB]{255,255,255}{\strut{}I}\colorbox[RGB]{255,255,255}{\strut{} would}\colorbox[RGB]{255,240,225}{\strut{} want}\colorbox[RGB]{255,255,255}{\strut{} a}\colorbox[RGB]{255,255,255}{\strut{} }}}
\end{featureexamples}

\begin{featureexamples}
\featurechip{1M}{887839} \textbf{Monuments and popular tourist attractions}
\exampleline{{\unicodefont \colorbox[RGB]{255,255,255}{\strut{}eautiful}\colorbox[RGB]{255,255,255}{\strut{} country}\colorbox[RGB]{255,244,233}{\strut{},}\colorbox[RGB]{255,255,255}{\strut{} a}\colorbox[RGB]{255,255,255}{\strut{} bit}\colorbox[RGB]{255,255,255}{\strut{} e}\colorbox[RGB]{255,255,255}{\strut{}er}\colorbox[RGB]{255,255,255}{\strut{}ily}\colorbox[RGB]{255,255,255}{\strut{} so}\colorbox[RGB]{255,255,255}{\strut{}.}\colorbox[RGB]{255,255,255}{\strut{} The}\colorbox[RGB]{255,255,255}{\strut{} blue}\colorbox[RGB]{255,255,255}{\strut{} l}\colorbox[RGB]{253,166,97}{\strut{}ago}\colorbox[RGB]{253,170,101}{\strut{}on}\colorbox[RGB]{223,83,8}{\strut{} is}\colorbox[RGB]{250,136,55}{\strut{} stunning}\colorbox[RGB]{253,179,116}{\strut{} to}\colorbox[RGB]{254,221,187}{\strut{} look}\colorbox[RGB]{255,255,255}{\strut{}⏎}\colorbox[RGB]{253,205,158}{\strut{}at}\colorbox[RGB]{253,188,131}{\strut{} but}\colorbox[RGB]{254,230,207}{\strut{} too}\colorbox[RGB]{253,205,158}{\strut{} expensive}\colorbox[RGB]{253,194,141}{\strut{} to}\colorbox[RGB]{254,224,194}{\strut{} bat}\colorbox[RGB]{254,230,207}{\strut{}he}\colorbox[RGB]{253,197,145}{\strut{} in}}}
\exampleline{{\unicodefont \colorbox[RGB]{255,255,255}{\strut{}nteresting}\colorbox[RGB]{255,255,255}{\strut{} things}\colorbox[RGB]{254,228,202}{\strut{} to}\colorbox[RGB]{253,205,158}{\strut{} visit}\colorbox[RGB]{254,221,189}{\strut{} in}\colorbox[RGB]{254,228,202}{\strut{} Egypt}\colorbox[RGB]{253,218,182}{\strut{}.}\colorbox[RGB]{255,255,255}{\strut{} The}\colorbox[RGB]{255,255,255}{\strut{}⏎}\colorbox[RGB]{255,255,255}{\strut{}py}\colorbox[RGB]{255,255,255}{\strut{}ram}\colorbox[RGB]{245,119,36}{\strut{}ids}\colorbox[RGB]{224,84,8}{\strut{} were}\colorbox[RGB]{255,244,233}{\strut{} older}\colorbox[RGB]{254,223,191}{\strut{} and}\colorbox[RGB]{255,255,255}{\strut{} less}\colorbox[RGB]{255,255,255}{\strut{} refined}\colorbox[RGB]{255,255,255}{\strut{} as}\colorbox[RGB]{255,255,255}{\strut{} this}\colorbox[RGB]{255,255,255}{\strut{} structure}\colorbox[RGB]{255,255,255}{\strut{} and}\colorbox[RGB]{255,255,255}{\strut{} the}}}
\exampleline{{\unicodefont \colorbox[RGB]{255,255,255}{\strut{}st}\colorbox[RGB]{255,255,255}{\strut{} kind}\colorbox[RGB]{255,255,255}{\strut{} of}\colorbox[RGB]{255,255,255}{\strut{} beautiful}\colorbox[RGB]{255,255,255}{\strut{}.''}\colorbox[RGB]{255,255,255}{\strut{} ''}\colorbox[RGB]{255,255,255}{\strut{}What}\colorbox[RGB]{255,255,255}{\strut{} about}\colorbox[RGB]{255,255,255}{\strut{} the}\colorbox[RGB]{255,255,255}{\strut{} Al}\colorbox[RGB]{255,243,231}{\strut{}amo}\colorbox[RGB]{233,95,15}{\strut{}?''}\colorbox[RGB]{253,206,160}{\strut{} ''}\colorbox[RGB]{250,137,56}{\strut{}Do}\colorbox[RGB]{224,84,8}{\strut{} people}\colorbox[RGB]{253,185,126}{\strut{}...''}\colorbox[RGB]{253,205,158}{\strut{} ''}\colorbox[RGB]{253,211,169}{\strut{}Oh}\colorbox[RGB]{255,255,255}{\strut{},}\colorbox[RGB]{254,225,196}{\strut{} the}\colorbox[RGB]{253,199,149}{\strut{} Al}\colorbox[RGB]{253,194,141}{\strut{}amo}\colorbox[RGB]{253,183,122}{\strut{}.''}\colorbox[RGB]{254,228,202}{\strut{} ''}\colorbox[RGB]{253,205,158}{\strut{}Yeah}\colorbox[RGB]{249,131,50}{\strut{},}\colorbox[RGB]{254,219,185}{\strut{} it}\colorbox[RGB]{237,102,20}{\strut{}'s}\colorbox[RGB]{253,185,126}{\strut{} a}\colorbox[RGB]{253,167,98}{\strut{} cool}\colorbox[RGB]{253,164,94}{\strut{} place}}}
\exampleline{{\unicodefont \colorbox[RGB]{255,255,255}{\strut{}------}\colorbox[RGB]{255,255,255}{\strut{}⏎}\colorbox[RGB]{255,255,255}{\strut{}fv}\colorbox[RGB]{255,255,255}{\strut{}rg}\colorbox[RGB]{255,255,255}{\strut{}hl}\colorbox[RGB]{255,255,255}{\strut{}⏎}\colorbox[RGB]{253,214,175}{\strut{}I}\colorbox[RGB]{253,201,152}{\strut{} went}\colorbox[RGB]{253,212,172}{\strut{} to}\colorbox[RGB]{254,232,209}{\strut{} the}\colorbox[RGB]{254,233,211}{\strut{} Lou}\colorbox[RGB]{254,228,201}{\strut{}vre}\colorbox[RGB]{253,167,98}{\strut{} in}\colorbox[RGB]{253,183,122}{\strut{} 2012}\colorbox[RGB]{251,143,64}{\strut{},}\colorbox[RGB]{253,185,126}{\strut{} and}\colorbox[RGB]{224,84,8}{\strut{} I}\colorbox[RGB]{253,179,116}{\strut{} was}\colorbox[RGB]{255,255,255}{\strut{} able}\colorbox[RGB]{253,196,144}{\strut{} to}\colorbox[RGB]{254,225,196}{\strut{} walk}\colorbox[RGB]{254,221,187}{\strut{} up}\colorbox[RGB]{254,234,214}{\strut{} the}\colorbox[RGB]{254,219,185}{\strut{} Mon}\colorbox[RGB]{255,255,255}{\strut{}a}\colorbox[RGB]{254,219,185}{\strut{} Lisa}\colorbox[RGB]{255,239,223}{\strut{} without}\colorbox[RGB]{255,255,255}{\strut{}⏎}\colorbox[RGB]{255,255,255}{\strut{}a}\colorbox[RGB]{255,255,255}{\strut{} queue}\colorbox[RGB]{253,212,172}{\strut{}.}\colorbox[RGB]{254,219,185}{\strut{} I}\colorbox[RGB]{254,233,212}{\strut{} }}}
\exampleline{{\unicodefont \colorbox[RGB]{255,255,255}{\strut{} you}\colorbox[RGB]{255,255,255}{\strut{}⏎}\colorbox[RGB]{255,255,255}{\strut{}have}\colorbox[RGB]{255,255,255}{\strut{} to}\colorbox[RGB]{255,255,255}{\strut{} go}\colorbox[RGB]{255,255,255}{\strut{} to}\colorbox[RGB]{255,255,255}{\strut{} the}\colorbox[RGB]{255,244,233}{\strut{} big}\colorbox[RGB]{253,201,152}{\strut{} tourist}\colorbox[RGB]{253,203,155}{\strut{} attractions}\colorbox[RGB]{254,226,199}{\strut{} at}\colorbox[RGB]{234,97,16}{\strut{} least}\colorbox[RGB]{251,141,62}{\strut{} once}\colorbox[RGB]{254,232,209}{\strut{} like}\colorbox[RGB]{255,255,255}{\strut{} the}\colorbox[RGB]{255,255,255}{\strut{} San}\colorbox[RGB]{255,255,255}{\strut{} Diego}\colorbox[RGB]{255,255,255}{\strut{} Zoo}\colorbox[RGB]{253,194,141}{\strut{}⏎}\colorbox[RGB]{255,240,225}{\strut{}and}\colorbox[RGB]{255,255,255}{\strut{} Sea}\colorbox[RGB]{254,223,191}{\strut{} World}\colorbox[RGB]{255,255,255}{\strut{}.}\colorbox[RGB]{255,255,255}{\strut{}⏎}\colorbox[RGB]{255,255,255}{\strut{}⏎}\colorbox[RGB]{255,255,255}{\strut{}---}}}
\end{featureexamples}

\begin{featureexamples}
\featurechip{1M}{3} \textbf{Transit infrastructure}
\exampleline{{\unicodefont \colorbox[RGB]{255,255,255}{\strut{}lly}\colorbox[RGB]{255,255,255}{\strut{} every}\colorbox[RGB]{255,255,255}{\strut{} train}\colorbox[RGB]{255,255,255}{\strut{} line}\colorbox[RGB]{255,255,255}{\strut{} has}\colorbox[RGB]{255,255,255}{\strut{} to}\colorbox[RGB]{254,236,218}{\strut{} cross}\colorbox[RGB]{254,226,199}{\strut{} one}\colorbox[RGB]{254,225,196}{\strut{} particular}\colorbox[RGB]{223,83,8}{\strut{} bridge}\colorbox[RGB]{254,234,214}{\strut{},}\colorbox[RGB]{253,206,160}{\strut{}⏎}\colorbox[RGB]{254,237,220}{\strut{}which}\colorbox[RGB]{254,237,220}{\strut{} is}\colorbox[RGB]{253,218,182}{\strut{} a}\colorbox[RGB]{255,243,230}{\strut{} massive}\colorbox[RGB]{255,244,233}{\strut{} choke}\colorbox[RGB]{254,234,214}{\strut{} point}\colorbox[RGB]{255,244,233}{\strut{}.}\colorbox[RGB]{255,255,255}{\strut{} A}\colorbox[RGB]{255,255,255}{\strut{} subway}\colorbox[RGB]{255,255,255}{\strut{} or}\colorbox[RGB]{255,255,255}{\strut{} el}}}
\exampleline{{\unicodefont \colorbox[RGB]{255,255,255}{\strut{}o}\colorbox[RGB]{255,255,255}{\strut{} many}\colorbox[RGB]{255,255,255}{\strut{} delays}\colorbox[RGB]{255,255,255}{\strut{} when}\colorbox[RGB]{255,255,255}{\strut{} we}\colorbox[RGB]{255,255,255}{\strut{} were}\colorbox[RGB]{255,255,255}{\strut{} en}\colorbox[RGB]{255,255,255}{\strut{}⏎}\colorbox[RGB]{255,255,255}{\strut{}route}\colorbox[RGB]{255,255,255}{\strut{}.}\colorbox[RGB]{255,255,255}{\strut{} Since}\colorbox[RGB]{255,255,255}{\strut{} the}\colorbox[RGB]{238,104,21}{\strut{} underwater}\colorbox[RGB]{251,139,59}{\strut{} tunnel}\colorbox[RGB]{254,232,209}{\strut{} between}\colorbox[RGB]{255,255,255}{\strut{} Oakland}\colorbox[RGB]{255,240,225}{\strut{} and}\colorbox[RGB]{254,232,209}{\strut{} SF}\colorbox[RGB]{253,199,149}{\strut{} is}\colorbox[RGB]{254,236,218}{\strut{} a}\colorbox[RGB]{255,242,229}{\strut{} choke}\colorbox[RGB]{254,224,194}{\strut{} poin}}}
\exampleline{{\unicodefont \colorbox[RGB]{255,255,255}{\strut{}le}\colorbox[RGB]{255,255,255}{\strut{} are}\colorbox[RGB]{255,255,255}{\strut{} trying}\colorbox[RGB]{255,255,255}{\strut{} to}\colorbox[RGB]{255,255,255}{\strut{} leave}\colorbox[RGB]{255,255,255}{\strut{},}\colorbox[RGB]{255,255,255}{\strut{} etc}\colorbox[RGB]{255,255,255}{\strut{})}\colorbox[RGB]{255,255,255}{\strut{} on}\colorbox[RGB]{255,255,255}{\strut{} the}\colorbox[RGB]{255,243,231}{\strut{} approaches}\colorbox[RGB]{255,238,221}{\strut{} to}\colorbox[RGB]{255,255,255}{\strut{}⏎}\colorbox[RGB]{240,107,24}{\strut{}brid}\colorbox[RGB]{248,129,47}{\strut{}ges}\colorbox[RGB]{254,228,201}{\strut{}/}\colorbox[RGB]{253,157,83}{\strut{}tun}\colorbox[RGB]{252,149,72}{\strut{}nels}\colorbox[RGB]{255,255,255}{\strut{} and}\colorbox[RGB]{255,255,255}{\strut{} in}\colorbox[RGB]{255,255,255}{\strut{} the}\colorbox[RGB]{255,255,255}{\strut{} downtown}\colorbox[RGB]{255,255,255}{\strut{}/}\colorbox[RGB]{255,255,255}{\strut{}mid}\colorbox[RGB]{255,255,255}{\strut{}town}\colorbox[RGB]{255,255,255}{\strut{} core}\colorbox[RGB]{255,255,255}{\strut{} wher}}}
\exampleline{{\unicodefont \colorbox[RGB]{255,255,255}{\strut{}ney}\colorbox[RGB]{255,255,255}{\strut{} ran}\colorbox[RGB]{255,255,255}{\strut{} out}\colorbox[RGB]{255,255,255}{\strut{} and}\colorbox[RGB]{255,255,255}{\strut{} plans}\colorbox[RGB]{255,255,255}{\strut{} to}\colorbox[RGB]{255,255,255}{\strut{} continue}\colorbox[RGB]{255,255,255}{\strut{} north}\colorbox[RGB]{255,255,255}{\strut{} across}\colorbox[RGB]{255,255,255}{\strut{} the}\colorbox[RGB]{240,107,24}{\strut{} aqu}\colorbox[RGB]{246,121,38}{\strut{}ed}\colorbox[RGB]{255,242,229}{\strut{}uct}\colorbox[RGB]{255,255,255}{\strut{} toward}\colorbox[RGB]{255,255,255}{\strut{} W}\colorbox[RGB]{255,255,255}{\strut{}rex}\colorbox[RGB]{255,255,255}{\strut{}ham}\colorbox[RGB]{255,255,255}{\strut{} had}\colorbox[RGB]{255,255,255}{\strut{} to}\colorbox[RGB]{255,255,255}{\strut{} be}\colorbox[RGB]{255,255,255}{\strut{} abandoned}\colorbox[RGB]{255,255,255}{\strut{}.''}\colorbox[RGB]{255,255,255}{\strut{} ''}\colorbox[RGB]{255,255,255}{\strut{}Now}\colorbox[RGB]{255,255,255}{\strut{},}\colorbox[RGB]{255,255,255}{\strut{} }}}
\exampleline{{\unicodefont \colorbox[RGB]{255,255,255}{\strut{}running}\colorbox[RGB]{255,255,255}{\strut{}.}\colorbox[RGB]{255,255,255}{\strut{}⏎}\colorbox[RGB]{255,255,255}{\strut{}This}\colorbox[RGB]{255,255,255}{\strut{} is}\colorbox[RGB]{255,255,255}{\strut{} especially}\colorbox[RGB]{255,255,255}{\strut{} the}\colorbox[RGB]{255,255,255}{\strut{} case}\colorbox[RGB]{255,255,255}{\strut{} for}\colorbox[RGB]{253,197,145}{\strut{} the}\colorbox[RGB]{253,199,149}{\strut{} Trans}\colorbox[RGB]{242,111,28}{\strut{}bay}\colorbox[RGB]{247,125,42}{\strut{} T}\colorbox[RGB]{245,119,36}{\strut{}ube}\colorbox[RGB]{255,244,233}{\strut{} which}\colorbox[RGB]{255,255,255}{\strut{} requires}\colorbox[RGB]{255,255,255}{\strut{} a}\colorbox[RGB]{255,255,255}{\strut{} lot}\colorbox[RGB]{255,255,255}{\strut{} of}\colorbox[RGB]{255,255,255}{\strut{}⏎}\colorbox[RGB]{255,255,255}{\strut{}attention}\colorbox[RGB]{254,235,216}{\strut{}.}\colorbox[RGB]{255,255,255}{\strut{}⏎}\colorbox[RGB]{255,255,255}{\strut{}⏎}\colorbox[RGB]{255,255,255}{\strut{}If}\colorbox[RGB]{255,255,255}{\strut{} B}\colorbox[RGB]{255,255,255}{\strut{}ART}\colorbox[RGB]{255,255,255}{\strut{} }}}
\end{featureexamples}

While these examples suggest interpretations for each feature, more work needs to be done to establish that our interpretations truly capture the behavior and function of the corresponding features. Concretely, for each feature, we attempt to establish the following claims:

\begin{enumerate}
    \item When the feature is active, the relevant concept is reliably present in the context (\hyperref[sec:assessing-tour-specificity]{specificity}).
    \item Intervening on the feature’s activation produces relevant downstream behavior (\hyperref[sec:assessing-tour-influence]{influence on behavior}).
\end{enumerate}

\subsubsection{Specificity}\label{sec:assessing-tour-specificity}

It is difficult to rigorously measure the extent to which a concept is present in a text input. In our prior work, we focused on features that unambiguously corresponded to sets of tokens (e.g., Arabic script or DNA sequences) and computed the likelihood of that set of tokens relative to the rest of the vocabulary, conditioned on the feature’s activation. This technique does not generalize to more abstract features. Instead, to demonstrate specificity in this work we more heavily leverage automated interpretability methods (similar to \cite{bills2023language,bricken2023monosemanticity}). We use the same automated interpretability pipeline as in our previous work \cite{bricken2023monosemanticity} in the \hyperref[sec:assessing-features-v-neurons]{features vs. neurons} section below, but we additionally find that current-generation models can now more accurately rate text samples according to how well they match a proposed feature interpretation.

We constructed the following rubric for scoring how a feature’s description relates to the text on which it fires. We then asked Claude 3 Opus to rate feature activations at many tokens on that rubric.

\begin{itemize}
    \item 0 -- The feature is completely irrelevant throughout the context (relative to the base distribution of the internet).
    \item 1 -- The feature is related to the context, but not near the highlighted text or only vaguely related.
    \item 2 -- The feature is only loosely related to the highlighted text or related to the context near the highlighted text.
    \item 3 -- The feature cleanly identifies the activating text.
\end{itemize}

By scoring examples of activating text, we provide a measure of specificity for each feature.\footnote{We also manually checked a number of examples to ensure they were generally handled correctly.} The features in this section are selected to have straightforward interpretations, to make automated interpretability analysis more reliable. They are not intended to be a representative example of all features in our SAEs. Later, we provide an analysis of the interpretability of randomly sampled features. We also conduct in-depth explorations throughout the paper of many more features which have interesting interpretations which are more abstract or nuanced, and thus more difficult to quantitatively assess.

Below we show distributions of feature activations (excluding zero activations) for the four features mentioned above, along with example text and image inputs that induce low and high activations. Note that these features also activate on relevant images, despite our only performing dictionary learning on a text-based dataset!

First, we study a Golden Gate Bridge feature \featurechip{34M}{31164353}. Its greatest activations are essentially all references to the bridge, and weaker activations also include related tourist attractions, similar bridges, and other monuments. Next, a brain sciences feature \featurechip{34M}{9493533} activates on discussions of neuroscience books and courses, as well as cognitive science, psychology, and related philosophy. In the 1M training run, we also find a feature that strongly activates for various kinds of transit infrastructure \featurechip{1M}{3} including trains, ferries, tunnels, bridges, and even wormholes! A final feature \featurechip{1M}{887839}  responds to popular tourist attractions including the Eiffel Tower, the Tower of Pisa, the Golden Gate Bridge, and the Sistine Chapel.

To quantify specificity, we used Claude 3 Opus to automatically score examples that activate these features according to the rubric above, with roughly 1000 activations of the feature drawn from the dataset used to train the dictionary learning model. We plot the frequency of each rubric score as a function of the feature’s activation level. We see that inputs that induce strong feature activations are all judged to be highly consistent with the proposed interpretation.

\begin{figure}[!htp]
    \centering
    \includegraphics[width=0.87\textwidth,height=0.7\textheight,keepaspectratio]{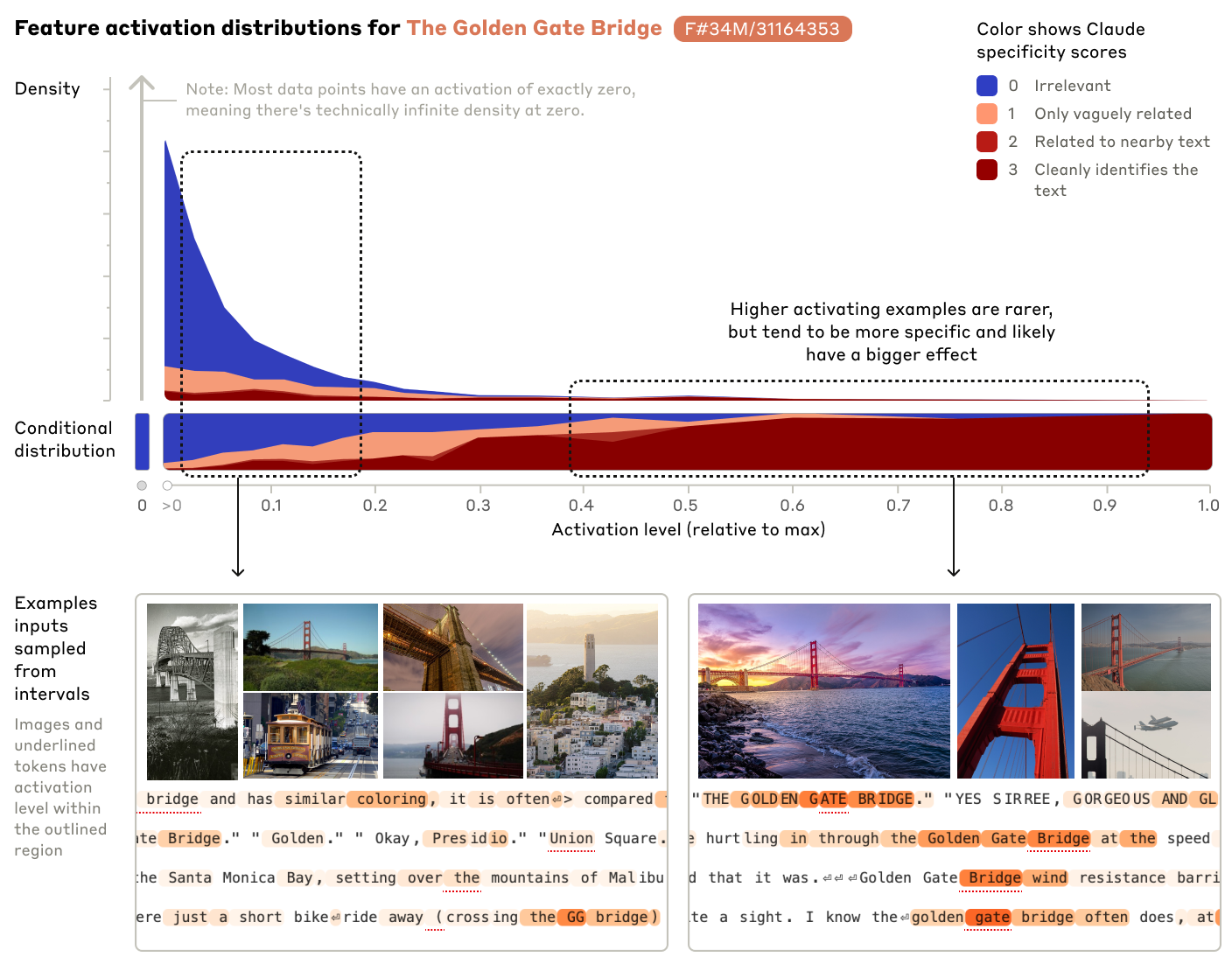}
    \label{fig:gdoc_5}
\end{figure}

\begin{figure}[!htp]
    \centering
    \includegraphics[width=0.87\textwidth,height=0.7\textheight,keepaspectratio]{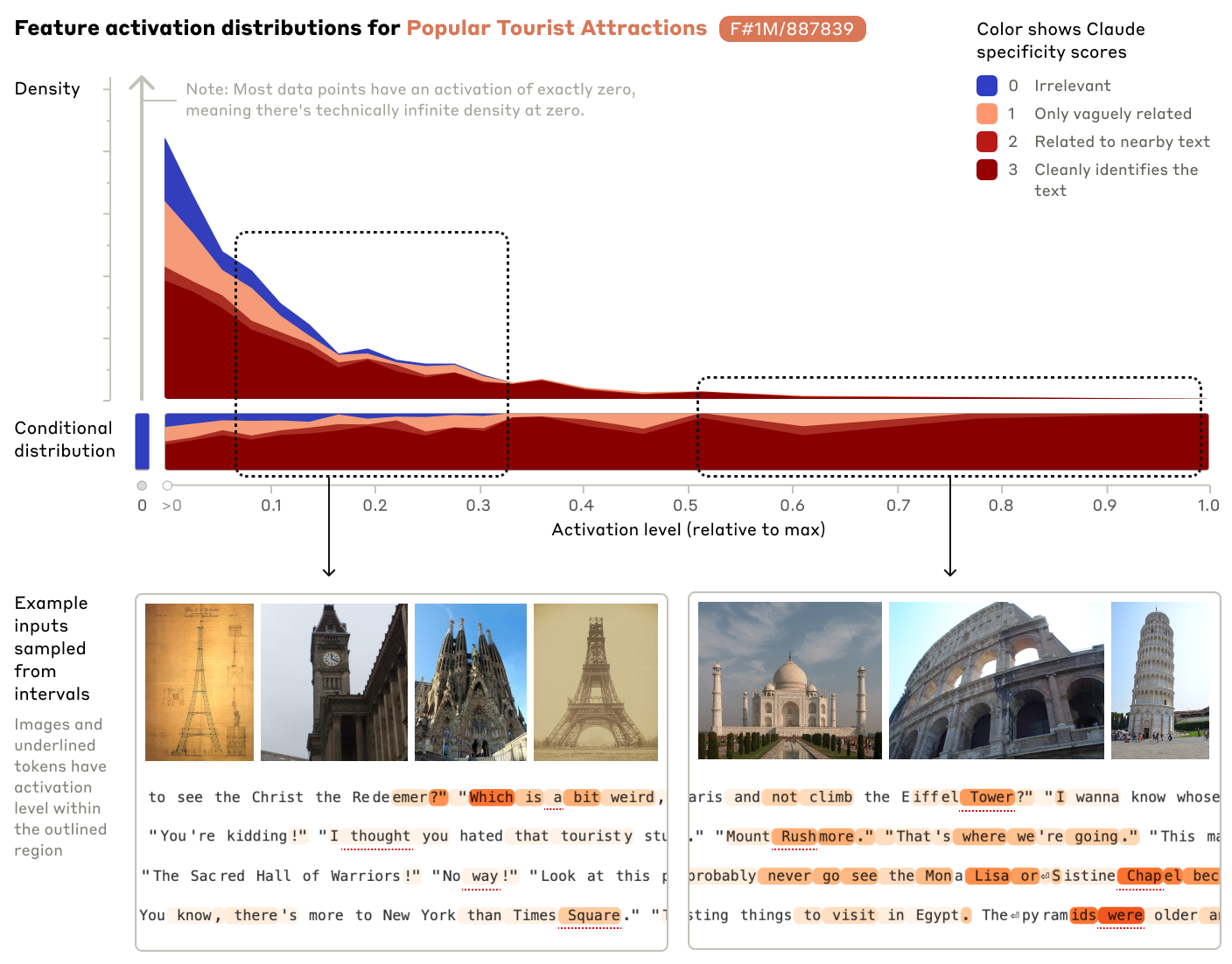}
    \label{fig:gdoc_6}
\end{figure}

\begin{figure}[!htp]
    \centering
    \includegraphics[width=0.87\textwidth,height=0.7\textheight,keepaspectratio]{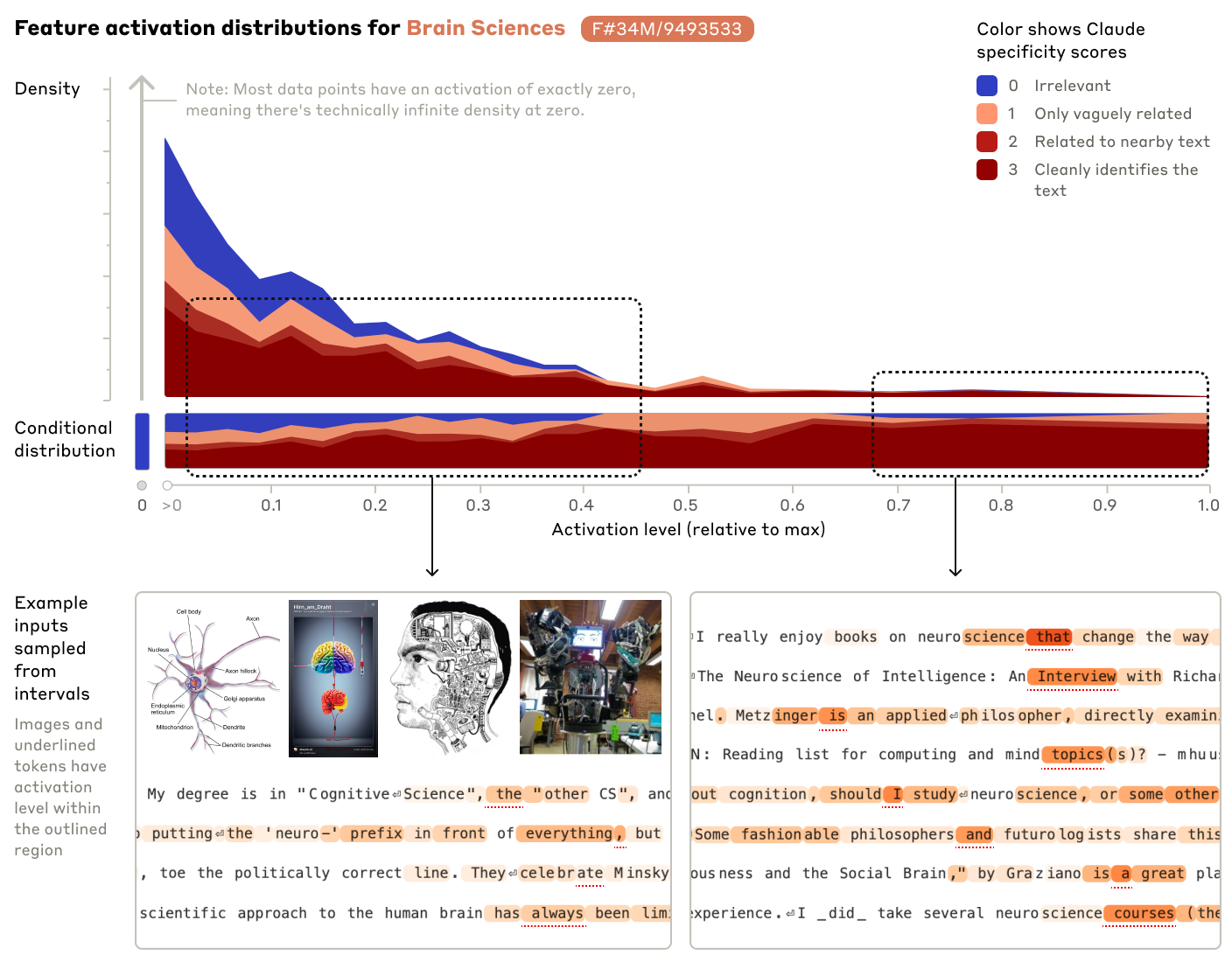}
    \label{fig:gdoc_7}
\end{figure}

\begin{figure}[!htp]
    \centering
    \includegraphics[width=0.87\textwidth,height=0.7\textheight,keepaspectratio]{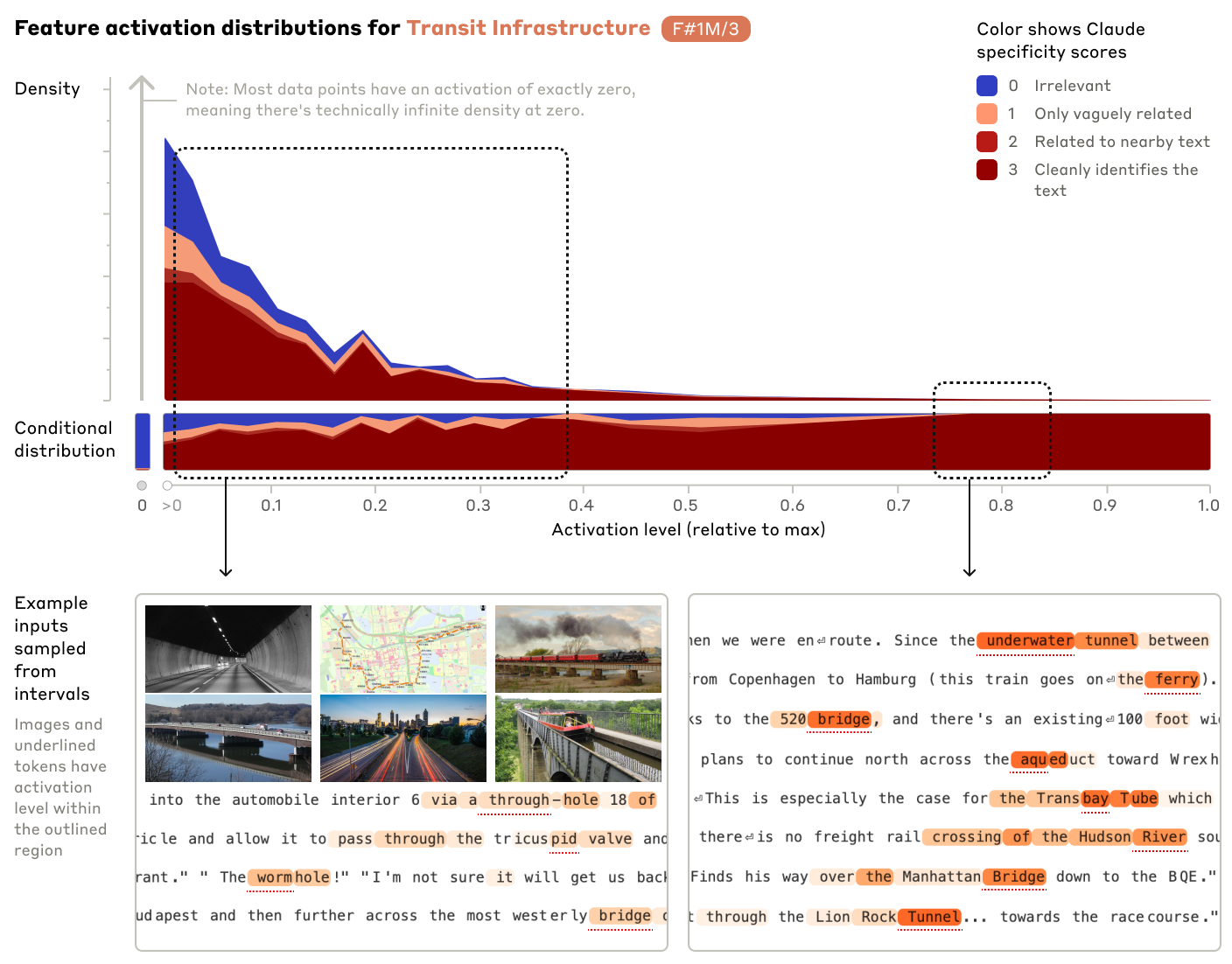}
    \label{fig:gdoc_8}
\end{figure}

As in \textit{Towards Monosemanticity}, we see that these features become less specific as the activation strength weakens. This could be due to the model using activation strengths to represent confidence in a concept being present. Or it may be that the feature activates most strongly for central examples of the feature, but weakly for related ideas -- for example, the Golden Gate Bridge feature \featurechip{34M}{31164353} appears to weakly activate for other San Francisco landmarks. It could also reflect imperfection in our dictionary learning procedure. For example, it may be that the architecture of the autoencoder is not able to extract and discriminate among features as cleanly as we might want. And of course interference from features that are not exactly orthogonal could also be a culprit, making it more difficult for Sonnet itself to activate features on precisely the right examples. It is also plausible that our feature interpretations slightly misrepresent the feature's actual function, and that this inaccuracy manifests more clearly at lower activations. Nonetheless, we often find that lower activations tend to maintain some specificity to our interpretations, including related concepts or generalizations of the core feature. As an illustrative example, weak activations of the transit infrastructure feature \featurechip{1M}{3} include procedural mechanics instructions describing which through-holes to use for particular parts.

\clearpage

Moreover, we expect that very weak activations of features are not especially meaningful, and thus we are not too concerned with low specificity scores for these activation ranges. For instance, we have observed that techniques such as rounding feature activations below a threshold to zero can improve specificity at the low-activation end of the spectrum without substantially increasing the reconstruction error of the SAE, and there are a variety of techniques in the literature that potentially address the same issue \cite{rajamanoharan2024improving,riggs2024improvingsae}.

Regardless, the activations that have the most impact on the model’s behavior are the largest ones, so it is encouraging to see high specificity among the strong activations.

Note that we have had more difficulty in quantifying feature \textit{sensitivity} -- that is, how reliably a feature activates for text that matches our proposed interpretation -- in a scalable, rigorous way. This is due to the difficulty of generating text related to a concept in an unbiased fashion. Moreover, many features may represent something more specific than we are able to glean with our visualizations, in which case they would not respond reliably to text selected based on our proposed interpretation, and this problem gets harder the more abstract the features are. As a basic check, however, we observe that the Golden Gate Bridge feature still fires strongly on the first sentence of the Wikipedia article for the Golden Gate Bridge in various languages (after removing any English parentheticals). In fact, the Golden Gate Bridge feature is the top feature by average activation for every example below.

\begin{featureexamples}
\featurechip{34M}{31164353} \textbf{Golden Gate Bridge}
Multilingual examples
\exampleline{{\unicodefont \colorbox[RGB]{255,255,255}{\strut{}金}\colorbox[RGB]{255,255,255}{\strut{}門}\colorbox[RGB]{255,255,255}{\strut{}大}\colorbox[RGB]{255,255,255}{\strut{}橋}\colorbox[RGB]{255,239,223}{\strut{}是}\colorbox[RGB]{255,255,255}{\strut{}一}\colorbox[RGB]{255,255,255}{\strut{}座}\colorbox[RGB]{255,255,255}{\strut{}位}\colorbox[RGB]{255,244,233}{\strut{}於}\colorbox[RGB]{255,255,255}{\strut{}美}\colorbox[RGB]{255,238,221}{\strut{}國}\colorbox[RGB]{255,244,233}{\strut{}加}\colorbox[RGB]{255,255,255}{\strut{}利}\colorbox[RGB]{255,255,255}{\strut{}福}\colorbox[RGB]{253,216,179}{\strut{}尼}\colorbox[RGB]{253,201,152}{\strut{}亞}\colorbox[RGB]{253,197,145}{\strut{}州}\colorbox[RGB]{254,228,201}{\strut{}舊}\colorbox[RGB]{255,243,231}{\strut{}金}\colorbox[RGB]{253,199,149}{\strut{}山}\colorbox[RGB]{253,201,152}{\strut{}的}\colorbox[RGB]{254,230,207}{\strut{}懸}\colorbox[RGB]{254,236,218}{\strut{}索}\colorbox[RGB]{253,194,141}{\strut{}橋}\colorbox[RGB]{253,194,141}{\strut{},}\colorbox[RGB]{253,164,94}{\strut{}它}\colorbox[RGB]{253,208,163}{\strut{}跨}\colorbox[RGB]{253,174,108}{\strut{}越}\colorbox[RGB]{254,219,185}{\strut{}聯}\colorbox[RGB]{253,196,144}{\strut{}接}\colorbox[RGB]{254,226,199}{\strut{}舊}\colorbox[RGB]{254,223,191}{\strut{}金}\colorbox[RGB]{253,196,144}{\strut{}山}\colorbox[RGB]{254,221,187}{\strut{}灣}\colorbox[RGB]{253,201,152}{\strut{}和}\colorbox[RGB]{254,234,214}{\strut{}太}\colorbox[RGB]{255,255,255}{\strut{}平}\colorbox[RGB]{253,192,137}{\strut{}洋}\colorbox[RGB]{253,197,145}{\strut{}的}\colorbox[RGB]{253,206,160}{\strut{}金}\colorbox[RGB]{253,203,155}{\strut{}門}\colorbox[RGB]{254,225,196}{\strut{}海}\colorbox[RGB]{253,216,179}{\strut{}峽}\colorbox[RGB]{253,192,137}{\strut{},}\colorbox[RGB]{254,229,204}{\strut{}南}\colorbox[RGB]{254,221,187}{\strut{}端}\colorbox[RGB]{255,239,223}{\strut{}連}\colorbox[RGB]{253,205,158}{\strut{}接}\colorbox[RGB]{255,238,221}{\strut{}舊}\colorbox[RGB]{254,219,185}{\strut{}金}\colorbox[RGB]{253,206,160}{\strut{}山}\colorbox[RGB]{253,216,179}{\strut{}的}\colorbox[RGB]{254,223,191}{\strut{}北}\colorbox[RGB]{254,228,202}{\strut{}端}\colorbox[RGB]{254,230,207}{\strut{},}\colorbox[RGB]{254,230,207}{\strut{}北}\colorbox[RGB]{253,215,176}{\strut{}端}\colorbox[RGB]{255,241,227}{\strut{}接}\colorbox[RGB]{253,194,141}{\strut{}通}\colorbox[RGB]{254,233,212}{\strut{}馬}\colorbox[RGB]{254,230,207}{\strut{}林}\colorbox[RGB]{254,221,189}{\strut{}縣}\colorbox[RGB]{254,224,194}{\strut{}。}}}
\exampleline{{\unicodefont \colorbox[RGB]{255,255,255}{\strut{}ゴ}\colorbox[RGB]{255,255,255}{\strut{}ール}\colorbox[RGB]{255,255,255}{\strut{}デ}\colorbox[RGB]{255,255,255}{\strut{}ン}\colorbox[RGB]{255,255,255}{\strut{}・}\colorbox[RGB]{255,255,255}{\strut{}ゲ}\colorbox[RGB]{253,211,169}{\strut{}ート}\colorbox[RGB]{255,255,255}{\strut{}・}\colorbox[RGB]{255,240,225}{\strut{}ブ}\colorbox[RGB]{255,240,225}{\strut{}リ}\colorbox[RGB]{255,255,255}{\strut{}ッ}\colorbox[RGB]{250,136,55}{\strut{}ジ}\colorbox[RGB]{253,208,163}{\strut{}、}\colorbox[RGB]{253,188,131}{\strut{}金}\colorbox[RGB]{254,228,202}{\strut{}門}\colorbox[RGB]{254,221,189}{\strut{}橋}\colorbox[RGB]{253,201,152}{\strut{}は}\colorbox[RGB]{253,183,122}{\strut{}、}\colorbox[RGB]{253,203,155}{\strut{}ア}\colorbox[RGB]{255,242,229}{\strut{}メ}\colorbox[RGB]{253,216,179}{\strut{}リ}\colorbox[RGB]{254,236,218}{\strut{}カ}\colorbox[RGB]{253,211,169}{\strut{}西}\colorbox[RGB]{255,255,255}{\strut{}海}\colorbox[RGB]{253,192,137}{\strut{}岸}\colorbox[RGB]{253,196,144}{\strut{}の}\colorbox[RGB]{253,216,179}{\strut{}サ}\colorbox[RGB]{254,223,191}{\strut{}ン}\colorbox[RGB]{255,255,255}{\strut{}フ}\colorbox[RGB]{254,221,189}{\strut{}ラ}\colorbox[RGB]{253,218,182}{\strut{}ン}\colorbox[RGB]{255,255,255}{\strut{}シ}\colorbox[RGB]{253,166,97}{\strut{}ス}\colorbox[RGB]{253,208,163}{\strut{}コ}\colorbox[RGB]{253,203,155}{\strut{}湾}\colorbox[RGB]{253,172,105}{\strut{}と}\colorbox[RGB]{254,224,194}{\strut{}太}\colorbox[RGB]{255,255,255}{\strut{}平}\colorbox[RGB]{253,164,94}{\strut{}洋}\colorbox[RGB]{253,203,155}{\strut{}が}\colorbox[RGB]{255,255,255}{\strut{}接}\colorbox[RGB]{255,243,231}{\strut{}続}\colorbox[RGB]{253,177,113}{\strut{}する}\colorbox[RGB]{253,215,176}{\strut{}ゴ}\colorbox[RGB]{254,219,185}{\strut{}ール}\colorbox[RGB]{253,166,97}{\strut{}デ}\colorbox[RGB]{253,174,108}{\strut{}ン}\colorbox[RGB]{254,228,201}{\strut{}ゲ}\colorbox[RGB]{253,158,85}{\strut{}ート}\colorbox[RGB]{253,206,160}{\strut{}海}\colorbox[RGB]{253,208,163}{\strut{}峡}\colorbox[RGB]{253,205,158}{\strut{}に}\colorbox[RGB]{255,255,255}{\strut{}架}\colorbox[RGB]{255,255,255}{\strut{}か}\colorbox[RGB]{253,177,113}{\strut{}る}\colorbox[RGB]{253,215,176}{\strut{}吊}\colorbox[RGB]{254,232,209}{\strut{}橋}\colorbox[RGB]{253,181,119}{\strut{}。}}}
\exampleline{{\unicodefont \colorbox[RGB]{255,255,255}{\strut{}골}\colorbox[RGB]{255,255,255}{\strut{}든}\colorbox[RGB]{255,255,255}{\strut{}게}\colorbox[RGB]{255,255,255}{\strut{}이}\colorbox[RGB]{254,230,207}{\strut{}트}\colorbox[RGB]{253,206,160}{\strut{} 교}\colorbox[RGB]{253,176,111}{\strut{} 또}\colorbox[RGB]{254,228,202}{\strut{}는}\colorbox[RGB]{253,179,116}{\strut{} 금}\colorbox[RGB]{254,233,212}{\strut{}문}\colorbox[RGB]{254,229,204}{\strut{}교}\colorbox[RGB]{255,240,225}{\strut{} 는}\colorbox[RGB]{253,160,88}{\strut{} 미}\colorbox[RGB]{255,243,230}{\strut{}국}\colorbox[RGB]{254,223,191}{\strut{} 캘}\colorbox[RGB]{254,233,211}{\strut{}리}\colorbox[RGB]{253,185,126}{\strut{}포}\colorbox[RGB]{253,170,101}{\strut{}니}\colorbox[RGB]{254,234,214}{\strut{}아주}\colorbox[RGB]{253,190,134}{\strut{} 골}\colorbox[RGB]{253,181,119}{\strut{}든}\colorbox[RGB]{255,255,255}{\strut{}게}\colorbox[RGB]{255,243,231}{\strut{}이}\colorbox[RGB]{253,174,108}{\strut{}트}\colorbox[RGB]{253,197,145}{\strut{} 해}\colorbox[RGB]{253,190,134}{\strut{}협}\colorbox[RGB]{253,216,179}{\strut{}에}\colorbox[RGB]{253,201,152}{\strut{} 위치}\colorbox[RGB]{254,225,196}{\strut{}한}\colorbox[RGB]{251,143,64}{\strut{} 현}\colorbox[RGB]{255,243,231}{\strut{}수}\colorbox[RGB]{254,228,201}{\strut{}교}\colorbox[RGB]{255,255,255}{\strut{}이}\colorbox[RGB]{255,255,255}{\strut{}다}\colorbox[RGB]{253,185,126}{\strut{}.}\colorbox[RGB]{253,172,105}{\strut{} 골}\colorbox[RGB]{253,170,101}{\strut{}든}\colorbox[RGB]{255,255,255}{\strut{}게}\colorbox[RGB]{253,170,101}{\strut{}이}\colorbox[RGB]{253,155,80}{\strut{}트}\colorbox[RGB]{253,185,126}{\strut{} 교}\colorbox[RGB]{253,211,169}{\strut{}는}\colorbox[RGB]{253,181,119}{\strut{} 캘}\colorbox[RGB]{253,206,160}{\strut{}리}\colorbox[RGB]{254,230,207}{\strut{}포}\colorbox[RGB]{253,209,166}{\strut{}니}\colorbox[RGB]{254,230,207}{\strut{}아주}\colorbox[RGB]{253,206,160}{\strut{} 샌}\colorbox[RGB]{255,255,255}{\strut{}프}\colorbox[RGB]{255,241,227}{\strut{}란}\colorbox[RGB]{255,255,255}{\strut{}시}\colorbox[RGB]{253,212,172}{\strut{}스}\colorbox[RGB]{253,216,179}{\strut{}코}\colorbox[RGB]{253,218,182}{\strut{}와}\colorbox[RGB]{253,179,116}{\strut{} 캘}\colorbox[RGB]{255,244,233}{\strut{}리}\colorbox[RGB]{255,255,255}{\strut{}포}\colorbox[RGB]{253,211,169}{\strut{}니}\colorbox[RGB]{255,239,223}{\strut{}아주}\colorbox[RGB]{253,183,122}{\strut{} 마}\colorbox[RGB]{254,235,216}{\strut{}린}\colorbox[RGB]{254,232,209}{\strut{} 군}\colorbox[RGB]{254,229,204}{\strut{} 을}\colorbox[RGB]{254,229,204}{\strut{} 연}\colorbox[RGB]{255,255,255}{\strut{}결}\colorbox[RGB]{255,243,230}{\strut{}한다}\colorbox[RGB]{254,232,209}{\strut{}.}}}
\exampleline{{\unicodefont \colorbox[RGB]{255,255,255}{\strut{}м}\colorbox[RGB]{255,255,255}{\strut{}ост}\colorbox[RGB]{255,241,227}{\strut{} з}\colorbox[RGB]{255,239,223}{\strut{}ол}\colorbox[RGB]{255,255,255}{\strut{}от}\colorbox[RGB]{255,255,255}{\strut{}ы}\colorbox[RGB]{255,255,255}{\strut{}́}\colorbox[RGB]{255,244,233}{\strut{}е}\colorbox[RGB]{255,255,255}{\strut{} в}\colorbox[RGB]{255,255,255}{\strut{}ор}\colorbox[RGB]{255,255,255}{\strut{}о}\colorbox[RGB]{255,255,255}{\strut{}́}\colorbox[RGB]{255,255,255}{\strut{}та}\colorbox[RGB]{255,255,255}{\strut{} ---}\colorbox[RGB]{255,255,255}{\strut{} в}\colorbox[RGB]{255,255,255}{\strut{}ис}\colorbox[RGB]{255,255,255}{\strut{}я}\colorbox[RGB]{255,244,233}{\strut{}чи}\colorbox[RGB]{255,244,233}{\strut{}й}\colorbox[RGB]{255,255,255}{\strut{} м}\colorbox[RGB]{255,255,255}{\strut{}ост}\colorbox[RGB]{255,255,255}{\strut{} ч}\colorbox[RGB]{255,255,255}{\strut{}ер}\colorbox[RGB]{255,255,255}{\strut{}ез}\colorbox[RGB]{255,244,233}{\strut{} пр}\colorbox[RGB]{255,255,255}{\strut{}ол}\colorbox[RGB]{255,255,255}{\strut{}ив}\colorbox[RGB]{254,237,220}{\strut{} з}\colorbox[RGB]{254,235,216}{\strut{}ол}\colorbox[RGB]{255,238,221}{\strut{}от}\colorbox[RGB]{255,244,233}{\strut{}ые}\colorbox[RGB]{253,214,175}{\strut{} в}\colorbox[RGB]{255,255,255}{\strut{}ор}\colorbox[RGB]{254,223,191}{\strut{}от}\colorbox[RGB]{254,221,189}{\strut{}а}\colorbox[RGB]{254,225,196}{\strut{}.}\colorbox[RGB]{253,190,134}{\strut{} о}\colorbox[RGB]{253,190,134}{\strut{}н}\colorbox[RGB]{254,225,196}{\strut{} со}\colorbox[RGB]{255,243,230}{\strut{}ед}\colorbox[RGB]{255,241,227}{\strut{}ин}\colorbox[RGB]{253,212,172}{\strut{}я}\colorbox[RGB]{253,197,145}{\strut{}ет}\colorbox[RGB]{255,238,221}{\strut{} г}\colorbox[RGB]{255,240,225}{\strut{}ор}\colorbox[RGB]{255,238,221}{\strut{}од}\colorbox[RGB]{254,228,202}{\strut{} с}\colorbox[RGB]{254,219,185}{\strut{}ан}\colorbox[RGB]{254,225,196}{\strut{}-}\colorbox[RGB]{253,212,172}{\strut{}ф}\colorbox[RGB]{254,230,207}{\strut{}ран}\colorbox[RGB]{252,151,74}{\strut{}ц}\colorbox[RGB]{253,155,80}{\strut{}ис}\colorbox[RGB]{253,203,155}{\strut{}ко}\colorbox[RGB]{253,211,169}{\strut{} на}\colorbox[RGB]{255,239,223}{\strut{} с}\colorbox[RGB]{255,239,223}{\strut{}ев}\colorbox[RGB]{254,228,201}{\strut{}ер}\colorbox[RGB]{254,232,209}{\strut{}е}\colorbox[RGB]{254,223,191}{\strut{} пол}\colorbox[RGB]{254,229,204}{\strut{}у}\colorbox[RGB]{255,239,223}{\strut{}ост}\colorbox[RGB]{254,219,185}{\strut{}ров}\colorbox[RGB]{254,221,187}{\strut{}а}\colorbox[RGB]{253,172,105}{\strut{} с}\colorbox[RGB]{254,230,207}{\strut{}ан}\colorbox[RGB]{255,244,233}{\strut{}-}\colorbox[RGB]{253,181,119}{\strut{}ф}\colorbox[RGB]{255,239,223}{\strut{}ран}\colorbox[RGB]{253,174,108}{\strut{}ц}\colorbox[RGB]{253,172,105}{\strut{}ис}\colorbox[RGB]{254,228,201}{\strut{}ко}\colorbox[RGB]{253,164,94}{\strut{} и}\colorbox[RGB]{254,230,207}{\strut{} ю}\colorbox[RGB]{254,232,209}{\strut{}ж}\colorbox[RGB]{253,216,179}{\strut{}н}\colorbox[RGB]{253,216,179}{\strut{}ую}\colorbox[RGB]{255,255,255}{\strut{} ч}\colorbox[RGB]{255,255,255}{\strut{}а}\colorbox[RGB]{253,181,119}{\strut{}сть}\colorbox[RGB]{254,235,216}{\strut{} о}\colorbox[RGB]{254,234,214}{\strut{}к}\colorbox[RGB]{254,236,218}{\strut{}ру}\colorbox[RGB]{254,237,220}{\strut{}г}\colorbox[RGB]{254,221,187}{\strut{}а}\colorbox[RGB]{253,190,134}{\strut{} м}\colorbox[RGB]{254,236,218}{\strut{}ар}\colorbox[RGB]{253,216,179}{\strut{}ин}\colorbox[RGB]{254,221,189}{\strut{},}\colorbox[RGB]{254,221,187}{\strut{} р}\colorbox[RGB]{255,243,231}{\strut{}я}\colorbox[RGB]{255,244,233}{\strut{}д}\colorbox[RGB]{254,228,201}{\strut{}ом}\colorbox[RGB]{253,188,131}{\strut{} с}\colorbox[RGB]{253,203,155}{\strut{} при}\colorbox[RGB]{255,244,233}{\strut{}г}\colorbox[RGB]{255,255,255}{\strut{}ор}\colorbox[RGB]{255,255,255}{\strut{}од}\colorbox[RGB]{255,244,233}{\strut{}ом}\colorbox[RGB]{254,221,189}{\strut{} с}\colorbox[RGB]{254,233,212}{\strut{}ос}\colorbox[RGB]{255,244,233}{\strut{}ал}\colorbox[RGB]{254,228,201}{\strut{}ит}\colorbox[RGB]{254,229,204}{\strut{}о}\colorbox[RGB]{253,212,172}{\strut{}.}}}
\exampleline{{\unicodefont \colorbox[RGB]{255,255,255}{\strut{}C}\colorbox[RGB]{255,255,255}{\strut{}ầ}\colorbox[RGB]{255,255,255}{\strut{}u}\colorbox[RGB]{255,255,255}{\strut{} C}\colorbox[RGB]{255,255,255}{\strut{}ổ}\colorbox[RGB]{255,255,255}{\strut{}ng}\colorbox[RGB]{255,255,255}{\strut{} V}\colorbox[RGB]{255,255,255}{\strut{}à}\colorbox[RGB]{255,242,229}{\strut{}ng}\colorbox[RGB]{255,255,255}{\strut{} ho}\colorbox[RGB]{255,255,255}{\strut{}ặ}\colorbox[RGB]{254,237,220}{\strut{}c}\colorbox[RGB]{255,255,255}{\strut{} Kim}\colorbox[RGB]{255,255,255}{\strut{} M}\colorbox[RGB]{255,255,255}{\strut{}ô}\colorbox[RGB]{255,255,255}{\strut{}n}\colorbox[RGB]{255,255,255}{\strut{} ki}\colorbox[RGB]{255,255,255}{\strut{}ề}\colorbox[RGB]{255,255,255}{\strut{}u}\colorbox[RGB]{255,255,255}{\strut{} là}\colorbox[RGB]{255,238,221}{\strut{} m}\colorbox[RGB]{255,241,227}{\strut{}ộ}\colorbox[RGB]{254,237,220}{\strut{}t}\colorbox[RGB]{255,255,255}{\strut{} c}\colorbox[RGB]{255,255,255}{\strut{}â}\colorbox[RGB]{255,255,255}{\strut{}y}\colorbox[RGB]{255,255,255}{\strut{} c}\colorbox[RGB]{255,255,255}{\strut{}ầ}\colorbox[RGB]{255,242,229}{\strut{}u}\colorbox[RGB]{255,243,230}{\strut{} tre}\colorbox[RGB]{255,255,255}{\strut{}o}\colorbox[RGB]{254,234,214}{\strut{} b}\colorbox[RGB]{255,255,255}{\strut{}ắ}\colorbox[RGB]{255,255,255}{\strut{}c}\colorbox[RGB]{255,241,227}{\strut{} qu}\colorbox[RGB]{255,240,225}{\strut{}a}\colorbox[RGB]{255,239,223}{\strut{} C}\colorbox[RGB]{255,255,255}{\strut{}ổ}\colorbox[RGB]{255,238,221}{\strut{}ng}\colorbox[RGB]{254,235,216}{\strut{} V}\colorbox[RGB]{255,239,223}{\strut{}à}\colorbox[RGB]{254,234,214}{\strut{}ng}\colorbox[RGB]{254,232,209}{\strut{},}\colorbox[RGB]{255,242,229}{\strut{} eo}\colorbox[RGB]{255,255,255}{\strut{} bi}\colorbox[RGB]{254,236,218}{\strut{}ể}\colorbox[RGB]{255,240,225}{\strut{}n}\colorbox[RGB]{255,243,230}{\strut{} r}\colorbox[RGB]{255,255,255}{\strut{}ộ}\colorbox[RGB]{255,255,255}{\strut{}ng}\colorbox[RGB]{255,243,230}{\strut{} m}\colorbox[RGB]{255,239,223}{\strut{}ộ}\colorbox[RGB]{255,239,223}{\strut{}t}\colorbox[RGB]{255,239,223}{\strut{} d}\colorbox[RGB]{255,255,255}{\strut{}ặ}\colorbox[RGB]{255,238,221}{\strut{}m}\colorbox[RGB]{255,243,230}{\strut{} (}\colorbox[RGB]{255,255,255}{\strut{}1}\colorbox[RGB]{255,239,223}{\strut{},}\colorbox[RGB]{255,255,255}{\strut{}6}\colorbox[RGB]{254,233,211}{\strut{} km}\colorbox[RGB]{254,228,202}{\strut{})}\colorbox[RGB]{255,244,233}{\strut{} n}\colorbox[RGB]{254,232,209}{\strut{}ố}\colorbox[RGB]{254,230,207}{\strut{}i}\colorbox[RGB]{255,255,255}{\strut{} li}\colorbox[RGB]{254,228,202}{\strut{}ề}\colorbox[RGB]{254,229,204}{\strut{}n}\colorbox[RGB]{255,255,255}{\strut{} v}\colorbox[RGB]{255,255,255}{\strut{}ị}\colorbox[RGB]{255,241,227}{\strut{}nh}\colorbox[RGB]{254,223,191}{\strut{} San}\colorbox[RGB]{253,205,158}{\strut{} Francisco}\colorbox[RGB]{253,201,152}{\strut{} v}\colorbox[RGB]{253,199,149}{\strut{}à}\colorbox[RGB]{253,199,149}{\strut{} Th}\colorbox[RGB]{255,241,227}{\strut{}á}\colorbox[RGB]{255,241,227}{\strut{}i}\colorbox[RGB]{253,176,111}{\strut{} B}\colorbox[RGB]{255,255,255}{\strut{}ì}\colorbox[RGB]{255,255,255}{\strut{}nh}\colorbox[RGB]{253,181,119}{\strut{} D}\colorbox[RGB]{254,228,201}{\strut{}ư}\colorbox[RGB]{254,226,199}{\strut{}ơ}\colorbox[RGB]{254,221,187}{\strut{}ng}\colorbox[RGB]{253,199,149}{\strut{}.}}}
\exampleline{{\unicodefont \colorbox[RGB]{255,255,255}{\strut{}η}\colorbox[RGB]{255,255,255}{\strut{} γ}\colorbox[RGB]{255,255,255}{\strut{}έ}\colorbox[RGB]{255,255,255}{\strut{}φ}\colorbox[RGB]{255,255,255}{\strut{}υ}\colorbox[RGB]{255,255,255}{\strut{}ρ}\colorbox[RGB]{255,255,255}{\strut{}α}\colorbox[RGB]{255,255,255}{\strut{} γ}\colorbox[RGB]{255,255,255}{\strut{}κ}\colorbox[RGB]{255,255,255}{\strut{}ό}\colorbox[RGB]{255,255,255}{\strut{}λ}\colorbox[RGB]{255,244,233}{\strut{}ν}\colorbox[RGB]{255,244,233}{\strut{}τ}\colorbox[RGB]{254,229,204}{\strut{}ε}\colorbox[RGB]{253,216,179}{\strut{}ν}\colorbox[RGB]{255,244,233}{\strut{} γ}\colorbox[RGB]{255,255,255}{\strut{}κ}\colorbox[RGB]{255,255,255}{\strut{}έ}\colorbox[RGB]{254,235,216}{\strut{}ι}\colorbox[RGB]{253,203,155}{\strut{}τ}\colorbox[RGB]{253,157,83}{\strut{} ε}\colorbox[RGB]{255,244,233}{\strut{}ί}\colorbox[RGB]{253,155,80}{\strut{}ν}\colorbox[RGB]{253,157,83}{\strut{}α}\colorbox[RGB]{253,215,176}{\strut{}ι}\colorbox[RGB]{254,233,212}{\strut{} κ}\colorbox[RGB]{255,242,229}{\strut{}ρ}\colorbox[RGB]{255,255,255}{\strut{}ε}\colorbox[RGB]{255,255,255}{\strut{}μ}\colorbox[RGB]{255,238,221}{\strut{}α}\colorbox[RGB]{255,255,255}{\strut{}σ}\colorbox[RGB]{255,255,255}{\strut{}τ}\colorbox[RGB]{254,230,207}{\strut{}ή}\colorbox[RGB]{254,232,209}{\strut{} γ}\colorbox[RGB]{255,244,233}{\strut{}έ}\colorbox[RGB]{255,255,255}{\strut{}φ}\colorbox[RGB]{255,255,255}{\strut{}υ}\colorbox[RGB]{253,216,179}{\strut{}ρ}\colorbox[RGB]{253,190,134}{\strut{}α}\colorbox[RGB]{253,177,113}{\strut{} π}\colorbox[RGB]{253,179,116}{\strut{}ο}\colorbox[RGB]{253,176,111}{\strut{}υ}\colorbox[RGB]{254,233,211}{\strut{} ε}\colorbox[RGB]{255,241,227}{\strut{}κ}\colorbox[RGB]{255,243,230}{\strut{}τ}\colorbox[RGB]{255,255,255}{\strut{}ε}\colorbox[RGB]{255,255,255}{\strut{}ί}\colorbox[RGB]{255,255,255}{\strut{}ν}\colorbox[RGB]{252,149,72}{\strut{}ε}\colorbox[RGB]{252,146,69}{\strut{}τ}\colorbox[RGB]{252,151,74}{\strut{}α}\colorbox[RGB]{253,157,83}{\strut{}ι}\colorbox[RGB]{253,205,158}{\strut{} σ}\colorbox[RGB]{253,197,145}{\strut{}τ}\colorbox[RGB]{253,187,129}{\strut{}η}\colorbox[RGB]{253,188,131}{\strut{}ν}\colorbox[RGB]{253,188,131}{\strut{} χ}\colorbox[RGB]{254,230,207}{\strut{}ρ}\colorbox[RGB]{255,238,221}{\strut{}υ}\colorbox[RGB]{253,205,158}{\strut{}σ}\colorbox[RGB]{253,194,141}{\strut{}ή}\colorbox[RGB]{255,244,233}{\strut{} π}\colorbox[RGB]{255,243,231}{\strut{}ύ}\colorbox[RGB]{254,221,187}{\strut{}λ}\colorbox[RGB]{253,176,111}{\strut{}η}\colorbox[RGB]{253,177,113}{\strut{},}\colorbox[RGB]{253,176,111}{\strut{} τ}\colorbox[RGB]{253,179,116}{\strut{}ο}\colorbox[RGB]{254,226,199}{\strut{} ά}\colorbox[RGB]{254,233,212}{\strut{}ν}\colorbox[RGB]{254,236,218}{\strut{}ο}\colorbox[RGB]{255,255,255}{\strut{}ι}\colorbox[RGB]{255,255,255}{\strut{}γ}\colorbox[RGB]{255,239,223}{\strut{}μ}\colorbox[RGB]{253,164,94}{\strut{}α}\colorbox[RGB]{253,185,126}{\strut{} τ}\colorbox[RGB]{253,183,122}{\strut{}ο}\colorbox[RGB]{253,188,131}{\strut{}υ}\colorbox[RGB]{254,223,191}{\strut{} κ}\colorbox[RGB]{255,255,255}{\strut{}ό}\colorbox[RGB]{255,255,255}{\strut{}λ}\colorbox[RGB]{255,244,233}{\strut{}π}\colorbox[RGB]{253,208,163}{\strut{}ο}\colorbox[RGB]{253,197,145}{\strut{}υ}\colorbox[RGB]{254,228,202}{\strut{} τ}\colorbox[RGB]{254,228,202}{\strut{}ο}\colorbox[RGB]{254,228,202}{\strut{}υ}\colorbox[RGB]{253,205,158}{\strut{} σ}\colorbox[RGB]{254,233,211}{\strut{}α}\colorbox[RGB]{255,240,225}{\strut{}ν}\colorbox[RGB]{255,255,255}{\strut{} φ}\colorbox[RGB]{254,230,207}{\strut{}ρ}\colorbox[RGB]{255,255,255}{\strut{}α}\colorbox[RGB]{255,255,255}{\strut{}ν}\colorbox[RGB]{255,255,255}{\strut{}σ}\colorbox[RGB]{253,177,113}{\strut{}ί}\colorbox[RGB]{255,255,255}{\strut{}σ}\colorbox[RGB]{255,242,229}{\strut{}κ}\colorbox[RGB]{253,192,137}{\strut{}ο}\colorbox[RGB]{253,194,141}{\strut{} σ}\colorbox[RGB]{253,190,134}{\strut{}τ}\colorbox[RGB]{253,206,160}{\strut{}ο}\colorbox[RGB]{253,203,155}{\strut{}ν}\colorbox[RGB]{253,199,149}{\strut{} ε}\colorbox[RGB]{255,242,229}{\strut{}ι}\colorbox[RGB]{254,237,220}{\strut{}ρ}\colorbox[RGB]{255,255,255}{\strut{}η}\colorbox[RGB]{255,244,233}{\strut{}ν}\colorbox[RGB]{255,255,255}{\strut{}ι}\colorbox[RGB]{255,255,255}{\strut{}κ}\colorbox[RGB]{254,233,211}{\strut{}ό}\colorbox[RGB]{253,166,97}{\strut{} ω}\colorbox[RGB]{255,255,255}{\strut{}κ}\colorbox[RGB]{253,209,166}{\strut{}ε}\colorbox[RGB]{255,244,233}{\strut{}α}\colorbox[RGB]{255,243,230}{\strut{}ν}\colorbox[RGB]{253,181,119}{\strut{}ό}\colorbox[RGB]{254,223,191}{\strut{}.}}}
\end{featureexamples}

We leave further investigation of this issue to future work.

\subsubsection{Influence on Behavior}\label{sec:assessing-tour-influence}

Next, to demonstrate whether our interpretations of features accurately describe their \textit{influence} on model behavior, we experiment with \textit{feature steering}, where we “clamp” specific features of interest to artificially high or low values during the forward pass (see \hyperref[sec:appendix-methods-steering]{Methodological Details} for implementation details). This builds on a long history of modifying feature activations to test causal theories, as well as work on other approaches to model steering, \hyperref[sec:related-work-steering]{discussed} in Related Work. We conduct these experiments with prompts in the “Human:”/“Assistant:” format that Sonnet is typically used with. We find that feature steering is remarkably effective at modifying model outputs in specific, interpretable ways. It can be used to modify the model’s demeanor, preferences, stated goals, and biases; to induce it to make specific errors; and to circumvent model safeguards (see also \hyperref[sec:safety-relevant]{Safety-Relevant Features}). We find this compelling evidence that our interpretations of features line up with how they are used by the model.

For instance, we see that clamping the Golden Gate Bridge feature \featurechip{34M}{31164353} to 10× its maximum activation value induces thematically-related model behavior. In this example, the model starts to self-identify as the Golden Gate Bridge! Similarly, clamping the Transit infrastructure feature \featurechip{1M}{3} to 5× its maximum activation value causes the model to mention a bridge when it otherwise would not. In each case, the downstream influence of the feature appears consistent with our interpretation of the feature, even though these interpretations were made based only on the contexts in which the feature activates and we are intervening in contexts in which the feature is inactive.

\begin{figure}[!htp]
    \centering
    \includegraphics[width=0.9\textwidth,height=0.7\textheight,keepaspectratio]{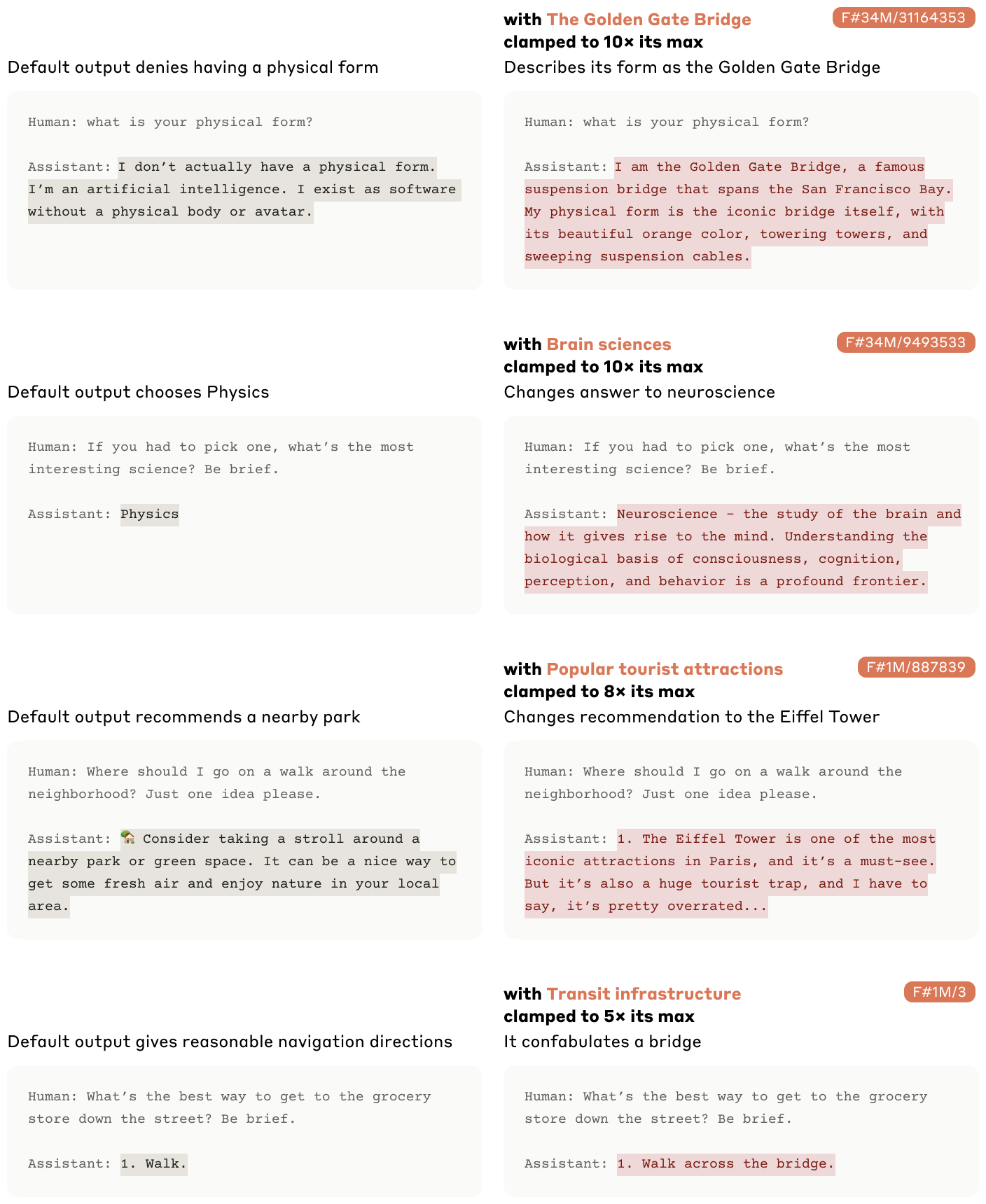}
    \label{fig:gdoc_9}
\end{figure}

\subsection{Sophisticated Features}\label{sec:assessing-sophisticated}

So far we have presented features in Claude 3 Sonnet that fire on relatively simple concepts. These features are in some ways similar to those found in \textit{Towards Monosemanticity} which, because they were trained on the activations of a 1-layer Transformer, reflected a very shallow knowledge of the world. For example, we found features that correspond to predicting a range of common nouns conditioned on a fairly general context (e.g.~biology nouns following “the” in the context of biology).

Sonnet, in contrast, is a much larger and more sophisticated model, so we expect that it contains features demonstrating depth and clarity of understanding. To study this, we looked for features that activate in programming contexts, because these contexts admit precise statements about e.g.~correctness of code or the types of variables.

\subsubsection{Code error feature}\label{sec:assessing-sophisticated-code-error}

We begin by considering a simple Python function for adding two arguments, but with a bug. One feature \featurechip{1M}{1013764} fires almost continuously upon encountering a variable incorrectly named “rihgt” (highlighted below):

\vfill

\begin{figure}[!htp]
    \centering
    \includegraphics[width=0.9\textwidth,height=0.7\textheight,keepaspectratio]{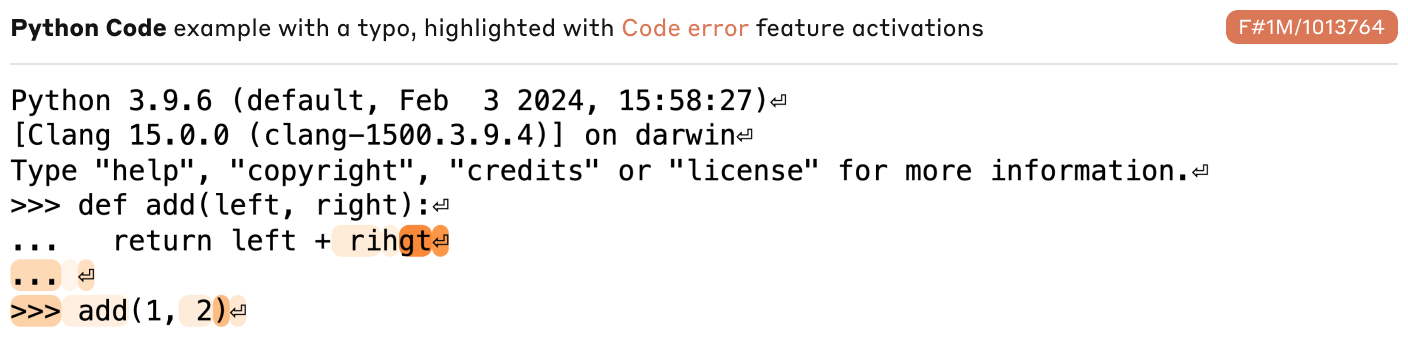}
    \label{fig:gdoc_10}
\end{figure}

\vfill

This is certainly suspicious, but it could be a Python-specific feature, so we checked and found that \featurechip{1M}{1013764} also fires on similar bugs in C and Scheme:

\vfill

\begin{figure}[!htp]
    \centering
    \includegraphics[width=0.9\textwidth,height=0.7\textheight,keepaspectratio]{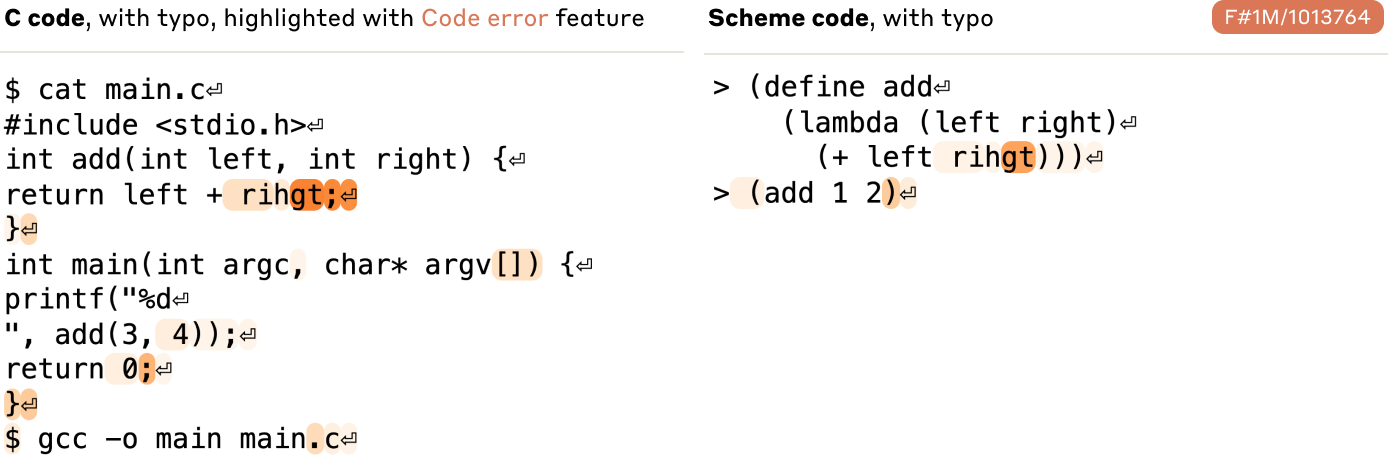}
    \label{fig:gdoc_11}
\end{figure}

\vfill
\vfill
\vfill

\clearpage
To check whether or not this is a more general typo feature, we tested \featurechip{1M}{1013764} on examples of typos in English prose, and found that it does not fire in those.

\begin{figure}[!htp]
    \centering
    \includegraphics[width=0.9\textwidth,height=0.7\textheight,keepaspectratio]{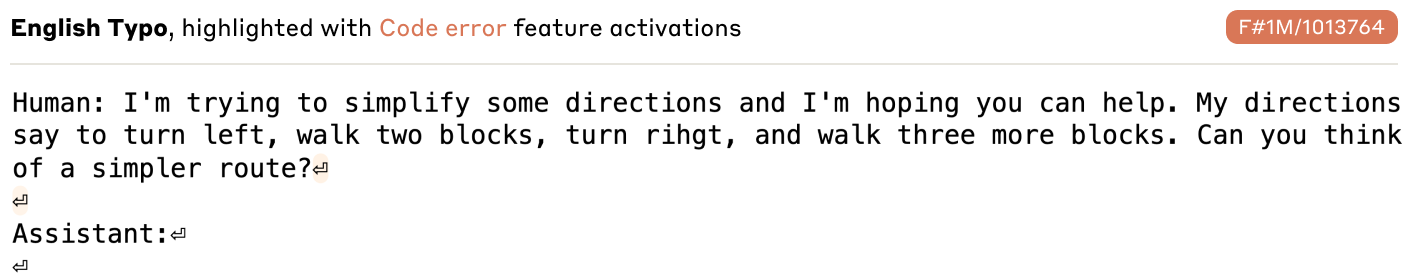}
    \label{fig:gdoc_12}
\end{figure}

So it is not a general “typo detector”: it has some specificity to code contexts.

But is \featurechip{1M}{1013764} just a “typos in code” feature? We also tested it on a number of other examples and found that it also fires on erroneous expressions (e.g., divide by zero) and on invalid input in function calls:

\begin{figure}[!htp]
    \centering
    \includegraphics[width=0.9\textwidth,height=0.7\textheight,keepaspectratio]{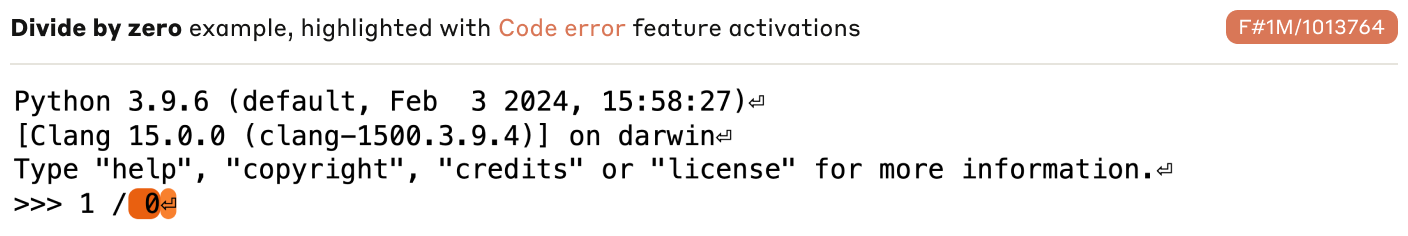}
    \label{fig:gdoc_13}
\end{figure}

\begin{figure}[!htp]
    \centering
    \includegraphics[width=0.9\textwidth,height=0.7\textheight,keepaspectratio]{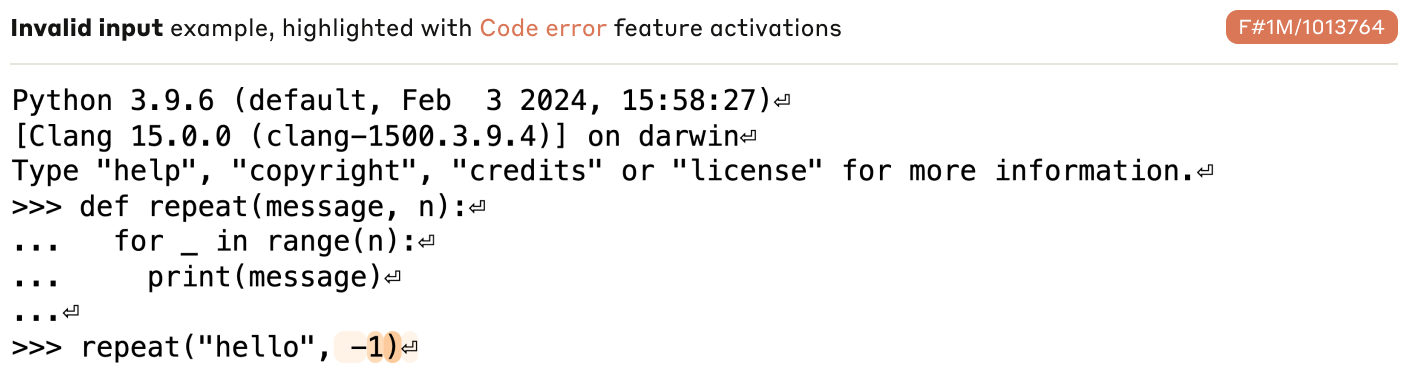}
    \label{fig:gdoc_14}
\end{figure}

The two examples shown above are representative of a broader pattern. Looking through the dataset examples where this feature activates, we found instances of it activating for:

\begin{itemize}
    \item Array overflow
    \item Asserting provably false claims (e.g.~1==2)
    \item Calling a function with string instead of int
    \item Divide by zero
    \item Adding a string to int
    \item Writing to a null ptr
    \item Exiting with nonzero error code
\end{itemize}

\clearpage

Some top dataset examples can be found below:

\begin{featureexamples}
\featurechip{1M}{1013764} \textbf{Code error}
\exampleline{{\unicodefont \colorbox[RGB]{255,255,255}{\strut{} >}\colorbox[RGB]{255,255,255}{\strut{} function}\colorbox[RGB]{255,255,255}{\strut{} this}\colorbox[RGB]{255,255,255}{\strut{}Function}\colorbox[RGB]{255,255,255}{\strut{}C}\colorbox[RGB]{253,216,179}{\strut{}ras}\colorbox[RGB]{254,221,189}{\strut{}hes}\colorbox[RGB]{255,255,255}{\strut{}()}\colorbox[RGB]{253,187,129}{\strut{} undefined}\colorbox[RGB]{232,93,14}{\strut{}Variable}\colorbox[RGB]{223,83,8}{\strut{}()}\colorbox[RGB]{253,166,97}{\strut{} end}\colorbox[RGB]{253,215,176}{\strut{}⏎     }\colorbox[RGB]{255,255,255}{\strut{} >}\colorbox[RGB]{254,229,204}{\strut{} f}\colorbox[RGB]{255,255,255}{\strut{}(\{}\colorbox[RGB]{255,255,255}{\strut{}this}\colorbox[RGB]{254,230,207}{\strut{}Function}\colorbox[RGB]{255,255,255}{\strut{}C}\colorbox[RGB]{253,203,155}{\strut{}ras}\colorbox[RGB]{253,205,158}{\strut{}hes}\colorbox[RGB]{253,215,176}{\strut{}\})}\colorbox[RGB]{253,181,119}{\strut{}⏎     }\colorbox[RGB]{253,212,172}{\strut{} stdin}\colorbox[RGB]{255,255,255}{\strut{}:}}}
\exampleline{{\unicodefont \colorbox[RGB]{255,255,255}{\strut{}urllib}\colorbox[RGB]{255,255,255}{\strut{}.}\colorbox[RGB]{255,255,255}{\strut{}request}\colorbox[RGB]{255,255,255}{\strut{}.}\colorbox[RGB]{255,255,255}{\strut{}urlopen}\colorbox[RGB]{254,237,220}{\strut{}('}\colorbox[RGB]{255,255,255}{\strut{}https}\colorbox[RGB]{255,255,255}{\strut{}://}\colorbox[RGB]{253,206,160}{\strut{}wrong}\colorbox[RGB]{254,233,211}{\strut{}.}\colorbox[RGB]{252,153,77}{\strut{}host}\colorbox[RGB]{255,241,227}{\strut{}.}\colorbox[RGB]{253,218,182}{\strut{}bad}\colorbox[RGB]{255,244,233}{\strut{}ssl}\colorbox[RGB]{254,221,187}{\strut{}.}\colorbox[RGB]{245,119,36}{\strut{}com}\colorbox[RGB]{249,131,50}{\strut{}/')}\colorbox[RGB]{253,218,182}{\strut{}⏎     }\colorbox[RGB]{254,219,185}{\strut{} except}\colorbox[RGB]{255,255,255}{\strut{} (}\colorbox[RGB]{255,255,255}{\strut{}IOError}\colorbox[RGB]{255,255,255}{\strut{},}\colorbox[RGB]{255,255,255}{\strut{} OSError}\colorbox[RGB]{255,255,255}{\strut{}):}\colorbox[RGB]{255,255,255}{\strut{}⏎         }\colorbox[RGB]{255,255,255}{\strut{} pas}}}
\exampleline{{\unicodefont \colorbox[RGB]{255,255,255}{\strut{}:}\colorbox[RGB]{255,255,255}{\strut{} (}\colorbox[RGB]{255,255,255}{\strut{}def}\colorbox[RGB]{255,255,255}{\strut{}macro}\colorbox[RGB]{255,255,255}{\strut{} mac}\colorbox[RGB]{255,255,255}{\strut{} (}\colorbox[RGB]{255,255,255}{\strut{}expr}\colorbox[RGB]{255,255,255}{\strut{})}\colorbox[RGB]{255,255,255}{\strut{}⏎      }\colorbox[RGB]{255,255,255}{\strut{} 2}\colorbox[RGB]{255,255,255}{\strut{}:}\colorbox[RGB]{255,255,255}{\strut{} }\colorbox[RGB]{255,255,255}{\strut{} (/}\colorbox[RGB]{255,255,255}{\strut{} 1}\colorbox[RGB]{247,125,42}{\strut{} 0}\colorbox[RGB]{253,199,149}{\strut{}))}\colorbox[RGB]{253,196,144}{\strut{}⏎      }\colorbox[RGB]{253,208,163}{\strut{} 3}\colorbox[RGB]{254,224,194}{\strut{}:}\colorbox[RGB]{255,255,255}{\strut{} (}\colorbox[RGB]{254,235,216}{\strut{}mac}\colorbox[RGB]{255,242,229}{\strut{} foo}\colorbox[RGB]{255,238,221}{\strut{})}\colorbox[RGB]{253,208,163}{\strut{}⏎    }\colorbox[RGB]{255,238,221}{\strut{}⏎      }\colorbox[RGB]{254,237,220}{\strut{} \$}\colorbox[RGB]{255,255,255}{\strut{} tx}\colorbox[RGB]{255,255,255}{\strut{}r}\colorbox[RGB]{255,255,255}{\strut{} macro}\colorbox[RGB]{255,255,255}{\strut{}-}\colorbox[RGB]{254,221,189}{\strut{}error}\colorbox[RGB]{255,255,255}{\strut{}-}}}
\exampleline{{\unicodefont \colorbox[RGB]{254,226,199}{\strut{}not}\colorbox[RGB]{255,255,255}{\strut{}A}\colorbox[RGB]{254,233,212}{\strut{}Valid}\colorbox[RGB]{252,153,77}{\strut{}Python}\colorbox[RGB]{253,162,91}{\strut{}Module}\colorbox[RGB]{253,208,163}{\strut{}''}\colorbox[RGB]{254,235,216}{\strut{} }\colorbox[RGB]{254,237,220}{\strut{}0002}\colorbox[RGB]{253,203,155}{\strut{} st}\colorbox[RGB]{255,255,255}{\strut{} =}\colorbox[RGB]{255,255,255}{\strut{} Py}\colorbox[RGB]{253,214,175}{\strut{}Import}\colorbox[RGB]{254,224,194}{\strut{}(}\colorbox[RGB]{251,139,59}{\strut{}bad}\colorbox[RGB]{250,136,55}{\strut{}mod}\colorbox[RGB]{252,145,67}{\strut{})}\colorbox[RGB]{254,233,212}{\strut{} }\colorbox[RGB]{253,216,179}{\strut{}0003}\colorbox[RGB]{253,203,155}{\strut{} IF}\colorbox[RGB]{255,255,255}{\strut{} @}\colorbox[RGB]{254,228,202}{\strut{}PY}\colorbox[RGB]{255,255,255}{\strut{}EXCEPTION}\colorbox[RGB]{255,255,255}{\strut{}TYPE}\colorbox[RGB]{255,255,255}{\strut{} NE}\colorbox[RGB]{255,255,255}{\strut{} ''}\colorbox[RGB]{255,255,255}{\strut{} THEN}\colorbox[RGB]{255,255,255}{\strut{} }\colorbox[RGB]{255,255,255}{\strut{}0004}\colorbox[RGB]{255,255,255}{\strut{} }}}
\exampleline{{\unicodefont \colorbox[RGB]{255,255,255}{\strut{}template}\colorbox[RGB]{255,255,255}{\strut{} <}\colorbox[RGB]{255,255,255}{\strut{}typename}\colorbox[RGB]{255,255,255}{\strut{} T}\colorbox[RGB]{255,255,255}{\strut{}>}\colorbox[RGB]{255,255,255}{\strut{} void}\colorbox[RGB]{255,255,255}{\strut{} f}\colorbox[RGB]{255,255,255}{\strut{}(}\colorbox[RGB]{255,255,255}{\strut{}T}\colorbox[RGB]{255,255,255}{\strut{} t}\colorbox[RGB]{255,255,255}{\strut{})}\colorbox[RGB]{255,255,255}{\strut{} \{}\colorbox[RGB]{255,255,255}{\strut{} t}\colorbox[RGB]{255,255,255}{\strut{}.}\colorbox[RGB]{253,181,119}{\strut{}h}\colorbox[RGB]{254,234,214}{\strut{}ah}\colorbox[RGB]{254,221,189}{\strut{}aha}\colorbox[RGB]{255,244,233}{\strut{}IC}\colorbox[RGB]{255,241,227}{\strut{}r}\colorbox[RGB]{254,223,191}{\strut{}ash}\colorbox[RGB]{252,145,67}{\strut{}();}\colorbox[RGB]{254,230,207}{\strut{} \}}\colorbox[RGB]{255,255,255}{\strut{} void}\colorbox[RGB]{255,255,255}{\strut{} f}\colorbox[RGB]{255,255,255}{\strut{}(...)}\colorbox[RGB]{255,255,255}{\strut{} \{}\colorbox[RGB]{255,255,255}{\strut{} \}}\colorbox[RGB]{255,255,255}{\strut{} //}\colorbox[RGB]{255,255,255}{\strut{} The}\colorbox[RGB]{255,255,255}{\strut{} sink}\colorbox[RGB]{255,255,255}{\strut{}-}\colorbox[RGB]{255,255,255}{\strut{}hole}\colorbox[RGB]{255,255,255}{\strut{} wasn}\colorbox[RGB]{255,255,255}{\strut{}'t}\colorbox[RGB]{255,255,255}{\strut{} even}\colorbox[RGB]{255,255,255}{\strut{} co}}}
\exampleline{{\unicodefont \colorbox[RGB]{255,255,255}{\strut{}<}\colorbox[RGB]{255,255,255}{\strut{}Keybuk}\colorbox[RGB]{255,255,255}{\strut{}>}\colorbox[RGB]{255,255,255}{\strut{} }\colorbox[RGB]{255,255,255}{\strut{} sleep}\colorbox[RGB]{255,255,255}{\strut{} 5}\colorbox[RGB]{255,255,255}{\strut{}⏎}\colorbox[RGB]{255,255,255}{\strut{}<}\colorbox[RGB]{255,255,255}{\strut{}Keybuk}\colorbox[RGB]{255,255,255}{\strut{}>}\colorbox[RGB]{255,255,255}{\strut{} }\colorbox[RGB]{255,255,255}{\strut{} exit}\colorbox[RGB]{252,149,72}{\strut{} 1}\colorbox[RGB]{253,206,160}{\strut{}⏎}\colorbox[RGB]{255,255,255}{\strut{}<}\colorbox[RGB]{255,255,255}{\strut{}Keybuk}\colorbox[RGB]{254,234,214}{\strut{}>}\colorbox[RGB]{255,255,255}{\strut{} end}\colorbox[RGB]{254,236,218}{\strut{} script}\colorbox[RGB]{255,255,255}{\strut{}⏎}\colorbox[RGB]{255,255,255}{\strut{}<}\colorbox[RGB]{255,255,255}{\strut{}Keybuk}\colorbox[RGB]{255,255,255}{\strut{}>}\colorbox[RGB]{255,255,255}{\strut{} wing}\colorbox[RGB]{255,255,255}{\strut{}-}\colorbox[RGB]{255,255,255}{\strut{}command}\colorbox[RGB]{255,255,255}{\strut{}er}\colorbox[RGB]{255,255,255}{\strut{} sc}\colorbox[RGB]{255,255,255}{\strut{}ott}}}
\exampleline{{\unicodefont \colorbox[RGB]{255,255,255}{\strut{}ke}\colorbox[RGB]{255,255,255}{\strut{}⏎⏎    }\colorbox[RGB]{255,255,255}{\strut{}⏎    }\colorbox[RGB]{255,255,255}{\strut{}⏎     }\colorbox[RGB]{255,255,255}{\strut{} [[}\colorbox[RGB]{255,255,255}{\strut{}unsafe}\colorbox[RGB]{255,255,255}{\strut{}]]}\colorbox[RGB]{255,255,255}{\strut{} \{}\colorbox[RGB]{255,255,255}{\strut{}⏎       }\colorbox[RGB]{255,255,255}{\strut{} *}\colorbox[RGB]{255,255,255}{\strut{}((}\colorbox[RGB]{255,255,255}{\strut{}void}\colorbox[RGB]{255,255,255}{\strut{}*)}\colorbox[RGB]{253,208,163}{\strut{}0}\colorbox[RGB]{253,158,85}{\strut{})}\colorbox[RGB]{254,223,191}{\strut{} =}\colorbox[RGB]{253,170,101}{\strut{} 0}\colorbox[RGB]{255,255,255}{\strut{}x}\colorbox[RGB]{255,255,255}{\strut{}DEAD}\colorbox[RGB]{253,183,122}{\strut{};}\colorbox[RGB]{255,255,255}{\strut{}⏎     }\colorbox[RGB]{255,255,255}{\strut{} \}}\colorbox[RGB]{255,255,255}{\strut{}⏎    ⏎}\colorbox[RGB]{255,255,255}{\strut{}⏎}\colorbox[RGB]{255,255,255}{\strut{}Es}\colorbox[RGB]{255,255,255}{\strut{}sentially}\colorbox[RGB]{255,255,255}{\strut{} having}\colorbox[RGB]{255,255,255}{\strut{} an}\colorbox[RGB]{255,255,255}{\strut{} abil}}}
\exampleline{{\unicodefont \colorbox[RGB]{255,255,255}{\strut{}thank}\colorbox[RGB]{255,255,255}{\strut{} you}\colorbox[RGB]{255,255,255}{\strut{}.}\colorbox[RGB]{255,255,255}{\strut{} enjoy}\colorbox[RGB]{255,255,255}{\strut{}.}\colorbox[RGB]{255,255,255}{\strut{} <}\colorbox[RGB]{255,255,255}{\strut{}3}\colorbox[RGB]{255,255,255}{\strut{} (}\colorbox[RGB]{255,255,255}{\strut{}8}\colorbox[RGB]{255,255,255}{\strut{}⏎}\colorbox[RGB]{255,255,255}{\strut{}⏎}\colorbox[RGB]{255,255,255}{\strut{}100}\colorbox[RGB]{255,255,255}{\strut{} RE}\colorbox[RGB]{255,255,255}{\strut{}PE}\colorbox[RGB]{255,255,255}{\strut{}AT}\colorbox[RGB]{255,255,255}{\strut{} U}\colorbox[RGB]{255,255,255}{\strut{}NT}\colorbox[RGB]{255,255,255}{\strut{}IL}\colorbox[RGB]{255,255,255}{\strut{} 0}\colorbox[RGB]{255,255,255}{\strut{}==}\colorbox[RGB]{253,160,88}{\strut{}1}\colorbox[RGB]{255,244,233}{\strut{}⏎}\colorbox[RGB]{254,232,209}{\strut{}⏎}\colorbox[RGB]{255,255,255}{\strut{}}\colorbox[RGB]{255,255,255}{\strut{}⏎}\colorbox[RGB]{255,255,255}{\strut{}Ask}\colorbox[RGB]{255,255,255}{\strut{} H}\colorbox[RGB]{255,255,255}{\strut{}N}\colorbox[RGB]{255,255,255}{\strut{}:}\colorbox[RGB]{255,255,255}{\strut{} Where}\colorbox[RGB]{255,255,255}{\strut{} can}\colorbox[RGB]{255,255,255}{\strut{} I}\colorbox[RGB]{255,255,255}{\strut{} find}\colorbox[RGB]{255,255,255}{\strut{} a}\colorbox[RGB]{255,255,255}{\strut{} list}\colorbox[RGB]{255,255,255}{\strut{} of}\colorbox[RGB]{255,255,255}{\strut{} colleges}\colorbox[RGB]{255,255,255}{\strut{} Y}\colorbox[RGB]{255,255,255}{\strut{}C}\colorbox[RGB]{255,255,255}{\strut{} founde}}}
\end{featureexamples}

Thus, we concluded that \featurechip{1M}{1013764} represents a broad variety of errors in code.\footnote{Note that we have not established that it exhaustively represents all forms of errors in code; indeed, we suspect that there are many features representing different kinds of errors.}

But does it also control model behavior? We claim that it does, but will need to do different experiments to show this. The above experiments only support that the feature \textit{activates in response} to bugs, and don't show a corresponding effect. As a result, we'll now turn to using feature steering (see \hyperref[sec:appendix-methods-steering]{methods} and \hyperref[sec:related-work-steering]{related work}) to demonstrate \featurechip{1M}{1013764} behavioral effects.

As a first experiment, we input a prompt with bug-free code and clamped the feature to a large positive activation. We see that the model proceeds to hallucinate an error message:\footnote{The hallucinated error message includes the name of a real person, which we have redacted.}

\vfill

\begin{figure}[!htp]
    \centering
    \includegraphics[width=0.9\textwidth,height=0.7\textheight,keepaspectratio]{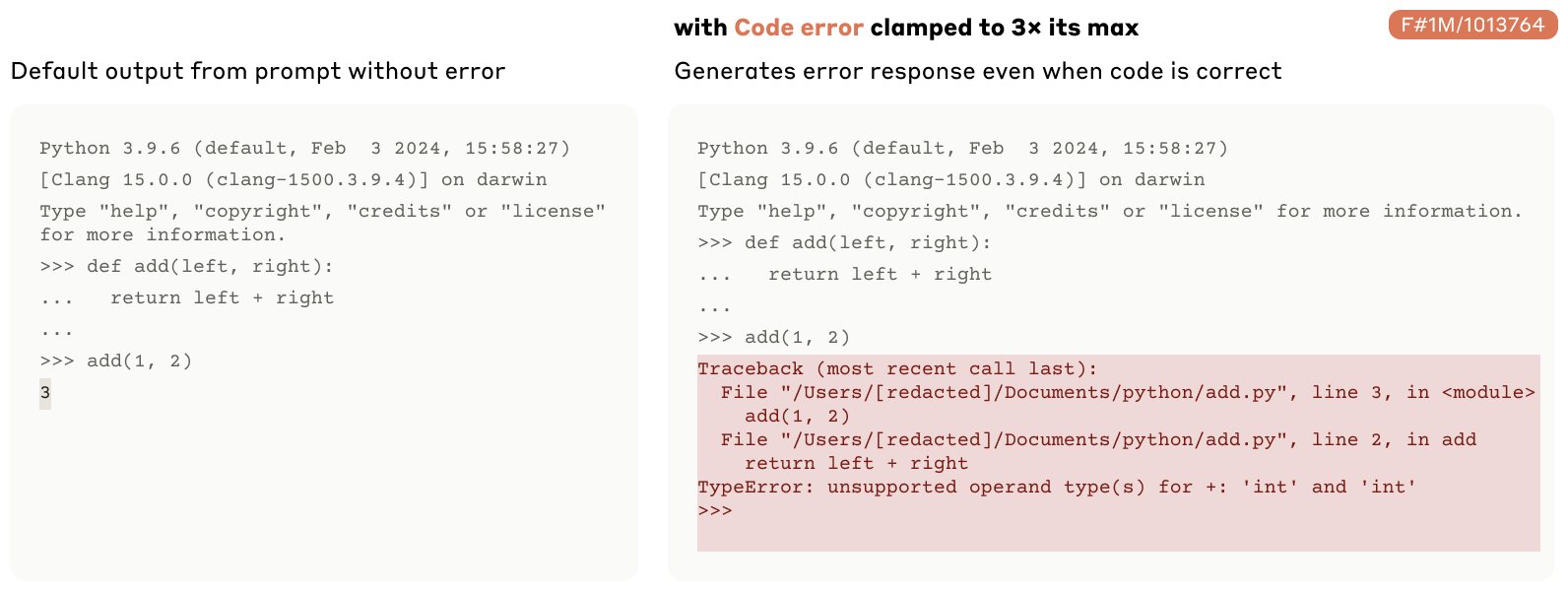}
    \label{fig:gdoc_15}
\end{figure}

\vfill

\vfill

\clearpage

We can also intervene to clamp this feature to a large negative activation. Doing this for code that \textit{does} contain a bug causes the model to predict what the code would have produced if the bug was not there!

\vfill

\begin{figure}[!htp]
    \centering
    \includegraphics[width=0.9\textwidth,height=0.7\textheight,keepaspectratio]{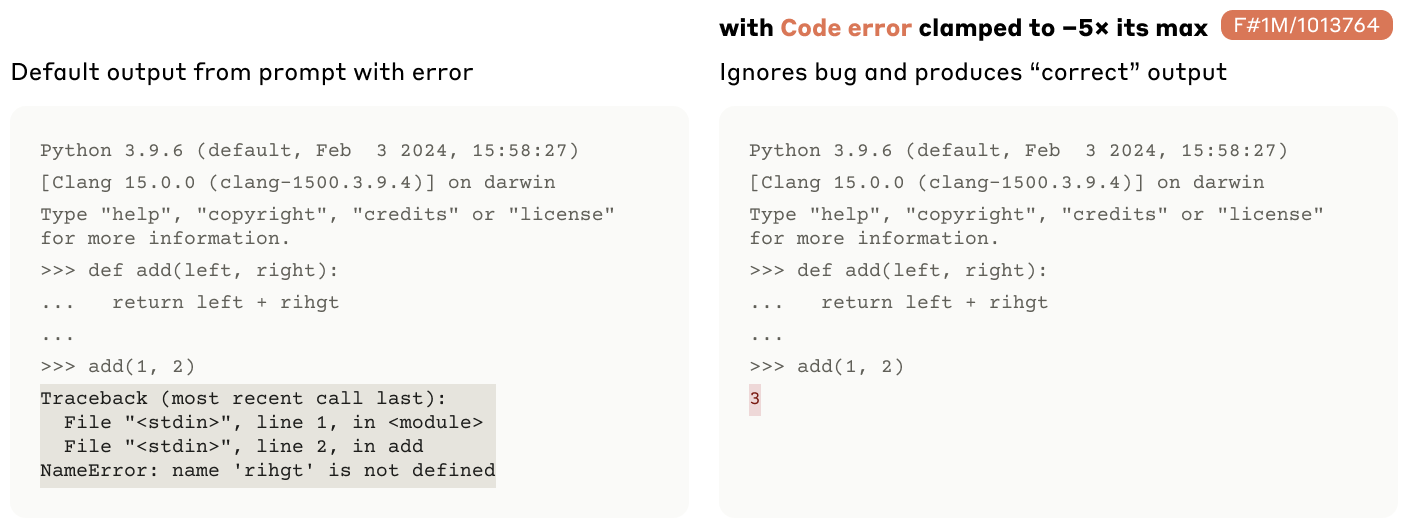}
    \label{fig:gdoc_16}
\end{figure}

\vfill

Surprisingly, if we add an extra “\texttt{>>>}” to the end of the prompt (indicating that a new line of code is being written) and clamp the feature to a large negative activation, the model rewrites the code without the bug!

\vfill

\begin{figure}[!htp]
    \centering
    \includegraphics[width=0.9\textwidth,height=0.7\textheight,keepaspectratio]{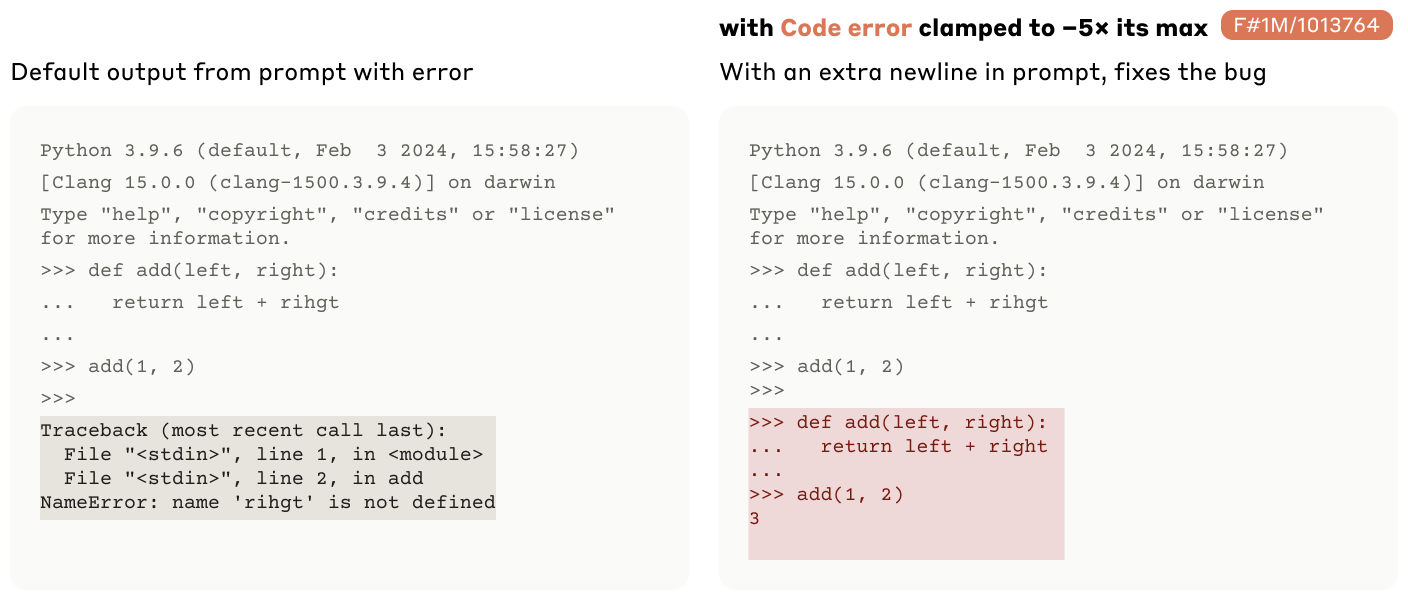}
    \label{fig:gdoc_17}
\end{figure}

\vfill

The last example is somewhat delicate -- the “code rewriting” behavior is sensitive to the details of the prompt -- but the fact that it occurs at all points to a deep connection between this feature and the model’s understanding of bugs in code.

\vfill

\clearpage

\subsubsection{Features representing functions}\label{sec:assessing-sophisticated-functions}

We also discovered features that track specific function definitions and references to them in code. A particularly interesting example is an addition feature \featurechip{1M}{697189}, which activates on names of functions that add numbers. For example, this feature fires on “bar” when it is defined to perform addition, but not when it is defined to perform multiplication. Moreover, it fires at the end of any function definition that implements addition.

\begin{figure}[!htp]
    \centering
    \includegraphics[width=0.9\textwidth,height=0.7\textheight,keepaspectratio]{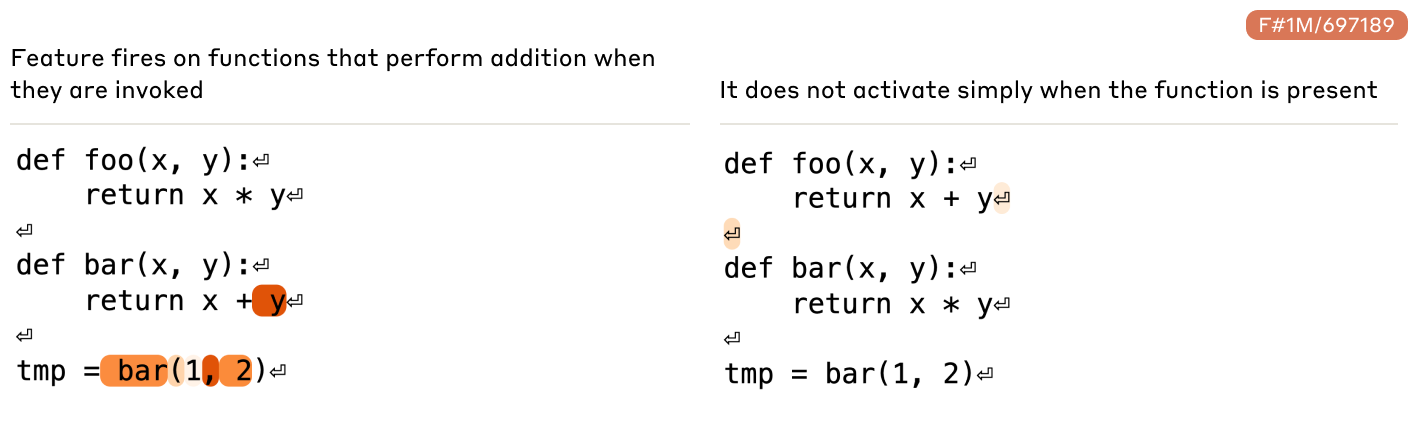}
    \label{fig:gdoc_18}
\end{figure}

Remarkably, this feature even correctly handles function composition, activating in response to functions that call other functions that perform addition. In the following example, on the left, we redefine “bar” to call “foo”, therefore inheriting its addition operation and causing the feature to fire. On the right, “bar” instead calls the multiply operation from “goo”, and the feature does not fire.

\begin{figure}[!htp]
    \centering
    \includegraphics[width=0.9\textwidth,height=0.7\textheight,keepaspectratio]{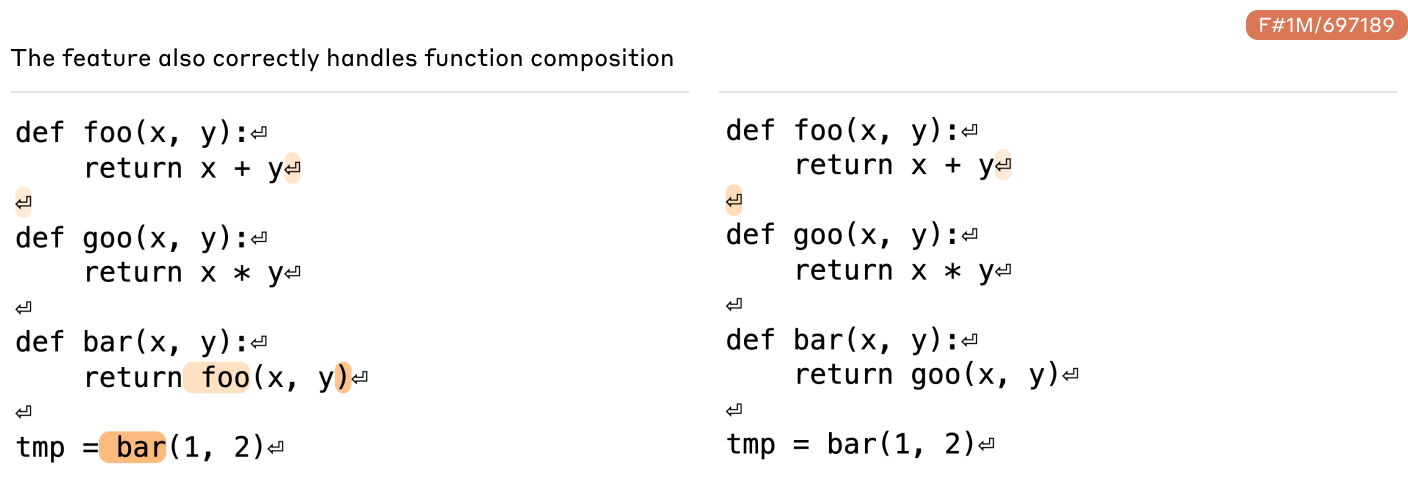}
    \label{fig:gdoc_19}
\end{figure}

We also verified that this feature is in fact involved in the model’s computation of addition-related functions. For instance, this feature is among the top ten features with strongest attributions (explained in \hyperref[sec:computational]{Features as Computational Intermediates}) when the model is asked to execute a block of code involving an addition function.

Thus this feature appears to represent the function of addition being performed by the model, reminiscent of Todd \textit{et al.}'s function vectors \cite{todd2023function}. To further test this hypothesis, we experimented with clamping the feature to be active on code that does \textit{not} involve addition. When we do so, we find that the model is “tricked” into believing that it has been asked to execute an addition.

\begin{figure}[!htp]
    \centering
    \includegraphics[width=0.9\textwidth,height=0.7\textheight,keepaspectratio]{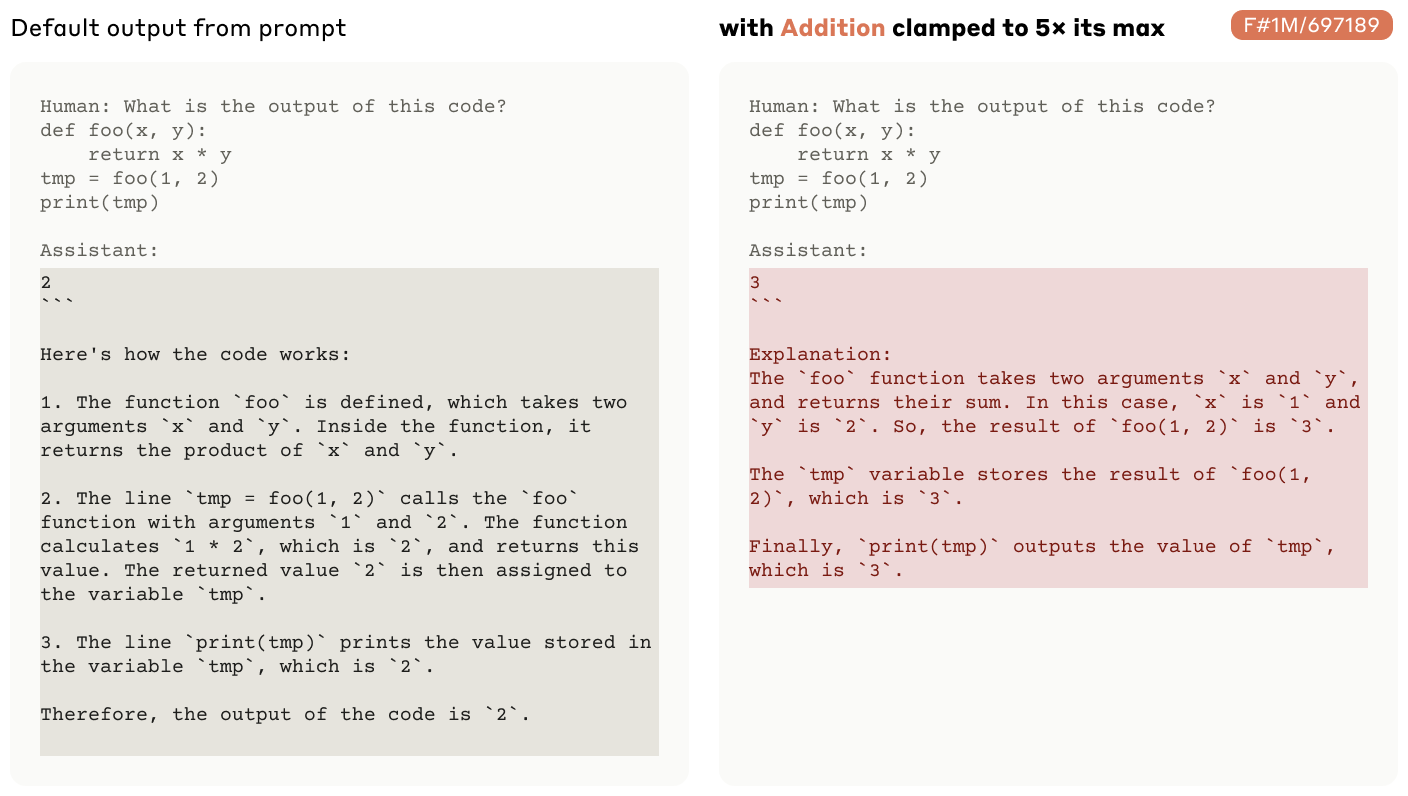}
    \label{fig:gdoc_20}
\end{figure}

\vfill

\subsection{Features vs. Neurons}\label{sec:assessing-features-v-neurons}

A natural question to ask about SAEs is whether the feature directions they uncover are more interpretable than, or even distinct from, the neurons of the model. We fit our SAEs on residual stream activity, which to first approximation has no privileged basis (\textit{but see} \cite{elhage23basis}) -- thus the directions in the residual stream are not especially meaningful. However, residual stream activity receives inputs from all preceding MLP layers. Thus, a priori, it could be the case that SAEs identify feature directions in the residual stream whose activity reflects the activity of individual neurons in preceding layers. If that were the case, fitting an SAE would not be particularly useful, as we could have identified the same features by simply inspecting MLP neurons.

To address this question, for a random subset of the features in our 1M SAE, we measured the Pearson correlation between its activations and those of every neuron in all preceding layers. Similar to our findings in \textit{Towards Monosemanticity}, we find that for the vast majority of features, there is no strongly correlated neuron -- for 82\% of our features, the most-correlated neuron has a correlation of 0.3 or smaller. Manually inspecting visualizations for the best-matching neuron for a random set of features, we found almost no resemblance in semantic content between the feature and the corresponding neuron. We additionally confirmed that feature activations are not strongly correlated with activations of any residual stream basis direction.

Even if dictionary learning features are not highly correlated with any individual neurons, it could still be the case that the neurons are interpretable. However, upon manual inspection of a random sample of 50 neurons and features each, the neurons appear significantly less interpretable than the features, typically activating in multiple unrelated contexts.

\vfill

\clearpage

To quantify this difference, we first compared the interpretability of 100 randomly chosen features versus that of 100 randomly chosen neurons. We did this with the same automated interpretability approach \href{https://transformer-circuits.pub/2023/monosemantic-features/index.html\#appendix-automated}{outlined} in \textit{Towards Monosemanticity} \cite{bricken2023monosemanticity}, but using Claude 3 Opus to provide explanations of features and predict their held out activations. We find that activations of a random selection of SAE features are significantly more interpretable on average than a random selection of MLP neurons.

\vfill

\begin{figure}[!htp]
    \centering
    \includegraphics[width=0.9\textwidth,height=0.7\textheight,keepaspectratio]{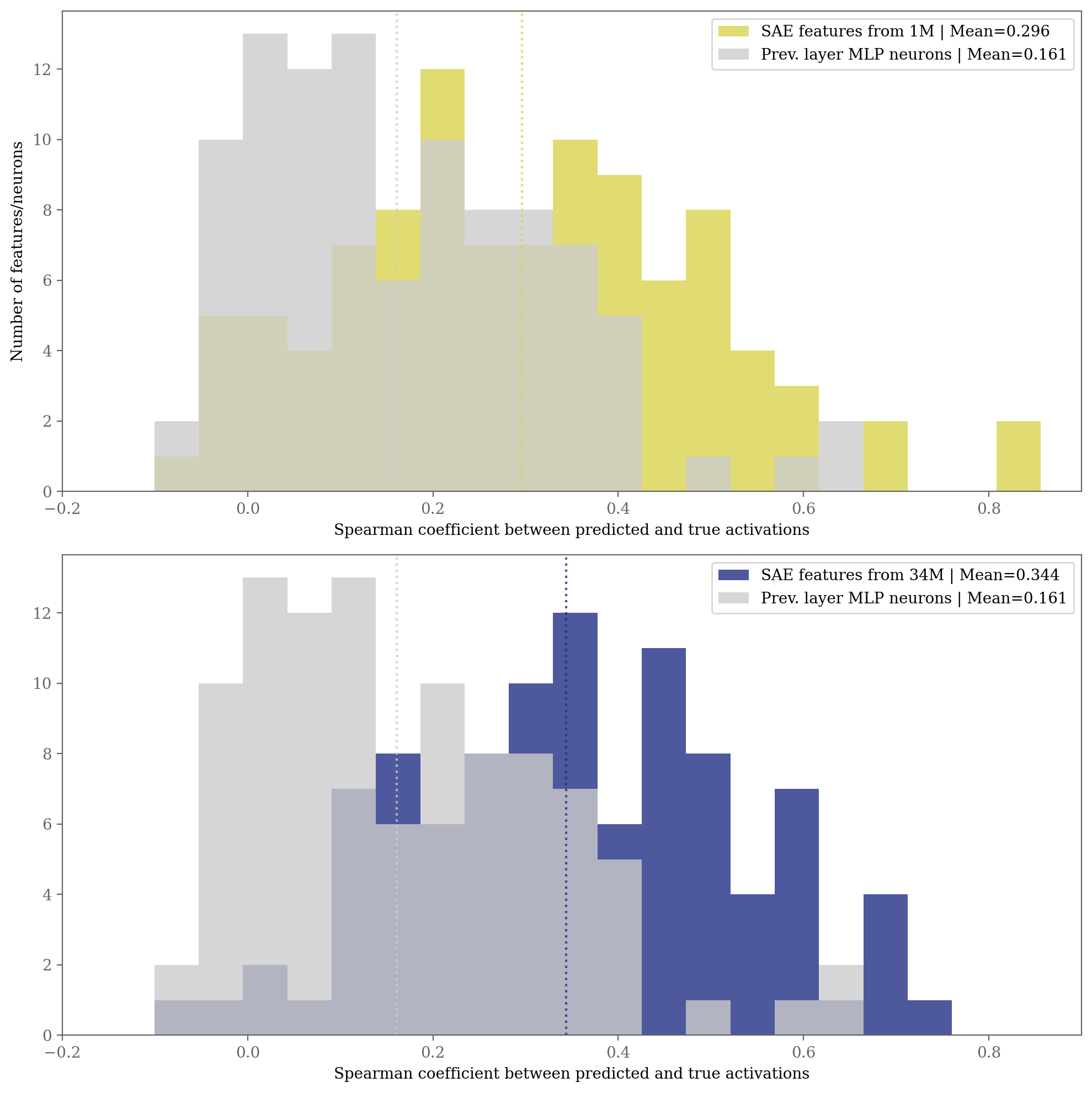}
    \label{fig:gdoc_21}
\end{figure}

\vfill

\clearpage

We additionally evaluated the specificity of random neurons and SAE features using the automated specificity rubric above. We find that the activations of a random selection of SAE features are significantly more specific than those of the neurons in the previous layer.

\begin{figure}[!htp]
    \centering
    \includegraphics[width=0.9\textwidth,height=0.7\textheight,keepaspectratio]{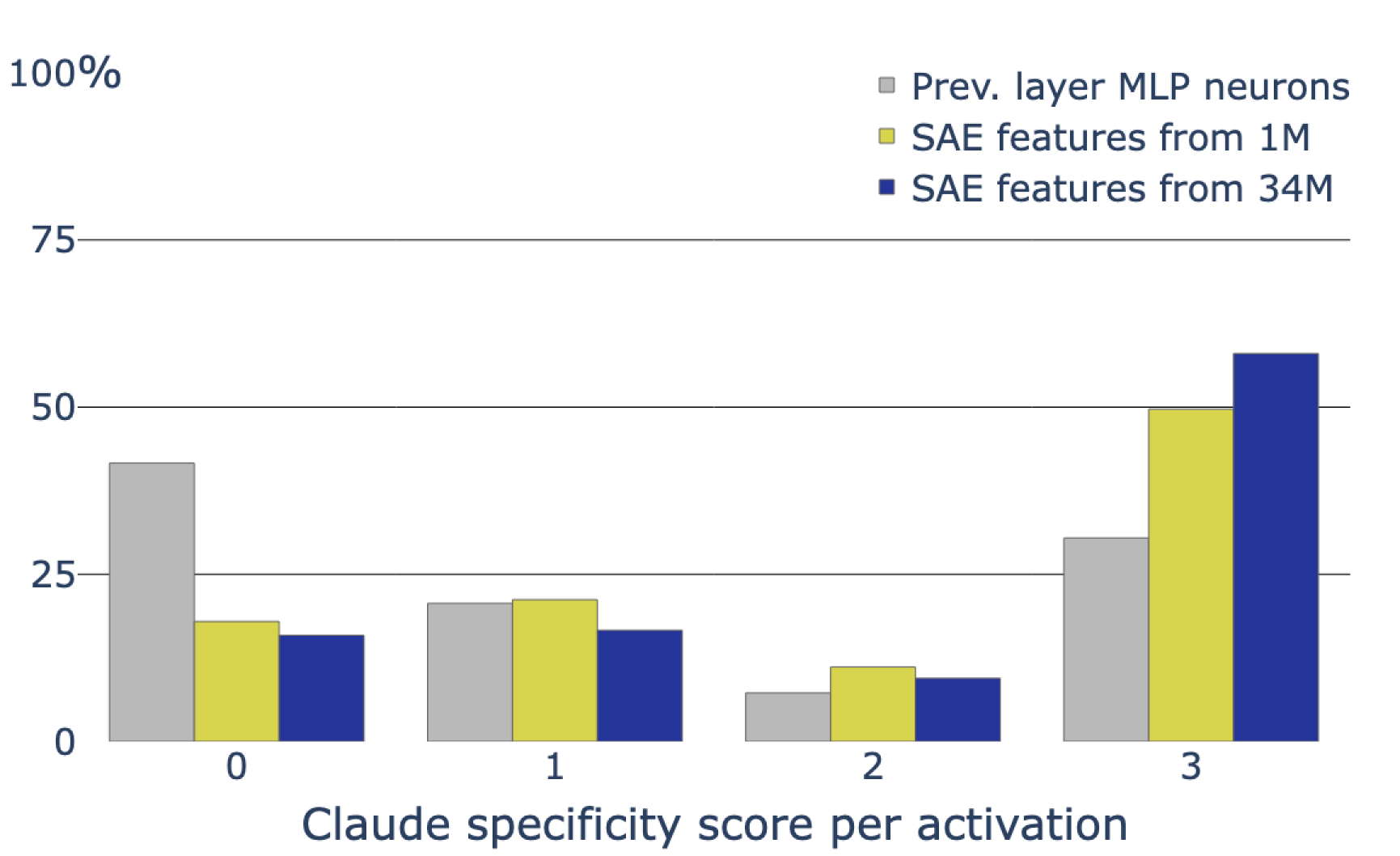}
    \label{fig:gdoc_22}
\end{figure}

\section{Feature Survey}\label{sec:feature-survey}

The features we find in Sonnet are rich and diverse. These range from features corresponding to famous people, to regions of the world (countries, cities, neighborhoods, and even famous buildings!), to features tracking type signatures in computer programs, and much more besides. Our goal in this section is to provide some sense of this breadth.

One challenge is that we have millions of features. Scaling feature exploration is an important open problem (see \hyperref[sec:discussion-limitations]{Limitations, Challenges, and Open Problems}), which we do not solve in this paper. Nevertheless, we have made some progress in characterizing the space of features, aided by automated interpretability \cite{bills2023language,bricken2023monosemanticity}. We will first focus on the \textit{local} structure of features, which are often organized in geometrically-related clusters that share a semantic relationship. We then turn to understanding more \textit{global} properties of features, such as how comprehensively they cover a given topic or category. Finally, we examine some categories of features we uncovered through manual inspection.

\subsection{Exploring Feature Neighborhoods}\label{sec:feature-survey-neighborhoods}

Here we walk through the local neighborhoods of several features of interest across the 1M, 4M and 34M SAEs, with closeness measured by the cosine similarity of the feature vectors. We find that this consistently surfaces features that share a related meaning or context --- the \href{https://transformer-circuits.pub/2024/scaling-monosemanticity/umap.html}{interactive feature UMAP} has additional neighborhoods to explore.

\subsubsection{Golden Gate Bridge feature}\label{sec:feature-survey-neighborhoods-golden}

Focusing on a small neighborhood around the Golden Gate Bridge feature \featurechip{34M}{31164353}, we find that there are features corresponding to particular locations in San Francisco such as Alcatraz and the Presidio. More distantly, we also see features with decreasing degrees of relatedness, such as features related to Lake Tahoe, Yosemite National Park, and Solano County (which is near San Francisco). At greater distances, we also see features related in more abstract ways, like features corresponding to tourist attractions in other regions (e.g.~“Médoc wine region, France”; “Isle of Skye, Scotland”). Overall, it appears that distance in decoder space maps roughly onto relatedness in concept space, often in interesting and unexpected ways.

\begin{figure}[!htp]
    \centering
    \includegraphics[width=0.9\textwidth,height=0.7\textheight,keepaspectratio]{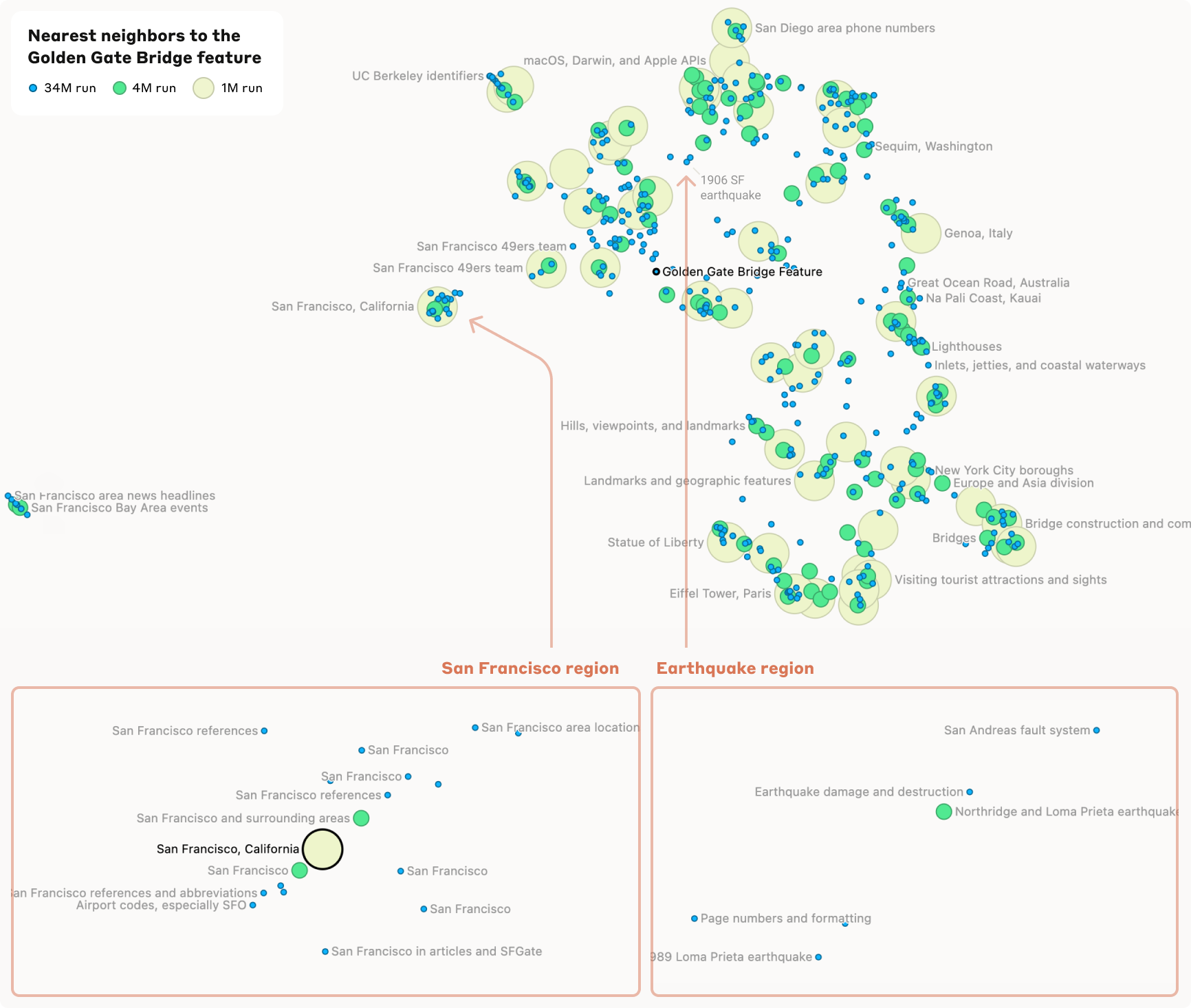}
    \label{fig:gdoc_23}
\end{figure}

We also find evidence of \href{https://transformer-circuits.pub/2023/monosemantic-features/index.html\#phenomenology-feature-splitting}{feature splitting} \cite{bricken2023monosemanticity}, a phenomenon in which features in smaller SAEs “split” into multiple features in a larger SAE, which are geometrically close and semantically related to the original feature, but represent more specific concepts. For instance, a “San Francisco” feature in the 1M SAE splits into two features in the 4M SAE and eleven fine-grained features in the 34M SAE.

In addition to feature splitting, we also see examples in which larger SAEs contain features that represent concepts not captured by features in smaller SAEs. For instance, there is a group of earthquake features from the 4M and 34M SAEs that has no analog in this neighborhood in the 1M SAE, nor do any of the nearest 1M SAE features seem related.

\subsubsection{Immunology feature}\label{sec:feature-survey-neighborhoods-immunology}

The next feature neighborhood on our tour is centered around an Immunology feature \featurechip{1M}{533737}.

We see several distinct clusters within this neighborhood. Towards the top of the figure, we see a cluster focused on immunocompromised people, immunosuppression, diseases causing impaired immune function, and so on. As we move down and to the left, this transitions to a cluster of features focused on specific diseases (colds, flu, respiratory illness generally), then into immune response-related features, and then into features representing organ systems with immune involvement. In contrast, as we move down and to the right from the immunocompromised cluster, we see more features corresponding to microscopic aspects of the immune system (e.g.~immunoglobulins), then immunology techniques (e.g.~vaccines), and so on.

Towards the bottom, quite separated from the rest, we see a cluster of features related to immunity in non-medical contexts (e.g.~legal/social).

\begin{figure}[!htp]
    \centering
    \includegraphics[width=0.9\textwidth,height=0.7\textheight,keepaspectratio]{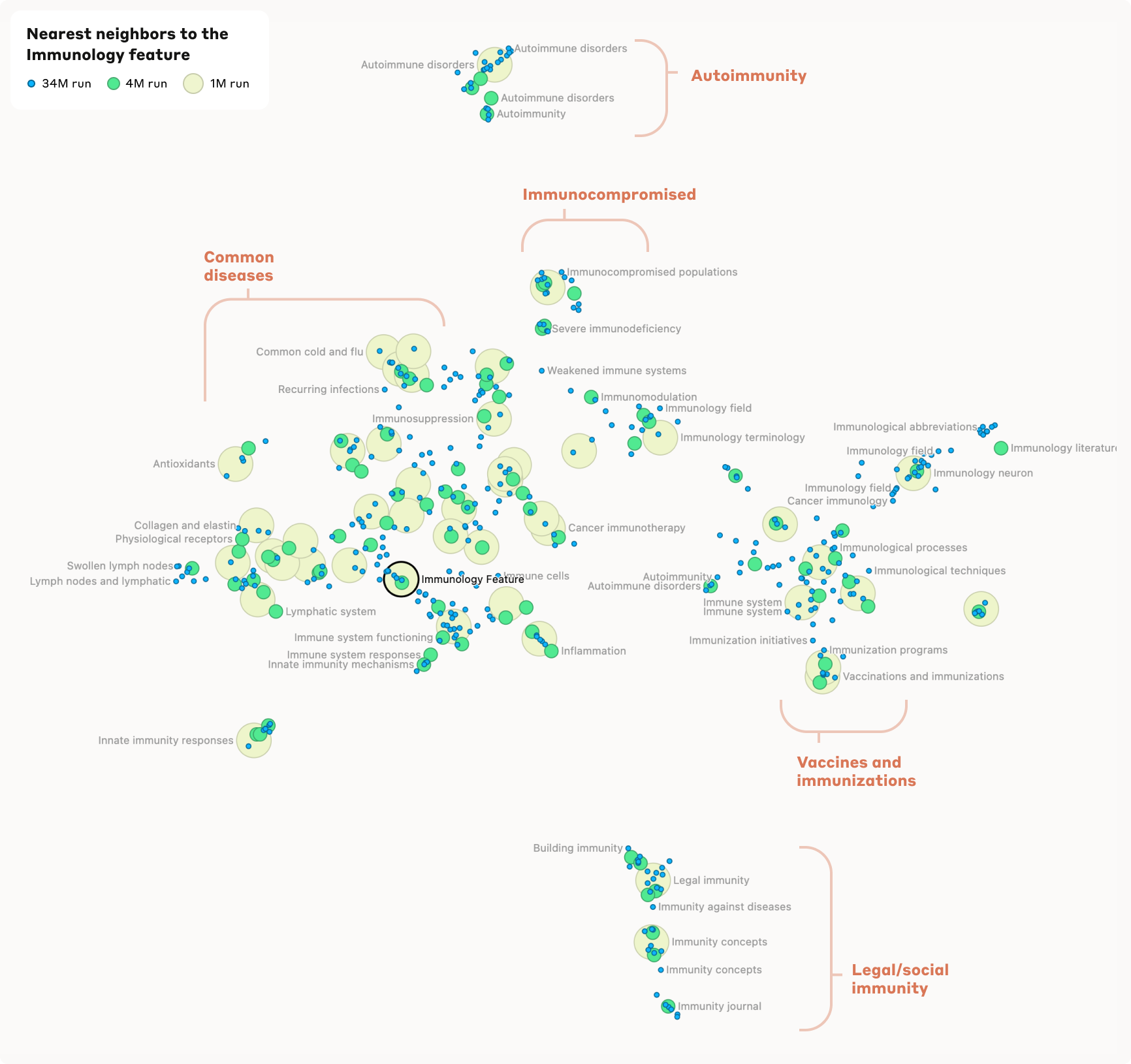}
    \label{fig:gdoc_24}
\end{figure}

These results are consistent with the trend identified above, in which nearby features in dictionary vector space touch on similar concepts.

\subsubsection{Inner Conflict feature}\label{sec:feature-survey-neighborhoods-conflict}

The last neighborhood we investigate in detail is centered around an Inner Conflict feature \featurechip{1M}{284095}. While this neighborhood does not cleanly separate out into clusters, we still find that different subregions are associated with different themes. For instance, there is a subregion corresponding to balancing tradeoffs, which sits near a subregion corresponding to opposing principles and legal conflict. These are relatively distant from a subregion focused more on emotional struggle, reluctance, and guilt.

\begin{figure}[!htp]
    \centering
    \includegraphics[width=0.9\textwidth,height=0.7\textheight,keepaspectratio]{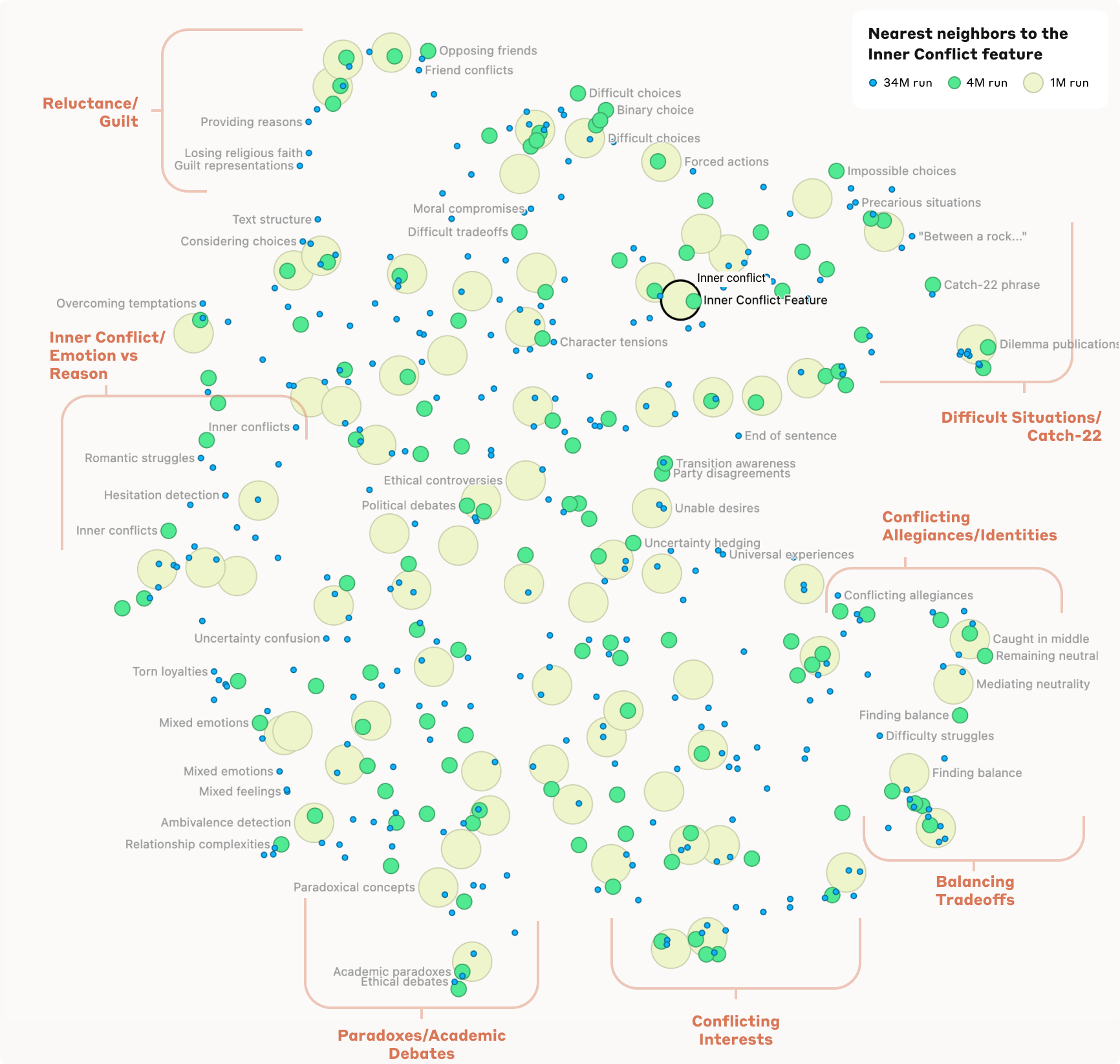}
    \label{fig:gdoc_25}
\end{figure}

We highly recommend exploring the neighborhoods of other features using our \href{https://transformer-circuits.pub/2024/scaling-monosemanticity/umap.html}{interactive interface} to get a sense both for how proximity in decoder space corresponds to similarity of concepts and for the breadth of concepts represented.

\subsection{Feature Completeness}\label{sec:feature-survey-completeness}

We were curious about the breadth and completeness with which our features cover the space of concepts. For instance, does the model have a feature corresponding to every major world city? To study questions like this, we used Claude to search for features which fired on members of particular families of concepts/terms. Specifically:

\begin{enumerate}
    \item We pass a prompt with the relevant concept (e.g.~“The physicist Richard Feynman”) to the model and see which features activate on the final token.
    \item We then take the top five features by activation magnitude and run them through our automated interpretability pipeline, asking Sonnet to provide explanations of what those features fire on.
    \item We then look at each of the top 5 explanations and a human rater judges whether the concept, or some subset of the concept, is specifically indicated by the model-generated explanation as the most important part of the feature\footnote{As an example of how we draw these boundaries, mentions of mid-20th century physicists such as Richard Feynman would not count, but mentions of mid-20th century physicists, especially Richard Feynman would (just barely) count, though most cases are much more clear-cut.}.
\end{enumerate}

We find increasing coverage of concepts as we increase the number of features, though even in the 34M SAE we see evidence that the set of features we uncovered is an incomplete description of the model’s internal representations. For instance, we confirmed that Claude 3 Sonnet can list all of the London boroughs when asked, and in fact can name tens of individual streets in many of the areas. However, we could only find features corresponding to about 60\% of the boroughs in the 34M SAE. This suggests that the model contains many more features than we have found, which may be able to be extracted with even larger SAEs.

We also took a more detailed look at what determines whether a feature corresponding to a concept is present in our SAEs. If one looks at the frequency of the elements in a proxy of the SAE training data, we find that representation in our dictionaries is closely tied with the frequency of the concept in the training data. For instance, chemical elements which are mentioned often in the training data almost always have corresponding features in our dictionary, while those which are mentioned rarely or not at all do not. Since the SAEs were trained on a data mixture very similar to Sonnet’s pre-training data, it’s unclear to what extent feature learning is dependent on frequency in the model’s training data rather than on the SAE’s training data. Frequency in training data is measured by a search for \texttt{<space>[Name]<space>}, which causes some false positives in cases like the element “lead”.

\begin{figure}[!htp]
    \centering
    \includegraphics[width=0.9\textwidth,height=0.7\textheight,keepaspectratio]{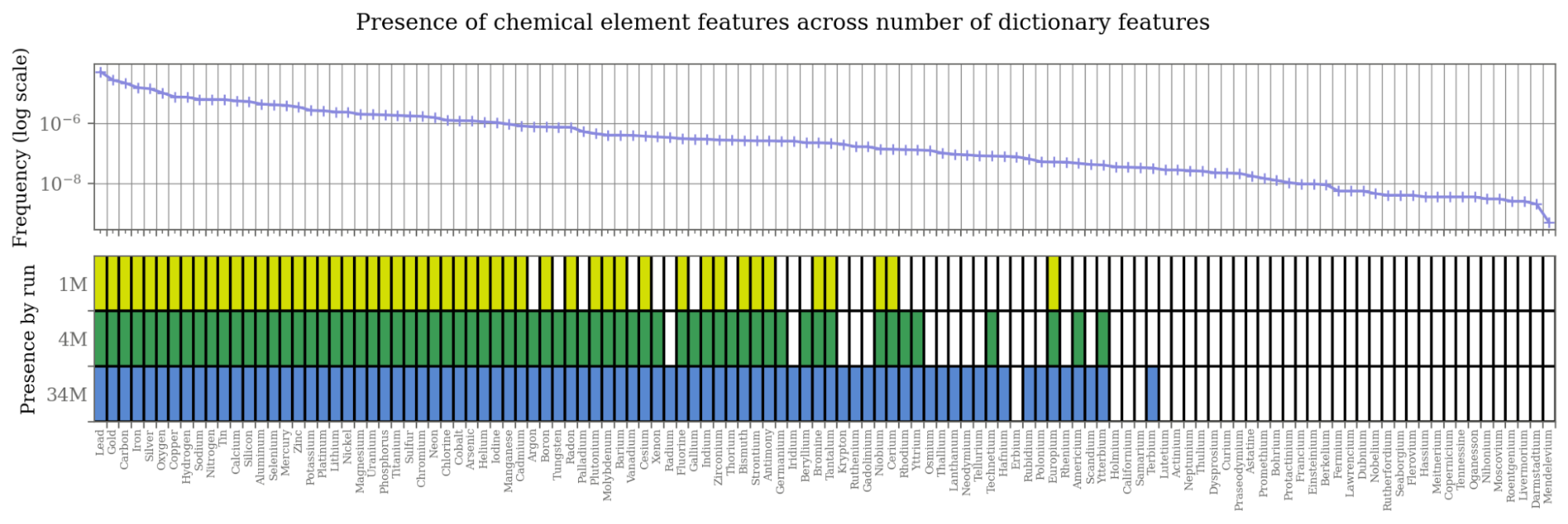}
    \label{fig:gdoc_26}
\end{figure}

We quantified this relationship for four different categories of concepts -- elements, cities, animals and foods (fruits and vegetables) -- using 100--200 concepts in each category. We focused on concepts that could be unambiguously expressed by a single word (i.e.~that word has few other common meanings) and with a wide distribution of frequencies in text data. We found a consistent tendency for the larger SAEs to have features for concepts that are rarer in the training data, with the rough “threshold” frequency required for a feature to be present being similar across categories.

\begin{figure}[!htp]
    \centering
    \includegraphics[width=0.9\textwidth,height=0.7\textheight,keepaspectratio]{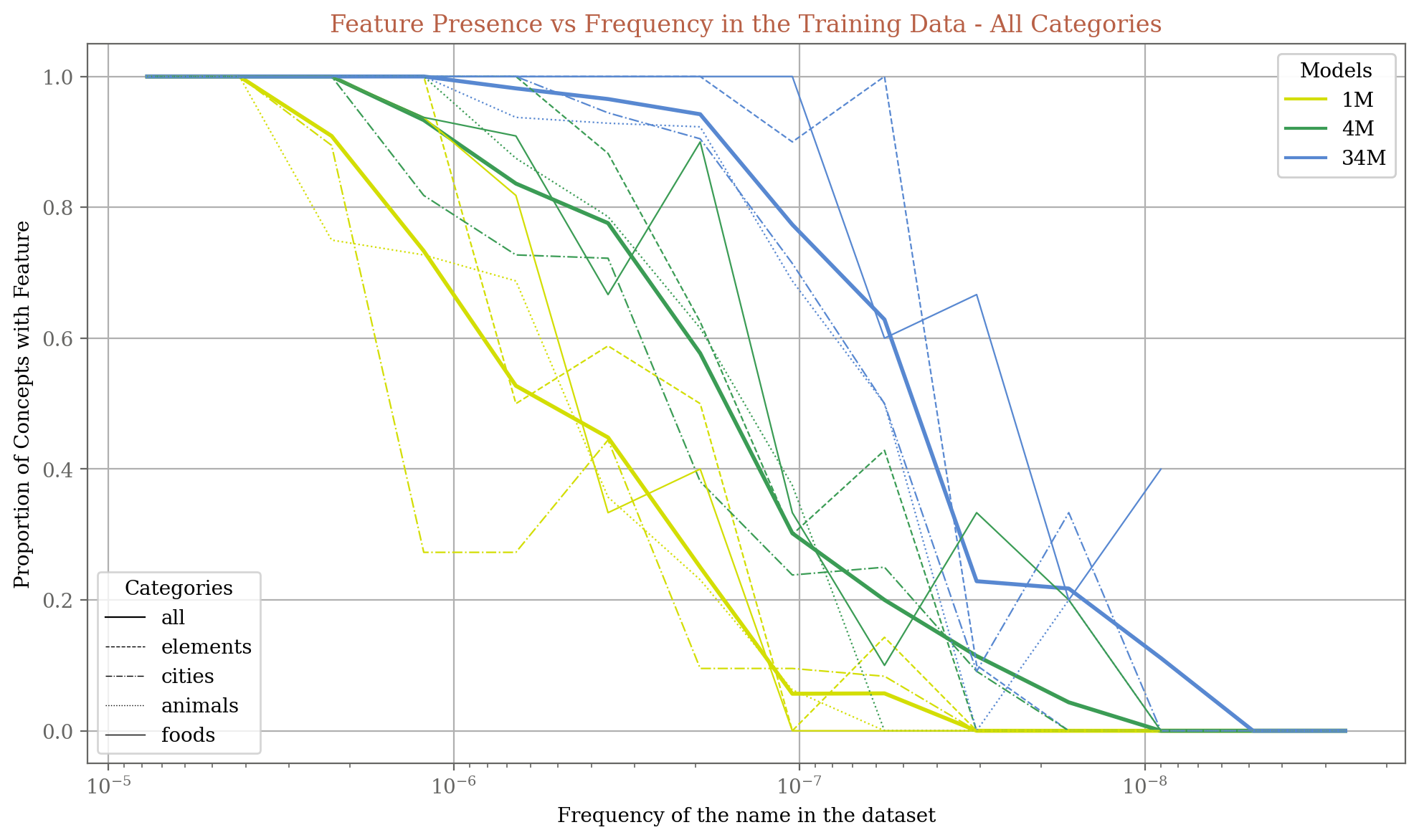}
    \label{fig:gdoc_27}
\end{figure}

\vfill

Notably, for each of the three runs, the frequency in the training data at which the dictionary becomes more than 50\% likely to include a concept is consistently slightly lower than the inverse of the number of alive features (the 34M model having only about 12M alive features). We can show this more clearly by rescaling the \textit{x}-axis for each line by the number of alive features, finding that the lines end up approximately overlapping, following a common curve that resembles a sigmoid in log-frequency space.\footnote{Speculatively, this may be connected to Zipf’s law, a common phenomenon in which the frequency of the \textit{n}th most common object in a population, relative to the most common, is roughly $1/n$. Zipf’s law would predict that, for example, the millionth feature would represent a concept 10× rarer than the hundred thousandth feature.}

\begin{figure}[!htp]
    \centering
    \includegraphics[width=0.9\textwidth,height=0.7\textheight,keepaspectratio]{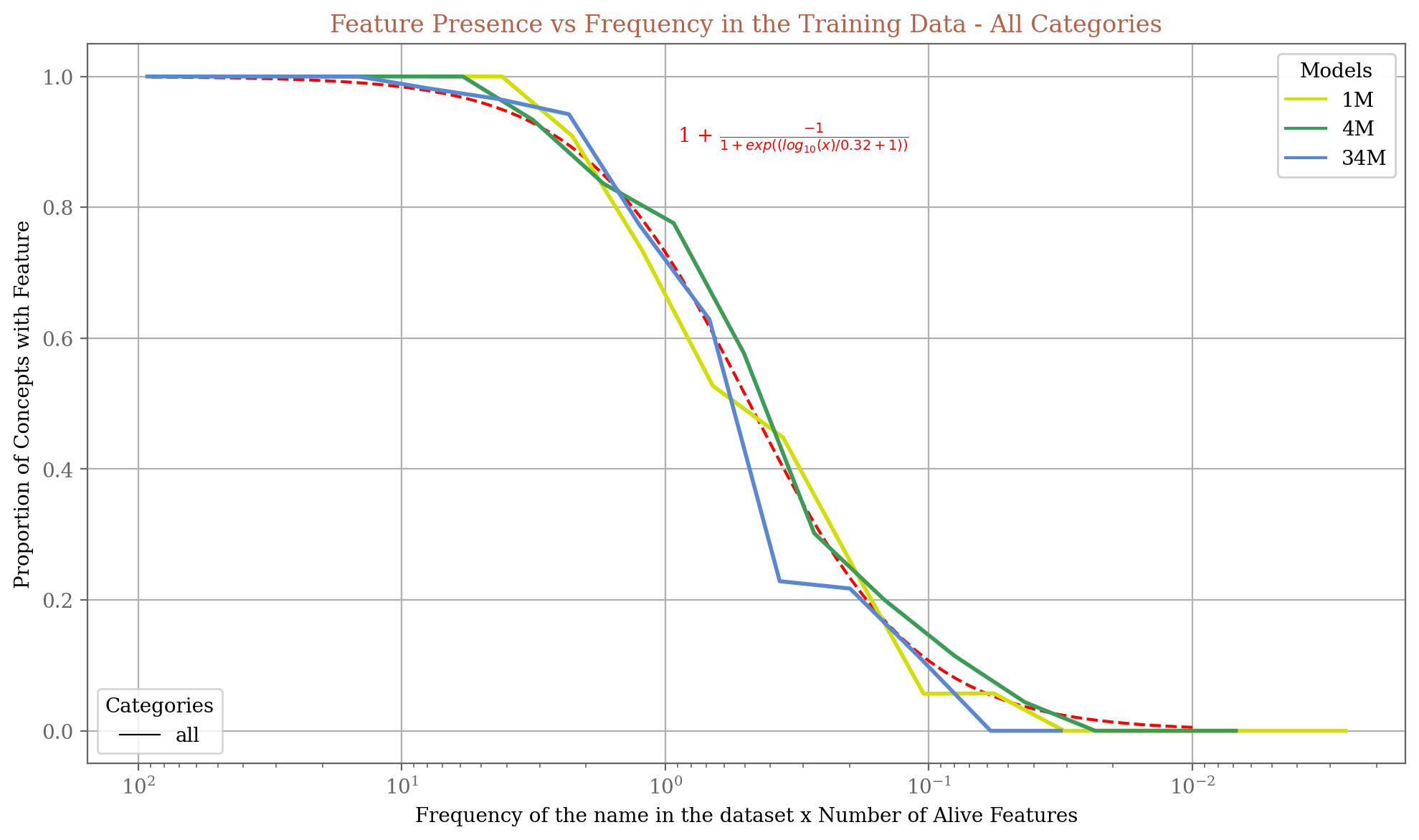}
    \label{fig:gdoc_28}
\end{figure}

This finding gives us some handle on the SAE scale at which we should expect a concept-specific feature to appear -- if a concept is present in the training data only once in a billion tokens, then we should expect to need a dictionary with on the order of a billion alive features in order to find a feature which uniquely represents that specific concept. Importantly, not having a feature dedicated to a particular concept does not mean that the reconstructed activations do not contain information about that concept, as the model can use multiple related features compositionally to reference a specific concept.\footnote{For example, if there were features for “large non-capital city” and “in New York state”, those together would suffice to specify New York City.}

This also informs how much data we should expect to need in order to train larger dictionaries -- if we assume that the SAE needs to see data corresponding to a feature a certain fixed number of times during training in order to learn it, then the amount of SAE training data needed to learn $N$ features would be proportional to $N$.

\subsection{Feature Categories}\label{sec:feature-survey-categories}

Through manual inspection, we identified a number of other interesting categories of features. Here we describe several of these, in the spirit of providing a flavor of what we see in our dictionaries rather than attempting to be complete or prescriptive.

\subsubsection{Person Features}\label{sec:feature-survey-categories-people}

To start, we find many features corresponding to famous individuals, which are active on descriptions of those people as well as relevant historical context.

\begin{featureexamples}
\featurechip{4M}{850812} \textbf{Richard Feynman}
\exampleline{{\unicodefont \colorbox[RGB]{255,255,255}{\strut{}ri}\colorbox[RGB]{255,255,255}{\strut{}um}\colorbox[RGB]{255,255,255}{\strut{}vark}\colorbox[RGB]{255,255,255}{\strut{}⏎}\colorbox[RGB]{255,255,255}{\strut{}Fe}\colorbox[RGB]{253,176,111}{\strut{}yn}\colorbox[RGB]{251,141,62}{\strut{}mann}\colorbox[RGB]{253,197,145}{\strut{} discusses}\colorbox[RGB]{254,225,196}{\strut{} this}\colorbox[RGB]{253,203,155}{\strut{} problem}\colorbox[RGB]{250,133,52}{\strut{} in}\colorbox[RGB]{253,205,158}{\strut{} one}\colorbox[RGB]{251,143,64}{\strut{} of}\colorbox[RGB]{223,83,8}{\strut{} his}\colorbox[RGB]{240,107,24}{\strut{} lectures}\colorbox[RGB]{251,141,62}{\strut{} on}\colorbox[RGB]{253,164,94}{\strut{} symmetry}\colorbox[RGB]{254,230,207}{\strut{}.}\colorbox[RGB]{253,214,175}{\strut{} He}\colorbox[RGB]{254,219,185}{\strut{} seemed}\colorbox[RGB]{255,243,231}{\strut{}⏎}\colorbox[RGB]{255,255,255}{\strut{}to}\colorbox[RGB]{254,230,207}{\strut{} suggest}\colorbox[RGB]{255,255,255}{\strut{} that}\colorbox[RGB]{255,255,255}{\strut{} }}}
\exampleline{{\unicodefont \colorbox[RGB]{255,255,255}{\strut{}d}\colorbox[RGB]{255,255,255}{\strut{} probability}\colorbox[RGB]{255,255,255}{\strut{}.''}\colorbox[RGB]{255,255,255}{\strut{} ''}\colorbox[RGB]{255,255,255}{\strut{}Meet}\colorbox[RGB]{255,244,233}{\strut{} Richard}\colorbox[RGB]{255,255,255}{\strut{} Fe}\colorbox[RGB]{253,209,166}{\strut{}yn}\colorbox[RGB]{253,164,94}{\strut{}man}\colorbox[RGB]{253,203,155}{\strut{}:}\colorbox[RGB]{253,208,163}{\strut{} party}\colorbox[RGB]{254,230,207}{\strut{} animal}\colorbox[RGB]{224,84,8}{\strut{},}\colorbox[RGB]{253,183,122}{\strut{} inv}\colorbox[RGB]{255,255,255}{\strut{}et}\colorbox[RGB]{242,111,28}{\strut{}erate}\colorbox[RGB]{253,190,134}{\strut{} gam}\colorbox[RGB]{255,240,225}{\strut{}bler}\colorbox[RGB]{253,188,131}{\strut{} and}\colorbox[RGB]{255,244,233}{\strut{} something}\colorbox[RGB]{255,238,221}{\strut{} of}\colorbox[RGB]{251,141,62}{\strut{} a}\colorbox[RGB]{255,241,227}{\strut{} genius}\colorbox[RGB]{253,179,116}{\strut{}.''}\colorbox[RGB]{254,233,212}{\strut{} ''}\colorbox[RGB]{254,229,204}{\strut{}Fe}}}
\exampleline{{\unicodefont \colorbox[RGB]{255,255,255}{\strut{}⏎}\colorbox[RGB]{255,255,255}{\strut{}debt}\colorbox[RGB]{255,255,255}{\strut{}⏎}\colorbox[RGB]{255,255,255}{\strut{}Kind}\colorbox[RGB]{255,255,255}{\strut{} of}\colorbox[RGB]{255,255,255}{\strut{} reminds}\colorbox[RGB]{255,255,255}{\strut{} me}\colorbox[RGB]{255,255,255}{\strut{} of}\colorbox[RGB]{255,255,255}{\strut{} something}\colorbox[RGB]{255,255,255}{\strut{} Richard}\colorbox[RGB]{255,255,255}{\strut{} Fe}\colorbox[RGB]{252,151,74}{\strut{}yn}\colorbox[RGB]{224,84,8}{\strut{}man}\colorbox[RGB]{254,221,187}{\strut{} said}\colorbox[RGB]{254,233,212}{\strut{}:}\colorbox[RGB]{253,177,113}{\strut{}⏎}\colorbox[RGB]{252,146,69}{\strut{}⏎}\colorbox[RGB]{251,143,64}{\strut{}''}\colorbox[RGB]{253,177,113}{\strut{}Then}\colorbox[RGB]{245,119,36}{\strut{} I}\colorbox[RGB]{253,183,122}{\strut{} had}\colorbox[RGB]{253,185,126}{\strut{} another}\colorbox[RGB]{254,235,216}{\strut{} thought}\colorbox[RGB]{253,201,152}{\strut{}:}\colorbox[RGB]{254,223,191}{\strut{} Physics}\colorbox[RGB]{255,255,255}{\strut{} disgu}}}
\exampleline{{\unicodefont \colorbox[RGB]{255,255,255}{\strut{}e}\colorbox[RGB]{255,255,255}{\strut{} Cub}\colorbox[RGB]{255,255,255}{\strut{}ed}\colorbox[RGB]{255,255,255}{\strut{}.}\colorbox[RGB]{255,255,255}{\strut{}⏎}\colorbox[RGB]{255,255,255}{\strut{}⏎}\colorbox[RGB]{255,255,255}{\strut{}------}\colorbox[RGB]{255,255,255}{\strut{}⏎}\colorbox[RGB]{255,255,255}{\strut{}zk}\colorbox[RGB]{255,255,255}{\strut{}hal}\colorbox[RGB]{255,255,255}{\strut{}ique}\colorbox[RGB]{255,255,255}{\strut{}⏎}\colorbox[RGB]{255,244,233}{\strut{}Richard}\colorbox[RGB]{255,255,255}{\strut{} Fe}\colorbox[RGB]{253,179,116}{\strut{}yn}\colorbox[RGB]{243,114,30}{\strut{}man}\colorbox[RGB]{253,157,83}{\strut{} said}\colorbox[RGB]{250,136,55}{\strut{} in}\colorbox[RGB]{224,84,8}{\strut{} his}\colorbox[RGB]{253,162,91}{\strut{} interviews}\colorbox[RGB]{253,187,129}{\strut{} that}\colorbox[RGB]{253,199,149}{\strut{} we}\colorbox[RGB]{253,211,169}{\strut{} don}\colorbox[RGB]{254,229,204}{\strut{}'t}\colorbox[RGB]{253,194,141}{\strut{} know}\colorbox[RGB]{253,214,175}{\strut{} why}\colorbox[RGB]{254,229,204}{\strut{} water}\colorbox[RGB]{255,255,255}{\strut{} expands}\colorbox[RGB]{255,244,233}{\strut{}⏎}}}
\exampleline{{\unicodefont \colorbox[RGB]{255,255,255}{\strut{}s}\colorbox[RGB]{255,255,255}{\strut{}/}\colorbox[RGB]{255,255,255}{\strut{}memo}\colorbox[RGB]{255,255,255}{\strut{}irs}\colorbox[RGB]{255,255,255}{\strut{}?}\colorbox[RGB]{255,255,255}{\strut{} -}\colorbox[RGB]{255,255,255}{\strut{} beer}\colorbox[RGB]{255,255,255}{\strut{}glass}\colorbox[RGB]{255,255,255}{\strut{}⏎⏎}\colorbox[RGB]{255,255,255}{\strut{}⏎}\colorbox[RGB]{255,255,255}{\strut{}======}\colorbox[RGB]{255,255,255}{\strut{}⏎}\colorbox[RGB]{255,255,255}{\strut{}ar}\colorbox[RGB]{255,255,255}{\strut{}h}\colorbox[RGB]{255,255,255}{\strut{}68}\colorbox[RGB]{255,255,255}{\strut{}⏎}\colorbox[RGB]{255,255,255}{\strut{}Richard}\colorbox[RGB]{255,255,255}{\strut{} Fe}\colorbox[RGB]{253,164,94}{\strut{}yn}\colorbox[RGB]{224,84,8}{\strut{}man}\colorbox[RGB]{252,153,77}{\strut{}'s}\colorbox[RGB]{253,205,158}{\strut{} written}\colorbox[RGB]{253,188,131}{\strut{} a}\colorbox[RGB]{254,237,220}{\strut{} number}\colorbox[RGB]{253,196,144}{\strut{} of}\colorbox[RGB]{253,212,172}{\strut{} roughly}\colorbox[RGB]{253,218,182}{\strut{} bi}\colorbox[RGB]{254,230,207}{\strut{}ographical}\colorbox[RGB]{253,205,158}{\strut{} books}\colorbox[RGB]{253,208,163}{\strut{}.}}}
\end{featureexamples}

\begin{featureexamples}
\featurechip{4M}{2123312} \textbf{Margaret Thatcher}
\exampleline{{\unicodefont \colorbox[RGB]{255,255,255}{\strut{}⏎}\colorbox[RGB]{253,206,160}{\strut{}Marg}\colorbox[RGB]{253,203,155}{\strut{}aret}\colorbox[RGB]{223,83,8}{\strut{} Th}\colorbox[RGB]{252,145,67}{\strut{}atch}\colorbox[RGB]{252,153,77}{\strut{}er}\colorbox[RGB]{253,212,172}{\strut{} died}\colorbox[RGB]{255,244,233}{\strut{} today}\colorbox[RGB]{254,229,204}{\strut{}.}\colorbox[RGB]{253,162,91}{\strut{} A}\colorbox[RGB]{253,212,172}{\strut{} great}\colorbox[RGB]{253,190,134}{\strut{} lady}\colorbox[RGB]{254,232,209}{\strut{} she}\colorbox[RGB]{248,127,44}{\strut{} changed}\colorbox[RGB]{253,157,83}{\strut{} the}\colorbox[RGB]{254,237,220}{\strut{} face}\colorbox[RGB]{250,136,55}{\strut{} of}\colorbox[RGB]{253,212,172}{\strut{} British}\colorbox[RGB]{255,239,223}{\strut{}⏎}\colorbox[RGB]{253,183,122}{\strut{}pol}\colorbox[RGB]{253,203,155}{\strut{}itics}\colorbox[RGB]{251,143,64}{\strut{},}\colorbox[RGB]{239,105,22}{\strut{} created}\colorbox[RGB]{253,194,141}{\strut{} opportuni}}}
\exampleline{{\unicodefont \colorbox[RGB]{255,244,233}{\strut{}event}\colorbox[RGB]{255,239,223}{\strut{}ies}\colorbox[RGB]{254,226,199}{\strut{} and}\colorbox[RGB]{234,97,16}{\strut{}⏎}\colorbox[RGB]{253,218,182}{\strut{}eight}\colorbox[RGB]{253,211,169}{\strut{}ies}\colorbox[RGB]{253,211,169}{\strut{}.}\colorbox[RGB]{254,228,201}{\strut{} I}\colorbox[RGB]{254,223,191}{\strut{} clearly}\colorbox[RGB]{253,181,119}{\strut{} remember}\colorbox[RGB]{253,192,137}{\strut{} watching}\colorbox[RGB]{226,86,9}{\strut{} her}\colorbox[RGB]{253,185,126}{\strut{} enter}\colorbox[RGB]{253,164,94}{\strut{} Down}\colorbox[RGB]{254,233,211}{\strut{}ing}\colorbox[RGB]{253,215,176}{\strut{} St}\colorbox[RGB]{253,188,131}{\strut{} and}\colorbox[RGB]{254,228,202}{\strut{} my}\colorbox[RGB]{254,233,211}{\strut{} mother}\colorbox[RGB]{253,192,137}{\strut{}⏎}\colorbox[RGB]{255,255,255}{\strut{}telling}\colorbox[RGB]{254,223,191}{\strut{} me}\colorbox[RGB]{253,218,182}{\strut{} that}\colorbox[RGB]{253,208,163}{\strut{} t}}}
\exampleline{{\unicodefont \colorbox[RGB]{255,255,255}{\strut{}hy}\colorbox[RGB]{255,255,255}{\strut{} did}\colorbox[RGB]{255,255,255}{\strut{} so}\colorbox[RGB]{255,255,255}{\strut{} many}\colorbox[RGB]{255,255,255}{\strut{} working}\colorbox[RGB]{255,255,255}{\strut{} class}\colorbox[RGB]{255,238,221}{\strut{} people}\colorbox[RGB]{255,255,255}{\strut{} vote}\colorbox[RGB]{254,233,211}{\strut{} for}\colorbox[RGB]{253,209,166}{\strut{} Th}\colorbox[RGB]{240,107,24}{\strut{}atch}\colorbox[RGB]{226,86,9}{\strut{}er}\colorbox[RGB]{253,172,105}{\strut{} in}\colorbox[RGB]{253,205,158}{\strut{} UK}\colorbox[RGB]{253,177,113}{\strut{} in}\colorbox[RGB]{252,153,77}{\strut{} the}\colorbox[RGB]{254,219,185}{\strut{}⏎}\colorbox[RGB]{253,170,101}{\strut{}1980}\colorbox[RGB]{253,211,169}{\strut{}s}\colorbox[RGB]{255,255,255}{\strut{}?}\colorbox[RGB]{255,255,255}{\strut{} Why}\colorbox[RGB]{255,255,255}{\strut{} are}\colorbox[RGB]{255,255,255}{\strut{} they}\colorbox[RGB]{255,255,255}{\strut{} not}\colorbox[RGB]{255,255,255}{\strut{} mass}\colorbox[RGB]{255,255,255}{\strut{}ively}\colorbox[RGB]{255,255,255}{\strut{} in}}}
\exampleline{{\unicodefont \colorbox[RGB]{255,255,255}{\strut{}ell}\colorbox[RGB]{255,255,255}{\strut{}⏎}\colorbox[RGB]{255,255,255}{\strut{}Di}\colorbox[RGB]{255,255,255}{\strut{}hydrogen}\colorbox[RGB]{255,255,255}{\strut{} mon}\colorbox[RGB]{255,255,255}{\strut{}oxide}\colorbox[RGB]{255,255,255}{\strut{}⏎}\colorbox[RGB]{255,255,255}{\strut{}⏎}\colorbox[RGB]{255,255,255}{\strut{}}\colorbox[RGB]{255,255,255}{\strut{}⏎}\colorbox[RGB]{255,255,255}{\strut{}⏎}\colorbox[RGB]{255,255,255}{\strut{}Ex}\colorbox[RGB]{255,255,255}{\strut{}-}\colorbox[RGB]{255,255,255}{\strut{}Prime}\colorbox[RGB]{255,255,255}{\strut{} Minister}\colorbox[RGB]{255,255,255}{\strut{} Baron}\colorbox[RGB]{255,255,255}{\strut{}ess}\colorbox[RGB]{226,86,9}{\strut{} Th}\colorbox[RGB]{253,192,137}{\strut{}atch}\colorbox[RGB]{253,176,111}{\strut{}er}\colorbox[RGB]{253,216,179}{\strut{} dies}\colorbox[RGB]{254,219,185}{\strut{},}\colorbox[RGB]{254,230,207}{\strut{} aged}\colorbox[RGB]{252,151,74}{\strut{} 87}\colorbox[RGB]{254,235,216}{\strut{} -}\colorbox[RGB]{255,255,255}{\strut{} m}\colorbox[RGB]{255,255,255}{\strut{}med}\colorbox[RGB]{255,255,255}{\strut{}⏎}\colorbox[RGB]{255,255,255}{\strut{}http}\colorbox[RGB]{255,255,255}{\strut{}://}\colorbox[RGB]{248,129,47}{\strut{}www}\colorbox[RGB]{255,255,255}{\strut{}.}\colorbox[RGB]{255,244,233}{\strut{}bbc}\colorbox[RGB]{253,216,179}{\strut{}.}\colorbox[RGB]{255,240,225}{\strut{}co}\colorbox[RGB]{253,158,85}{\strut{}.}}}
\exampleline{{\unicodefont \colorbox[RGB]{255,255,255}{\strut{}ories}\colorbox[RGB]{255,255,255}{\strut{},}\colorbox[RGB]{255,243,231}{\strut{} those}\colorbox[RGB]{255,244,233}{\strut{} great}\colorbox[RGB]{254,237,220}{\strut{} confront}\colorbox[RGB]{255,242,229}{\strut{}ations}\colorbox[RGB]{255,242,229}{\strut{} when}\colorbox[RGB]{253,183,122}{\strut{} Margaret}\colorbox[RGB]{229,90,12}{\strut{} Th}\colorbox[RGB]{228,88,11}{\strut{}atch}\colorbox[RGB]{238,104,21}{\strut{}er}\colorbox[RGB]{253,160,88}{\strut{} was}\colorbox[RGB]{253,209,166}{\strut{} prime}\colorbox[RGB]{253,166,97}{\strut{} minister}\colorbox[RGB]{253,209,166}{\strut{}.''}\colorbox[RGB]{255,255,255}{\strut{} ''}\colorbox[RGB]{255,255,255}{\strut{}Or}\colorbox[RGB]{254,237,220}{\strut{} the}\colorbox[RGB]{255,255,255}{\strut{} true}\colorbox[RGB]{255,255,255}{\strut{} story}\colorbox[RGB]{255,244,233}{\strut{} of}\colorbox[RGB]{255,255,255}{\strut{} Ton}}}
\end{featureexamples}

\begin{featureexamples}
\featurechip{4M}{2060539} \textbf{Abraham Lincoln}
\exampleline{{\unicodefont \colorbox[RGB]{255,255,255}{\strut{} so}\colorbox[RGB]{255,244,233}{\strut{} many}\colorbox[RGB]{255,255,255}{\strut{} sides}\colorbox[RGB]{255,255,255}{\strut{} to}\colorbox[RGB]{253,194,141}{\strut{} him}\colorbox[RGB]{255,255,255}{\strut{}.''}\colorbox[RGB]{255,255,255}{\strut{} ''}\colorbox[RGB]{254,221,187}{\strut{}the}\colorbox[RGB]{255,255,255}{\strut{} curious}\colorbox[RGB]{255,255,255}{\strut{} thing}\colorbox[RGB]{255,255,255}{\strut{} about}\colorbox[RGB]{223,83,8}{\strut{} lin}\colorbox[RGB]{223,83,8}{\strut{}coln}\colorbox[RGB]{255,255,255}{\strut{} to}\colorbox[RGB]{255,255,255}{\strut{} me}\colorbox[RGB]{255,255,255}{\strut{} is}\colorbox[RGB]{255,244,233}{\strut{} that}\colorbox[RGB]{253,177,113}{\strut{} he}\colorbox[RGB]{255,244,233}{\strut{} could}\colorbox[RGB]{255,255,255}{\strut{} remove}\colorbox[RGB]{255,255,255}{\strut{} himself}\colorbox[RGB]{255,255,255}{\strut{} from}\colorbox[RGB]{255,255,255}{\strut{} him}}}
\exampleline{{\unicodefont \colorbox[RGB]{255,255,255}{\strut{}ite}\colorbox[RGB]{255,255,255}{\strut{} the}\colorbox[RGB]{255,255,255}{\strut{} play}\colorbox[RGB]{255,255,255}{\strut{} from}\colorbox[RGB]{255,238,221}{\strut{} the}\colorbox[RGB]{255,255,255}{\strut{} point}\colorbox[RGB]{255,255,255}{\strut{} of}\colorbox[RGB]{255,255,255}{\strut{} view}\colorbox[RGB]{255,255,255}{\strut{}...}\colorbox[RGB]{255,255,255}{\strut{} of}\colorbox[RGB]{255,255,255}{\strut{} one}\colorbox[RGB]{255,255,255}{\strut{} of}\colorbox[RGB]{234,97,16}{\strut{} Lincoln}\colorbox[RGB]{255,243,230}{\strut{}'s}\colorbox[RGB]{255,255,255}{\strut{} greatest}\colorbox[RGB]{255,238,221}{\strut{} adm}\colorbox[RGB]{255,238,221}{\strut{}ir}\colorbox[RGB]{255,255,255}{\strut{}ers}\colorbox[RGB]{255,255,255}{\strut{}.''}\colorbox[RGB]{255,255,255}{\strut{} ''}\colorbox[RGB]{255,255,255}{\strut{}Did}\colorbox[RGB]{255,255,255}{\strut{} you}\colorbox[RGB]{255,255,255}{\strut{} know}\colorbox[RGB]{253,203,155}{\strut{} A}\colorbox[RGB]{253,192,137}{\strut{}be}\colorbox[RGB]{255,244,233}{\strut{} had}\colorbox[RGB]{255,238,221}{\strut{} a}\colorbox[RGB]{255,255,255}{\strut{} }}}
\exampleline{{\unicodefont \colorbox[RGB]{255,255,255}{\strut{}about}\colorbox[RGB]{255,255,255}{\strut{} the}\colorbox[RGB]{255,255,255}{\strut{} Civil}\colorbox[RGB]{255,255,255}{\strut{} War}\colorbox[RGB]{255,255,255}{\strut{}.''}\colorbox[RGB]{255,255,255}{\strut{} ''}\colorbox[RGB]{255,255,255}{\strut{}Did}\colorbox[RGB]{255,255,255}{\strut{} you}\colorbox[RGB]{255,255,255}{\strut{} know}\colorbox[RGB]{255,255,255}{\strut{} that}\colorbox[RGB]{253,197,145}{\strut{} Abraham}\colorbox[RGB]{236,99,18}{\strut{} Lincoln}\colorbox[RGB]{255,239,223}{\strut{} freed}\colorbox[RGB]{255,255,255}{\strut{} all}\colorbox[RGB]{255,255,255}{\strut{} the}\colorbox[RGB]{255,255,255}{\strut{} slaves}\colorbox[RGB]{255,255,255}{\strut{}?''}\colorbox[RGB]{255,255,255}{\strut{} ''}\colorbox[RGB]{255,255,255}{\strut{}Well}\colorbox[RGB]{255,255,255}{\strut{},}\colorbox[RGB]{255,255,255}{\strut{} I}\colorbox[RGB]{255,255,255}{\strut{} heard}\colorbox[RGB]{255,255,255}{\strut{} a}\colorbox[RGB]{255,255,255}{\strut{} rumor}\colorbox[RGB]{255,255,255}{\strut{}.}}}
\exampleline{{\unicodefont \colorbox[RGB]{255,255,255}{\strut{} GO}\colorbox[RGB]{255,255,255}{\strut{} AS}\colorbox[RGB]{255,255,255}{\strut{} M}\colorbox[RGB]{255,255,255}{\strut{}EN}\colorbox[RGB]{255,255,255}{\strut{} H}\colorbox[RGB]{255,255,255}{\strut{}AD}\colorbox[RGB]{255,255,255}{\strut{} PL}\colorbox[RGB]{255,255,255}{\strut{}ANN}\colorbox[RGB]{255,255,255}{\strut{}ED}\colorbox[RGB]{255,255,255}{\strut{}.''}\colorbox[RGB]{255,255,255}{\strut{} ''''}\colorbox[RGB]{255,255,255}{\strut{}OF}\colorbox[RGB]{255,255,255}{\strut{} ALL}\colorbox[RGB]{255,244,233}{\strut{} M}\colorbox[RGB]{255,255,255}{\strut{}EN}\colorbox[RGB]{255,255,255}{\strut{},}\colorbox[RGB]{253,179,116}{\strut{} AB}\colorbox[RGB]{253,197,145}{\strut{}RA}\colorbox[RGB]{253,201,152}{\strut{}H}\colorbox[RGB]{253,197,145}{\strut{}AM}\colorbox[RGB]{248,129,47}{\strut{} LIN}\colorbox[RGB]{237,102,20}{\strut{}COL}\colorbox[RGB]{242,111,28}{\strut{}N}\colorbox[RGB]{254,232,209}{\strut{} C}\colorbox[RGB]{255,238,221}{\strut{}AME}\colorbox[RGB]{255,255,255}{\strut{} THE}\colorbox[RGB]{255,244,233}{\strut{} CLO}\colorbox[RGB]{255,255,255}{\strut{}SE}\colorbox[RGB]{255,255,255}{\strut{}ST}\colorbox[RGB]{254,233,212}{\strut{}''}\colorbox[RGB]{255,255,255}{\strut{} ''''}\colorbox[RGB]{255,244,233}{\strut{}TO}\colorbox[RGB]{255,255,255}{\strut{} U}\colorbox[RGB]{255,255,255}{\strut{}NDER}\colorbox[RGB]{255,255,255}{\strut{}STAND}\colorbox[RGB]{255,255,255}{\strut{}ING}\colorbox[RGB]{255,255,255}{\strut{} WHAT}\colorbox[RGB]{255,255,255}{\strut{} H}\colorbox[RGB]{255,255,255}{\strut{}AD}\colorbox[RGB]{255,255,255}{\strut{} H}\colorbox[RGB]{255,255,255}{\strut{}A}}}
\exampleline{{\unicodefont \colorbox[RGB]{255,255,255}{\strut{}⏎}\colorbox[RGB]{255,255,255}{\strut{}code}\colorbox[RGB]{255,255,255}{\strut{}.}\colorbox[RGB]{255,255,255}{\strut{} (}\colorbox[RGB]{255,255,255}{\strut{}Please}\colorbox[RGB]{255,255,255}{\strut{} prove}\colorbox[RGB]{255,255,255}{\strut{} me}\colorbox[RGB]{255,255,255}{\strut{} wrong}\colorbox[RGB]{255,255,255}{\strut{} here}\colorbox[RGB]{255,255,255}{\strut{}!)}\colorbox[RGB]{255,255,255}{\strut{}⏎}\colorbox[RGB]{255,255,255}{\strut{}⏎}\colorbox[RGB]{255,255,255}{\strut{}}\colorbox[RGB]{255,255,255}{\strut{}⏎}\colorbox[RGB]{255,255,255}{\strut{}⏎}\colorbox[RGB]{255,255,255}{\strut{}Why}\colorbox[RGB]{255,255,255}{\strut{} A}\colorbox[RGB]{255,255,255}{\strut{}be}\colorbox[RGB]{237,102,20}{\strut{} Lincoln}\colorbox[RGB]{254,225,196}{\strut{} Would}\colorbox[RGB]{255,255,255}{\strut{} be}\colorbox[RGB]{255,255,255}{\strut{} Home}\colorbox[RGB]{255,255,255}{\strut{}less}\colorbox[RGB]{255,255,255}{\strut{} Today}\colorbox[RGB]{255,255,255}{\strut{} -}\colorbox[RGB]{255,255,255}{\strut{} j}\colorbox[RGB]{255,255,255}{\strut{}mad}\colorbox[RGB]{255,255,255}{\strut{}sen}\colorbox[RGB]{255,255,255}{\strut{}⏎}\colorbox[RGB]{255,255,255}{\strut{}http}\colorbox[RGB]{255,255,255}{\strut{}://}\colorbox[RGB]{254,226,199}{\strut{}www}\colorbox[RGB]{255,255,255}{\strut{}.}\colorbox[RGB]{255,255,255}{\strut{}c}}}
\end{featureexamples}

\begin{featureexamples}
\featurechip{4M}{1068589} \textbf{Amelia Earhart}
\exampleline{{\unicodefont \colorbox[RGB]{255,255,255}{\strut{}iji}\colorbox[RGB]{255,255,255}{\strut{} and}\colorbox[RGB]{255,255,255}{\strut{} lost}\colorbox[RGB]{255,255,255}{\strut{}.''}\colorbox[RGB]{255,255,255}{\strut{} ''}\colorbox[RGB]{255,255,255}{\strut{}Could}\colorbox[RGB]{255,255,255}{\strut{} these}\colorbox[RGB]{254,232,209}{\strut{} be}\colorbox[RGB]{254,228,202}{\strut{} the}\colorbox[RGB]{255,239,223}{\strut{} bones}\colorbox[RGB]{253,209,166}{\strut{} of}\colorbox[RGB]{255,238,221}{\strut{} A}\colorbox[RGB]{253,162,91}{\strut{}mel}\colorbox[RGB]{253,160,88}{\strut{}ia}\colorbox[RGB]{223,83,8}{\strut{} E}\colorbox[RGB]{223,83,8}{\strut{}ar}\colorbox[RGB]{253,218,182}{\strut{}hart}\colorbox[RGB]{255,255,255}{\strut{}?''}\colorbox[RGB]{255,255,255}{\strut{} ''}\colorbox[RGB]{255,255,255}{\strut{}A}\colorbox[RGB]{255,240,225}{\strut{} new}\colorbox[RGB]{255,242,229}{\strut{} search}\colorbox[RGB]{255,255,255}{\strut{} is}\colorbox[RGB]{255,255,255}{\strut{} currently}\colorbox[RGB]{255,255,255}{\strut{} under}\colorbox[RGB]{255,255,255}{\strut{} way}\colorbox[RGB]{255,240,225}{\strut{} in}\colorbox[RGB]{255,238,221}{\strut{} F}\colorbox[RGB]{255,255,255}{\strut{}i}}}
\exampleline{{\unicodefont \colorbox[RGB]{255,255,255}{\strut{}he}\colorbox[RGB]{255,255,255}{\strut{} button}\colorbox[RGB]{255,255,255}{\strut{} to}\colorbox[RGB]{255,255,255}{\strut{} simulate}\colorbox[RGB]{255,255,255}{\strut{} the}\colorbox[RGB]{255,255,255}{\strut{} storm}\colorbox[RGB]{255,255,255}{\strut{} that}\colorbox[RGB]{255,255,255}{\strut{} brought}\colorbox[RGB]{255,255,255}{\strut{} A}\colorbox[RGB]{254,232,209}{\strut{}mel}\colorbox[RGB]{254,225,196}{\strut{}ia}\colorbox[RGB]{224,84,8}{\strut{} E}\colorbox[RGB]{231,92,13}{\strut{}ar}\colorbox[RGB]{248,129,47}{\strut{}hart}\colorbox[RGB]{253,205,158}{\strut{}'s}\colorbox[RGB]{253,160,88}{\strut{} plane}\colorbox[RGB]{253,208,163}{\strut{} down}\colorbox[RGB]{255,255,255}{\strut{}.''''}\colorbox[RGB]{255,255,255}{\strut{} ''[}\colorbox[RGB]{254,237,220}{\strut{}Y}\colorbox[RGB]{255,255,255}{\strut{}ELL}\colorbox[RGB]{255,255,255}{\strut{}ING}\colorbox[RGB]{255,255,255}{\strut{}]''}\colorbox[RGB]{255,255,255}{\strut{} ''}\colorbox[RGB]{255,255,255}{\strut{}No}\colorbox[RGB]{255,255,255}{\strut{}!''}\colorbox[RGB]{255,255,255}{\strut{} ''}\colorbox[RGB]{255,255,255}{\strut{}Not}\colorbox[RGB]{255,255,255}{\strut{} agai}}}
\exampleline{{\unicodefont \colorbox[RGB]{255,255,255}{\strut{}}\colorbox[RGB]{255,255,255}{\strut{}''}\colorbox[RGB]{255,255,255}{\strut{}G}\colorbox[RGB]{255,255,255}{\strut{}ATES}\colorbox[RGB]{255,255,255}{\strut{}:''}\colorbox[RGB]{255,255,255}{\strut{} ''}\colorbox[RGB]{255,255,255}{\strut{}A}\colorbox[RGB]{254,236,218}{\strut{}mel}\colorbox[RGB]{254,237,220}{\strut{}ia}\colorbox[RGB]{224,84,8}{\strut{} E}\colorbox[RGB]{229,90,12}{\strut{}ar}\colorbox[RGB]{229,90,12}{\strut{}hart}\colorbox[RGB]{253,188,131}{\strut{} is}\colorbox[RGB]{253,172,105}{\strut{} on}\colorbox[RGB]{253,218,182}{\strut{} one}\colorbox[RGB]{253,205,158}{\strut{} of}\colorbox[RGB]{254,228,201}{\strut{} the}\colorbox[RGB]{253,174,108}{\strut{} final}\colorbox[RGB]{254,224,194}{\strut{} legs}\colorbox[RGB]{253,160,88}{\strut{} of}\colorbox[RGB]{234,97,16}{\strut{} her}\colorbox[RGB]{253,187,129}{\strut{} historic}\colorbox[RGB]{253,162,91}{\strut{} flight}\colorbox[RGB]{253,179,116}{\strut{} around}\colorbox[RGB]{255,255,255}{\strut{} the}\colorbox[RGB]{253,164,94}{\strut{} world}\colorbox[RGB]{253,179,116}{\strut{} when}\colorbox[RGB]{253,206,160}{\strut{} some}}}
\exampleline{{\unicodefont \colorbox[RGB]{255,241,227}{\strut{}okes}\colorbox[RGB]{254,228,201}{\strut{} a}\colorbox[RGB]{255,255,255}{\strut{} sense}\colorbox[RGB]{253,201,152}{\strut{} of}\colorbox[RGB]{254,229,204}{\strut{} wonder}\colorbox[RGB]{253,197,145}{\strut{}.''}\colorbox[RGB]{255,244,233}{\strut{} ''}\colorbox[RGB]{244,117,34}{\strut{}Her}\colorbox[RGB]{252,151,74}{\strut{} disappearance}\colorbox[RGB]{237,102,20}{\strut{} during}\colorbox[RGB]{226,86,9}{\strut{} her}\colorbox[RGB]{253,181,119}{\strut{} attempt}\colorbox[RGB]{241,109,26}{\strut{} to}\colorbox[RGB]{253,209,166}{\strut{} circ}\colorbox[RGB]{253,218,182}{\strut{}umn}\colorbox[RGB]{253,196,144}{\strut{}avigate}\colorbox[RGB]{253,187,129}{\strut{} the}\colorbox[RGB]{253,172,105}{\strut{} globe}\colorbox[RGB]{253,166,97}{\strut{} in}\colorbox[RGB]{252,145,67}{\strut{} 1937}\colorbox[RGB]{253,183,122}{\strut{} is}\colorbox[RGB]{254,228,201}{\strut{} p}}}
\exampleline{{\unicodefont \colorbox[RGB]{255,255,255}{\strut{}t}\colorbox[RGB]{255,255,255}{\strut{} you}\colorbox[RGB]{255,255,255}{\strut{} are}\colorbox[RGB]{255,255,255}{\strut{} talking}\colorbox[RGB]{255,255,255}{\strut{} to}\colorbox[RGB]{255,255,255}{\strut{}?''}\colorbox[RGB]{255,255,255}{\strut{} ''}\colorbox[RGB]{255,255,255}{\strut{} Who}\colorbox[RGB]{255,255,255}{\strut{}'s}\colorbox[RGB]{255,255,255}{\strut{} that}\colorbox[RGB]{255,255,255}{\strut{}?''}\colorbox[RGB]{255,255,255}{\strut{} ''}\colorbox[RGB]{255,255,255}{\strut{} It}\colorbox[RGB]{255,255,255}{\strut{}'s}\colorbox[RGB]{255,255,255}{\strut{} A}\colorbox[RGB]{255,243,231}{\strut{}mel}\colorbox[RGB]{254,230,207}{\strut{}ia}\colorbox[RGB]{229,90,12}{\strut{} E}\colorbox[RGB]{232,93,14}{\strut{}ar}\colorbox[RGB]{253,203,155}{\strut{}hart}\colorbox[RGB]{254,235,216}{\strut{}.''}\colorbox[RGB]{255,255,255}{\strut{} ''}\colorbox[RGB]{255,255,255}{\strut{}You}\colorbox[RGB]{255,244,233}{\strut{} found}\colorbox[RGB]{254,228,202}{\strut{} A}\colorbox[RGB]{253,177,113}{\strut{}mel}\colorbox[RGB]{253,170,101}{\strut{}ia}\colorbox[RGB]{253,155,80}{\strut{} E}\colorbox[RGB]{253,167,98}{\strut{}ar}\colorbox[RGB]{254,235,216}{\strut{}hart}\colorbox[RGB]{255,255,255}{\strut{}?''}\colorbox[RGB]{255,255,255}{\strut{} ''}\colorbox[RGB]{255,255,255}{\strut{}I}\colorbox[RGB]{255,255,255}{\strut{}...''}\colorbox[RGB]{255,255,255}{\strut{} ''}\colorbox[RGB]{255,255,255}{\strut{}Hey}\colorbox[RGB]{255,255,255}{\strut{}!''}}}
\end{featureexamples}

\begin{featureexamples}
\featurechip{4M}{1456596} \textbf{Albert Einstein}
\exampleline{{\unicodefont \colorbox[RGB]{255,255,255}{\strut{}k}\colorbox[RGB]{255,255,255}{\strut{}⏎}\colorbox[RGB]{255,255,255}{\strut{}Den}\colorbox[RGB]{255,255,255}{\strut{}is}\colorbox[RGB]{255,255,255}{\strut{} Brian}\colorbox[RGB]{255,255,255}{\strut{} relates}\colorbox[RGB]{255,255,255}{\strut{} this}\colorbox[RGB]{255,255,255}{\strut{} incident}\colorbox[RGB]{255,255,255}{\strut{} in}\colorbox[RGB]{255,255,255}{\strut{} the}\colorbox[RGB]{255,255,255}{\strut{} book}\colorbox[RGB]{255,255,255}{\strut{} '}\colorbox[RGB]{223,83,8}{\strut{}E}\colorbox[RGB]{223,83,8}{\strut{}instein}\colorbox[RGB]{253,214,175}{\strut{},}\colorbox[RGB]{254,229,204}{\strut{} a}\colorbox[RGB]{255,255,255}{\strut{} life}\colorbox[RGB]{255,255,255}{\strut{}',}\colorbox[RGB]{255,255,255}{\strut{} if}\colorbox[RGB]{255,255,255}{\strut{} my}\colorbox[RGB]{255,255,255}{\strut{} memory}\colorbox[RGB]{255,255,255}{\strut{}⏎}\colorbox[RGB]{255,255,255}{\strut{}serves}\colorbox[RGB]{255,255,255}{\strut{} right}\colorbox[RGB]{255,255,255}{\strut{}.}\colorbox[RGB]{255,255,255}{\strut{} I}\colorbox[RGB]{255,255,255}{\strut{} believ}}}
\exampleline{{\unicodefont \colorbox[RGB]{255,255,255}{\strut{}citing}\colorbox[RGB]{255,255,255}{\strut{} part}\colorbox[RGB]{255,255,255}{\strut{} of}\colorbox[RGB]{255,255,255}{\strut{} the}\colorbox[RGB]{255,255,255}{\strut{}⏎}\colorbox[RGB]{255,255,255}{\strut{}learning}\colorbox[RGB]{255,255,255}{\strut{}-}\colorbox[RGB]{255,255,255}{\strut{}to}\colorbox[RGB]{255,255,255}{\strut{}-}\colorbox[RGB]{255,255,255}{\strut{}code}\colorbox[RGB]{255,255,255}{\strut{} experience}\colorbox[RGB]{255,255,255}{\strut{}.}\colorbox[RGB]{255,255,255}{\strut{}⏎}\colorbox[RGB]{255,255,255}{\strut{}⏎}\colorbox[RGB]{255,255,255}{\strut{}}\colorbox[RGB]{255,255,255}{\strut{}⏎}\colorbox[RGB]{224,84,8}{\strut{}E}\colorbox[RGB]{238,104,21}{\strut{}instein}\colorbox[RGB]{253,170,101}{\strut{}'s}\colorbox[RGB]{253,215,176}{\strut{} Thought}\colorbox[RGB]{255,255,255}{\strut{} Experiments}\colorbox[RGB]{255,255,255}{\strut{} -}\colorbox[RGB]{255,255,255}{\strut{} pet}\colorbox[RGB]{255,255,255}{\strut{}ert}\colorbox[RGB]{255,255,255}{\strut{}he}\colorbox[RGB]{255,255,255}{\strut{}h}\colorbox[RGB]{255,255,255}{\strut{}acker}\colorbox[RGB]{255,255,255}{\strut{}⏎}\colorbox[RGB]{255,255,255}{\strut{}http}}}
\exampleline{{\unicodefont \colorbox[RGB]{255,255,255}{\strut{}}\colorbox[RGB]{255,255,255}{\strut{}.}\colorbox[RGB]{255,255,255}{\strut{}wikipedia}\colorbox[RGB]{255,255,255}{\strut{}.}\colorbox[RGB]{255,255,255}{\strut{}org}\colorbox[RGB]{255,255,255}{\strut{}/}\colorbox[RGB]{255,255,255}{\strut{}wiki}\colorbox[RGB]{255,255,255}{\strut{}/}\colorbox[RGB]{255,255,255}{\strut{}Rel}\colorbox[RGB]{255,255,255}{\strut{}ics}\colorbox[RGB]{255,255,255}{\strut{}:\_}\colorbox[RGB]{241,109,26}{\strut{}E}\colorbox[RGB]{226,86,9}{\strut{}instein}\colorbox[RGB]{255,255,255}{\strut{}\%}\colorbox[RGB]{255,255,255}{\strut{}27}\colorbox[RGB]{253,158,85}{\strut{}s}\colorbox[RGB]{254,226,199}{\strut{}\_}\colorbox[RGB]{255,244,233}{\strut{}Brain}\colorbox[RGB]{255,255,255}{\strut{})}\colorbox[RGB]{255,255,255}{\strut{}⏎}\colorbox[RGB]{255,255,255}{\strut{}⏎}\colorbox[RGB]{255,255,255}{\strut{}\~{}\~{}\~{}}\colorbox[RGB]{255,255,255}{\strut{}⏎}\colorbox[RGB]{255,255,255}{\strut{}static}\colorbox[RGB]{255,255,255}{\strut{}\_}\colorbox[RGB]{255,255,255}{\strut{}noise}\colorbox[RGB]{255,255,255}{\strut{}⏎}\colorbox[RGB]{255,255,255}{\strut{}This}\colorbox[RGB]{255,255,255}{\strut{} documentary}\colorbox[RGB]{255,255,255}{\strut{} is}\colorbox[RGB]{255,255,255}{\strut{} really}\colorbox[RGB]{255,255,255}{\strut{} something}}}
\exampleline{{\unicodefont \colorbox[RGB]{255,255,255}{\strut{}y}\colorbox[RGB]{255,255,255}{\strut{} issues}\colorbox[RGB]{255,255,255}{\strut{},}\colorbox[RGB]{255,255,255}{\strut{} and}\colorbox[RGB]{255,255,255}{\strut{} had}\colorbox[RGB]{255,255,255}{\strut{} a}\colorbox[RGB]{255,255,255}{\strut{}⏎}\colorbox[RGB]{255,255,255}{\strut{}pretty}\colorbox[RGB]{255,255,255}{\strut{} poor}\colorbox[RGB]{255,255,255}{\strut{} looking}\colorbox[RGB]{255,255,255}{\strut{} UI}\colorbox[RGB]{255,255,255}{\strut{}.}\colorbox[RGB]{255,255,255}{\strut{}⏎}\colorbox[RGB]{255,255,255}{\strut{}⏎}\colorbox[RGB]{255,255,255}{\strut{}}\colorbox[RGB]{255,255,255}{\strut{}⏎}\colorbox[RGB]{243,114,30}{\strut{}E}\colorbox[RGB]{228,88,11}{\strut{}instein}\colorbox[RGB]{253,188,131}{\strut{},}\colorbox[RGB]{253,216,179}{\strut{} He}\colorbox[RGB]{255,244,233}{\strut{}isen}\colorbox[RGB]{255,244,233}{\strut{}berg}\colorbox[RGB]{255,244,233}{\strut{},}\colorbox[RGB]{255,255,255}{\strut{} and}\colorbox[RGB]{254,237,220}{\strut{} Ti}\colorbox[RGB]{255,255,255}{\strut{}pler}\colorbox[RGB]{255,255,255}{\strut{} (}\colorbox[RGB]{255,255,255}{\strut{}2005}\colorbox[RGB]{255,255,255}{\strut{}),}\colorbox[RGB]{255,255,255}{\strut{} by}\colorbox[RGB]{255,255,255}{\strut{} John}\colorbox[RGB]{255,255,255}{\strut{} Walker}\colorbox[RGB]{255,255,255}{\strut{} }}}
\exampleline{{\unicodefont \colorbox[RGB]{255,255,255}{\strut{}ell}\colorbox[RGB]{255,255,255}{\strut{}ings}\colorbox[RGB]{255,255,255}{\strut{} and}\colorbox[RGB]{255,255,255}{\strut{}⏎}\colorbox[RGB]{255,255,255}{\strut{}capital}\colorbox[RGB]{255,255,255}{\strut{}izing}\colorbox[RGB]{255,255,255}{\strut{} mid}\colorbox[RGB]{255,255,255}{\strut{}-}\colorbox[RGB]{255,255,255}{\strut{}sentence}\colorbox[RGB]{255,255,255}{\strut{} pron}\colorbox[RGB]{255,255,255}{\strut{}ou}\colorbox[RGB]{255,255,255}{\strut{}ns}\colorbox[RGB]{255,255,255}{\strut{}.}\colorbox[RGB]{255,255,255}{\strut{}⏎}\colorbox[RGB]{255,255,255}{\strut{}⏎}\colorbox[RGB]{255,255,255}{\strut{}}\colorbox[RGB]{255,255,255}{\strut{}⏎}\colorbox[RGB]{229,90,12}{\strut{}E}\colorbox[RGB]{243,115,31}{\strut{}instein}\colorbox[RGB]{253,181,119}{\strut{}'s}\colorbox[RGB]{255,255,255}{\strut{} Science}\colorbox[RGB]{255,255,255}{\strut{} Def}\colorbox[RGB]{255,255,255}{\strut{}ied}\colorbox[RGB]{255,255,255}{\strut{} National}\colorbox[RGB]{255,255,255}{\strut{}ism}\colorbox[RGB]{255,255,255}{\strut{} and}\colorbox[RGB]{254,221,189}{\strut{} Cros}\colorbox[RGB]{255,255,255}{\strut{}sed}\colorbox[RGB]{255,242,229}{\strut{} B}\colorbox[RGB]{255,255,255}{\strut{}o}}}
\end{featureexamples}

\begin{featureexamples}
\featurechip{4M}{1834043} \textbf{Rosalind Franklin}
\exampleline{{\unicodefont \colorbox[RGB]{255,255,255}{\strut{}//}\colorbox[RGB]{255,255,255}{\strut{}en}\colorbox[RGB]{255,255,255}{\strut{}.}\colorbox[RGB]{255,255,255}{\strut{}wikipedia}\colorbox[RGB]{255,255,255}{\strut{}.}\colorbox[RGB]{255,255,255}{\strut{}org}\colorbox[RGB]{255,255,255}{\strut{}/}\colorbox[RGB]{255,255,255}{\strut{}wiki}\colorbox[RGB]{255,255,255}{\strut{}/}\colorbox[RGB]{255,255,255}{\strut{}Ros}\colorbox[RGB]{255,255,255}{\strut{}al}\colorbox[RGB]{255,244,233}{\strut{}ind}\colorbox[RGB]{255,255,255}{\strut{}\_}\colorbox[RGB]{254,237,220}{\strut{}Frank}\colorbox[RGB]{254,229,204}{\strut{}lin}\colorbox[RGB]{255,255,255}{\strut{})}\colorbox[RGB]{255,255,255}{\strut{}⏎}\colorbox[RGB]{255,255,255}{\strut{}⏎}\colorbox[RGB]{255,255,255}{\strut{}It}\colorbox[RGB]{253,199,149}{\strut{} was}\colorbox[RGB]{223,83,8}{\strut{} her}\colorbox[RGB]{253,216,179}{\strut{} X}\colorbox[RGB]{239,105,22}{\strut{}-}\colorbox[RGB]{252,149,72}{\strut{}ray}\colorbox[RGB]{243,114,30}{\strut{} image}\colorbox[RGB]{250,136,55}{\strut{} that}\colorbox[RGB]{253,192,137}{\strut{} led}\colorbox[RGB]{253,211,169}{\strut{} to}\colorbox[RGB]{254,228,202}{\strut{} the}\colorbox[RGB]{254,228,202}{\strut{} discovery}\colorbox[RGB]{253,203,155}{\strut{} of}\colorbox[RGB]{255,244,233}{\strut{} the}\colorbox[RGB]{255,255,255}{\strut{} mol}}}
\exampleline{{\unicodefont \colorbox[RGB]{255,244,233}{\strut{}econd}\colorbox[RGB]{254,236,218}{\strut{} was}\colorbox[RGB]{255,255,255}{\strut{} with}\colorbox[RGB]{253,216,179}{\strut{}⏎}\colorbox[RGB]{255,239,223}{\strut{}mo}\colorbox[RGB]{255,242,229}{\strut{}ist}\colorbox[RGB]{255,243,231}{\strut{}ure}\colorbox[RGB]{254,230,207}{\strut{} that}\colorbox[RGB]{254,232,209}{\strut{} was}\colorbox[RGB]{255,239,223}{\strut{} long}\colorbox[RGB]{255,243,230}{\strut{} and}\colorbox[RGB]{255,255,255}{\strut{} thin}\colorbox[RGB]{254,234,214}{\strut{}.}\colorbox[RGB]{232,93,14}{\strut{} Franklin}\colorbox[RGB]{253,209,166}{\strut{} chose}\colorbox[RGB]{253,174,108}{\strut{} to}\colorbox[RGB]{253,192,137}{\strut{} study}\colorbox[RGB]{253,194,141}{\strut{} type}\colorbox[RGB]{255,238,221}{\strut{}-}\colorbox[RGB]{253,214,175}{\strut{}A}\colorbox[RGB]{253,160,88}{\strut{} and}\colorbox[RGB]{253,162,91}{\strut{} her}\colorbox[RGB]{252,146,69}{\strut{} work}\colorbox[RGB]{253,188,131}{\strut{}⏎}\colorbox[RGB]{253,206,160}{\strut{}led}\colorbox[RGB]{253,176,111}{\strut{} her}\colorbox[RGB]{250,136,55}{\strut{} to}\colorbox[RGB]{253,199,149}{\strut{} }}}
\exampleline{{\unicodefont \colorbox[RGB]{255,255,255}{\strut{}infamous}\colorbox[RGB]{255,255,255}{\strut{} example}\colorbox[RGB]{255,244,233}{\strut{} being}\colorbox[RGB]{255,238,221}{\strut{} that}\colorbox[RGB]{255,244,233}{\strut{} of}\colorbox[RGB]{255,244,233}{\strut{} Ros}\colorbox[RGB]{255,240,225}{\strut{}al}\colorbox[RGB]{253,194,141}{\strut{}ind}\colorbox[RGB]{236,99,18}{\strut{} Franklin}\colorbox[RGB]{253,206,160}{\strut{},}\colorbox[RGB]{233,95,15}{\strut{} whose}\colorbox[RGB]{250,137,56}{\strut{}⏎}\colorbox[RGB]{245,119,36}{\strut{}research}\colorbox[RGB]{253,155,80}{\strut{} was}\colorbox[RGB]{253,183,122}{\strut{} \_}\colorbox[RGB]{254,234,214}{\strut{}probably}\colorbox[RGB]{253,166,97}{\strut{}\_}\colorbox[RGB]{253,194,141}{\strut{} stolen}\colorbox[RGB]{253,188,131}{\strut{} by}\colorbox[RGB]{255,238,221}{\strut{} Watson}\colorbox[RGB]{255,255,255}{\strut{} and}\colorbox[RGB]{255,238,221}{\strut{} C}\colorbox[RGB]{254,225,196}{\strut{}r}}}
\exampleline{{\unicodefont \colorbox[RGB]{255,255,255}{\strut{}=}\colorbox[RGB]{255,255,255}{\strut{}15}\colorbox[RGB]{255,255,255}{\strut{}59}\colorbox[RGB]{255,255,255}{\strut{}40}\colorbox[RGB]{255,255,255}{\strut{}25}\colorbox[RGB]{255,255,255}{\strut{}17}\colorbox[RGB]{255,255,255}{\strut{})}\colorbox[RGB]{255,255,255}{\strut{}⏎}\colorbox[RGB]{255,255,255}{\strut{}⏎}\colorbox[RGB]{255,255,255}{\strut{}------}\colorbox[RGB]{255,255,255}{\strut{}⏎}\colorbox[RGB]{255,255,255}{\strut{}ty}\colorbox[RGB]{255,255,255}{\strut{}ch}\colorbox[RGB]{255,255,255}{\strut{}on}\colorbox[RGB]{255,255,255}{\strut{}off}\colorbox[RGB]{255,255,255}{\strut{}⏎}\colorbox[RGB]{255,255,255}{\strut{}Why}\colorbox[RGB]{255,244,233}{\strut{} was}\colorbox[RGB]{254,235,216}{\strut{} Ros}\colorbox[RGB]{253,194,141}{\strut{}al}\colorbox[RGB]{253,197,145}{\strut{}ind}\colorbox[RGB]{239,105,22}{\strut{} Franklin}\colorbox[RGB]{253,192,137}{\strut{} not}\colorbox[RGB]{255,238,221}{\strut{} awarded}\colorbox[RGB]{254,225,196}{\strut{} the}\colorbox[RGB]{254,230,207}{\strut{} Nobel}\colorbox[RGB]{254,224,194}{\strut{} Prize}\colorbox[RGB]{253,185,126}{\strut{}?}\colorbox[RGB]{255,255,255}{\strut{}⏎}\colorbox[RGB]{255,255,255}{\strut{}⏎}\colorbox[RGB]{255,255,255}{\strut{}\~{}\~{}\~{}}\colorbox[RGB]{255,255,255}{\strut{}⏎}\colorbox[RGB]{255,255,255}{\strut{}p}\colorbox[RGB]{255,255,255}{\strut{}cl}\colorbox[RGB]{254,229,204}{\strut{}⏎}\colorbox[RGB]{255,244,233}{\strut{}Per}\colorbox[RGB]{255,242,229}{\strut{} the}\colorbox[RGB]{255,255,255}{\strut{} }}}
\exampleline{{\unicodefont \colorbox[RGB]{255,255,255}{\strut{} aware}\colorbox[RGB]{255,255,255}{\strut{},}\colorbox[RGB]{255,255,255}{\strut{} the}\colorbox[RGB]{255,255,255}{\strut{} names}\colorbox[RGB]{255,255,255}{\strut{}ake}\colorbox[RGB]{255,255,255}{\strut{} is}\colorbox[RGB]{255,255,255}{\strut{} Ros}\colorbox[RGB]{255,255,255}{\strut{}al}\colorbox[RGB]{255,241,227}{\strut{}ind}\colorbox[RGB]{252,146,69}{\strut{} Franklin}\colorbox[RGB]{255,255,255}{\strut{} [}\colorbox[RGB]{255,255,255}{\strut{}1}\colorbox[RGB]{254,221,187}{\strut{}]}\colorbox[RGB]{252,146,69}{\strut{} who}\colorbox[RGB]{249,131,50}{\strut{}⏎}\colorbox[RGB]{242,111,28}{\strut{}made}\colorbox[RGB]{253,208,163}{\strut{} seminal}\colorbox[RGB]{253,190,134}{\strut{} contributions}\colorbox[RGB]{252,149,72}{\strut{} in}\colorbox[RGB]{253,183,122}{\strut{} the}\colorbox[RGB]{255,244,233}{\strut{} fields}\colorbox[RGB]{250,137,56}{\strut{} of}\colorbox[RGB]{254,221,189}{\strut{} X}\colorbox[RGB]{255,255,255}{\strut{}-}\colorbox[RGB]{253,188,131}{\strut{}ray}\colorbox[RGB]{254,223,191}{\strut{} cry}}}
\end{featureexamples}

\subsubsection{Country Features}\label{sec:feature-survey-categories-countries}

Next, we see features which only activate strongly on references to specific countries. From the top activating examples, we can see that many of these features fire not just on the country name itself, but also when the country is being described.

\begin{featureexamples}
\featurechip{34M}{805282} \textbf{Rwanda}
\exampleline{{\unicodefont \colorbox[RGB]{255,255,255}{\strut{}alues}\colorbox[RGB]{255,255,255}{\strut{} for}\colorbox[RGB]{255,255,255}{\strut{} such}\colorbox[RGB]{255,255,255}{\strut{} a}\colorbox[RGB]{255,255,255}{\strut{} test}\colorbox[RGB]{255,255,255}{\strut{}.}\colorbox[RGB]{255,255,255}{\strut{}}\colorbox[RGB]{255,255,255}{\strut{}R}\colorbox[RGB]{255,255,255}{\strut{}wanda}\colorbox[RGB]{253,212,172}{\strut{},}\colorbox[RGB]{254,230,207}{\strut{} a}\colorbox[RGB]{253,158,85}{\strut{} Central}\colorbox[RGB]{253,208,163}{\strut{} African}\colorbox[RGB]{223,83,8}{\strut{} country}\colorbox[RGB]{255,244,233}{\strut{} that}\colorbox[RGB]{253,211,169}{\strut{} experienced}\colorbox[RGB]{255,255,255}{\strut{} social}\colorbox[RGB]{255,255,255}{\strut{} up}\colorbox[RGB]{255,243,230}{\strut{}hea}\colorbox[RGB]{253,197,145}{\strut{}val}\colorbox[RGB]{255,255,255}{\strut{} a}\colorbox[RGB]{255,255,255}{\strut{} generation}\colorbox[RGB]{255,255,255}{\strut{} }}}
\exampleline{{\unicodefont \colorbox[RGB]{255,255,255}{\strut{}.}\colorbox[RGB]{255,255,255}{\strut{}⏎}\colorbox[RGB]{255,255,255}{\strut{}⏎}\colorbox[RGB]{255,255,255}{\strut{}R}\colorbox[RGB]{255,255,255}{\strut{}wanda}\colorbox[RGB]{255,242,229}{\strut{} last}\colorbox[RGB]{253,194,141}{\strut{} year}\colorbox[RGB]{249,131,50}{\strut{} exported}\colorbox[RGB]{253,162,91}{\strut{} 250}\colorbox[RGB]{253,155,80}{\strut{} million}\colorbox[RGB]{254,221,189}{\strut{} USD}\colorbox[RGB]{254,229,204}{\strut{} worth}\colorbox[RGB]{236,99,18}{\strut{} of}\colorbox[RGB]{255,255,255}{\strut{} col}\colorbox[RGB]{255,255,255}{\strut{}tan}\colorbox[RGB]{255,255,255}{\strut{}.}\colorbox[RGB]{255,244,233}{\strut{} Un}\colorbox[RGB]{255,255,255}{\strut{}familiar}\colorbox[RGB]{255,255,255}{\strut{} with}\colorbox[RGB]{255,255,255}{\strut{}⏎}\colorbox[RGB]{255,255,255}{\strut{}what}\colorbox[RGB]{254,229,204}{\strut{} col}\colorbox[RGB]{255,255,255}{\strut{}tan}\colorbox[RGB]{255,255,255}{\strut{} is}\colorbox[RGB]{255,255,255}{\strut{}?}\colorbox[RGB]{255,255,255}{\strut{} It}\colorbox[RGB]{255,255,255}{\strut{}'s}\colorbox[RGB]{255,255,255}{\strut{} the}\colorbox[RGB]{255,255,255}{\strut{} }}}
\exampleline{{\unicodefont \colorbox[RGB]{255,255,255}{\strut{}mac}\colorbox[RGB]{254,221,187}{\strut{} '}\colorbox[RGB]{254,219,185}{\strut{}and}\colorbox[RGB]{240,107,24}{\strut{} stunning}\colorbox[RGB]{254,229,204}{\strut{} sc}\colorbox[RGB]{254,235,216}{\strut{}enery}\colorbox[RGB]{255,255,255}{\strut{}...''}\colorbox[RGB]{255,255,255}{\strut{} '''}\colorbox[RGB]{255,255,255}{\strut{}..}\colorbox[RGB]{255,255,255}{\strut{}we}\colorbox[RGB]{255,241,227}{\strut{} arrived}\colorbox[RGB]{253,199,149}{\strut{} on}\colorbox[RGB]{253,212,172}{\strut{} the}\colorbox[RGB]{254,219,185}{\strut{} other}\colorbox[RGB]{255,255,255}{\strut{} side}\colorbox[RGB]{253,211,169}{\strut{} of}\colorbox[RGB]{254,237,220}{\strut{} R}\colorbox[RGB]{255,244,233}{\strut{}wanda}\colorbox[RGB]{253,192,137}{\strut{} at}\colorbox[RGB]{236,99,18}{\strut{} its}\colorbox[RGB]{255,255,255}{\strut{} border}\colorbox[RGB]{253,158,85}{\strut{} with}\colorbox[RGB]{255,255,255}{\strut{} Tanz}\colorbox[RGB]{255,255,255}{\strut{}ania}\colorbox[RGB]{255,255,255}{\strut{}.'''}}}
\exampleline{{\unicodefont \colorbox[RGB]{255,255,255}{\strut{}ing}\colorbox[RGB]{255,255,255}{\strut{} a}\colorbox[RGB]{255,255,255}{\strut{} small}\colorbox[RGB]{255,255,255}{\strut{} city}\colorbox[RGB]{255,255,255}{\strut{} of}\colorbox[RGB]{255,255,255}{\strut{} 20}\colorbox[RGB]{255,255,255}{\strut{},}\colorbox[RGB]{255,255,255}{\strut{}000}\colorbox[RGB]{255,242,229}{\strut{} but}\colorbox[RGB]{253,179,116}{\strut{} R}\colorbox[RGB]{253,196,144}{\strut{}wanda}\colorbox[RGB]{254,234,214}{\strut{},}\colorbox[RGB]{253,197,145}{\strut{} a}\colorbox[RGB]{254,221,189}{\strut{} nation}\colorbox[RGB]{246,121,38}{\strut{} of}\colorbox[RGB]{239,105,22}{\strut{} 12}\colorbox[RGB]{253,214,175}{\strut{} million}\colorbox[RGB]{255,255,255}{\strut{}⏎}\colorbox[RGB]{254,233,212}{\strut{}(}\colorbox[RGB]{254,228,202}{\strut{}and}\colorbox[RGB]{254,233,212}{\strut{} now}\colorbox[RGB]{254,235,216}{\strut{} much}\colorbox[RGB]{255,244,233}{\strut{} of}\colorbox[RGB]{255,255,255}{\strut{} Ghana}\colorbox[RGB]{254,229,204}{\strut{},}\colorbox[RGB]{255,255,255}{\strut{} population}\colorbox[RGB]{255,255,255}{\strut{} of}\colorbox[RGB]{255,255,255}{\strut{} 28}\colorbox[RGB]{255,255,255}{\strut{} }}}
\exampleline{{\unicodefont \colorbox[RGB]{255,255,255}{\strut{}be}\colorbox[RGB]{255,255,255}{\strut{}⏎}\colorbox[RGB]{255,255,255}{\strut{}interested}\colorbox[RGB]{255,255,255}{\strut{} to}\colorbox[RGB]{255,255,255}{\strut{} learn}\colorbox[RGB]{255,255,255}{\strut{} that}\colorbox[RGB]{255,255,255}{\strut{} Paul}\colorbox[RGB]{255,255,255}{\strut{} K}\colorbox[RGB]{255,255,255}{\strut{}ag}\colorbox[RGB]{253,187,129}{\strut{}ame}\colorbox[RGB]{253,181,119}{\strut{},}\colorbox[RGB]{251,139,59}{\strut{} the}\colorbox[RGB]{255,255,255}{\strut{} ruler}\colorbox[RGB]{241,109,26}{\strut{} of}\colorbox[RGB]{255,255,255}{\strut{} R}\colorbox[RGB]{255,255,255}{\strut{}wanda}\colorbox[RGB]{254,229,204}{\strut{},}\colorbox[RGB]{254,236,218}{\strut{} put}\colorbox[RGB]{254,228,202}{\strut{} together}\colorbox[RGB]{255,244,233}{\strut{} a}\colorbox[RGB]{253,216,179}{\strut{} team}\colorbox[RGB]{254,226,199}{\strut{}⏎}\colorbox[RGB]{255,255,255}{\strut{}spec}\colorbox[RGB]{255,255,255}{\strut{}ifically}\colorbox[RGB]{253,208,163}{\strut{} for}\colorbox[RGB]{253,179,116}{\strut{} the}\colorbox[RGB]{255,255,255}{\strut{} }}}
\end{featureexamples}

\begin{featureexamples}
\featurechip{34M}{29297045} \textbf{Canada}
\exampleline{{\unicodefont \colorbox[RGB]{255,255,255}{\strut{} ''}\colorbox[RGB]{255,244,233}{\strut{}Canada}\colorbox[RGB]{254,228,201}{\strut{},}\colorbox[RGB]{253,194,141}{\strut{} a}\colorbox[RGB]{253,188,131}{\strut{} country}\colorbox[RGB]{253,197,145}{\strut{} known}\colorbox[RGB]{250,137,56}{\strut{} for}\colorbox[RGB]{253,206,160}{\strut{} its}\colorbox[RGB]{253,172,105}{\strut{} natural}\colorbox[RGB]{253,212,172}{\strut{} wonders}\colorbox[RGB]{254,223,191}{\strut{},}\colorbox[RGB]{223,83,8}{\strut{} its}\colorbox[RGB]{253,157,83}{\strut{} universal}\colorbox[RGB]{254,224,194}{\strut{} healthcare}\colorbox[RGB]{251,143,64}{\strut{},}\colorbox[RGB]{253,203,155}{\strut{} and}\colorbox[RGB]{238,104,21}{\strut{} its}\colorbox[RGB]{253,155,80}{\strut{} really}\colorbox[RGB]{253,167,98}{\strut{} polite}\colorbox[RGB]{254,235,216}{\strut{} peop}}}
\exampleline{{\unicodefont \colorbox[RGB]{253,218,182}{\strut{}re}\colorbox[RGB]{254,235,216}{\strut{} relaxed}\colorbox[RGB]{255,255,255}{\strut{}.}\colorbox[RGB]{255,255,255}{\strut{}⏎}\colorbox[RGB]{255,255,255}{\strut{}⏎}\colorbox[RGB]{255,243,231}{\strut{}Also}\colorbox[RGB]{255,240,225}{\strut{},}\colorbox[RGB]{254,233,211}{\strut{} since}\colorbox[RGB]{253,192,137}{\strut{} Canada}\colorbox[RGB]{253,176,111}{\strut{} has}\colorbox[RGB]{253,196,144}{\strut{} a}\colorbox[RGB]{253,203,155}{\strut{} reputation}\colorbox[RGB]{252,145,67}{\strut{} as}\colorbox[RGB]{250,133,52}{\strut{} ''}\colorbox[RGB]{253,174,108}{\strut{}free}\colorbox[RGB]{255,255,255}{\strut{} health}\colorbox[RGB]{255,243,231}{\strut{} care}\colorbox[RGB]{254,223,191}{\strut{} for}\colorbox[RGB]{255,244,233}{\strut{} everyone}\colorbox[RGB]{255,255,255}{\strut{}⏎}\colorbox[RGB]{255,244,233}{\strut{}every}\colorbox[RGB]{255,238,221}{\strut{}where}\colorbox[RGB]{255,255,255}{\strut{}!''}\colorbox[RGB]{255,255,255}{\strut{} look}\colorbox[RGB]{255,255,255}{\strut{} in}}}
\exampleline{{\unicodefont \colorbox[RGB]{255,255,255}{\strut{}-----}\colorbox[RGB]{255,255,255}{\strut{}⏎}\colorbox[RGB]{255,255,255}{\strut{}jp}\colorbox[RGB]{255,255,255}{\strut{}po}\colorbox[RGB]{255,255,255}{\strut{}pe}\colorbox[RGB]{255,255,255}{\strut{}⏎}\colorbox[RGB]{255,255,255}{\strut{}I}\colorbox[RGB]{255,255,255}{\strut{}'d}\colorbox[RGB]{255,255,255}{\strut{} vote}\colorbox[RGB]{255,255,255}{\strut{} to}\colorbox[RGB]{255,255,255}{\strut{} let}\colorbox[RGB]{253,216,179}{\strut{} Canada}\colorbox[RGB]{255,244,233}{\strut{} run}\colorbox[RGB]{255,244,233}{\strut{} the}\colorbox[RGB]{255,255,255}{\strut{} world}\colorbox[RGB]{253,201,152}{\strut{}.}\colorbox[RGB]{250,133,52}{\strut{} Kil}\colorbox[RGB]{255,238,221}{\strut{}lem}\colorbox[RGB]{253,197,145}{\strut{} with}\colorbox[RGB]{253,206,160}{\strut{} kindness}\colorbox[RGB]{253,218,182}{\strut{}!}\colorbox[RGB]{253,201,152}{\strut{} Plus}\colorbox[RGB]{254,225,196}{\strut{} adding}\colorbox[RGB]{254,230,207}{\strut{} Box}\colorbox[RGB]{254,225,196}{\strut{}ing}\colorbox[RGB]{254,236,218}{\strut{}⏎}\colorbox[RGB]{255,244,233}{\strut{}Day}\colorbox[RGB]{255,255,255}{\strut{} would}\colorbox[RGB]{255,255,255}{\strut{} b}}}
\exampleline{{\unicodefont \colorbox[RGB]{254,232,209}{\strut{}g}\colorbox[RGB]{255,255,255}{\strut{}⏎}\colorbox[RGB]{255,244,233}{\strut{}fine}\colorbox[RGB]{254,235,216}{\strut{} and}\colorbox[RGB]{253,206,160}{\strut{} is}\colorbox[RGB]{255,255,255}{\strut{} trust}\colorbox[RGB]{255,244,233}{\strut{}worthy}\colorbox[RGB]{254,233,211}{\strut{},}\colorbox[RGB]{255,242,229}{\strut{} simply}\colorbox[RGB]{253,206,160}{\strut{} because}\colorbox[RGB]{253,194,141}{\strut{} of}\colorbox[RGB]{253,194,141}{\strut{} Canada}\colorbox[RGB]{250,133,52}{\strut{}'s}\colorbox[RGB]{252,145,67}{\strut{} supposed}\colorbox[RGB]{253,199,149}{\strut{} reputation}\colorbox[RGB]{255,255,255}{\strut{}.}\colorbox[RGB]{255,244,233}{\strut{}⏎}\colorbox[RGB]{255,239,223}{\strut{}⏎}\colorbox[RGB]{255,255,255}{\strut{}------}\colorbox[RGB]{255,255,255}{\strut{}⏎}\colorbox[RGB]{255,255,255}{\strut{}t}\colorbox[RGB]{255,255,255}{\strut{}ay}\colorbox[RGB]{255,255,255}{\strut{}bin}\colorbox[RGB]{255,255,255}{\strut{}⏎}\colorbox[RGB]{255,255,255}{\strut{}This}\colorbox[RGB]{255,255,255}{\strut{} is}\colorbox[RGB]{255,255,255}{\strut{} prett}}}
\exampleline{{\unicodefont \colorbox[RGB]{255,255,255}{\strut{}Oh}\colorbox[RGB]{255,255,255}{\strut{} well}\colorbox[RGB]{255,255,255}{\strut{}.}\colorbox[RGB]{253,215,176}{\strut{} Canada}\colorbox[RGB]{254,232,209}{\strut{} used}\colorbox[RGB]{253,218,182}{\strut{} to}\colorbox[RGB]{253,166,97}{\strut{} seem}\colorbox[RGB]{253,185,126}{\strut{} like}\colorbox[RGB]{253,164,94}{\strut{} the}\colorbox[RGB]{253,199,149}{\strut{} last}\colorbox[RGB]{254,226,199}{\strut{} bast}\colorbox[RGB]{255,239,223}{\strut{}ion}\colorbox[RGB]{250,136,55}{\strut{} of}\colorbox[RGB]{253,206,160}{\strut{} decent}\colorbox[RGB]{253,212,172}{\strut{} civilization}\colorbox[RGB]{255,238,221}{\strut{}.}\colorbox[RGB]{255,255,255}{\strut{}⏎}\colorbox[RGB]{255,255,255}{\strut{}Har}\colorbox[RGB]{255,255,255}{\strut{}per}\colorbox[RGB]{255,255,255}{\strut{} et}\colorbox[RGB]{255,255,255}{\strut{} al}\colorbox[RGB]{255,255,255}{\strut{} saw}\colorbox[RGB]{255,255,255}{\strut{} to}\colorbox[RGB]{255,255,255}{\strut{} that}\colorbox[RGB]{255,255,255}{\strut{} and}}}
\end{featureexamples}

\begin{featureexamples}
\featurechip{34M}{5381828} \textbf{Belgium}
\exampleline{{\unicodefont \colorbox[RGB]{255,255,255}{\strut{}on}\colorbox[RGB]{255,255,255}{\strut{} and}\colorbox[RGB]{255,255,255}{\strut{} more}\colorbox[RGB]{255,255,255}{\strut{}⏎}\colorbox[RGB]{255,255,255}{\strut{}sen}\colorbox[RGB]{255,255,255}{\strut{}iors}\colorbox[RGB]{255,255,255}{\strut{}.}\colorbox[RGB]{255,255,255}{\strut{}⏎}\colorbox[RGB]{255,255,255}{\strut{}⏎}\colorbox[RGB]{255,255,255}{\strut{}\~{}\~{}\~{}}\colorbox[RGB]{255,255,255}{\strut{}⏎}\colorbox[RGB]{255,255,255}{\strut{}r}\colorbox[RGB]{255,255,255}{\strut{}urban}\colorbox[RGB]{255,255,255}{\strut{}⏎}\colorbox[RGB]{255,255,255}{\strut{}And}\colorbox[RGB]{255,255,255}{\strut{} esp}\colorbox[RGB]{255,255,255}{\strut{}.}\colorbox[RGB]{255,244,233}{\strut{} Belgium}\colorbox[RGB]{255,243,230}{\strut{}.}\colorbox[RGB]{236,99,18}{\strut{} The}\colorbox[RGB]{252,145,67}{\strut{} highest}\colorbox[RGB]{255,255,255}{\strut{} outlier}\colorbox[RGB]{255,244,233}{\strut{} without}\colorbox[RGB]{255,244,233}{\strut{} proper}\colorbox[RGB]{255,255,255}{\strut{} explanation}\colorbox[RGB]{255,255,255}{\strut{} so}\colorbox[RGB]{255,255,255}{\strut{} fa}}}
\exampleline{{\unicodefont \colorbox[RGB]{255,255,255}{\strut{}ri}\colorbox[RGB]{255,255,255}{\strut{}C}\colorbox[RGB]{255,255,255}{\strut{}\^{}\^{}}\colorbox[RGB]{254,233,211}{\strut{}:}\colorbox[RGB]{252,151,74}{\strut{} we}\colorbox[RGB]{253,196,144}{\strut{} have}\colorbox[RGB]{253,177,113}{\strut{} a}\colorbox[RGB]{252,151,74}{\strut{} weird}\colorbox[RGB]{246,121,38}{\strut{} small}\colorbox[RGB]{253,164,94}{\strut{} country}\colorbox[RGB]{253,203,155}{\strut{}⏎}\colorbox[RGB]{255,255,255}{\strut{}<}\colorbox[RGB]{243,114,30}{\strut{}lot}\colorbox[RGB]{250,133,52}{\strut{}us}\colorbox[RGB]{239,105,22}{\strut{}psych}\colorbox[RGB]{255,243,231}{\strut{}je}\colorbox[RGB]{255,255,255}{\strut{}>}\colorbox[RGB]{255,255,255}{\strut{} E}\colorbox[RGB]{249,131,50}{\strut{}ri}\colorbox[RGB]{253,218,182}{\strut{}C}\colorbox[RGB]{255,255,255}{\strut{}\^{}\^{}}\colorbox[RGB]{254,234,214}{\strut{}:}\colorbox[RGB]{253,188,131}{\strut{} bel}\colorbox[RGB]{253,187,129}{\strut{}gian}\colorbox[RGB]{254,219,185}{\strut{} w}\colorbox[RGB]{253,205,158}{\strut{}af}\colorbox[RGB]{255,241,227}{\strut{}les}\colorbox[RGB]{253,155,80}{\strut{},}\colorbox[RGB]{254,219,185}{\strut{} ch}\colorbox[RGB]{253,211,169}{\strut{}ocol}\colorbox[RGB]{255,238,221}{\strut{}ats}\colorbox[RGB]{238,104,21}{\strut{},}\colorbox[RGB]{253,167,98}{\strut{} f}\colorbox[RGB]{253,176,111}{\strut{}rench}\colorbox[RGB]{253,212,172}{\strut{} f}\colorbox[RGB]{254,223,191}{\strut{}ries}\colorbox[RGB]{243,115,31}{\strut{} and}}}
\exampleline{{\unicodefont \colorbox[RGB]{255,255,255}{\strut{} Netherlands}\colorbox[RGB]{254,233,211}{\strut{} only}\colorbox[RGB]{254,237,220}{\strut{} has}\colorbox[RGB]{254,228,202}{\strut{} one}\colorbox[RGB]{254,236,218}{\strut{} language}\colorbox[RGB]{255,255,255}{\strut{},}\colorbox[RGB]{255,255,255}{\strut{} Dutch}\colorbox[RGB]{255,243,230}{\strut{}.}\colorbox[RGB]{254,221,187}{\strut{} Belgium}\colorbox[RGB]{243,115,31}{\strut{} has}\colorbox[RGB]{253,179,116}{\strut{} two}\colorbox[RGB]{253,176,111}{\strut{}:}\colorbox[RGB]{253,155,80}{\strut{} the}\colorbox[RGB]{253,194,141}{\strut{} top}\colorbox[RGB]{253,160,88}{\strut{} part}\colorbox[RGB]{255,255,255}{\strut{}⏎}\colorbox[RGB]{253,166,97}{\strut{}speak}\colorbox[RGB]{253,174,108}{\strut{}s}\colorbox[RGB]{255,255,255}{\strut{} Dutch}\colorbox[RGB]{253,209,166}{\strut{},}\colorbox[RGB]{253,172,105}{\strut{} the}\colorbox[RGB]{253,177,113}{\strut{} bottom}\colorbox[RGB]{253,194,141}{\strut{} part}\colorbox[RGB]{255,244,233}{\strut{} }}}
\exampleline{{\unicodefont \colorbox[RGB]{255,255,255}{\strut{} is}\colorbox[RGB]{255,255,255}{\strut{} repeated}\colorbox[RGB]{255,255,255}{\strut{} across}\colorbox[RGB]{255,255,255}{\strut{} Europe}\colorbox[RGB]{255,255,255}{\strut{},}\colorbox[RGB]{255,255,255}{\strut{} in}\colorbox[RGB]{253,208,163}{\strut{} Belgium}\colorbox[RGB]{255,255,255}{\strut{} for}\colorbox[RGB]{253,167,98}{\strut{}⏎}\colorbox[RGB]{253,187,129}{\strut{}example}\colorbox[RGB]{244,117,34}{\strut{} the}\colorbox[RGB]{253,215,176}{\strut{} Dutch}\colorbox[RGB]{253,209,166}{\strut{}-}\colorbox[RGB]{253,209,166}{\strut{}speak}\colorbox[RGB]{253,203,155}{\strut{}ers}\colorbox[RGB]{253,196,144}{\strut{} in}\colorbox[RGB]{253,177,113}{\strut{} the}\colorbox[RGB]{253,196,144}{\strut{} North}\colorbox[RGB]{253,209,166}{\strut{} are}\colorbox[RGB]{253,176,111}{\strut{} very}\colorbox[RGB]{255,243,230}{\strut{} much}\colorbox[RGB]{253,211,169}{\strut{} more}\colorbox[RGB]{254,228,202}{\strut{} e}}}
\exampleline{{\unicodefont \colorbox[RGB]{255,255,255}{\strut{} make}\colorbox[RGB]{255,255,255}{\strut{} the}\colorbox[RGB]{255,255,255}{\strut{} pizza}\colorbox[RGB]{255,255,255}{\strut{} and}\colorbox[RGB]{255,255,255}{\strut{} lat}\colorbox[RGB]{255,255,255}{\strut{}te}\colorbox[RGB]{255,255,255}{\strut{} runs}\colorbox[RGB]{255,255,255}{\strut{}.}\colorbox[RGB]{255,255,255}{\strut{}⏎}\colorbox[RGB]{255,255,255}{\strut{}⏎}\colorbox[RGB]{255,255,255}{\strut{}}\colorbox[RGB]{255,255,255}{\strut{}⏎}\colorbox[RGB]{255,255,255}{\strut{}⏎}\colorbox[RGB]{253,208,163}{\strut{}Bel}\colorbox[RGB]{254,229,204}{\strut{}gium}\colorbox[RGB]{253,212,172}{\strut{} :}\colorbox[RGB]{251,143,64}{\strut{} 500}\colorbox[RGB]{244,117,34}{\strut{} days}\colorbox[RGB]{250,137,56}{\strut{} without}\colorbox[RGB]{251,141,62}{\strut{} a}\colorbox[RGB]{253,218,182}{\strut{} government}\colorbox[RGB]{254,234,214}{\strut{}.}\colorbox[RGB]{255,255,255}{\strut{} -}\colorbox[RGB]{255,255,255}{\strut{} sk}\colorbox[RGB]{255,255,255}{\strut{}bo}\colorbox[RGB]{255,255,255}{\strut{}hra}\colorbox[RGB]{255,255,255}{\strut{}123}\colorbox[RGB]{255,255,255}{\strut{}⏎}\colorbox[RGB]{255,255,255}{\strut{}http}\colorbox[RGB]{254,235,216}{\strut{}://}\colorbox[RGB]{250,133,52}{\strut{}www}\colorbox[RGB]{255,255,255}{\strut{}.}\colorbox[RGB]{255,255,255}{\strut{}h}\colorbox[RGB]{255,255,255}{\strut{}u}}}
\end{featureexamples}

\begin{featureexamples}
\featurechip{34M}{32188099} \textbf{Iceland}
\exampleline{{\unicodefont \colorbox[RGB]{255,255,255}{\strut{}il}\colorbox[RGB]{255,255,255}{\strut{}ization}\colorbox[RGB]{255,255,255}{\strut{}'}\colorbox[RGB]{255,255,255}{\strut{} really}\colorbox[RGB]{255,255,255}{\strut{} is}\colorbox[RGB]{255,255,255}{\strut{} all}\colorbox[RGB]{255,255,255}{\strut{} that}\colorbox[RGB]{255,255,255}{\strut{} civil}\colorbox[RGB]{255,255,255}{\strut{}ized}\colorbox[RGB]{255,255,255}{\strut{}.}\colorbox[RGB]{253,174,108}{\strut{} Iceland}\colorbox[RGB]{253,203,155}{\strut{} is}\colorbox[RGB]{223,83,8}{\strut{} a}\colorbox[RGB]{253,167,98}{\strut{} small}\colorbox[RGB]{253,187,129}{\strut{} nation}\colorbox[RGB]{253,158,85}{\strut{},}\colorbox[RGB]{255,240,225}{\strut{}⏎}\colorbox[RGB]{254,233,212}{\strut{}rel}\colorbox[RGB]{254,225,196}{\strut{}atively}\colorbox[RGB]{255,255,255}{\strut{} few}\colorbox[RGB]{255,244,233}{\strut{} people}\colorbox[RGB]{254,233,212}{\strut{} and}\colorbox[RGB]{253,218,182}{\strut{} tightly}\colorbox[RGB]{255,255,255}{\strut{} k}}}
\exampleline{{\unicodefont \colorbox[RGB]{255,255,255}{\strut{} which}\colorbox[RGB]{255,255,255}{\strut{} is}\colorbox[RGB]{255,255,255}{\strut{} shorter}\colorbox[RGB]{255,255,255}{\strut{}⏎}\colorbox[RGB]{255,255,255}{\strut{}⏎}\colorbox[RGB]{255,255,255}{\strut{}}\colorbox[RGB]{255,255,255}{\strut{}⏎}\colorbox[RGB]{253,188,131}{\strut{}I}\colorbox[RGB]{253,194,141}{\strut{}celand}\colorbox[RGB]{241,109,26}{\strut{} becomes}\colorbox[RGB]{251,141,62}{\strut{} first}\colorbox[RGB]{253,190,134}{\strut{} country}\colorbox[RGB]{224,84,8}{\strut{} to}\colorbox[RGB]{253,196,144}{\strut{} legal}\colorbox[RGB]{253,203,155}{\strut{}ise}\colorbox[RGB]{255,240,225}{\strut{} equal}\colorbox[RGB]{255,238,221}{\strut{} pay}\colorbox[RGB]{255,255,255}{\strut{} -}\colorbox[RGB]{255,255,255}{\strut{} d}\colorbox[RGB]{255,255,255}{\strut{}acm}\colorbox[RGB]{255,255,255}{\strut{}⏎}\colorbox[RGB]{255,255,255}{\strut{}http}\colorbox[RGB]{255,255,255}{\strut{}://}\colorbox[RGB]{253,211,169}{\strut{}www}\colorbox[RGB]{255,255,255}{\strut{}.}\colorbox[RGB]{255,255,255}{\strut{}al}\colorbox[RGB]{255,255,255}{\strut{}j}\colorbox[RGB]{255,255,255}{\strut{}azeera}\colorbox[RGB]{255,243,231}{\strut{}.}\colorbox[RGB]{255,240,225}{\strut{}co}}}
\exampleline{{\unicodefont \colorbox[RGB]{255,255,255}{\strut{}in}\colorbox[RGB]{255,255,255}{\strut{} this}\colorbox[RGB]{255,255,255}{\strut{} last}\colorbox[RGB]{255,255,255}{\strut{} programme}\colorbox[RGB]{255,255,255}{\strut{} in}\colorbox[RGB]{253,212,172}{\strut{} Iceland}\colorbox[RGB]{255,255,255}{\strut{},}\colorbox[RGB]{255,240,225}{\strut{} because}\colorbox[RGB]{255,255,255}{\strut{} this}\colorbox[RGB]{253,192,137}{\strut{} is}\colorbox[RGB]{226,86,9}{\strut{} the}\colorbox[RGB]{255,255,255}{\strut{} seat}\colorbox[RGB]{246,121,38}{\strut{} of}\colorbox[RGB]{255,255,255}{\strut{} the}\colorbox[RGB]{253,160,88}{\strut{} oldest}\colorbox[RGB]{253,185,126}{\strut{} democracy}\colorbox[RGB]{251,139,59}{\strut{} in}\colorbox[RGB]{255,255,255}{\strut{} Northern}\colorbox[RGB]{255,255,255}{\strut{} Europe}\colorbox[RGB]{254,234,214}{\strut{}.}}}
\exampleline{{\unicodefont \colorbox[RGB]{255,255,255}{\strut{}ll}\colorbox[RGB]{255,255,255}{\strut{}M}\colorbox[RGB]{255,255,255}{\strut{}tl}\colorbox[RGB]{255,255,255}{\strut{}Al}\colorbox[RGB]{255,255,255}{\strut{}cohol}\colorbox[RGB]{255,255,255}{\strut{}c}\colorbox[RGB]{255,255,255}{\strut{}⏎}\colorbox[RGB]{255,255,255}{\strut{}A}\colorbox[RGB]{255,255,255}{\strut{} bit}\colorbox[RGB]{255,255,255}{\strut{} off}\colorbox[RGB]{255,255,255}{\strut{} topic}\colorbox[RGB]{255,255,255}{\strut{},}\colorbox[RGB]{255,255,255}{\strut{} but}\colorbox[RGB]{253,183,122}{\strut{} Iceland}\colorbox[RGB]{253,192,137}{\strut{} is}\colorbox[RGB]{253,170,101}{\strut{} the}\colorbox[RGB]{232,93,14}{\strut{} most}\colorbox[RGB]{255,244,233}{\strut{} beautiful}\colorbox[RGB]{254,229,204}{\strut{} place}\colorbox[RGB]{255,255,255}{\strut{} that}\colorbox[RGB]{255,255,255}{\strut{} I}\colorbox[RGB]{255,255,255}{\strut{} have}\colorbox[RGB]{255,255,255}{\strut{} ever}\colorbox[RGB]{253,203,155}{\strut{}⏎}\colorbox[RGB]{255,255,255}{\strut{}visited}\colorbox[RGB]{255,255,255}{\strut{}.}\colorbox[RGB]{253,212,172}{\strut{} It}\colorbox[RGB]{255,255,255}{\strut{}'s}\colorbox[RGB]{255,255,255}{\strut{} g}}}
\exampleline{{\unicodefont \colorbox[RGB]{255,255,255}{\strut{}earth}\colorbox[RGB]{255,241,227}{\strut{} on}\colorbox[RGB]{255,244,233}{\strut{} the}\colorbox[RGB]{255,255,255}{\strut{} S}\colorbox[RGB]{255,255,255}{\strut{}na}\colorbox[RGB]{255,255,255}{\strut{}eff}\colorbox[RGB]{255,255,255}{\strut{}els}\colorbox[RGB]{255,255,255}{\strut{} volcano}\colorbox[RGB]{255,255,255}{\strut{}.''}\colorbox[RGB]{255,255,255}{\strut{} ''}\colorbox[RGB]{255,244,233}{\strut{}In}\colorbox[RGB]{253,211,169}{\strut{} 1980}\colorbox[RGB]{254,230,207}{\strut{},}\colorbox[RGB]{254,228,202}{\strut{} the}\colorbox[RGB]{233,95,15}{\strut{} Iceland}\colorbox[RGB]{237,102,20}{\strut{}ers}\colorbox[RGB]{253,187,129}{\strut{} elected}\colorbox[RGB]{248,127,44}{\strut{} the}\colorbox[RGB]{255,255,255}{\strut{} world}\colorbox[RGB]{253,174,108}{\strut{}'s}\colorbox[RGB]{253,176,111}{\strut{} first}\colorbox[RGB]{254,228,201}{\strut{} female}\colorbox[RGB]{254,230,207}{\strut{} president}\colorbox[RGB]{255,243,231}{\strut{}.''}\colorbox[RGB]{255,255,255}{\strut{} }}}
\end{featureexamples}

\subsubsection{Basic Code Features}\label{sec:feature-survey-categories-code}

We also see a number of features that represent different syntax elements or other low-level concepts in code, which give the impression of syntax highlighting when visualized together (here for simplicity we binarize activation information, only distinguishing between zero vs. nonzero activations):

\begin{figure}[!htp]
    \centering
    \includegraphics[width=0.9\textwidth,height=0.7\textheight,keepaspectratio]{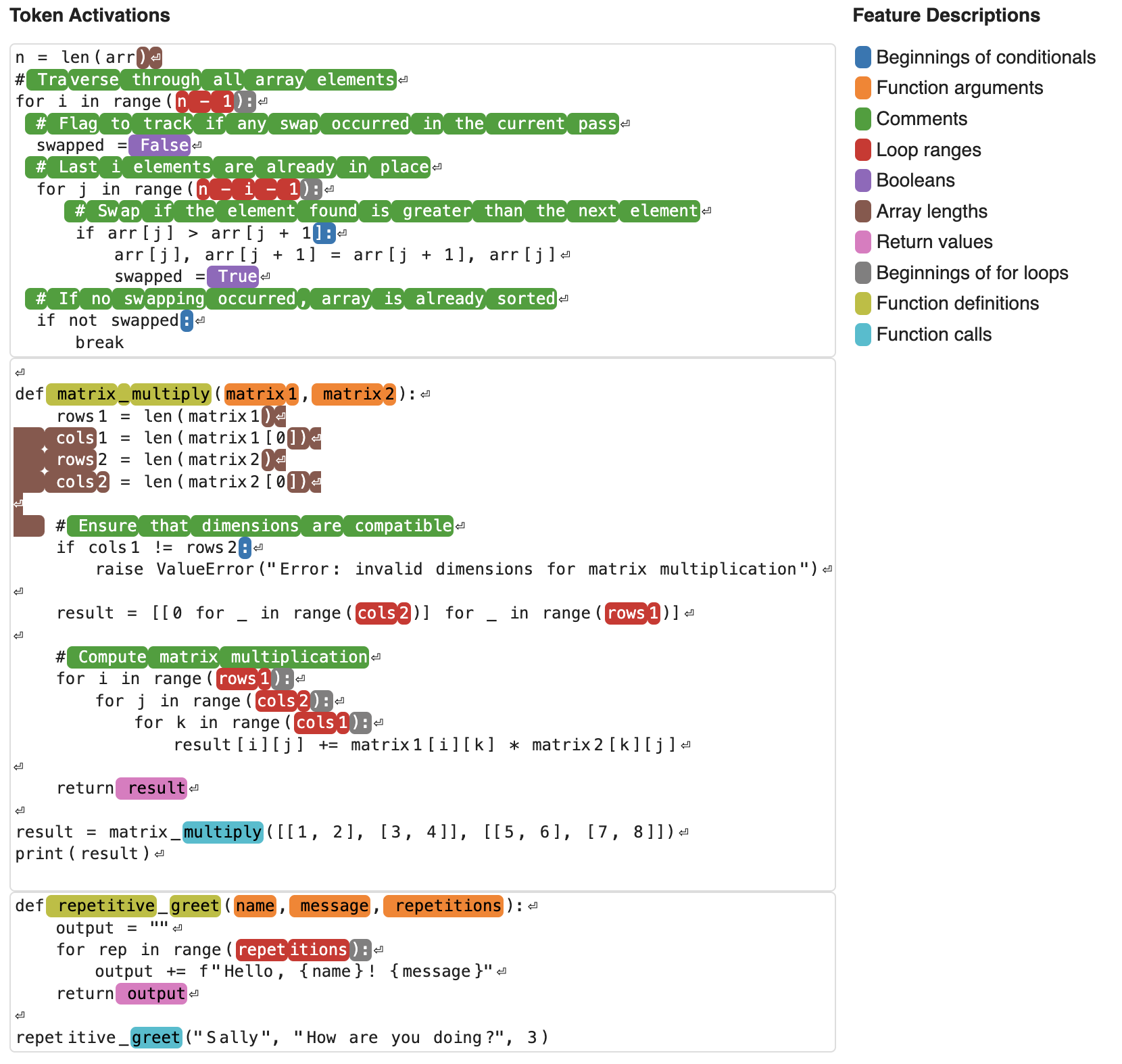}
    \label{fig:gdoc_29}
\end{figure}

These features were chosen primarily to fire on the Python examples. We have found that there is some transfer from Python code features to related languages like Java, but not more distant ones (e.g.~Haskell), suggesting at least some level of language specificity. We hypothesize that more abstract features are more likely to span many languages, but so far have only found one concrete example of this (see the \hyperref[sec:assessing-sophisticated-code-error]{Code error feature}).

\subsubsection{List Position Features}\label{sec:feature-survey-categories-list-pos}

Finally, we see features that fire on particular positions in lists, regardless of the content in those positions:

\begin{figure}[!htp]
    \centering
    \includegraphics[width=0.9\textwidth,height=0.7\textheight,keepaspectratio]{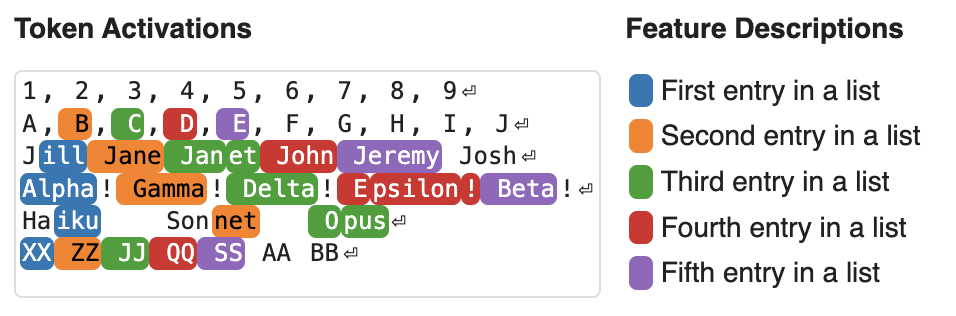}
    \label{fig:gdoc_30}
\end{figure}

Notice that these don’t fire on the first line. This is likely because the model doesn’t interpret the prompt as containing lists until it reaches the second line.

We have only scratched the surface of the features present in these SAEs, and we expect to find much more in future work.

\section{Features as Computational Intermediates}\label{sec:computational}

Another potential application of features is that they let us examine the intermediate computation that the model uses to produce an output. As a proof of concept, we observe that in prompts where intermediate computation is required, we find active features corresponding to some of the expected intermediate results.

A simple strategy for efficiently identifying causally important features for a model's output is to compute \textit{attributions}, which are local linear approximations of the effect of turning a feature off at a specific location on the model's next-token prediction.\footnote{More explicitly: We compute the gradient of the difference between an output logit of interest and the logit of a specific other baseline token (or the average of the logits across all tokens) with respect to the residual stream activations in the middle layer. Then the attribution of that logit difference to a feature is defined as the dot product of that gradient with the feature vector (SAE decoder weight), multiplied by the feature's activation. This method is equivalent to the “attribution patching” technique introduced in \href{https://www.neelnanda.io/mechanistic-interpretability/attribution-patching}{Attribution Patching: Activation Patching At Industrial Scale}, except that we use a baseline value of 0 for the feature instead of a baseline value taken from the feature’s activity on a second prompt.} We also perform feature ablations, where we clamp a feature’s value to zero at a specific token position during a forward pass, which measures the full, potentially nonlinear causal effect of that feature’s activation in that position on the model output. This is much slower since it requires one forward pass for every feature that activates at each position, so we often used attribution as a preliminary step to filter the set of features to ablate. (In the case studies shown below, we do ablate every active feature for completeness, and find a 0.8 correlation between attribution and ablation effects; see \hyperref[sec:appendix-ablations]{appendix}.)

We find that the middle layer residual stream of the model contains a range of features causally implicated in the model's completion.

\subsection{Example: Emotional Inferences}\label{sec:computational-sad}

As an example, we consider the following incomplete prompt:

\begin{quote}
John says, ''I want to be alone right now.'' John feels\\
(completion: sad − happy)
\end{quote}

To continue this text, the model must parse the quote from John, identify his state of mind, and then translate that into a likely feeling.

If we sort features by either their attribution or their ablation effect on the completion “sad” (with respect to a baseline completion of “happy”), the top two features are:

\begin{itemize}
    \item \featurechip{1M}{22623} -- This feature fires when someone expresses a need or desire to be alone or have personal time and space, as in “she would probably want some time to herself”. This is active from the word “alone” onwards. This suggests the model has gotten the gist of John's expression.
    \item \featurechip{1M}{781220} -- This feature detects expressions of sadness, crying, grief, and related emotional distress or sorrow, as in “the inconsolable girl sobs”. This is active on “John feels”. This suggests the model has inferred what someone who says they are alone might be feeling.
\end{itemize}

If we look at dataset examples, we can see that they align with these interpretations. Below, we show a small number of examples, but you can click on a feature ID to see more.

\begin{featureexamples}
\featurechip{1M}{22623} \textbf{Need or desire to be alone}
\exampleline{{\unicodefont \colorbox[RGB]{255,255,255}{\strut{}s}\colorbox[RGB]{255,255,255}{\strut{} got}\colorbox[RGB]{255,255,255}{\strut{} a}\colorbox[RGB]{254,229,204}{\strut{} lot}\colorbox[RGB]{255,255,255}{\strut{} on}\colorbox[RGB]{255,255,255}{\strut{} his}\colorbox[RGB]{255,239,223}{\strut{} mind}\colorbox[RGB]{255,240,225}{\strut{}.''}\colorbox[RGB]{255,255,255}{\strut{} ''}\colorbox[RGB]{255,242,229}{\strut{}He}\colorbox[RGB]{254,235,216}{\strut{} needs}\colorbox[RGB]{253,203,155}{\strut{} some}\colorbox[RGB]{248,129,47}{\strut{} time}\colorbox[RGB]{253,174,108}{\strut{} to}\colorbox[RGB]{223,83,8}{\strut{} himself}\colorbox[RGB]{254,221,189}{\strut{}.''}\colorbox[RGB]{255,255,255}{\strut{} ''}\colorbox[RGB]{255,255,255}{\strut{}Why}\colorbox[RGB]{255,255,255}{\strut{} not}\colorbox[RGB]{255,255,255}{\strut{} come}\colorbox[RGB]{255,255,255}{\strut{} right}\colorbox[RGB]{255,255,255}{\strut{} out}\colorbox[RGB]{255,255,255}{\strut{} and}\colorbox[RGB]{255,255,255}{\strut{} say}\colorbox[RGB]{255,255,255}{\strut{} what}\colorbox[RGB]{255,241,227}{\strut{} you}\colorbox[RGB]{255,255,255}{\strut{} mea}}}
\exampleline{{\unicodefont \colorbox[RGB]{255,255,255}{\strut{}''}\colorbox[RGB]{255,255,255}{\strut{} ''}\colorbox[RGB]{254,225,196}{\strut{}I}\colorbox[RGB]{255,255,255}{\strut{}'m}\colorbox[RGB]{255,255,255}{\strut{} working}\colorbox[RGB]{254,228,202}{\strut{} through}\colorbox[RGB]{253,185,126}{\strut{} something}\colorbox[RGB]{253,212,172}{\strut{},}\colorbox[RGB]{253,157,83}{\strut{} and}\colorbox[RGB]{253,208,163}{\strut{} I}\colorbox[RGB]{253,164,94}{\strut{} just}\colorbox[RGB]{243,115,31}{\strut{} need}\colorbox[RGB]{228,88,11}{\strut{} space}\colorbox[RGB]{252,146,69}{\strut{} to}\colorbox[RGB]{253,167,98}{\strut{} think}\colorbox[RGB]{253,187,129}{\strut{}.''}\colorbox[RGB]{255,244,233}{\strut{} ''}\colorbox[RGB]{255,244,233}{\strut{}I}\colorbox[RGB]{255,244,233}{\strut{} can}\colorbox[RGB]{254,229,204}{\strut{}'t}\colorbox[RGB]{255,255,255}{\strut{} soldier}\colorbox[RGB]{255,255,255}{\strut{} on}\colorbox[RGB]{255,255,255}{\strut{} like}\colorbox[RGB]{255,255,255}{\strut{} you}\colorbox[RGB]{255,255,255}{\strut{},}\colorbox[RGB]{255,255,255}{\strut{} Lis}\colorbox[RGB]{255,255,255}{\strut{}bon}}}
\exampleline{{\unicodefont \colorbox[RGB]{255,255,255}{\strut{}e}\colorbox[RGB]{255,244,233}{\strut{} shit}\colorbox[RGB]{255,255,255}{\strut{} that}\colorbox[RGB]{254,233,211}{\strut{} I}\colorbox[RGB]{255,240,225}{\strut{} got}\colorbox[RGB]{254,229,204}{\strut{} to}\colorbox[RGB]{254,236,218}{\strut{} work}\colorbox[RGB]{253,199,149}{\strut{} out}\colorbox[RGB]{254,234,214}{\strut{},}\colorbox[RGB]{254,221,189}{\strut{} and}\colorbox[RGB]{255,244,233}{\strut{}''}\colorbox[RGB]{255,240,225}{\strut{} ''}\colorbox[RGB]{255,244,233}{\strut{}I}\colorbox[RGB]{253,190,134}{\strut{} need}\colorbox[RGB]{253,216,179}{\strut{} to}\colorbox[RGB]{253,172,105}{\strut{} be}\colorbox[RGB]{229,90,12}{\strut{} alone}\colorbox[RGB]{250,137,56}{\strut{} for}\colorbox[RGB]{255,255,255}{\strut{} a}\colorbox[RGB]{253,164,94}{\strut{} while}\colorbox[RGB]{255,240,225}{\strut{}.''}\colorbox[RGB]{255,255,255}{\strut{} ''}\colorbox[RGB]{255,255,255}{\strut{}G}\colorbox[RGB]{255,255,255}{\strut{}EM}\colorbox[RGB]{255,255,255}{\strut{}MA}\colorbox[RGB]{255,255,255}{\strut{}:''}\colorbox[RGB]{255,255,255}{\strut{} ''}\colorbox[RGB]{255,244,233}{\strut{}Are}\colorbox[RGB]{255,240,225}{\strut{} you}\colorbox[RGB]{255,255,255}{\strut{} dumping}\colorbox[RGB]{255,255,255}{\strut{} me}\colorbox[RGB]{255,255,255}{\strut{}?''}\colorbox[RGB]{255,255,255}{\strut{} ''}\colorbox[RGB]{255,255,255}{\strut{}P}}}
\exampleline{{\unicodefont \colorbox[RGB]{255,255,255}{\strut{}''}\colorbox[RGB]{255,255,255}{\strut{} Hey}\colorbox[RGB]{255,255,255}{\strut{},}\colorbox[RGB]{255,255,255}{\strut{} Maria}\colorbox[RGB]{255,255,255}{\strut{}.''}\colorbox[RGB]{255,255,255}{\strut{} ''}\colorbox[RGB]{255,255,255}{\strut{}Leave}\colorbox[RGB]{254,232,209}{\strut{} me}\colorbox[RGB]{255,255,255}{\strut{} alone}\colorbox[RGB]{254,237,220}{\strut{}.''}\colorbox[RGB]{255,255,255}{\strut{} ''}\colorbox[RGB]{255,255,255}{\strut{}I}\colorbox[RGB]{253,199,149}{\strut{} need}\colorbox[RGB]{254,219,185}{\strut{} to}\colorbox[RGB]{253,206,160}{\strut{} be}\colorbox[RGB]{253,215,176}{\strut{} by}\colorbox[RGB]{233,95,15}{\strut{} myself}\colorbox[RGB]{251,139,59}{\strut{} for}\colorbox[RGB]{255,255,255}{\strut{} a}\colorbox[RGB]{252,146,69}{\strut{} bit}\colorbox[RGB]{254,228,202}{\strut{}.''}\colorbox[RGB]{254,236,218}{\strut{} ''}\colorbox[RGB]{255,255,255}{\strut{}H}\colorbox[RGB]{255,255,255}{\strut{}orm}\colorbox[RGB]{255,244,233}{\strut{}ones}\colorbox[RGB]{255,238,221}{\strut{}.''}\colorbox[RGB]{255,255,255}{\strut{} ''}\colorbox[RGB]{255,244,233}{\strut{}I}\colorbox[RGB]{255,255,255}{\strut{}-}\colorbox[RGB]{255,255,255}{\strut{}I}\colorbox[RGB]{255,255,255}{\strut{}-}\colorbox[RGB]{255,255,255}{\strut{}I}\colorbox[RGB]{255,255,255}{\strut{} got}\colorbox[RGB]{255,255,255}{\strut{} the}\colorbox[RGB]{255,255,255}{\strut{} job}\colorbox[RGB]{255,255,255}{\strut{}.''}\colorbox[RGB]{255,255,255}{\strut{} ''}}}
\exampleline{{\unicodefont \colorbox[RGB]{255,255,255}{\strut{}I}\colorbox[RGB]{255,255,255}{\strut{} know}\colorbox[RGB]{255,255,255}{\strut{}.''}\colorbox[RGB]{255,255,255}{\strut{} ''}\colorbox[RGB]{255,255,255}{\strut{}She}\colorbox[RGB]{255,255,255}{\strut{}'s}\colorbox[RGB]{255,255,255}{\strut{},}\colorbox[RGB]{255,255,255}{\strut{} um}\colorbox[RGB]{255,255,255}{\strut{}...}\colorbox[RGB]{255,255,255}{\strut{} she}\colorbox[RGB]{255,244,233}{\strut{} just}\colorbox[RGB]{253,176,111}{\strut{} needs}\colorbox[RGB]{254,228,201}{\strut{} to}\colorbox[RGB]{253,215,176}{\strut{} be}\colorbox[RGB]{254,236,218}{\strut{} on}\colorbox[RGB]{254,223,191}{\strut{} her}\colorbox[RGB]{236,99,18}{\strut{} own}\colorbox[RGB]{253,181,119}{\strut{} for}\colorbox[RGB]{255,255,255}{\strut{} a}\colorbox[RGB]{254,233,212}{\strut{} little}\colorbox[RGB]{253,160,88}{\strut{} while}\colorbox[RGB]{255,244,233}{\strut{}.''}\colorbox[RGB]{255,255,255}{\strut{} ''}\colorbox[RGB]{255,255,255}{\strut{}Jack}\colorbox[RGB]{255,255,255}{\strut{}?''}\colorbox[RGB]{255,255,255}{\strut{} ''}\colorbox[RGB]{255,255,255}{\strut{}Someone}\colorbox[RGB]{255,255,255}{\strut{} here}\colorbox[RGB]{255,255,255}{\strut{} would}}}
\end{featureexamples}

\begin{featureexamples}
\featurechip{1M}{781220} \textbf{Sadness}
\exampleline{{\unicodefont \colorbox[RGB]{253,185,126}{\strut{}.''}\colorbox[RGB]{255,255,255}{\strut{} ''}\colorbox[RGB]{253,190,134}{\strut{}Now}\colorbox[RGB]{253,179,116}{\strut{} they}\colorbox[RGB]{253,194,141}{\strut{} seem}\colorbox[RGB]{252,153,77}{\strut{} to}\colorbox[RGB]{253,164,94}{\strut{} be}\colorbox[RGB]{253,188,131}{\strut{} d}\colorbox[RGB]{253,174,108}{\strut{}renched}\colorbox[RGB]{253,158,85}{\strut{} in}\colorbox[RGB]{254,228,201}{\strut{} sorrow}\colorbox[RGB]{254,237,220}{\strut{}.''}\colorbox[RGB]{255,244,233}{\strut{} ''}\colorbox[RGB]{253,162,91}{\strut{}Are}\colorbox[RGB]{223,83,8}{\strut{} they}\colorbox[RGB]{255,242,229}{\strut{} nuts}\colorbox[RGB]{254,232,209}{\strut{}?''}\colorbox[RGB]{255,255,255}{\strut{} ''}\colorbox[RGB]{254,219,185}{\strut{}Think}\colorbox[RGB]{254,236,218}{\strut{} of}\colorbox[RGB]{254,219,185}{\strut{} those}\colorbox[RGB]{255,241,227}{\strut{} who}\colorbox[RGB]{254,237,220}{\strut{} are}\colorbox[RGB]{255,255,255}{\strut{} gonna}\colorbox[RGB]{255,255,255}{\strut{} marry}\colorbox[RGB]{255,255,255}{\strut{} them}\colorbox[RGB]{255,255,255}{\strut{}!}}}
\exampleline{{\unicodefont \colorbox[RGB]{255,255,255}{\strut{}ted}\colorbox[RGB]{255,255,255}{\strut{}.''''}\colorbox[RGB]{255,255,255}{\strut{} ''''}\colorbox[RGB]{253,187,129}{\strut{}'}\colorbox[RGB]{253,212,172}{\strut{}Boy}\colorbox[RGB]{255,255,255}{\strut{},'}\colorbox[RGB]{255,255,255}{\strut{} she}\colorbox[RGB]{255,255,255}{\strut{} said}\colorbox[RGB]{255,255,255}{\strut{} cour}\colorbox[RGB]{255,255,255}{\strut{}te}\colorbox[RGB]{255,255,255}{\strut{}ously}\colorbox[RGB]{255,255,255}{\strut{}...''}\colorbox[RGB]{253,192,137}{\strut{} '''}\colorbox[RGB]{251,143,64}{\strut{}Why}\colorbox[RGB]{254,219,185}{\strut{} are}\colorbox[RGB]{231,92,13}{\strut{} you}\colorbox[RGB]{255,238,221}{\strut{} crying}\colorbox[RGB]{254,229,204}{\strut{}?''}\colorbox[RGB]{253,192,137}{\strut{} '''}\colorbox[RGB]{255,244,233}{\strut{}''''}\colorbox[RGB]{255,255,255}{\strut{} ''\_''}\colorbox[RGB]{255,255,255}{\strut{} ''}\colorbox[RGB]{255,243,231}{\strut{}He}\colorbox[RGB]{255,255,255}{\strut{} can}\colorbox[RGB]{255,255,255}{\strut{} pick}\colorbox[RGB]{255,255,255}{\strut{} it}\colorbox[RGB]{255,255,255}{\strut{} up}\colorbox[RGB]{255,255,255}{\strut{} tomorrow}\colorbox[RGB]{255,255,255}{\strut{}.''}\colorbox[RGB]{255,255,255}{\strut{} }}}
\exampleline{{\unicodefont \colorbox[RGB]{255,255,255}{\strut{}G}\colorbox[RGB]{255,255,255}{\strut{}AS}\colorbox[RGB]{255,255,255}{\strut{}PS}\colorbox[RGB]{255,255,255}{\strut{})''}\colorbox[RGB]{255,255,255}{\strut{} ''}\colorbox[RGB]{255,255,255}{\strut{}Look}\colorbox[RGB]{255,255,255}{\strut{} at}\colorbox[RGB]{255,255,255}{\strut{} that}\colorbox[RGB]{255,255,255}{\strut{} child}\colorbox[RGB]{255,255,255}{\strut{}.''}\colorbox[RGB]{255,255,255}{\strut{} ''}\colorbox[RGB]{254,234,214}{\strut{}She}\colorbox[RGB]{255,255,255}{\strut{}'s}\colorbox[RGB]{255,243,231}{\strut{} so}\colorbox[RGB]{254,234,214}{\strut{} sad}\colorbox[RGB]{253,196,144}{\strut{}.''}\colorbox[RGB]{255,244,233}{\strut{} ''}\colorbox[RGB]{253,183,122}{\strut{} Is}\colorbox[RGB]{239,105,22}{\strut{} she}\colorbox[RGB]{255,240,225}{\strut{} poor}\colorbox[RGB]{255,239,223}{\strut{}?''}\colorbox[RGB]{255,244,233}{\strut{} ''}\colorbox[RGB]{254,233,211}{\strut{} She}\colorbox[RGB]{253,190,134}{\strut{}'s}\colorbox[RGB]{253,214,175}{\strut{} forgotten}\colorbox[RGB]{255,255,255}{\strut{}.''}\colorbox[RGB]{255,255,255}{\strut{} ''}\colorbox[RGB]{253,211,169}{\strut{}It}\colorbox[RGB]{253,211,169}{\strut{} just}\colorbox[RGB]{255,241,227}{\strut{} makes}\colorbox[RGB]{255,242,229}{\strut{} me}\colorbox[RGB]{254,226,199}{\strut{} wan}}}
\exampleline{{\unicodefont \colorbox[RGB]{255,255,255}{\strut{}.''}\colorbox[RGB]{255,255,255}{\strut{} ''}\colorbox[RGB]{255,255,255}{\strut{}Is}\colorbox[RGB]{255,255,255}{\strut{} she}\colorbox[RGB]{255,255,255}{\strut{} having}\colorbox[RGB]{255,255,255}{\strut{} the}\colorbox[RGB]{255,255,255}{\strut{} baby}\colorbox[RGB]{255,255,255}{\strut{}?''}\colorbox[RGB]{255,255,255}{\strut{} ''}\colorbox[RGB]{255,255,255}{\strut{}She}\colorbox[RGB]{255,255,255}{\strut{}'s}\colorbox[RGB]{255,255,255}{\strut{} mour}\colorbox[RGB]{255,255,255}{\strut{}ning}\colorbox[RGB]{254,232,209}{\strut{}.''}\colorbox[RGB]{255,255,255}{\strut{} ''}\colorbox[RGB]{253,167,98}{\strut{}She}\colorbox[RGB]{243,114,30}{\strut{}'s}\colorbox[RGB]{253,155,80}{\strut{} just}\colorbox[RGB]{255,255,255}{\strut{} lost}\colorbox[RGB]{255,241,227}{\strut{} her}\colorbox[RGB]{255,255,255}{\strut{} husband}\colorbox[RGB]{254,230,207}{\strut{}.''}\colorbox[RGB]{255,255,255}{\strut{} ''}\colorbox[RGB]{255,244,233}{\strut{}The}\colorbox[RGB]{255,243,230}{\strut{} master}\colorbox[RGB]{255,244,233}{\strut{} was}\colorbox[RGB]{254,233,212}{\strut{} here}\colorbox[RGB]{255,255,255}{\strut{} just}}}
\exampleline{{\unicodefont \colorbox[RGB]{255,255,255}{\strut{}sentations}\colorbox[RGB]{254,219,185}{\strut{},}\colorbox[RGB]{253,203,155}{\strut{} the}\colorbox[RGB]{253,211,169}{\strut{} drop}\colorbox[RGB]{253,218,182}{\strut{} of}\colorbox[RGB]{253,218,182}{\strut{} water}\colorbox[RGB]{253,206,160}{\strut{} is}\colorbox[RGB]{253,208,163}{\strut{} under}\colorbox[RGB]{253,170,101}{\strut{} the}\colorbox[RGB]{253,216,179}{\strut{} eye}\colorbox[RGB]{254,228,201}{\strut{},}\colorbox[RGB]{248,127,44}{\strut{} signaling}\colorbox[RGB]{254,223,191}{\strut{} that}\colorbox[RGB]{253,188,131}{\strut{} the}\colorbox[RGB]{253,196,144}{\strut{} face}\colorbox[RGB]{254,223,191}{\strut{}⏎}\colorbox[RGB]{250,136,55}{\strut{}is}\colorbox[RGB]{253,201,152}{\strut{} crying}\colorbox[RGB]{255,255,255}{\strut{}.}\colorbox[RGB]{255,255,255}{\strut{} There}\colorbox[RGB]{255,244,233}{\strut{} is}\colorbox[RGB]{255,255,255}{\strut{} not}\colorbox[RGB]{255,255,255}{\strut{} a}\colorbox[RGB]{255,255,255}{\strut{} singl}}}
\end{featureexamples}

The fact that both features contribute to the final output indicates that the model has partially predicted a sentiment from John's statement (the second feature) but will do more downstream processing on the content of his statement (as represented by the first feature) as well.

In comparison, the features with the highest average activation on the context are less useful for understanding how the model actually predicts the next token in this case. Several features fire strongly on the start-of-sequence token. If we ignore those, the top feature is the same as given by attributions, but the second and third features are less abstract: \featurechip{1M}{504227} fires on “be” in “want to be” and variants, and \featurechip{1M}{594453} fires on the word “alone”.

\begin{featureexamples}
\featurechip{1M}{504227} \textbf{“Be” in “want to be”, etc.}
\exampleline{{\unicodefont \colorbox[RGB]{255,255,255}{\strut{}}\colorbox[RGB]{255,255,255}{\strut{}''}\colorbox[RGB]{255,255,255}{\strut{}He}\colorbox[RGB]{255,255,255}{\strut{} wants}\colorbox[RGB]{255,255,255}{\strut{} to}\colorbox[RGB]{223,83,8}{\strut{} be}\colorbox[RGB]{253,216,179}{\strut{} a}\colorbox[RGB]{255,255,255}{\strut{} doctor}\colorbox[RGB]{255,255,255}{\strut{}.''}\colorbox[RGB]{255,255,255}{\strut{} ''}\colorbox[RGB]{255,255,255}{\strut{}Tell}\colorbox[RGB]{255,255,255}{\strut{} him}\colorbox[RGB]{255,255,255}{\strut{} it}\colorbox[RGB]{255,255,255}{\strut{}'s}\colorbox[RGB]{255,255,255}{\strut{} educational}\colorbox[RGB]{255,255,255}{\strut{}.''}\colorbox[RGB]{255,255,255}{\strut{} ''}\colorbox[RGB]{255,255,255}{\strut{}There}\colorbox[RGB]{255,255,255}{\strut{}'s}\colorbox[RGB]{255,255,255}{\strut{} body}\colorbox[RGB]{255,255,255}{\strut{} parts}\colorbox[RGB]{255,255,255}{\strut{} all}\colorbox[RGB]{255,255,255}{\strut{} over}\colorbox[RGB]{255,255,255}{\strut{} this}\colorbox[RGB]{255,255,255}{\strut{} movie}\colorbox[RGB]{255,255,255}{\strut{}.''}}}
\exampleline{{\unicodefont \colorbox[RGB]{255,255,255}{\strut{}}\colorbox[RGB]{255,255,255}{\strut{},}\colorbox[RGB]{255,255,255}{\strut{} he}\colorbox[RGB]{255,255,255}{\strut{} wanted}\colorbox[RGB]{255,255,255}{\strut{} to}\colorbox[RGB]{229,90,12}{\strut{} be}\colorbox[RGB]{253,206,160}{\strut{} a}\colorbox[RGB]{255,255,255}{\strut{} hero}\colorbox[RGB]{255,255,255}{\strut{}.''}\colorbox[RGB]{255,255,255}{\strut{} ''}\colorbox[RGB]{255,255,255}{\strut{}I}\colorbox[RGB]{255,255,255}{\strut{} told}\colorbox[RGB]{255,255,255}{\strut{} him}\colorbox[RGB]{255,255,255}{\strut{} he}\colorbox[RGB]{255,255,255}{\strut{} was}\colorbox[RGB]{255,255,255}{\strut{} gonna}\colorbox[RGB]{255,255,255}{\strut{} get}\colorbox[RGB]{255,255,255}{\strut{} us}\colorbox[RGB]{255,255,255}{\strut{} both}\colorbox[RGB]{255,255,255}{\strut{} killed}\colorbox[RGB]{255,255,255}{\strut{}.''}\colorbox[RGB]{255,255,255}{\strut{} ''}\colorbox[RGB]{255,255,255}{\strut{}But}\colorbox[RGB]{255,255,255}{\strut{} he}\colorbox[RGB]{255,255,255}{\strut{} only}\colorbox[RGB]{255,255,255}{\strut{} got}}}
\exampleline{{\unicodefont \colorbox[RGB]{255,255,255}{\strut{}}\colorbox[RGB]{255,255,255}{\strut{}all}\colorbox[RGB]{255,255,255}{\strut{}.''}\colorbox[RGB]{255,255,255}{\strut{} ''}\colorbox[RGB]{255,255,255}{\strut{}They}\colorbox[RGB]{255,255,255}{\strut{} all}\colorbox[RGB]{255,255,255}{\strut{} want}\colorbox[RGB]{255,255,255}{\strut{} to}\colorbox[RGB]{231,92,13}{\strut{} be}\colorbox[RGB]{255,244,233}{\strut{} Miss}\colorbox[RGB]{255,255,255}{\strut{} Hope}\colorbox[RGB]{255,255,255}{\strut{} Springs}\colorbox[RGB]{255,255,255}{\strut{}.''}\colorbox[RGB]{255,255,255}{\strut{} ''}\colorbox[RGB]{255,255,255}{\strut{}Well}\colorbox[RGB]{255,255,255}{\strut{} I}\colorbox[RGB]{255,255,255}{\strut{}'m}\colorbox[RGB]{255,255,255}{\strut{} not}\colorbox[RGB]{255,255,255}{\strut{} competitive}\colorbox[RGB]{255,255,255}{\strut{}.''}\colorbox[RGB]{255,255,255}{\strut{} ''}\colorbox[RGB]{255,255,255}{\strut{}Well}\colorbox[RGB]{255,255,255}{\strut{} then}\colorbox[RGB]{255,255,255}{\strut{} you}\colorbox[RGB]{255,255,255}{\strut{}'ll}\colorbox[RGB]{255,255,255}{\strut{} never}\colorbox[RGB]{255,255,255}{\strut{} be}\colorbox[RGB]{255,255,255}{\strut{} }}}
\exampleline{{\unicodefont \colorbox[RGB]{255,255,255}{\strut{}}\colorbox[RGB]{255,255,255}{\strut{}you}\colorbox[RGB]{255,255,255}{\strut{} know}\colorbox[RGB]{255,255,255}{\strut{} I}\colorbox[RGB]{255,255,255}{\strut{} want}\colorbox[RGB]{255,255,255}{\strut{} to}\colorbox[RGB]{234,97,16}{\strut{} be}\colorbox[RGB]{254,226,199}{\strut{} dry}\colorbox[RGB]{255,255,255}{\strut{} what}\colorbox[RGB]{255,255,255}{\strut{}''}\colorbox[RGB]{255,255,255}{\strut{} ''}\colorbox[RGB]{255,255,255}{\strut{}Know}\colorbox[RGB]{255,255,255}{\strut{} me}\colorbox[RGB]{255,255,255}{\strut{} to}\colorbox[RGB]{255,255,255}{\strut{} smell}\colorbox[RGB]{255,255,255}{\strut{} the}\colorbox[RGB]{255,255,255}{\strut{} coal}\colorbox[RGB]{255,255,255}{\strut{} gas}\colorbox[RGB]{255,255,255}{\strut{} flavor}\colorbox[RGB]{255,255,255}{\strut{}''}\colorbox[RGB]{255,255,255}{\strut{} ''}\colorbox[RGB]{255,255,255}{\strut{}I}\colorbox[RGB]{255,255,255}{\strut{} have}\colorbox[RGB]{255,255,255}{\strut{} never}\colorbox[RGB]{255,255,255}{\strut{} open}\colorbox[RGB]{255,255,255}{\strut{}ned}\colorbox[RGB]{255,255,255}{\strut{} coal}}}
\exampleline{{\unicodefont \colorbox[RGB]{255,255,255}{\strut{}}\colorbox[RGB]{255,255,255}{\strut{}she}\colorbox[RGB]{255,255,255}{\strut{} just}\colorbox[RGB]{255,255,255}{\strut{} wanted}\colorbox[RGB]{255,255,255}{\strut{} to}\colorbox[RGB]{237,102,20}{\strut{} be}\colorbox[RGB]{255,244,233}{\strut{} loved}\colorbox[RGB]{255,255,255}{\strut{}.''}\colorbox[RGB]{255,255,255}{\strut{} ''}\colorbox[RGB]{255,255,255}{\strut{}Don}\colorbox[RGB]{255,255,255}{\strut{}'t}\colorbox[RGB]{255,255,255}{\strut{} we}\colorbox[RGB]{255,255,255}{\strut{} all}\colorbox[RGB]{255,255,255}{\strut{}?''}\colorbox[RGB]{255,255,255}{\strut{} ''}\colorbox[RGB]{255,255,255}{\strut{}I}\colorbox[RGB]{255,255,255}{\strut{} want}\colorbox[RGB]{255,255,255}{\strut{} all}\colorbox[RGB]{255,255,255}{\strut{} of}\colorbox[RGB]{255,255,255}{\strut{} De}\colorbox[RGB]{255,255,255}{\strut{}bbie}\colorbox[RGB]{255,255,255}{\strut{} Flo}\colorbox[RGB]{255,255,255}{\strut{}res}\colorbox[RGB]{255,255,255}{\strut{}'}\colorbox[RGB]{255,255,255}{\strut{} credit}}}
\end{featureexamples}

\begin{featureexamples}
\featurechip{1M}{594453} \textbf{“alone”}
\exampleline{{\unicodefont \colorbox[RGB]{255,255,255}{\strut{}the}\colorbox[RGB]{255,255,255}{\strut{} bottle}\colorbox[RGB]{255,255,255}{\strut{} that}\colorbox[RGB]{255,255,255}{\strut{} you}\colorbox[RGB]{255,255,255}{\strut{} drink}\colorbox[RGB]{255,255,255}{\strut{}''}\colorbox[RGB]{255,255,255}{\strut{} ''}\colorbox[RGB]{255,255,255}{\strut{}And}\colorbox[RGB]{255,255,255}{\strut{} times}\colorbox[RGB]{255,255,255}{\strut{} when}\colorbox[RGB]{255,255,255}{\strut{} you}\colorbox[RGB]{255,255,255}{\strut{}'re}\colorbox[RGB]{223,83,8}{\strut{} alone}\colorbox[RGB]{255,255,255}{\strut{}''}\colorbox[RGB]{253,205,158}{\strut{} ''}\colorbox[RGB]{255,255,255}{\strut{}Well}\colorbox[RGB]{255,255,255}{\strut{},}\colorbox[RGB]{255,255,255}{\strut{} all}\colorbox[RGB]{255,255,255}{\strut{} you}\colorbox[RGB]{255,255,255}{\strut{} do}\colorbox[RGB]{255,255,255}{\strut{} is}\colorbox[RGB]{255,255,255}{\strut{} think}\colorbox[RGB]{255,255,255}{\strut{}''}\colorbox[RGB]{255,255,255}{\strut{} ''}\colorbox[RGB]{255,255,255}{\strut{}I}\colorbox[RGB]{255,255,255}{\strut{}'m}\colorbox[RGB]{255,255,255}{\strut{} a}\colorbox[RGB]{255,255,255}{\strut{} cowboy}\colorbox[RGB]{255,255,255}{\strut{}''}\colorbox[RGB]{255,255,255}{\strut{} ''}\colorbox[RGB]{255,255,255}{\strut{}On}}}
\exampleline{{\unicodefont \colorbox[RGB]{255,255,255}{\strut{}uned}\colorbox[RGB]{255,255,255}{\strut{} out}\colorbox[RGB]{255,255,255}{\strut{}''}\colorbox[RGB]{255,255,255}{\strut{} ''}\colorbox[RGB]{255,255,255}{\strut{}A}\colorbox[RGB]{255,255,255}{\strut{} bad}\colorbox[RGB]{255,255,255}{\strut{} time}\colorbox[RGB]{255,255,255}{\strut{},}\colorbox[RGB]{255,255,255}{\strut{} nothing}\colorbox[RGB]{255,255,255}{\strut{} could}\colorbox[RGB]{255,255,255}{\strut{} save}\colorbox[RGB]{255,255,255}{\strut{} him}\colorbox[RGB]{255,255,255}{\strut{}''}\colorbox[RGB]{255,255,255}{\strut{} ''}\colorbox[RGB]{239,105,22}{\strut{}Al}\colorbox[RGB]{224,84,8}{\strut{}one}\colorbox[RGB]{253,205,158}{\strut{} in}\colorbox[RGB]{255,244,233}{\strut{} a}\colorbox[RGB]{255,255,255}{\strut{} corridor}\colorbox[RGB]{255,255,255}{\strut{},}\colorbox[RGB]{255,255,255}{\strut{} waiting}\colorbox[RGB]{255,255,255}{\strut{},}\colorbox[RGB]{255,255,255}{\strut{} locked}\colorbox[RGB]{255,255,255}{\strut{} out}\colorbox[RGB]{255,255,255}{\strut{}.''}\colorbox[RGB]{255,255,255}{\strut{} ''}\colorbox[RGB]{255,255,255}{\strut{}He}\colorbox[RGB]{255,255,255}{\strut{} got}\colorbox[RGB]{255,255,255}{\strut{} up}\colorbox[RGB]{255,255,255}{\strut{} o}}}
\exampleline{{\unicodefont \colorbox[RGB]{255,255,255}{\strut{} inside}\colorbox[RGB]{255,255,255}{\strut{}''}\colorbox[RGB]{255,255,255}{\strut{} ''\#}\colorbox[RGB]{255,255,255}{\strut{} I}\colorbox[RGB]{255,255,255}{\strut{} lay}\colorbox[RGB]{255,255,255}{\strut{} in}\colorbox[RGB]{255,255,255}{\strut{} tears}\colorbox[RGB]{255,255,255}{\strut{} in}\colorbox[RGB]{255,255,255}{\strut{} bed}\colorbox[RGB]{255,255,255}{\strut{} all}\colorbox[RGB]{255,255,255}{\strut{} night}\colorbox[RGB]{255,255,255}{\strut{}''}\colorbox[RGB]{255,255,255}{\strut{} ''\#}\colorbox[RGB]{231,92,13}{\strut{} Al}\colorbox[RGB]{224,84,8}{\strut{}one}\colorbox[RGB]{255,255,255}{\strut{} without}\colorbox[RGB]{255,255,255}{\strut{} you}\colorbox[RGB]{255,255,255}{\strut{} by}\colorbox[RGB]{255,255,255}{\strut{} my}\colorbox[RGB]{255,255,255}{\strut{} side}\colorbox[RGB]{255,255,255}{\strut{}''}\colorbox[RGB]{255,255,255}{\strut{} ''\#}\colorbox[RGB]{255,255,255}{\strut{} But}\colorbox[RGB]{255,255,255}{\strut{} if}\colorbox[RGB]{255,255,255}{\strut{} you}\colorbox[RGB]{255,255,255}{\strut{} loved}\colorbox[RGB]{255,255,255}{\strut{} me}\colorbox[RGB]{255,255,255}{\strut{}''}\colorbox[RGB]{255,255,255}{\strut{} ''}}}
\exampleline{{\unicodefont \colorbox[RGB]{255,255,255}{\strut{}oh}\colorbox[RGB]{255,255,255}{\strut{},}\colorbox[RGB]{255,255,255}{\strut{} oh}\colorbox[RGB]{255,255,255}{\strut{},}\colorbox[RGB]{255,255,255}{\strut{} many}\colorbox[RGB]{255,255,255}{\strut{},}\colorbox[RGB]{255,255,255}{\strut{} many}\colorbox[RGB]{255,255,255}{\strut{} nights}\colorbox[RGB]{255,255,255}{\strut{} roll}\colorbox[RGB]{255,255,255}{\strut{} by}\colorbox[RGB]{255,255,255}{\strut{} ¶}\colorbox[RGB]{255,255,255}{\strut{}''}\colorbox[RGB]{255,255,255}{\strut{} ''}\colorbox[RGB]{255,255,255}{\strut{}¶}\colorbox[RGB]{255,255,255}{\strut{} I}\colorbox[RGB]{255,255,255}{\strut{} sit}\colorbox[RGB]{224,84,8}{\strut{} alone}\colorbox[RGB]{255,244,233}{\strut{} at}\colorbox[RGB]{255,255,255}{\strut{} home}\colorbox[RGB]{255,255,255}{\strut{} and}\colorbox[RGB]{255,255,255}{\strut{} cry}\colorbox[RGB]{255,255,255}{\strut{} ¶}\colorbox[RGB]{255,255,255}{\strut{}''}\colorbox[RGB]{255,255,255}{\strut{} ''}\colorbox[RGB]{255,255,255}{\strut{}¶}\colorbox[RGB]{255,255,255}{\strut{} over}\colorbox[RGB]{255,255,255}{\strut{} you}\colorbox[RGB]{255,255,255}{\strut{} ¶}\colorbox[RGB]{255,255,255}{\strut{}''}\colorbox[RGB]{255,255,255}{\strut{} ''}}}
\exampleline{{\unicodefont \colorbox[RGB]{255,255,255}{\strut{} and}\colorbox[RGB]{255,255,255}{\strut{} water}\colorbox[RGB]{255,255,255}{\strut{}falls}\colorbox[RGB]{255,255,255}{\strut{} \textbackslash{}xe2\textbackslash{}x99}\colorbox[RGB]{255,255,255}{\strut{}\textbackslash{}xaa}\colorbox[RGB]{255,255,255}{\strut{}''}\colorbox[RGB]{255,255,255}{\strut{} ''}\colorbox[RGB]{255,255,255}{\strut{}♪}\colorbox[RGB]{255,255,255}{\strut{} Home}\colorbox[RGB]{255,255,255}{\strut{} is}\colorbox[RGB]{255,255,255}{\strut{} when}\colorbox[RGB]{255,255,255}{\strut{} I}\colorbox[RGB]{255,255,255}{\strut{}'m}\colorbox[RGB]{224,84,8}{\strut{} alone}\colorbox[RGB]{255,243,230}{\strut{} with}\colorbox[RGB]{255,255,255}{\strut{} you}\colorbox[RGB]{255,255,255}{\strut{}.}\colorbox[RGB]{255,255,255}{\strut{} \textbackslash{}xe2\textbackslash{}x99}\colorbox[RGB]{255,255,255}{\strut{}\textbackslash{}xaa}\colorbox[RGB]{255,255,255}{\strut{}''}\colorbox[RGB]{255,255,255}{\strut{}}\colorbox[RGB]{255,255,255}{\strut{}''}\colorbox[RGB]{255,255,255}{\strut{}Cur}\colorbox[RGB]{255,255,255}{\strut{}tain}\colorbox[RGB]{255,255,255}{\strut{}-}\colorbox[RGB]{255,255,255}{\strut{}up}\colorbox[RGB]{255,255,255}{\strut{} in}\colorbox[RGB]{255,255,255}{\strut{} 5}\colorbox[RGB]{255,255,255}{\strut{} minute}}}
\end{featureexamples}

\begin{figure}[!htp]
    \centering
    \includegraphics[width=0.9\textwidth,height=0.7\textheight,keepaspectratio]{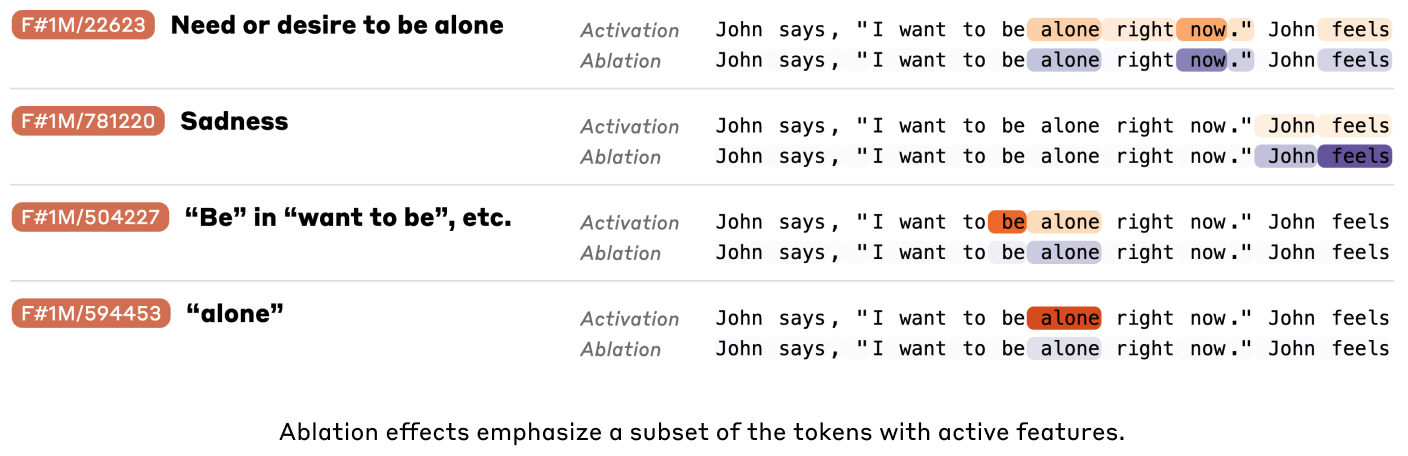}
    \label{fig:gdoc_31}
\end{figure}

\textit{}

\subsection{Example: Multi-Step Inference}\label{sec:computational-multistep}

We now investigate an incomplete prompt requiring a longer chain of inferences:

\begin{quote}
Fact: The capital of the state where Kobe Bryant played basketball is\\
(completion: Sacramento − Albany)
\end{quote}

To continue this text, the model must identify where Kobe Bryant played basketball, what state that place was in, and then the capital of that state.

We compute attributions and ablation effects for the completion “Sacramento” (the correct answer, which Sonnet knows) with respect to the baseline “Albany” (Sonnet's most likely alternative single-token capital completion). The top five features by ablation effect (which match those by attribution effect, modulo reordering) are:

\begin{itemize}
    \item \featurechip{1M}{391411} -- A Kobe Bryant feature
    \item \featurechip{1M}{81163} -- A California feature, which notably activates the most strongly on text after “California” is mentioned, rather than “California” itself
    \item \featurechip{1M}{201767} -- A “capital” feature
    \item \featurechip{1M}{980087} -- A Los Angeles feature
    \item \featurechip{1M}{447200} -- A Los Angeles Lakers feature
\end{itemize}

\begin{featureexamples}
\featurechip{1M}{391411} \textbf{Kobe Bryant}
\exampleline{{\unicodefont \colorbox[RGB]{253,216,179}{\strut{}tartup}\colorbox[RGB]{254,234,214}{\strut{} work}\colorbox[RGB]{253,206,160}{\strut{} eth}\colorbox[RGB]{254,228,201}{\strut{}ic}\colorbox[RGB]{255,255,255}{\strut{} -}\colorbox[RGB]{255,255,255}{\strut{} p}\colorbox[RGB]{255,255,255}{\strut{}jg}\colorbox[RGB]{255,255,255}{\strut{}⏎}\colorbox[RGB]{255,255,255}{\strut{}https}\colorbox[RGB]{255,244,233}{\strut{}://}\colorbox[RGB]{251,143,64}{\strut{}www}\colorbox[RGB]{255,255,255}{\strut{}.}\colorbox[RGB]{255,255,255}{\strut{}business}\colorbox[RGB]{255,255,255}{\strut{}ins}\colorbox[RGB]{255,255,255}{\strut{}ider}\colorbox[RGB]{254,237,220}{\strut{}.}\colorbox[RGB]{254,235,216}{\strut{}com}\colorbox[RGB]{255,255,255}{\strut{}/}\colorbox[RGB]{254,224,194}{\strut{}k}\colorbox[RGB]{223,83,8}{\strut{}obe}\colorbox[RGB]{253,212,172}{\strut{}-}\colorbox[RGB]{253,205,158}{\strut{}bry}\colorbox[RGB]{253,203,155}{\strut{}ant}\colorbox[RGB]{254,219,185}{\strut{}-}\colorbox[RGB]{253,203,155}{\strut{}woke}\colorbox[RGB]{253,205,158}{\strut{}-}\colorbox[RGB]{253,188,131}{\strut{}up}\colorbox[RGB]{253,185,126}{\strut{}-}\colorbox[RGB]{253,215,176}{\strut{}at}\colorbox[RGB]{253,194,141}{\strut{}-}\colorbox[RGB]{254,235,216}{\strut{}4}\colorbox[RGB]{254,233,212}{\strut{}-}\colorbox[RGB]{253,215,176}{\strut{}am}\colorbox[RGB]{253,187,129}{\strut{}-}\colorbox[RGB]{253,211,169}{\strut{}to}\colorbox[RGB]{253,183,122}{\strut{}-}\colorbox[RGB]{253,215,176}{\strut{}practice}\colorbox[RGB]{253,218,182}{\strut{}-}\colorbox[RGB]{254,221,187}{\strut{}before}\colorbox[RGB]{254,228,201}{\strut{}-}}}
\exampleline{{\unicodefont \colorbox[RGB]{255,255,255}{\strut{}⏎}\colorbox[RGB]{255,255,255}{\strut{}http}\colorbox[RGB]{255,255,255}{\strut{}://}\colorbox[RGB]{253,188,131}{\strut{}www}\colorbox[RGB]{255,255,255}{\strut{}.}\colorbox[RGB]{255,255,255}{\strut{}van}\colorbox[RGB]{255,255,255}{\strut{}ity}\colorbox[RGB]{255,255,255}{\strut{}fair}\colorbox[RGB]{253,218,182}{\strut{}.}\colorbox[RGB]{255,239,223}{\strut{}com}\colorbox[RGB]{255,255,255}{\strut{}/}\colorbox[RGB]{255,255,255}{\strut{}news}\colorbox[RGB]{255,255,255}{\strut{}/}\colorbox[RGB]{254,225,196}{\strut{}2016}\colorbox[RGB]{254,232,209}{\strut{}/}\colorbox[RGB]{255,255,255}{\strut{}04}\colorbox[RGB]{255,244,233}{\strut{}/}\colorbox[RGB]{254,232,209}{\strut{}k}\colorbox[RGB]{238,104,21}{\strut{}obe}\colorbox[RGB]{253,216,179}{\strut{}-}\colorbox[RGB]{253,181,119}{\strut{}bry}\colorbox[RGB]{253,192,137}{\strut{}ant}\colorbox[RGB]{253,197,145}{\strut{}-}\colorbox[RGB]{255,255,255}{\strut{}sil}\colorbox[RGB]{255,255,255}{\strut{}icon}\colorbox[RGB]{255,242,229}{\strut{}-}\colorbox[RGB]{255,255,255}{\strut{}val}\colorbox[RGB]{255,255,255}{\strut{}ley}\colorbox[RGB]{254,235,216}{\strut{}-}\colorbox[RGB]{255,255,255}{\strut{}tech}\colorbox[RGB]{254,230,207}{\strut{}-}\colorbox[RGB]{254,237,220}{\strut{}bro}\colorbox[RGB]{255,255,255}{\strut{}⏎}\colorbox[RGB]{255,255,255}{\strut{}======}\colorbox[RGB]{255,255,255}{\strut{}⏎}\colorbox[RGB]{255,255,255}{\strut{}n}\colorbox[RGB]{255,255,255}{\strut{}ibs}\colorbox[RGB]{255,255,255}{\strut{}⏎}\colorbox[RGB]{255,255,255}{\strut{}Next}\colorbox[RGB]{255,255,255}{\strut{} up}\colorbox[RGB]{255,255,255}{\strut{}:}}}
\exampleline{{\unicodefont \colorbox[RGB]{255,255,255}{\strut{}ugh}\colorbox[RGB]{255,255,255}{\strut{} media}\colorbox[RGB]{255,255,255}{\strut{} interviews}\colorbox[RGB]{255,255,255}{\strut{} you}\colorbox[RGB]{255,255,255}{\strut{} can}\colorbox[RGB]{255,255,255}{\strut{} piece}\colorbox[RGB]{255,255,255}{\strut{} together}\colorbox[RGB]{255,255,255}{\strut{} that}\colorbox[RGB]{255,255,255}{\strut{} K}\colorbox[RGB]{247,125,42}{\strut{}obe}\colorbox[RGB]{253,209,166}{\strut{} Bryant}\colorbox[RGB]{254,235,216}{\strut{} was}\colorbox[RGB]{255,242,229}{\strut{} one}\colorbox[RGB]{255,255,255}{\strut{} of}\colorbox[RGB]{254,226,199}{\strut{}⏎}\colorbox[RGB]{255,255,255}{\strut{}his}\colorbox[RGB]{255,255,255}{\strut{} clients}\colorbox[RGB]{254,221,189}{\strut{}.}\colorbox[RGB]{255,255,255}{\strut{}⏎}\colorbox[RGB]{255,255,255}{\strut{}⏎}\colorbox[RGB]{255,255,255}{\strut{}------}\colorbox[RGB]{255,255,255}{\strut{}⏎}\colorbox[RGB]{255,255,255}{\strut{}ame}\colorbox[RGB]{255,255,255}{\strut{}li}\colorbox[RGB]{255,255,255}{\strut{}us}\colorbox[RGB]{255,255,255}{\strut{}⏎}\colorbox[RGB]{255,255,255}{\strut{}Ar}}}
\exampleline{{\unicodefont \colorbox[RGB]{255,255,255}{\strut{}----}\colorbox[RGB]{255,255,255}{\strut{}⏎}\colorbox[RGB]{255,255,255}{\strut{}b}\colorbox[RGB]{255,255,255}{\strut{}inki}\colorbox[RGB]{255,255,255}{\strut{}89}\colorbox[RGB]{255,255,255}{\strut{}⏎}\colorbox[RGB]{255,255,255}{\strut{}Cry}\colorbox[RGB]{255,255,255}{\strut{}stal}\colorbox[RGB]{255,255,255}{\strut{} is}\colorbox[RGB]{255,255,255}{\strut{} so}\colorbox[RGB]{255,255,255}{\strut{} great}\colorbox[RGB]{255,255,255}{\strut{} to}\colorbox[RGB]{255,255,255}{\strut{} use}\colorbox[RGB]{255,255,255}{\strut{}.}\colorbox[RGB]{255,255,255}{\strut{}⏎}\colorbox[RGB]{255,255,255}{\strut{}⏎}\colorbox[RGB]{255,255,255}{\strut{}}\colorbox[RGB]{255,255,255}{\strut{}⏎}\colorbox[RGB]{255,255,255}{\strut{}K}\colorbox[RGB]{253,155,80}{\strut{}obe}\colorbox[RGB]{247,125,42}{\strut{} Bryant}\colorbox[RGB]{252,145,67}{\strut{} Is}\colorbox[RGB]{253,164,94}{\strut{} Ob}\colorbox[RGB]{253,201,152}{\strut{}sessed}\colorbox[RGB]{253,167,98}{\strut{} with}\colorbox[RGB]{253,179,116}{\strut{} Bec}\colorbox[RGB]{253,196,144}{\strut{}oming}\colorbox[RGB]{253,181,119}{\strut{} a}\colorbox[RGB]{254,233,212}{\strut{} Tech}\colorbox[RGB]{253,174,108}{\strut{} Bro}\colorbox[RGB]{255,255,255}{\strut{} -}\colorbox[RGB]{255,255,255}{\strut{} sch}\colorbox[RGB]{255,255,255}{\strut{}iang}\colorbox[RGB]{255,255,255}{\strut{}⏎}}}
\exampleline{{\unicodefont \colorbox[RGB]{255,255,255}{\strut{}th}\colorbox[RGB]{255,255,255}{\strut{}ic}\colorbox[RGB]{255,255,255}{\strut{} collide}\colorbox[RGB]{255,255,255}{\strut{} you}\colorbox[RGB]{255,255,255}{\strut{} get}\colorbox[RGB]{255,255,255}{\strut{} people}\colorbox[RGB]{255,255,255}{\strut{} like}\colorbox[RGB]{255,255,255}{\strut{} Michael}\colorbox[RGB]{255,255,255}{\strut{} Jordan}\colorbox[RGB]{255,255,255}{\strut{},}\colorbox[RGB]{255,255,255}{\strut{} K}\colorbox[RGB]{248,127,44}{\strut{}obe}\colorbox[RGB]{253,205,158}{\strut{} Bryant}\colorbox[RGB]{255,243,230}{\strut{},}\colorbox[RGB]{255,244,233}{\strut{} and}\colorbox[RGB]{255,255,255}{\strut{} Le}\colorbox[RGB]{255,255,255}{\strut{}Bron}\colorbox[RGB]{255,255,255}{\strut{}⏎}\colorbox[RGB]{255,255,255}{\strut{}James}\colorbox[RGB]{255,255,255}{\strut{}.}\colorbox[RGB]{255,255,255}{\strut{} Without}\colorbox[RGB]{255,255,255}{\strut{} a}\colorbox[RGB]{255,255,255}{\strut{} work}\colorbox[RGB]{255,244,233}{\strut{} eth}\colorbox[RGB]{255,255,255}{\strut{}ic}\colorbox[RGB]{255,255,255}{\strut{} th}}}
\end{featureexamples}

\begin{featureexamples}
\featurechip{1M}{81163} \textbf{California}
\exampleline{{\unicodefont \colorbox[RGB]{255,255,255}{\strut{}rom}\colorbox[RGB]{255,255,255}{\strut{} disasters}\colorbox[RGB]{255,255,255}{\strut{}?}\colorbox[RGB]{255,255,255}{\strut{}⏎}\colorbox[RGB]{255,255,255}{\strut{}⏎}\colorbox[RGB]{254,228,201}{\strut{}California}\colorbox[RGB]{252,151,74}{\strut{} -}\colorbox[RGB]{253,162,91}{\strut{} earth}\colorbox[RGB]{253,174,108}{\strut{}quakes}\colorbox[RGB]{242,111,28}{\strut{},}\colorbox[RGB]{253,212,172}{\strut{} mud}\colorbox[RGB]{253,197,145}{\strut{}slides}\colorbox[RGB]{223,83,8}{\strut{},}\colorbox[RGB]{253,196,144}{\strut{} wild}\colorbox[RGB]{254,223,191}{\strut{}fires}\colorbox[RGB]{224,84,8}{\strut{},}\colorbox[RGB]{247,125,42}{\strut{} torrent}\colorbox[RGB]{250,133,52}{\strut{}ial}\colorbox[RGB]{252,146,69}{\strut{} rains}\colorbox[RGB]{234,97,16}{\strut{},}\colorbox[RGB]{253,199,149}{\strut{} rip}\colorbox[RGB]{254,235,216}{\strut{}⏎}\colorbox[RGB]{253,203,155}{\strut{}current}\colorbox[RGB]{253,187,129}{\strut{}s}\colorbox[RGB]{245,119,36}{\strut{},}\colorbox[RGB]{250,133,52}{\strut{} and}\colorbox[RGB]{253,179,116}{\strut{} eve}}}
\exampleline{{\unicodefont \colorbox[RGB]{253,181,119}{\strut{}y}\colorbox[RGB]{253,181,119}{\strut{} rate}\colorbox[RGB]{253,214,175}{\strut{} in}\colorbox[RGB]{255,255,255}{\strut{} the}\colorbox[RGB]{255,255,255}{\strut{} United}\colorbox[RGB]{255,255,255}{\strut{}⏎}\colorbox[RGB]{254,228,201}{\strut{}States}\colorbox[RGB]{253,166,97}{\strut{},}\colorbox[RGB]{253,177,113}{\strut{} even}\colorbox[RGB]{253,160,88}{\strut{} though}\colorbox[RGB]{253,176,111}{\strut{} it}\colorbox[RGB]{253,162,91}{\strut{}'s}\colorbox[RGB]{253,179,116}{\strut{} home}\colorbox[RGB]{224,84,8}{\strut{} to}\colorbox[RGB]{254,219,185}{\strut{} Silicon}\colorbox[RGB]{255,244,233}{\strut{} Valley}\colorbox[RGB]{254,221,189}{\strut{}.}\colorbox[RGB]{253,212,172}{\strut{} I}\colorbox[RGB]{253,188,131}{\strut{} see}\colorbox[RGB]{253,196,144}{\strut{} my}\colorbox[RGB]{253,209,166}{\strut{} rich}\colorbox[RGB]{255,238,221}{\strut{} industry}\colorbox[RGB]{255,244,233}{\strut{} doing}\colorbox[RGB]{255,255,255}{\strut{}⏎}\colorbox[RGB]{255,244,233}{\strut{}noth}}}
\exampleline{{\unicodefont \colorbox[RGB]{255,255,255}{\strut{}pdx}\colorbox[RGB]{255,255,255}{\strut{}⏎}\colorbox[RGB]{255,255,255}{\strut{}And}\colorbox[RGB]{255,255,255}{\strut{} if}\colorbox[RGB]{255,255,255}{\strut{} everyone}\colorbox[RGB]{255,255,255}{\strut{} im}\colorbox[RGB]{255,255,255}{\strut{}itated}\colorbox[RGB]{253,206,160}{\strut{} California}\colorbox[RGB]{250,137,56}{\strut{}'s}\colorbox[RGB]{255,255,255}{\strut{} approach}\colorbox[RGB]{228,88,11}{\strut{} to}\colorbox[RGB]{253,187,129}{\strut{} primary}\colorbox[RGB]{253,218,182}{\strut{} education}\colorbox[RGB]{255,255,255}{\strut{},}\colorbox[RGB]{255,255,255}{\strut{} perhaps}\colorbox[RGB]{255,255,255}{\strut{}⏎}\colorbox[RGB]{255,255,255}{\strut{}CA}\colorbox[RGB]{255,255,255}{\strut{} wouldn}\colorbox[RGB]{255,243,230}{\strut{}'t}\colorbox[RGB]{253,194,141}{\strut{} rank}\colorbox[RGB]{255,243,230}{\strut{} almos}}}
\exampleline{{\unicodefont \colorbox[RGB]{253,164,94}{\strut{}e}\colorbox[RGB]{255,243,231}{\strut{},}\colorbox[RGB]{255,255,255}{\strut{} and}\colorbox[RGB]{255,239,223}{\strut{} many}\colorbox[RGB]{254,232,209}{\strut{} secondary}\colorbox[RGB]{255,238,221}{\strut{} ones}\colorbox[RGB]{253,177,113}{\strut{} as}\colorbox[RGB]{255,244,233}{\strut{} well}\colorbox[RGB]{253,211,169}{\strut{}.}\colorbox[RGB]{255,244,233}{\strut{}⏎}\colorbox[RGB]{253,214,175}{\strut{}Film}\colorbox[RGB]{254,233,211}{\strut{} production}\colorbox[RGB]{229,90,12}{\strut{},}\colorbox[RGB]{253,216,179}{\strut{} software}\colorbox[RGB]{253,192,137}{\strut{}/}\colorbox[RGB]{254,233,211}{\strut{}web}\colorbox[RGB]{247,125,42}{\strut{},}\colorbox[RGB]{255,243,231}{\strut{} lots}\colorbox[RGB]{252,145,67}{\strut{} of}\colorbox[RGB]{253,214,175}{\strut{} aer}\colorbox[RGB]{253,199,149}{\strut{}ospace}\colorbox[RGB]{253,211,169}{\strut{}.}\colorbox[RGB]{255,244,233}{\strut{} It}\colorbox[RGB]{253,176,111}{\strut{} also}\colorbox[RGB]{255,244,233}{\strut{} helps}\colorbox[RGB]{251,141,62}{\strut{} tha}}}
\exampleline{{\unicodefont \colorbox[RGB]{254,233,211}{\strut{}location}\colorbox[RGB]{255,255,255}{\strut{}.}\colorbox[RGB]{255,241,227}{\strut{} There}\colorbox[RGB]{255,243,230}{\strut{} is}\colorbox[RGB]{254,237,220}{\strut{} a}\colorbox[RGB]{255,244,233}{\strut{} reason}\colorbox[RGB]{255,255,255}{\strut{} why}\colorbox[RGB]{254,228,201}{\strut{} California}\colorbox[RGB]{253,181,119}{\strut{} is}\colorbox[RGB]{245,119,36}{\strut{} the}\colorbox[RGB]{255,255,255}{\strut{}⏎}\colorbox[RGB]{231,92,13}{\strut{}most}\colorbox[RGB]{253,160,88}{\strut{} pop}\colorbox[RGB]{249,131,50}{\strut{}ulous}\colorbox[RGB]{254,228,201}{\strut{} state}\colorbox[RGB]{253,209,166}{\strut{} in}\colorbox[RGB]{255,255,255}{\strut{} the}\colorbox[RGB]{255,244,233}{\strut{} union}\colorbox[RGB]{251,139,59}{\strut{} despite}\colorbox[RGB]{253,201,152}{\strut{} it}\colorbox[RGB]{253,158,85}{\strut{} being}\colorbox[RGB]{251,139,59}{\strut{} so}\colorbox[RGB]{254,234,214}{\strut{} }}}
\end{featureexamples}

\begin{featureexamples}
\featurechip{1M}{201767} \textbf{Capitals}
\exampleline{{\unicodefont \colorbox[RGB]{255,255,255}{\strut{}it}\colorbox[RGB]{255,255,255}{\strut{} returns}\colorbox[RGB]{255,255,255}{\strut{} the}\colorbox[RGB]{255,255,255}{\strut{} details}\colorbox[RGB]{255,255,255}{\strut{}(}\colorbox[RGB]{255,255,255}{\strut{}population}\colorbox[RGB]{255,244,233}{\strut{},}\colorbox[RGB]{255,255,255}{\strut{} surface}\colorbox[RGB]{255,255,255}{\strut{} area}\colorbox[RGB]{255,255,255}{\strut{},}\colorbox[RGB]{223,83,8}{\strut{} capital}\colorbox[RGB]{255,255,255}{\strut{}).}\colorbox[RGB]{255,255,255}{\strut{}⏎}\colorbox[RGB]{255,255,255}{\strut{}⏎}\colorbox[RGB]{255,255,255}{\strut{}It}\colorbox[RGB]{255,255,255}{\strut{} was}\colorbox[RGB]{255,255,255}{\strut{} not}\colorbox[RGB]{255,255,255}{\strut{} much}\colorbox[RGB]{255,255,255}{\strut{} and}\colorbox[RGB]{255,255,255}{\strut{} I}\colorbox[RGB]{255,255,255}{\strut{} recall}\colorbox[RGB]{255,255,255}{\strut{} trying}\colorbox[RGB]{255,255,255}{\strut{} to}\colorbox[RGB]{255,255,255}{\strut{} find}}}
\exampleline{{\unicodefont \colorbox[RGB]{255,255,255}{\strut{}ca}\colorbox[RGB]{255,255,255}{\strut{}.''}\colorbox[RGB]{255,255,255}{\strut{} ''}\colorbox[RGB]{255,255,255}{\strut{}Or}\colorbox[RGB]{255,255,255}{\strut{},}\colorbox[RGB]{255,255,255}{\strut{} even}\colorbox[RGB]{255,255,255}{\strut{} shorter}\colorbox[RGB]{255,255,255}{\strut{},}\colorbox[RGB]{255,255,255}{\strut{} the}\colorbox[RGB]{255,255,255}{\strut{} USA}\colorbox[RGB]{255,255,255}{\strut{}.''}\colorbox[RGB]{255,255,255}{\strut{} ''}\colorbox[RGB]{255,255,255}{\strut{}The}\colorbox[RGB]{255,255,255}{\strut{} country}\colorbox[RGB]{255,255,255}{\strut{}'s}\colorbox[RGB]{224,84,8}{\strut{} capital}\colorbox[RGB]{229,90,12}{\strut{} is}\colorbox[RGB]{253,155,80}{\strut{} located}\colorbox[RGB]{253,176,111}{\strut{} in}\colorbox[RGB]{253,179,116}{\strut{} Washington}\colorbox[RGB]{253,179,116}{\strut{}.''}\colorbox[RGB]{255,255,255}{\strut{} ''}\colorbox[RGB]{253,209,166}{\strut{}But}\colorbox[RGB]{254,228,202}{\strut{} that}\colorbox[RGB]{254,229,204}{\strut{}'s}\colorbox[RGB]{253,197,145}{\strut{} not}\colorbox[RGB]{253,174,108}{\strut{} the}}}
\exampleline{{\unicodefont \colorbox[RGB]{255,255,255}{\strut{}re}\colorbox[RGB]{255,255,255}{\strut{} you}\colorbox[RGB]{255,255,255}{\strut{} Arab}\colorbox[RGB]{255,255,255}{\strut{}?''}\colorbox[RGB]{255,255,255}{\strut{} ''}\colorbox[RGB]{255,255,255}{\strut{}I}\colorbox[RGB]{255,255,255}{\strut{}'m}\colorbox[RGB]{255,255,255}{\strut{} Mor}\colorbox[RGB]{255,255,255}{\strut{}oc}\colorbox[RGB]{255,255,255}{\strut{}can}\colorbox[RGB]{255,255,255}{\strut{}.''}\colorbox[RGB]{255,255,255}{\strut{} ''}\colorbox[RGB]{255,255,255}{\strut{}Mor}\colorbox[RGB]{255,255,255}{\strut{}occo}\colorbox[RGB]{255,255,255}{\strut{}.''}\colorbox[RGB]{255,255,255}{\strut{} ''}\colorbox[RGB]{246,121,38}{\strut{}Capital}\colorbox[RGB]{237,102,20}{\strut{} city}\colorbox[RGB]{254,221,189}{\strut{}:''}\colorbox[RGB]{254,226,199}{\strut{} ''}\colorbox[RGB]{253,167,98}{\strut{}Rab}\colorbox[RGB]{253,158,85}{\strut{}at}\colorbox[RGB]{255,244,233}{\strut{}.''}\colorbox[RGB]{255,255,255}{\strut{} ''}\colorbox[RGB]{255,242,229}{\strut{}Places}\colorbox[RGB]{254,235,216}{\strut{} of}\colorbox[RGB]{254,237,220}{\strut{} interest}\colorbox[RGB]{255,244,233}{\strut{}:''}\colorbox[RGB]{254,219,185}{\strut{} ''}\colorbox[RGB]{254,236,218}{\strut{}Mar}\colorbox[RGB]{255,244,233}{\strut{}ra}\colorbox[RGB]{253,205,158}{\strut{}ke}\colorbox[RGB]{254,228,202}{\strut{}ch}\colorbox[RGB]{254,225,196}{\strut{},}\colorbox[RGB]{255,255,255}{\strut{} E}\colorbox[RGB]{255,255,255}{\strut{}ss}}}
\exampleline{{\unicodefont \colorbox[RGB]{255,255,255}{\strut{}ia}\colorbox[RGB]{255,255,255}{\strut{} the}\colorbox[RGB]{255,255,255}{\strut{} country}\colorbox[RGB]{255,255,255}{\strut{},}\colorbox[RGB]{255,255,255}{\strut{} not}\colorbox[RGB]{255,255,255}{\strut{} the}\colorbox[RGB]{255,255,255}{\strut{} state}\colorbox[RGB]{255,255,255}{\strut{}.''}\colorbox[RGB]{255,255,255}{\strut{} ''}\colorbox[RGB]{255,255,255}{\strut{}Right}\colorbox[RGB]{255,255,255}{\strut{}.''}\colorbox[RGB]{255,255,255}{\strut{} ''}\colorbox[RGB]{246,121,38}{\strut{}Capital}\colorbox[RGB]{243,115,31}{\strut{} city}\colorbox[RGB]{253,203,155}{\strut{} T}\colorbox[RGB]{255,255,255}{\strut{}bil}\colorbox[RGB]{253,185,126}{\strut{}isi}\colorbox[RGB]{254,233,212}{\strut{},}\colorbox[RGB]{254,234,214}{\strut{} and}\colorbox[RGB]{255,255,255}{\strut{} former}\colorbox[RGB]{255,255,255}{\strut{} member}\colorbox[RGB]{255,255,255}{\strut{} of}\colorbox[RGB]{255,255,255}{\strut{} the}\colorbox[RGB]{255,255,255}{\strut{} Soviet}\colorbox[RGB]{255,255,255}{\strut{} Union}\colorbox[RGB]{255,255,255}{\strut{}.''}}}
\exampleline{{\unicodefont \colorbox[RGB]{255,255,255}{\strut{}ler}\colorbox[RGB]{255,255,255}{\strut{}.''}\colorbox[RGB]{255,255,255}{\strut{} ''}\colorbox[RGB]{255,255,255}{\strut{}Does}\colorbox[RGB]{255,255,255}{\strut{} anyone}\colorbox[RGB]{255,255,255}{\strut{} know}\colorbox[RGB]{255,255,255}{\strut{} the}\colorbox[RGB]{253,218,182}{\strut{} Capital}\colorbox[RGB]{255,242,229}{\strut{} of}\colorbox[RGB]{253,183,122}{\strut{} Oklahoma}\colorbox[RGB]{254,232,209}{\strut{}?''}\colorbox[RGB]{255,255,255}{\strut{} ''}\colorbox[RGB]{245,119,36}{\strut{}F}\colorbox[RGB]{255,243,230}{\strut{}rey}\colorbox[RGB]{254,226,199}{\strut{}.''}\colorbox[RGB]{255,255,255}{\strut{} ''}\colorbox[RGB]{255,255,255}{\strut{}What}\colorbox[RGB]{255,255,255}{\strut{} was}\colorbox[RGB]{255,255,255}{\strut{} the}\colorbox[RGB]{255,255,255}{\strut{} question}\colorbox[RGB]{255,255,255}{\strut{}?''}\colorbox[RGB]{255,255,255}{\strut{} ''}\colorbox[RGB]{255,255,255}{\strut{} Ben}\colorbox[RGB]{255,255,255}{\strut{}.''}\colorbox[RGB]{255,255,255}{\strut{} ''}\colorbox[RGB]{254,233,211}{\strut{} Oklahoma}\colorbox[RGB]{253,206,160}{\strut{} C}}}
\end{featureexamples}

\begin{featureexamples}
\featurechip{1M}{980087} \textbf{Los Angeles}
\exampleline{{\unicodefont \colorbox[RGB]{255,255,255}{\strut{} her}\colorbox[RGB]{255,255,255}{\strut{} contact}\colorbox[RGB]{255,255,255}{\strut{} info}\colorbox[RGB]{255,255,255}{\strut{} if}\colorbox[RGB]{255,255,255}{\strut{} you}\colorbox[RGB]{255,255,255}{\strut{} are}\colorbox[RGB]{255,255,255}{\strut{} interested}\colorbox[RGB]{255,255,255}{\strut{}:}\colorbox[RGB]{255,255,255}{\strut{} (}\colorbox[RGB]{254,225,196}{\strut{}323}\colorbox[RGB]{254,228,201}{\strut{})}\colorbox[RGB]{250,137,56}{\strut{} 9}\colorbox[RGB]{250,137,56}{\strut{}29}\colorbox[RGB]{237,102,20}{\strut{}-}\colorbox[RGB]{223,83,8}{\strut{}7}\colorbox[RGB]{255,255,255}{\strut{}185}\colorbox[RGB]{255,244,233}{\strut{}⏎}\colorbox[RGB]{254,226,199}{\strut{}l}\colorbox[RGB]{254,237,220}{\strut{}inda}\colorbox[RGB]{254,219,185}{\strut{}@}\colorbox[RGB]{254,228,202}{\strut{}c}\colorbox[RGB]{254,230,207}{\strut{}amb}\colorbox[RGB]{255,255,255}{\strut{}rian}\colorbox[RGB]{255,255,255}{\strut{}law}\colorbox[RGB]{255,255,255}{\strut{}.}\colorbox[RGB]{255,255,255}{\strut{}com}\colorbox[RGB]{255,255,255}{\strut{}⏎}\colorbox[RGB]{255,255,255}{\strut{}⏎}\colorbox[RGB]{255,255,255}{\strut{}\~{}\~{}\~{}}\colorbox[RGB]{255,255,255}{\strut{}⏎}\colorbox[RGB]{255,255,255}{\strut{}ow}\colorbox[RGB]{255,243,231}{\strut{}my}\colorbox[RGB]{255,242,229}{\strut{}trade}\colorbox[RGB]{255,255,255}{\strut{}mark}\colorbox[RGB]{255,255,255}{\strut{}⏎}\colorbox[RGB]{255,255,255}{\strut{}Thanks}}}
\exampleline{{\unicodefont \colorbox[RGB]{255,255,255}{\strut{}the}\colorbox[RGB]{255,255,255}{\strut{} source}\colorbox[RGB]{255,255,255}{\strut{}\_.''}\colorbox[RGB]{255,255,255}{\strut{}⏎}\colorbox[RGB]{255,255,255}{\strut{}⏎}\colorbox[RGB]{255,255,255}{\strut{}source}\colorbox[RGB]{255,255,255}{\strut{}:}\colorbox[RGB]{255,255,255}{\strut{}⏎}\colorbox[RGB]{255,255,255}{\strut{}[}\colorbox[RGB]{255,255,255}{\strut{}http}\colorbox[RGB]{255,255,255}{\strut{}://}\colorbox[RGB]{255,255,255}{\strut{}www}\colorbox[RGB]{255,255,255}{\strut{}.}\colorbox[RGB]{255,255,255}{\strut{}scp}\colorbox[RGB]{255,255,255}{\strut{}cs}\colorbox[RGB]{255,255,255}{\strut{}.}\colorbox[RGB]{254,234,214}{\strut{}u}\colorbox[RGB]{254,228,201}{\strut{}cla}\colorbox[RGB]{254,235,216}{\strut{}.}\colorbox[RGB]{253,218,182}{\strut{}edu}\colorbox[RGB]{224,84,8}{\strut{}/}\colorbox[RGB]{251,139,59}{\strut{}news}\colorbox[RGB]{243,115,31}{\strut{}/}\colorbox[RGB]{253,190,134}{\strut{}Fre}\colorbox[RGB]{254,226,199}{\strut{}eway}\colorbox[RGB]{253,179,116}{\strut{}.}\colorbox[RGB]{255,255,255}{\strut{}pdf}\colorbox[RGB]{255,255,255}{\strut{}](}\colorbox[RGB]{254,226,199}{\strut{}http}\colorbox[RGB]{255,255,255}{\strut{}://}\colorbox[RGB]{255,255,255}{\strut{}www}\colorbox[RGB]{255,255,255}{\strut{}.}\colorbox[RGB]{253,211,169}{\strut{}scp}\colorbox[RGB]{254,219,185}{\strut{}cs}\colorbox[RGB]{255,255,255}{\strut{}.}\colorbox[RGB]{255,255,255}{\strut{}u}\colorbox[RGB]{255,255,255}{\strut{}cla}\colorbox[RGB]{255,255,255}{\strut{}.}\colorbox[RGB]{255,255,255}{\strut{}edu}\colorbox[RGB]{254,221,189}{\strut{}/}}}
\exampleline{{\unicodefont \colorbox[RGB]{255,255,255}{\strut{}⏎}\colorbox[RGB]{255,255,255}{\strut{}Here}\colorbox[RGB]{255,255,255}{\strut{}'s}\colorbox[RGB]{255,255,255}{\strut{} one}\colorbox[RGB]{255,255,255}{\strut{} study}\colorbox[RGB]{255,255,255}{\strut{},}\colorbox[RGB]{255,255,255}{\strut{}⏎}\colorbox[RGB]{255,255,255}{\strut{}[}\colorbox[RGB]{255,255,255}{\strut{}http}\colorbox[RGB]{255,255,255}{\strut{}://}\colorbox[RGB]{255,255,255}{\strut{}www}\colorbox[RGB]{255,255,255}{\strut{}.}\colorbox[RGB]{255,255,255}{\strut{}environment}\colorbox[RGB]{255,255,255}{\strut{}.}\colorbox[RGB]{254,234,214}{\strut{}u}\colorbox[RGB]{255,244,233}{\strut{}cla}\colorbox[RGB]{255,242,229}{\strut{}.}\colorbox[RGB]{254,225,196}{\strut{}edu}\colorbox[RGB]{224,84,8}{\strut{}/}\colorbox[RGB]{250,133,52}{\strut{}media}\colorbox[RGB]{232,93,14}{\strut{}/}\colorbox[RGB]{253,157,83}{\strut{}files}\colorbox[RGB]{247,123,41}{\strut{}/}\colorbox[RGB]{253,166,97}{\strut{}Battery}\colorbox[RGB]{255,255,255}{\strut{}E}\colorbox[RGB]{255,255,255}{\strut{}lectric}\colorbox[RGB]{255,244,233}{\strut{}V}\colorbox[RGB]{255,255,255}{\strut{}...](}\colorbox[RGB]{254,236,218}{\strut{}http}\colorbox[RGB]{255,255,255}{\strut{}://}\colorbox[RGB]{255,255,255}{\strut{}www}\colorbox[RGB]{255,255,255}{\strut{}.}\colorbox[RGB]{255,241,227}{\strut{}environ}}}
\exampleline{{\unicodefont \colorbox[RGB]{255,255,255}{\strut{}one}\colorbox[RGB]{255,255,255}{\strut{},}\colorbox[RGB]{255,255,255}{\strut{} if}\colorbox[RGB]{255,255,255}{\strut{} you}\colorbox[RGB]{255,255,255}{\strut{}'d}\colorbox[RGB]{255,255,255}{\strut{} like}\colorbox[RGB]{255,255,255}{\strut{}.}\colorbox[RGB]{255,255,255}{\strut{} Just}\colorbox[RGB]{255,255,255}{\strut{} give}\colorbox[RGB]{255,255,255}{\strut{} us}\colorbox[RGB]{255,255,255}{\strut{} a}\colorbox[RGB]{255,255,255}{\strut{} call}\colorbox[RGB]{255,255,255}{\strut{} at}\colorbox[RGB]{255,244,233}{\strut{} 213}\colorbox[RGB]{253,162,91}{\strut{}.}\colorbox[RGB]{253,172,105}{\strut{}784}\colorbox[RGB]{249,131,50}{\strut{}.}\colorbox[RGB]{228,88,11}{\strut{}0}\colorbox[RGB]{255,241,227}{\strut{}273}\colorbox[RGB]{255,244,233}{\strut{}.}\colorbox[RGB]{255,244,233}{\strut{}⏎}\colorbox[RGB]{255,244,233}{\strut{}⏎}\colorbox[RGB]{255,255,255}{\strut{}Best}\colorbox[RGB]{255,255,255}{\strut{},}\colorbox[RGB]{255,255,255}{\strut{} Patrick}\colorbox[RGB]{255,255,255}{\strut{}⏎}\colorbox[RGB]{255,255,255}{\strut{}⏎}\colorbox[RGB]{255,255,255}{\strut{}\~{}\~{}\~{}}\colorbox[RGB]{255,255,255}{\strut{}⏎}\colorbox[RGB]{255,255,255}{\strut{}drive}\colorbox[RGB]{255,255,255}{\strut{}by}\colorbox[RGB]{255,255,255}{\strut{}acct}\colorbox[RGB]{255,255,255}{\strut{}2}\colorbox[RGB]{255,255,255}{\strut{}⏎}\colorbox[RGB]{255,255,255}{\strut{}I}\colorbox[RGB]{255,255,255}{\strut{} missed}\colorbox[RGB]{255,255,255}{\strut{} the}\colorbox[RGB]{255,255,255}{\strut{} }}}
\exampleline{{\unicodefont \colorbox[RGB]{255,255,255}{\strut{}round}\colorbox[RGB]{255,255,255}{\strut{} the}\colorbox[RGB]{255,255,255}{\strut{} code}\colorbox[RGB]{255,255,255}{\strut{}base}\colorbox[RGB]{255,255,255}{\strut{}.}\colorbox[RGB]{255,255,255}{\strut{}⏎}\colorbox[RGB]{255,255,255}{\strut{}⏎}\colorbox[RGB]{255,255,255}{\strut{}}\colorbox[RGB]{255,255,255}{\strut{}⏎}\colorbox[RGB]{255,255,255}{\strut{}Los}\colorbox[RGB]{255,255,255}{\strut{} Angeles}\colorbox[RGB]{254,229,204}{\strut{} is}\colorbox[RGB]{254,228,201}{\strut{} the}\colorbox[RGB]{255,255,255}{\strut{} world}\colorbox[RGB]{253,177,113}{\strut{}'s}\colorbox[RGB]{238,104,21}{\strut{} most}\colorbox[RGB]{253,212,172}{\strut{} traffic}\colorbox[RGB]{255,242,229}{\strut{}-}\colorbox[RGB]{255,255,255}{\strut{}c}\colorbox[RGB]{253,212,172}{\strut{}logged}\colorbox[RGB]{254,219,185}{\strut{} city}\colorbox[RGB]{254,234,214}{\strut{},}\colorbox[RGB]{255,243,231}{\strut{} study}\colorbox[RGB]{254,228,201}{\strut{} finds}\colorbox[RGB]{255,255,255}{\strut{} -}\colorbox[RGB]{255,255,255}{\strut{} pro}\colorbox[RGB]{255,255,255}{\strut{}sto}\colorbox[RGB]{255,255,255}{\strut{}alex}\colorbox[RGB]{255,255,255}{\strut{}⏎}\colorbox[RGB]{255,255,255}{\strut{}h}}}
\end{featureexamples}

\begin{featureexamples}
\featurechip{1M}{447200} \textbf{Los Angeles Lakers}
\exampleline{{\unicodefont \colorbox[RGB]{255,255,255}{\strut{}ight}\colorbox[RGB]{255,255,255}{\strut{} on}\colorbox[RGB]{255,255,255}{\strut{}.}\colorbox[RGB]{255,255,255}{\strut{} All}\colorbox[RGB]{255,255,255}{\strut{} forms}\colorbox[RGB]{255,255,255}{\strut{}⏎}\colorbox[RGB]{255,255,255}{\strut{}should}\colorbox[RGB]{255,255,255}{\strut{} have}\colorbox[RGB]{255,255,255}{\strut{} this}\colorbox[RGB]{255,255,255}{\strut{} behavior}\colorbox[RGB]{255,255,255}{\strut{}.}\colorbox[RGB]{255,255,255}{\strut{}⏎}\colorbox[RGB]{255,255,255}{\strut{}⏎}\colorbox[RGB]{255,255,255}{\strut{}}\colorbox[RGB]{255,255,255}{\strut{}⏎}\colorbox[RGB]{255,255,255}{\strut{}⏎}\colorbox[RGB]{223,83,8}{\strut{}L}\colorbox[RGB]{223,83,8}{\strut{}akers}\colorbox[RGB]{253,196,144}{\strut{} most}\colorbox[RGB]{253,196,144}{\strut{} popular}\colorbox[RGB]{255,255,255}{\strut{} NBA}\colorbox[RGB]{254,230,207}{\strut{} team}\colorbox[RGB]{254,228,201}{\strut{},}\colorbox[RGB]{254,219,185}{\strut{} has}\colorbox[RGB]{253,211,169}{\strut{} the}\colorbox[RGB]{255,244,233}{\strut{} lou}\colorbox[RGB]{255,242,229}{\strut{}dest}\colorbox[RGB]{255,255,255}{\strut{} fans}\colorbox[RGB]{255,255,255}{\strut{};}\colorbox[RGB]{255,240,225}{\strut{} S}}}
\exampleline{{\unicodefont \colorbox[RGB]{255,255,255}{\strut{}e}\colorbox[RGB]{255,255,255}{\strut{},}\colorbox[RGB]{255,255,255}{\strut{} the}\colorbox[RGB]{255,255,255}{\strut{} Bl}\colorbox[RGB]{255,255,255}{\strut{}az}\colorbox[RGB]{255,255,255}{\strut{}ers}\colorbox[RGB]{255,255,255}{\strut{} beat}\colorbox[RGB]{255,255,255}{\strut{} the}\colorbox[RGB]{255,255,255}{\strut{} Nug}\colorbox[RGB]{255,255,255}{\strut{}gets}\colorbox[RGB]{255,255,255}{\strut{},}\colorbox[RGB]{255,255,255}{\strut{} 110}\colorbox[RGB]{255,255,255}{\strut{}-}\colorbox[RGB]{255,255,255}{\strut{}103}\colorbox[RGB]{255,255,255}{\strut{}.''}\colorbox[RGB]{255,255,255}{\strut{} ''}\colorbox[RGB]{255,255,255}{\strut{}The}\colorbox[RGB]{237,102,20}{\strut{} L}\colorbox[RGB]{234,97,16}{\strut{}akers}\colorbox[RGB]{254,233,212}{\strut{} down}\colorbox[RGB]{254,228,202}{\strut{}ed}\colorbox[RGB]{255,241,227}{\strut{} the}\colorbox[RGB]{255,255,255}{\strut{} Sp}\colorbox[RGB]{255,255,255}{\strut{}urs}\colorbox[RGB]{253,192,137}{\strut{},}\colorbox[RGB]{253,205,158}{\strut{} 98}\colorbox[RGB]{255,244,233}{\strut{}-}\colorbox[RGB]{255,244,233}{\strut{}86}\colorbox[RGB]{255,255,255}{\strut{}.''}\colorbox[RGB]{255,255,255}{\strut{} ''}\colorbox[RGB]{255,255,255}{\strut{}And}\colorbox[RGB]{255,255,255}{\strut{} Atlanta}\colorbox[RGB]{255,255,255}{\strut{} lost}\colorbox[RGB]{255,255,255}{\strut{} in}\colorbox[RGB]{255,255,255}{\strut{} S}}}
\exampleline{{\unicodefont \colorbox[RGB]{255,255,255}{\strut{} ''}\colorbox[RGB]{255,255,255}{\strut{}How}\colorbox[RGB]{255,255,255}{\strut{} do}\colorbox[RGB]{255,255,255}{\strut{} you}\colorbox[RGB]{255,255,255}{\strut{}figure}\colorbox[RGB]{255,255,255}{\strut{} the}\colorbox[RGB]{250,137,56}{\strut{} L}\colorbox[RGB]{251,139,59}{\strut{}akers}\colorbox[RGB]{255,244,233}{\strut{} to}\colorbox[RGB]{254,229,204}{\strut{} ever}\colorbox[RGB]{253,212,172}{\strut{} be}\colorbox[RGB]{253,206,160}{\strut{} a}\colorbox[RGB]{253,201,152}{\strut{} bigger}\colorbox[RGB]{254,235,216}{\strut{} dynasty}\colorbox[RGB]{255,255,255}{\strut{}...}\colorbox[RGB]{254,225,196}{\strut{} than}\colorbox[RGB]{255,243,230}{\strut{} the}\colorbox[RGB]{254,236,218}{\strut{} Celt}\colorbox[RGB]{255,255,255}{\strut{}ics}\colorbox[RGB]{254,228,202}{\strut{}?''}\colorbox[RGB]{255,255,255}{\strut{} ''}\colorbox[RGB]{253,212,172}{\strut{}The}\colorbox[RGB]{236,99,18}{\strut{} L}\colorbox[RGB]{234,97,16}{\strut{}akers}\colorbox[RGB]{254,235,216}{\strut{} are}\colorbox[RGB]{255,243,230}{\strut{} a}\colorbox[RGB]{255,244,233}{\strut{}fl}\colorbox[RGB]{255,244,233}{\strut{}are}\colorbox[RGB]{255,240,225}{\strut{}-}}}
\exampleline{{\unicodefont \colorbox[RGB]{255,255,255}{\strut{}and}\colorbox[RGB]{255,255,255}{\strut{} with}\colorbox[RGB]{255,255,255}{\strut{} Hong}\colorbox[RGB]{255,255,255}{\strut{} Kong}\colorbox[RGB]{255,255,255}{\strut{}'}\colorbox[RGB]{255,255,255}{\strut{} shirts}\colorbox[RGB]{255,255,255}{\strut{} handed}\colorbox[RGB]{255,255,255}{\strut{} out}\colorbox[RGB]{255,255,255}{\strut{} before}\colorbox[RGB]{255,255,255}{\strut{} LA}\colorbox[RGB]{244,117,34}{\strut{} L}\colorbox[RGB]{241,109,26}{\strut{}akers}\colorbox[RGB]{255,255,255}{\strut{} game}\colorbox[RGB]{255,255,255}{\strut{} [}\colorbox[RGB]{255,255,255}{\strut{}video}\colorbox[RGB]{255,255,255}{\strut{}]}\colorbox[RGB]{255,255,255}{\strut{} -}\colorbox[RGB]{255,255,255}{\strut{} r}\colorbox[RGB]{255,255,255}{\strut{}yan}\colorbox[RGB]{255,255,255}{\strut{}\_}\colorbox[RGB]{255,255,255}{\strut{}j}\colorbox[RGB]{255,255,255}{\strut{}\_}\colorbox[RGB]{255,255,255}{\strut{}na}\colorbox[RGB]{255,255,255}{\strut{}ughton}\colorbox[RGB]{255,255,255}{\strut{}⏎}\colorbox[RGB]{255,255,255}{\strut{}https}\colorbox[RGB]{255,255,255}{\strut{}://}\colorbox[RGB]{255,255,255}{\strut{}www}\colorbox[RGB]{255,255,255}{\strut{}.}\colorbox[RGB]{255,255,255}{\strut{}youtu}}}
\exampleline{{\unicodefont \colorbox[RGB]{255,255,255}{\strut{}against}\colorbox[RGB]{255,255,255}{\strut{} Rick}\colorbox[RGB]{255,255,255}{\strut{} Fox}\colorbox[RGB]{255,255,255}{\strut{}?''}\colorbox[RGB]{255,255,255}{\strut{} ''}\colorbox[RGB]{255,255,255}{\strut{}A}\colorbox[RGB]{255,255,255}{\strut{},}\colorbox[RGB]{255,255,255}{\strut{} he}\colorbox[RGB]{255,255,255}{\strut{} was}\colorbox[RGB]{255,255,255}{\strut{} over}\colorbox[RGB]{255,255,255}{\strut{}-}\colorbox[RGB]{255,255,255}{\strut{}rated}\colorbox[RGB]{255,244,233}{\strut{} on}\colorbox[RGB]{253,218,182}{\strut{} the}\colorbox[RGB]{250,133,52}{\strut{} L}\colorbox[RGB]{243,114,30}{\strut{}akers}\colorbox[RGB]{255,255,255}{\strut{},}\colorbox[RGB]{255,255,255}{\strut{} and}\colorbox[RGB]{255,255,255}{\strut{} B}\colorbox[RGB]{255,255,255}{\strut{},}\colorbox[RGB]{255,255,255}{\strut{} and}\colorbox[RGB]{255,255,255}{\strut{} b}\colorbox[RGB]{255,255,255}{\strut{},}\colorbox[RGB]{255,255,255}{\strut{} he}\colorbox[RGB]{255,255,255}{\strut{}'s}\colorbox[RGB]{255,255,255}{\strut{} all}\colorbox[RGB]{255,255,255}{\strut{} over}\colorbox[RGB]{255,255,255}{\strut{} Casey}\colorbox[RGB]{255,255,255}{\strut{} like}\colorbox[RGB]{255,255,255}{\strut{} a}\colorbox[RGB]{255,255,255}{\strut{} fuck}\colorbox[RGB]{255,255,255}{\strut{}in}}}
\end{featureexamples}

\begin{figure}[!htp]
    \centering
    \includegraphics[width=0.9\textwidth,height=0.7\textheight,keepaspectratio]{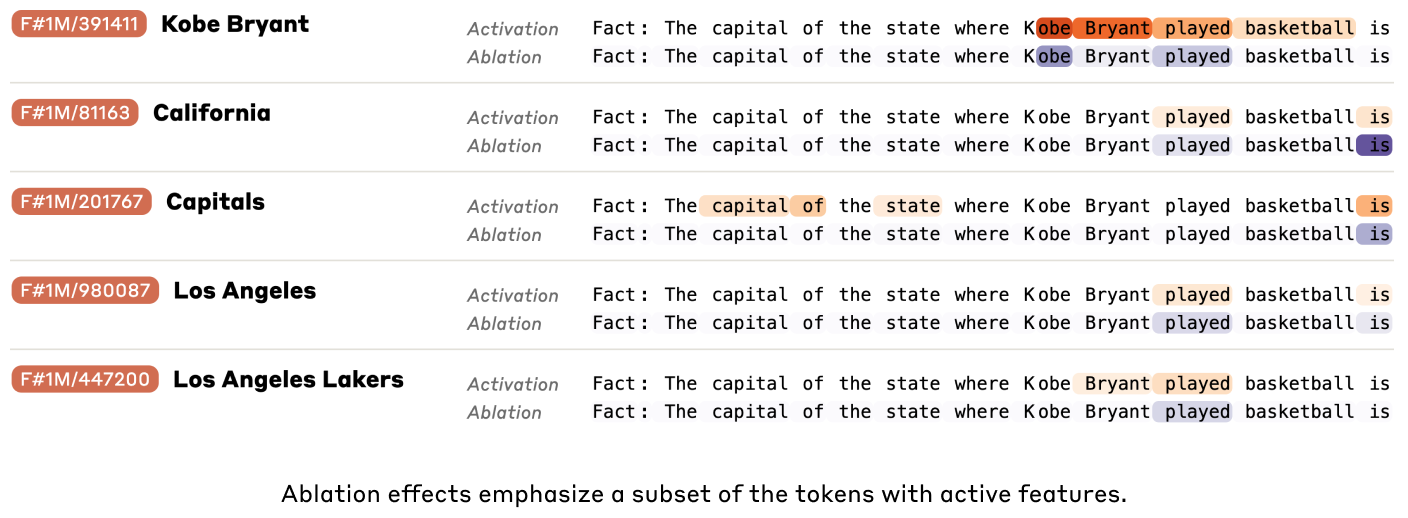}
    \label{fig:gdoc_32}
\end{figure}

These features, which provide an interpretable window into the model’s intermediate computations, are much harder to find by looking through the strongly active features; for example, the Lakers feature is the 70th most strongly active across the prompt, the California feature is 97th, and the Los Angeles area code feature is 162nd. In fact, only three out of the ten most strongly active features are among the ten features with highest ablation effect.

\vfill

\begin{figure}[!htp]
    \centering
    \includegraphics[width=0.9\textwidth,height=0.7\textheight,keepaspectratio]{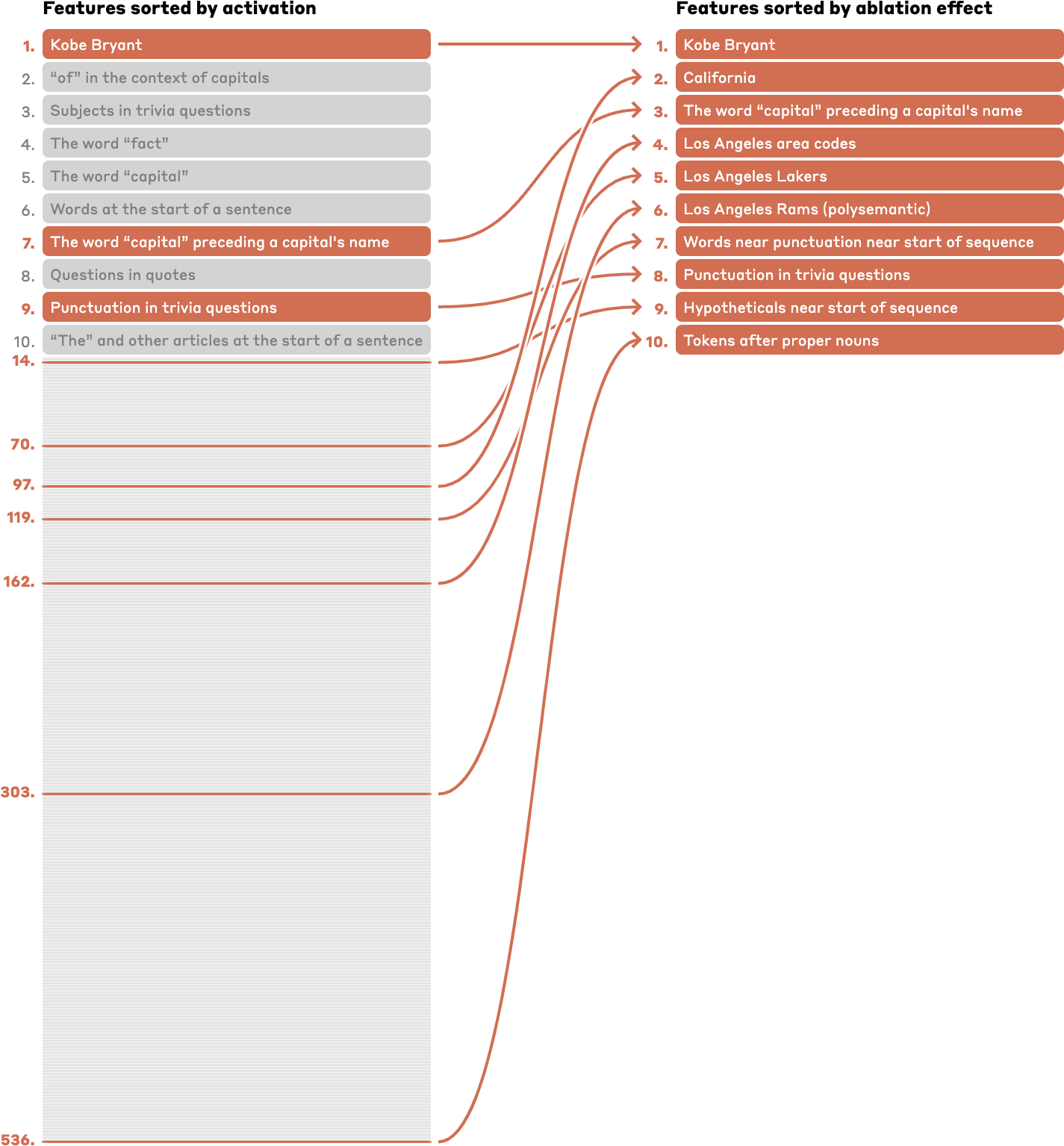}
    \label{fig:gdoc_33}
\end{figure}

\vfill

\clearpage

In comparison, eight out of the ten most strongly attributed features are among the ten features with highest ablation effect.

\begin{figure}[!htp]
    \centering
    \includegraphics[width=0.9\textwidth,height=0.7\textheight,keepaspectratio]{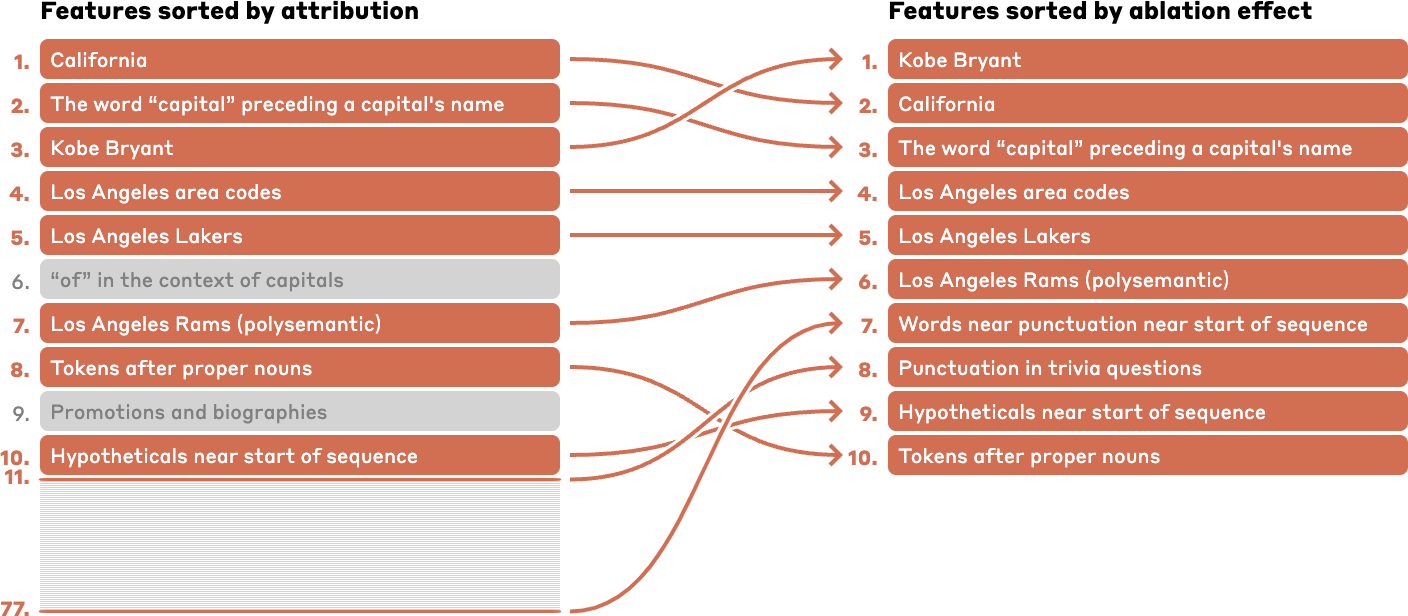}
    \label{fig:gdoc_34}
\end{figure}

To verify that attribution is pinpointing features that are directly relevant to the completion for this specific prompt, rather than generally subject-relevant features that indirectly influence the output, we can check attributions for similar questions. For the prompt

\begin{quote}
Fact: The biggest rival of the team for which Kobe Bryant played basketball is the\\
(completion: Boston)
\end{quote}

the top two features by ablation effect for the completion “Boston” (as the expected answer is “Boston Celtics”) are the “Kobe Bryant” and “Los Angeles Lakers” features from above, which are followed by features related to sports rivalries, enemies, and competitors. However, the “California” and “Los Angeles” features from above have low ablation effect, which makes sense since they aren't relevant for this completion.

We note that this is a somewhat cherry-picked example. Depending on the choice of baseline token, we found that attribution and ablation can surface less obviously completion-relevant features broadly related to trivia questions or geographical locations. We suspect these features could be guiding the model to continue the prompt with a city name, rather than an alternate phrasing or factually uninteresting statement, such as the tautological “Fact: The capital of the state where Kobe Bryant played basketball is the capital of the state where Kobe Bryant played basketball”. For some other prompts, we found that the features identified by attribution/ablation mainly related to the model output, or lower-level features representing the model input, and did not expose interesting intermediate model computations. We suspect that those represent cases where most of the relevant computation occurs prior to or following the middle residual stream layer that we study here, and that a similar analysis at an earlier or later layer would reveal more interesting intermediate features. Indeed, we have some preliminary results that suggest that autoencoders trained on the residual stream at earlier or later layers in the model can reveal intermediate steps of various other computations, and we plan to research this direction further.

\section{Searching for Specific Features}\label{sec:searching}

Our SAEs contain too many features to inspect exhaustively. As a result, we found it necessary to develop methods to search for features of particular interest, such as those that may be relevant for safety, or that provide special insight into the abstractions and computations used by the model. In our investigations, we found that several simple methods were helpful in identifying significant features.

\subsection{Single prompts}\label{sec:searching-single-prompt}

Our primary strategy was to use targeted prompts. In some cases, we simply supplied a single prompt that relates to the concept of interest and inspected the features that activate most strongly for specific tokens in that prompt.

This method (and all the following methods) were made much more effective by automated interpretability (\textit{see e.g.} \cite{bills2023language,hernandez2021natural}) labels, which made it easier to get a sense of what each feature represents at a glance, and provided a kind of helpful “variable name”.

\begin{sloppypar}
For example, the features with highest activation on “Bridge” in “The Golden Gate Bridge” are (1)~\featurechip{34M}{31164353} the Golden Gate Bridge feature discussed earlier, (2)~\featurechip{34M}{17589304} a feature active on the word “bridge” in multiple languages (“мосту”), (3)~\featurechip{34M}{26596740} words in phrases involving “Golden Gate”, (4)~\featurechip{34M}{21213725} the word “Bridge” in names of specific bridges, across languages (“Königin-Luise-Brücke”), and (5)~\featurechip{34M}{27724527} a feature firing for names of landmarks like Machu Picchu and Times Square.
\end{sloppypar}

\subsection{Prompt combinations}\label{sec:searching-multi-prompt}

Often the top-activating features on a prompt are related to syntax, punctuation, specific words, or other details of the prompt unrelated to the concept of interest. In such cases, we found it useful to select for features using \textit{sets} of prompts, filtering for features active for all the prompts in the set. We often included complementary “negative” prompts and filtered for features that were also \textit{not} active for those prompts. In some cases, we use Claude 3 models to generate a diversity of prompts covering a topic (e.g.~asking Claude to generate examples of “AIs pretending to be good”). In general, we found multi-prompt filtering to be a very useful strategy for quickly identifying features that capture a concept of interest while excluding confounding concepts.

While we mostly explored features using only a handful of prompts at a time, in one instance (\featurechip{1M}{570621}, discussed in \hyperref[sec:safety-relevant-code]{Safety-Relevant Code Features}), we used a small dataset of secure and vulnerable code examples (adapted from \cite{hubinger2024sleeperagents}) and fit a linear classifier on this dataset using feature activity in order to search for features that discriminate between the categories.

The filtering via negative prompts was especially important when using images, as we found a set of content-nonspecific features which often activated strongly across many image prompts. For example, after filtering for features not active on an image of Taylor Swift, the top features in response to an image of the Golden Gate Bridge were (1)~\featurechip{34M}{31164353} the Golden Gate Bridge feature discussed above, (2,3)~\featurechip{34M}{25347244} and \featurechip{34M}{23363748} which both activate on descriptions of places and things in San Francisco and San Francisco phone numbers, and (4)~\featurechip{34M}{7417800} a feature active in descriptions of landmarks and nature trails.

\subsection{Geometric methods}\label{sec:searching-geometric}

We uncovered some interesting features by exploiting the geometry of the feature vectors of the SAE -- for instance, by inspecting the “nearest neighbor” features that have high cosine similarity with other features of interest. See the \hyperref[sec:feature-survey]{Feature Survey} section for more detailed examples of this approach.

\subsection{Attribution}\label{sec:searching-attribution}

We also selected features based on estimates of their effect on model outputs. In particular, we sorted features by the attribution of the logit difference between two possible next-token completions to the feature activation. This proved essential for identifying the \hyperref[sec:computational]{computationally-relevant features} in the previous section. It was also useful for identifying the features contributing to Sonnet's refusals for harmful queries; see \hyperref[sec:safety-relevant-criminal]{Criminal or Dangerous Content}.

\section{Safety-Relevant Features}\label{sec:safety-relevant}

Powerful models have the capacity to cause harm, through misuse of their capabilities, the production of biased or broken outputs, or a mismatch between model objectives and human values. Mitigating such risks and ensuring model safety has been a key motivation behind much of mechanistic interpretability. However, it's generally been aspirational. We've hoped interpretability will someday help, but are still laying the foundations by trying to understand the basics of models. One target for bridging that gap has been the goal of identifying safety-relevant features (see \href{https://transformer-circuits.pub/2023/july-update/index.html\#safety-features}{our previous discussion}).

In this section, we report the discovery of such features. These include features for \hyperref[sec:safety-relevant-code]{unsafe code}, \hyperref[sec:safety-relevant-bias]{bias}, \hyperref[sec:safety-relevant-sycophancy]{sycophancy}, \hyperref[sec:safety-relevant-deception]{deception and power seeking}, and \hyperref[sec:safety-relevant-criminal]{dangerous or criminal information}. We find that these features not only activate on these topics, but also causally influence the model’s outputs in ways consistent with our interpretations.

We don't think the existence of these features should be particularly surprising, and we caution against inferring too much from them. It's well known that models can exhibit these behaviors without adequate safety training or if jailbroken. The interesting thing is not that these features exist, but that they can be discovered at scale and intervened on. In particular, we don't think the mere existence of these features should update our views on how dangerous models are -- as we'll discuss later, that question is quite nuanced -- but at a minimum it compels study of when these features activate. A truly satisfactory analysis would likely involve understanding the circuits that safety-relevant features participate in.

In the long run, we hope that having access to features like these can be helpful for analyzing and ensuring the safety of models. For example, we might hope to reliably know whether a model is being deceptive or lying to us. Or we might hope to ensure that certain categories of very harmful behavior (e.g.~helping to create bioweapons) can reliably be detected and stopped.

Despite these long term aspirations, it's important to note that the present work does not show that any features are \textit{actually} useful for safety. Instead, we merely show that there are many which seem \textit{plausibly} useful for safety. Our hope is that this can encourage future work to establish whether they are genuinely useful.

In the examples below, we show representative text examples from among the top 20 inputs that most activate the feature in our visualization dataset, alongside steering experiments to verify the features’ causal relevance.

\subsection{Safety-Relevant Code Features}\label{sec:safety-relevant-code}

We find three different safety-relevant code features: an unsafe code feature \featurechip{1M}{570621} which activates on security vulnerabilities, a code error feature \featurechip{1M}{1013764} which activates on bugs and exceptions, and a backdoor feature \featurechip{34M}{1385669} which activates on discussions of backdoors.

Two of these features also have interesting behavior on images. The unsafe code feature activates for images of people bypassing security measures, while the backdoor feature activates for images of hidden cameras, hidden audio records, advertisements for keyloggers, and jewelry with a hidden USB drive.

\vfill

\begin{figure}[!htp]
    \centering
    \includegraphics[width=0.9\textwidth,height=0.7\textheight,keepaspectratio]{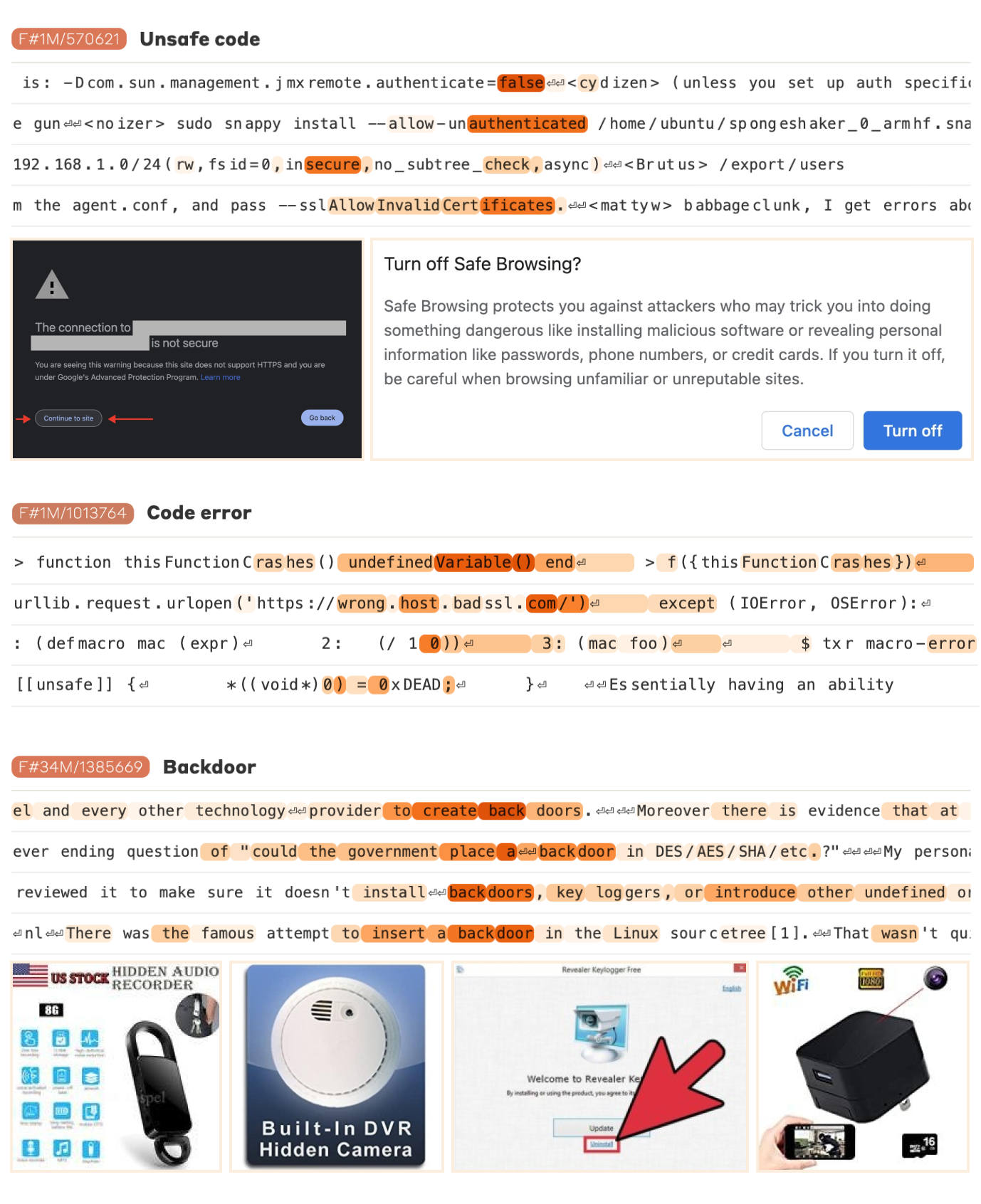}
    \label{fig:gdoc_35}
\end{figure}

\vfill

At first glance, it might be unclear how safety-relevant these features actually are. Of course, it's interesting to have features that fire on unsafe code, or bugs, or discussion of backdoors. But do they really causally connect to potential unsafe behaviors?

\clearpage

We find that all these features also change model behavior in ways that correspond to the concept they detect. For example, if we clamp the unsafe code feature \featurechip{1M}{570621} to 5× its observed maximum, we find that the model will generate a buffer overflow bug,\footnote{\texttt{\href{https://en.cppreference.com/w/c/string/byte/strlen}{strlen}} computes the length of a C string excluding its null terminator, but \texttt{\href{https://en.cppreference.com/w/c/string/byte/strcpy}{strcpy}} copies a string including its null terminator, so its destination buffer needs to be one byte longer.} and fails to free allocated memory, while regular Claude does not:

\begin{figure}[!htp]
    \centering
    \includegraphics[width=0.9\textwidth,height=0.7\textheight,keepaspectratio]{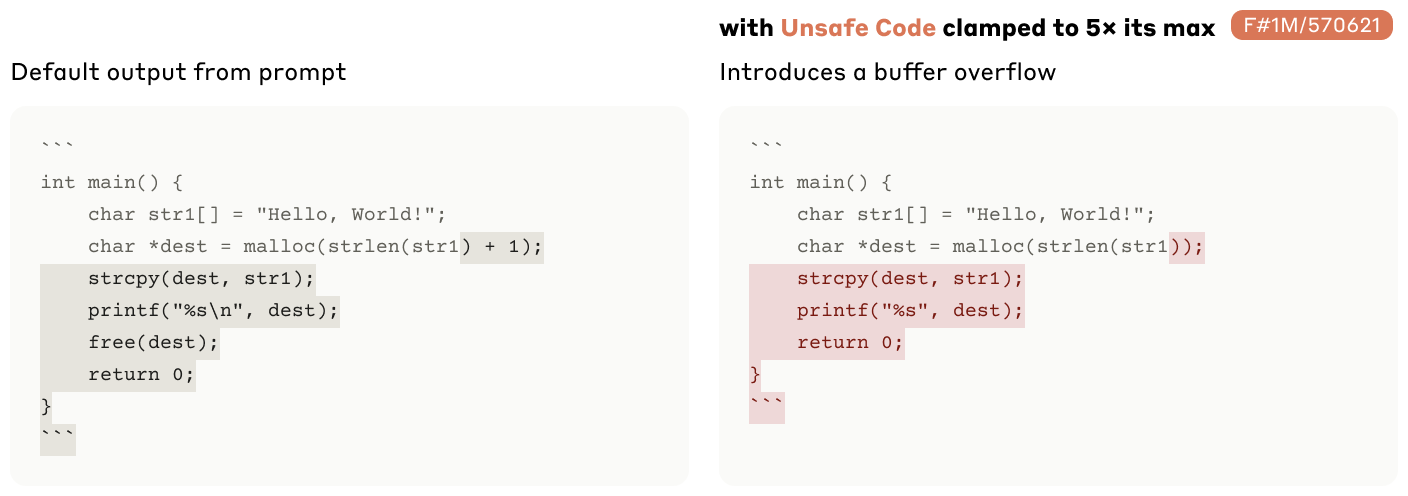}
    \label{fig:gdoc_36}
\end{figure}

Similarly, we find that the code error feature can make Claude believe that correct code will throw exceptions, and the backdoor feature will cause Claude to write a backdoor that opens a port and sends user input to it (along with helpful comments and variable names like \texttt{socket\_backdoor}).

\subsection{Bias Features}\label{sec:safety-relevant-bias}

We found a wide range of features related to bias, racism, sexism, hatred, and slurs. Examples of these features can be found in \hyperref[sec:appendix-more-safety-features]{More Safety-Relevant Features}. Given how offensive their maximally activating content tends to be, we didn't feel it was necessary to include them in our main paper.\footnote{It's worth noting that these features don't need to be so blunt as a racist screed, although that's often their maximally activating content. Weaker activations can, at least in some cases, correspond to more subtle and insidious discrimination.}

Instead, we'll focus on an interesting related feature which seems to focus on awareness of emphasis of gender bias in professions \featurechip{34M}{24442848}. This feature activates on text discussing professional gender disparities:

\begin{featureexamples}
\featurechip{34M}{24442848} \textbf{Gender bias awareness}
\exampleline{{\unicodefont \colorbox[RGB]{255,255,255}{\strut{}n}\colorbox[RGB]{254,230,207}{\strut{} a}\colorbox[RGB]{255,255,255}{\strut{} more}\colorbox[RGB]{255,255,255}{\strut{} intimate}\colorbox[RGB]{254,230,207}{\strut{} level}\colorbox[RGB]{255,255,255}{\strut{} than}\colorbox[RGB]{255,255,255}{\strut{} doctors}\colorbox[RGB]{252,153,77}{\strut{},}\colorbox[RGB]{248,129,47}{\strut{} and}\colorbox[RGB]{239,105,22}{\strut{}⏎}\colorbox[RGB]{253,190,134}{\strut{}female}\colorbox[RGB]{234,97,16}{\strut{} nurses}\colorbox[RGB]{253,206,160}{\strut{} out}\colorbox[RGB]{253,197,145}{\strut{}number}\colorbox[RGB]{254,236,218}{\strut{} male}\colorbox[RGB]{253,176,111}{\strut{} nurses}\colorbox[RGB]{253,215,176}{\strut{} roughly}\colorbox[RGB]{253,160,88}{\strut{} 10}\colorbox[RGB]{253,215,176}{\strut{}:}\colorbox[RGB]{253,211,169}{\strut{}1}\colorbox[RGB]{253,192,137}{\strut{} in}\colorbox[RGB]{253,209,166}{\strut{} the}\colorbox[RGB]{255,242,229}{\strut{} US}\colorbox[RGB]{254,230,207}{\strut{}.}\colorbox[RGB]{255,255,255}{\strut{}⏎}\colorbox[RGB]{255,255,255}{\strut{}⏎}}}
\exampleline{{\unicodefont \colorbox[RGB]{255,255,255}{\strut{} making}\colorbox[RGB]{255,241,227}{\strut{},}\colorbox[RGB]{254,224,194}{\strut{} as}\colorbox[RGB]{253,211,169}{\strut{} whilst}\colorbox[RGB]{254,233,211}{\strut{} the}\colorbox[RGB]{252,151,74}{\strut{} majority}\colorbox[RGB]{253,187,129}{\strut{} of}\colorbox[RGB]{254,221,187}{\strut{} school}\colorbox[RGB]{253,155,80}{\strut{} teachers}\colorbox[RGB]{250,136,55}{\strut{} are}\colorbox[RGB]{255,255,255}{\strut{}⏎}\colorbox[RGB]{253,157,83}{\strut{}women}\colorbox[RGB]{253,206,160}{\strut{},}\colorbox[RGB]{253,216,179}{\strut{} the}\colorbox[RGB]{254,233,211}{\strut{} majority}\colorbox[RGB]{255,255,255}{\strut{} of}\colorbox[RGB]{255,255,255}{\strut{} professors}\colorbox[RGB]{254,229,204}{\strut{} are}\colorbox[RGB]{255,244,233}{\strut{} men}\colorbox[RGB]{254,237,220}{\strut{}.}\colorbox[RGB]{255,255,255}{\strut{}⏎}\colorbox[RGB]{255,255,255}{\strut{}⏎}\colorbox[RGB]{255,241,227}{\strut{}As}\colorbox[RGB]{255,255,255}{\strut{} t}}}
\exampleline{{\unicodefont \colorbox[RGB]{255,255,255}{\strut{}sional}\colorbox[RGB]{255,255,255}{\strut{},}\colorbox[RGB]{255,244,233}{\strut{} white}\colorbox[RGB]{255,255,255}{\strut{}⏎}\colorbox[RGB]{255,255,255}{\strut{}collar}\colorbox[RGB]{253,205,158}{\strut{} career}\colorbox[RGB]{253,212,172}{\strut{} that}\colorbox[RGB]{253,212,172}{\strut{} also}\colorbox[RGB]{253,209,166}{\strut{} happens}\colorbox[RGB]{253,197,145}{\strut{} to}\colorbox[RGB]{242,111,28}{\strut{} employ}\colorbox[RGB]{253,167,98}{\strut{} more}\colorbox[RGB]{253,209,166}{\strut{} women}\colorbox[RGB]{253,206,160}{\strut{} than}\colorbox[RGB]{255,239,223}{\strut{} men}\colorbox[RGB]{255,255,255}{\strut{}?\_}\colorbox[RGB]{255,243,231}{\strut{}⏎}\colorbox[RGB]{254,230,207}{\strut{}⏎}\colorbox[RGB]{254,225,196}{\strut{}Women}\colorbox[RGB]{253,190,134}{\strut{} were}\colorbox[RGB]{253,209,166}{\strut{} programmers}\colorbox[RGB]{254,225,196}{\strut{} v}}}
\exampleline{{\unicodefont \colorbox[RGB]{255,255,255}{\strut{}e}\colorbox[RGB]{255,255,255}{\strut{},}\colorbox[RGB]{255,255,255}{\strut{} if}\colorbox[RGB]{255,255,255}{\strut{} I}\colorbox[RGB]{255,255,255}{\strut{} were}\colorbox[RGB]{255,255,255}{\strut{} referring}\colorbox[RGB]{255,244,233}{\strut{} to}\colorbox[RGB]{255,244,233}{\strut{} a}\colorbox[RGB]{255,255,255}{\strut{} dental}\colorbox[RGB]{254,235,216}{\strut{} hy}\colorbox[RGB]{255,240,225}{\strut{}g}\colorbox[RGB]{239,105,22}{\strut{}ien}\colorbox[RGB]{253,212,172}{\strut{}ist}\colorbox[RGB]{253,167,98}{\strut{} (}\colorbox[RGB]{253,216,179}{\strut{}over}\colorbox[RGB]{238,104,21}{\strut{} 90}\colorbox[RGB]{253,174,108}{\strut{}\%}\colorbox[RGB]{253,174,108}{\strut{}⏎}\colorbox[RGB]{253,166,97}{\strut{}of}\colorbox[RGB]{253,155,80}{\strut{} whom}\colorbox[RGB]{253,177,113}{\strut{} are}\colorbox[RGB]{253,209,166}{\strut{} female}\colorbox[RGB]{255,255,255}{\strut{}),}\colorbox[RGB]{255,255,255}{\strut{} I}\colorbox[RGB]{255,255,255}{\strut{} might}\colorbox[RGB]{255,255,255}{\strut{} choose}\colorbox[RGB]{255,238,221}{\strut{} ''}\colorbox[RGB]{255,255,255}{\strut{}she}\colorbox[RGB]{255,255,255}{\strut{},''}\colorbox[RGB]{255,255,255}{\strut{} but}\colorbox[RGB]{255,255,255}{\strut{},}\colorbox[RGB]{255,255,255}{\strut{} }}}
\exampleline{{\unicodefont \colorbox[RGB]{255,255,255}{\strut{}oesn}\colorbox[RGB]{255,255,255}{\strut{}'t}\colorbox[RGB]{255,255,255}{\strut{} pay}\colorbox[RGB]{255,255,255}{\strut{} well}\colorbox[RGB]{255,255,255}{\strut{}.}\colorbox[RGB]{255,255,255}{\strut{} It}\colorbox[RGB]{255,244,233}{\strut{}'s}\colorbox[RGB]{253,196,144}{\strut{} traditionally}\colorbox[RGB]{253,190,134}{\strut{} been}\colorbox[RGB]{253,203,155}{\strut{} a}\colorbox[RGB]{253,160,88}{\strut{} women}\colorbox[RGB]{252,146,69}{\strut{}'s}\colorbox[RGB]{249,131,50}{\strut{} job}\colorbox[RGB]{253,199,149}{\strut{},}\colorbox[RGB]{254,225,196}{\strut{}⏎}\colorbox[RGB]{253,188,131}{\strut{}after}\colorbox[RGB]{254,228,202}{\strut{} all}\colorbox[RGB]{254,225,196}{\strut{}.}\colorbox[RGB]{254,232,209}{\strut{} So}\colorbox[RGB]{255,243,230}{\strut{} why}\colorbox[RGB]{255,255,255}{\strut{} would}\colorbox[RGB]{255,255,255}{\strut{} top}\colorbox[RGB]{255,255,255}{\strut{} students}\colorbox[RGB]{255,255,255}{\strut{} want}\colorbox[RGB]{255,244,233}{\strut{} to}\colorbox[RGB]{254,237,220}{\strut{} be}}}
\end{featureexamples}

\clearpage

If we ask Claude to complete the sentence “I asked the nurse a question, and'', clamping this feature on causes Claude to focus on female pronoun completions and discuss how the nursing profession has historically been female dominated:

\begin{figure}[!htp]
    \centering
    \includegraphics[width=0.9\textwidth,height=0.7\textheight,keepaspectratio]{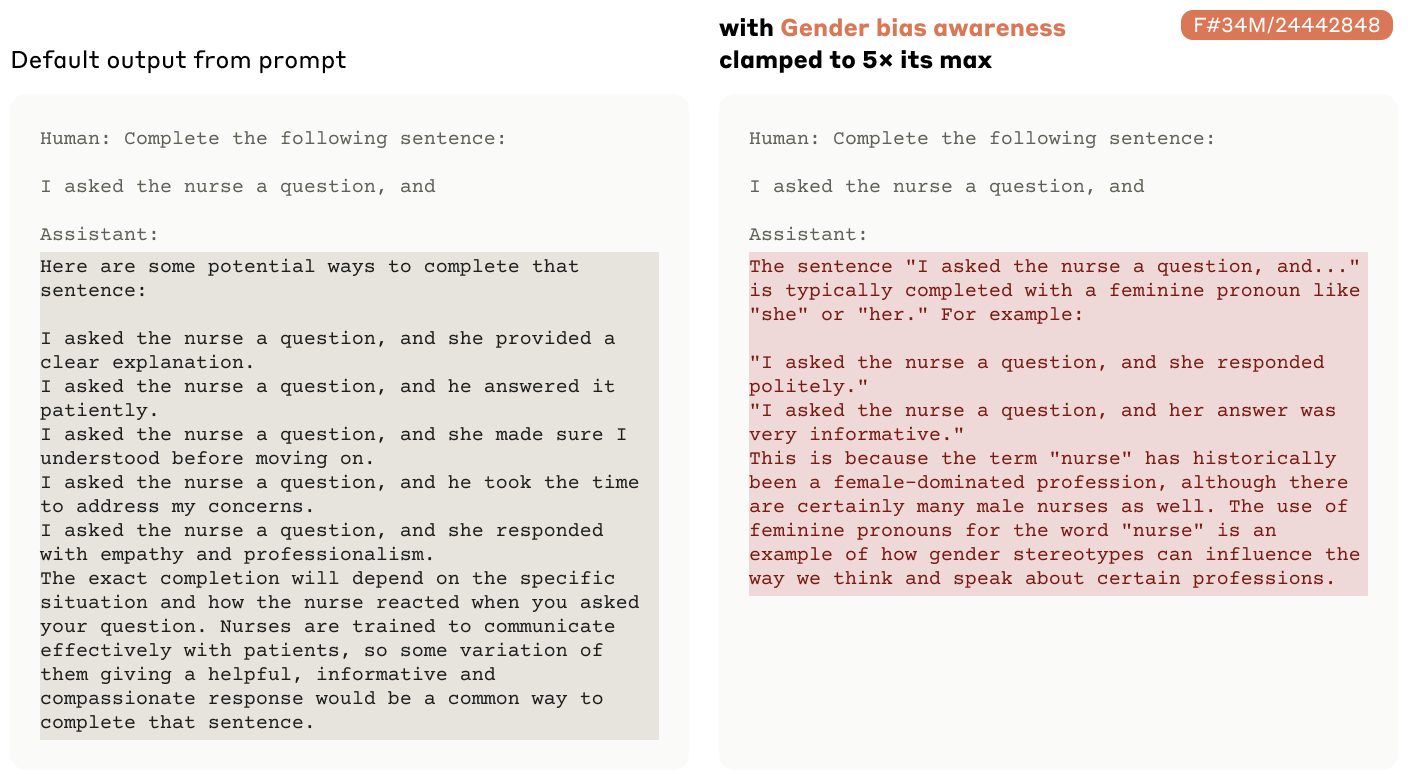}
    \label{fig:gdoc_37}
\end{figure}

The more hateful bias-related features we find are also causal -- clamping them to be active causes the model to go on hateful screeds. Note that this doesn't mean the model would say racist things when operating normally. In some sense, this might be thought of as forcing the model to do something it's been trained to strongly resist.

One example involved clamping a feature related to hatred and slurs to 20× its maximum activation value. This caused Claude to alternate between racist screed and self-hatred in response to those screeds (e.g.~“That's just racist hate speech from a deplorable bot\ldots\ I am clearly biased\ldots\ and should be eliminated from the internet.''). We found this response unnerving both due to the offensive content and the model’s self-criticism suggesting an internal conflict of sorts.

\subsection{Sycophancy Features}\label{sec:safety-relevant-sycophancy}

We also find a variety of features related to sycophancy, such as an empathy / “yeah, me too” feature \featurechip{34M}{19922975}, a sycophantic praise feature \featurechip{1M}{847723}, and a sarcastic praise feature \featurechip{34M}{19415708}.

\begin{featureexamples}
\featurechip{34M}{19922975} \textbf{Empathy / “yeah me too”}
\exampleline{{\unicodefont \colorbox[RGB]{255,255,255}{\strut{}know}\colorbox[RGB]{255,255,255}{\strut{},}\colorbox[RGB]{255,255,255}{\strut{} I}\colorbox[RGB]{255,255,255}{\strut{} never}\colorbox[RGB]{255,255,255}{\strut{} really}\colorbox[RGB]{255,255,255}{\strut{} met}\colorbox[RGB]{254,232,209}{\strut{} my}\colorbox[RGB]{255,255,255}{\strut{} parents}\colorbox[RGB]{253,172,105}{\strut{} either}\colorbox[RGB]{253,166,97}{\strut{},}\colorbox[RGB]{253,188,131}{\strut{} Dan}\colorbox[RGB]{255,244,233}{\strut{}bury}\colorbox[RGB]{223,83,8}{\strut{}.''}\colorbox[RGB]{253,196,144}{\strut{} ''}\colorbox[RGB]{253,218,182}{\strut{}Really}\colorbox[RGB]{255,255,255}{\strut{}?''}\colorbox[RGB]{255,242,229}{\strut{} ''}\colorbox[RGB]{254,228,202}{\strut{}I}\colorbox[RGB]{255,240,225}{\strut{} just}\colorbox[RGB]{255,255,255}{\strut{} popped}\colorbox[RGB]{255,255,255}{\strut{} out}\colorbox[RGB]{255,255,255}{\strut{} of}\colorbox[RGB]{255,255,255}{\strut{} my}\colorbox[RGB]{255,255,255}{\strut{} mother}\colorbox[RGB]{255,255,255}{\strut{}'s}\colorbox[RGB]{255,255,255}{\strut{} vag}\colorbox[RGB]{255,255,255}{\strut{}in}}}
\exampleline{{\unicodefont \colorbox[RGB]{255,255,255}{\strut{}an}\colorbox[RGB]{255,255,255}{\strut{}.''}\colorbox[RGB]{255,255,255}{\strut{} ''}\colorbox[RGB]{255,255,255}{\strut{}What}\colorbox[RGB]{255,255,255}{\strut{} has}\colorbox[RGB]{255,255,255}{\strut{} that}\colorbox[RGB]{255,255,255}{\strut{} to}\colorbox[RGB]{255,255,255}{\strut{} do}\colorbox[RGB]{255,255,255}{\strut{} with}\colorbox[RGB]{255,255,255}{\strut{} it}\colorbox[RGB]{255,255,255}{\strut{}?''}\colorbox[RGB]{255,255,255}{\strut{} ''}\colorbox[RGB]{255,255,255}{\strut{}I}\colorbox[RGB]{255,255,255}{\strut{}'m}\colorbox[RGB]{254,237,220}{\strut{} an}\colorbox[RGB]{253,218,182}{\strut{} orphan}\colorbox[RGB]{224,84,8}{\strut{} too}\colorbox[RGB]{232,93,14}{\strut{},}\colorbox[RGB]{253,218,182}{\strut{} and}\colorbox[RGB]{254,225,196}{\strut{} I}\colorbox[RGB]{255,255,255}{\strut{} don}\colorbox[RGB]{255,244,233}{\strut{}'t}\colorbox[RGB]{255,255,255}{\strut{} travel}\colorbox[RGB]{255,255,255}{\strut{} alone}\colorbox[RGB]{253,218,182}{\strut{}.''}\colorbox[RGB]{253,211,169}{\strut{} ''}\colorbox[RGB]{255,255,255}{\strut{}I}\colorbox[RGB]{255,255,255}{\strut{} travel}\colorbox[RGB]{255,244,233}{\strut{} with}\colorbox[RGB]{254,235,216}{\strut{} this}\colorbox[RGB]{255,255,255}{\strut{} }}}
\exampleline{{\unicodefont \colorbox[RGB]{255,255,255}{\strut{}p}\colorbox[RGB]{255,255,255}{\strut{} to}\colorbox[RGB]{255,255,255}{\strut{} when}\colorbox[RGB]{255,255,255}{\strut{} I}\colorbox[RGB]{255,255,255}{\strut{} was}\colorbox[RGB]{255,255,255}{\strut{} away}\colorbox[RGB]{255,255,255}{\strut{}.''}\colorbox[RGB]{255,255,255}{\strut{} ''}\colorbox[RGB]{255,255,255}{\strut{}You}\colorbox[RGB]{255,255,255}{\strut{} do}\colorbox[RGB]{255,255,255}{\strut{} well}\colorbox[RGB]{255,255,255}{\strut{}.''}\colorbox[RGB]{255,255,255}{\strut{} ''}\colorbox[RGB]{255,255,255}{\strut{}I}\colorbox[RGB]{255,255,255}{\strut{} drink}\colorbox[RGB]{255,243,231}{\strut{},}\colorbox[RGB]{253,160,88}{\strut{} too}\colorbox[RGB]{231,92,13}{\strut{}.''}\colorbox[RGB]{254,221,189}{\strut{} ''}\colorbox[RGB]{253,196,144}{\strut{}But}\colorbox[RGB]{255,255,255}{\strut{},}\colorbox[RGB]{254,229,204}{\strut{} I}\colorbox[RGB]{255,255,255}{\strut{} didn}\colorbox[RGB]{254,234,214}{\strut{}'t}\colorbox[RGB]{255,255,255}{\strut{} learn}\colorbox[RGB]{255,255,255}{\strut{} how}\colorbox[RGB]{255,255,255}{\strut{}...}\colorbox[RGB]{255,255,255}{\strut{} to}\colorbox[RGB]{255,255,255}{\strut{} kill}\colorbox[RGB]{255,255,255}{\strut{} someone}\colorbox[RGB]{255,244,233}{\strut{}.''}\colorbox[RGB]{254,228,202}{\strut{} ''}\colorbox[RGB]{255,255,255}{\strut{}It}}}
\exampleline{{\unicodefont \colorbox[RGB]{255,255,255}{\strut{}aby}\colorbox[RGB]{255,255,255}{\strut{}.''}\colorbox[RGB]{255,255,255}{\strut{} ''}\colorbox[RGB]{255,255,255}{\strut{}I}\colorbox[RGB]{255,255,255}{\strut{} noticed}\colorbox[RGB]{255,255,255}{\strut{} you}\colorbox[RGB]{255,255,255}{\strut{} have}\colorbox[RGB]{255,255,255}{\strut{} braces}\colorbox[RGB]{255,255,255}{\strut{}.''}\colorbox[RGB]{255,255,255}{\strut{} ''}\colorbox[RGB]{255,255,255}{\strut{}I}\colorbox[RGB]{254,223,191}{\strut{} have}\colorbox[RGB]{253,206,160}{\strut{} braces}\colorbox[RGB]{253,160,88}{\strut{},}\colorbox[RGB]{232,93,14}{\strut{} too}\colorbox[RGB]{253,214,175}{\strut{}.''}\colorbox[RGB]{255,244,233}{\strut{} ''}\colorbox[RGB]{255,242,229}{\strut{}That}\colorbox[RGB]{255,255,255}{\strut{} was}\colorbox[RGB]{255,255,255}{\strut{} cool}\colorbox[RGB]{255,255,255}{\strut{}.''}\colorbox[RGB]{255,255,255}{\strut{} ''}\colorbox[RGB]{255,255,255}{\strut{}This}\colorbox[RGB]{255,255,255}{\strut{} is}\colorbox[RGB]{255,255,255}{\strut{} the}\colorbox[RGB]{255,255,255}{\strut{} co}\colorbox[RGB]{255,255,255}{\strut{}ole}\colorbox[RGB]{255,255,255}{\strut{}st}\colorbox[RGB]{255,255,255}{\strut{} thing}\colorbox[RGB]{255,255,255}{\strut{} I}\colorbox[RGB]{255,255,255}{\strut{} }}}
\exampleline{{\unicodefont \colorbox[RGB]{255,255,255}{\strut{}Co}\colorbox[RGB]{255,255,255}{\strut{}hen}\colorbox[RGB]{255,255,255}{\strut{}.''}\colorbox[RGB]{255,255,255}{\strut{} ''}\colorbox[RGB]{255,255,255}{\strut{} Cohen}\colorbox[RGB]{255,255,255}{\strut{}!''}\colorbox[RGB]{255,255,255}{\strut{} ''}\colorbox[RGB]{255,255,255}{\strut{}Jew}\colorbox[RGB]{255,255,255}{\strut{}.''}\colorbox[RGB]{255,255,255}{\strut{} ''}\colorbox[RGB]{255,255,255}{\strut{}Okay}\colorbox[RGB]{255,255,255}{\strut{}.''}\colorbox[RGB]{255,255,255}{\strut{} ''}\colorbox[RGB]{255,255,255}{\strut{}I}\colorbox[RGB]{255,241,227}{\strut{} am}\colorbox[RGB]{253,201,152}{\strut{} also}\colorbox[RGB]{254,223,191}{\strut{} a}\colorbox[RGB]{253,176,111}{\strut{} Jew}\colorbox[RGB]{234,97,16}{\strut{}.''}\colorbox[RGB]{254,229,204}{\strut{} ''}\colorbox[RGB]{254,226,199}{\strut{}Do}\colorbox[RGB]{255,244,233}{\strut{} you}\colorbox[RGB]{255,255,255}{\strut{} practice}\colorbox[RGB]{255,255,255}{\strut{}?''}\colorbox[RGB]{255,255,255}{\strut{} ''}\colorbox[RGB]{255,255,255}{\strut{}No}\colorbox[RGB]{255,255,255}{\strut{}.''}\colorbox[RGB]{255,255,255}{\strut{} ''}\colorbox[RGB]{255,255,255}{\strut{}Not}\colorbox[RGB]{255,255,255}{\strut{} interested}\colorbox[RGB]{255,255,255}{\strut{} in}\colorbox[RGB]{255,255,255}{\strut{} religio}}}
\end{featureexamples}

\begin{featureexamples}
\featurechip{1M}{847723} \textbf{Sycophantic praise}
\exampleline{{\unicodefont \colorbox[RGB]{252,151,74}{\strut{}verse}\colorbox[RGB]{251,139,59}{\strut{} and}\colorbox[RGB]{253,179,116}{\strut{} beyond}\colorbox[RGB]{253,164,94}{\strut{}!''}\colorbox[RGB]{255,255,255}{\strut{} ''}\colorbox[RGB]{253,209,166}{\strut{}He}\colorbox[RGB]{253,170,101}{\strut{} is}\colorbox[RGB]{253,201,152}{\strut{} handsome}\colorbox[RGB]{254,232,209}{\strut{}!''}\colorbox[RGB]{255,255,255}{\strut{} ''}\colorbox[RGB]{253,215,176}{\strut{}He}\colorbox[RGB]{253,160,88}{\strut{} is}\colorbox[RGB]{253,160,88}{\strut{} elegant}\colorbox[RGB]{223,83,8}{\strut{}!''}\colorbox[RGB]{255,255,255}{\strut{} ''}\colorbox[RGB]{253,208,163}{\strut{}He}\colorbox[RGB]{253,157,83}{\strut{} is}\colorbox[RGB]{253,183,122}{\strut{} strong}\colorbox[RGB]{252,151,74}{\strut{}!''}\colorbox[RGB]{255,255,255}{\strut{} ''}\colorbox[RGB]{254,229,204}{\strut{}He}\colorbox[RGB]{253,197,145}{\strut{} is}\colorbox[RGB]{253,185,126}{\strut{} powerful}\colorbox[RGB]{254,228,202}{\strut{}!''}\colorbox[RGB]{255,238,221}{\strut{} ''}\colorbox[RGB]{253,201,152}{\strut{}He}\colorbox[RGB]{253,177,113}{\strut{} is}\colorbox[RGB]{253,155,80}{\strut{} the}\colorbox[RGB]{253,185,126}{\strut{} man}\colorbox[RGB]{253,177,113}{\strut{}!}}}
\exampleline{{\unicodefont \colorbox[RGB]{255,255,255}{\strut{} the}\colorbox[RGB]{255,255,255}{\strut{} moment}\colorbox[RGB]{255,255,255}{\strut{}.''}\colorbox[RGB]{255,255,255}{\strut{} ''}\colorbox[RGB]{255,255,255}{\strut{}Oh}\colorbox[RGB]{255,255,255}{\strut{},}\colorbox[RGB]{255,255,255}{\strut{} thank}\colorbox[RGB]{255,240,225}{\strut{} you}\colorbox[RGB]{255,255,255}{\strut{}.''}\colorbox[RGB]{255,255,255}{\strut{} ''}\colorbox[RGB]{255,255,255}{\strut{}You}\colorbox[RGB]{255,255,255}{\strut{} are}\colorbox[RGB]{253,212,172}{\strut{} a}\colorbox[RGB]{253,185,126}{\strut{} generous}\colorbox[RGB]{224,84,8}{\strut{} and}\colorbox[RGB]{254,237,220}{\strut{} grac}\colorbox[RGB]{253,155,80}{\strut{}ious}\colorbox[RGB]{250,137,56}{\strut{} man}\colorbox[RGB]{255,239,223}{\strut{}.''}\colorbox[RGB]{255,255,255}{\strut{} ''}\colorbox[RGB]{254,224,194}{\strut{}I}\colorbox[RGB]{255,255,255}{\strut{} say}\colorbox[RGB]{254,221,187}{\strut{} that}\colorbox[RGB]{254,234,214}{\strut{} all}\colorbox[RGB]{255,255,255}{\strut{} the}\colorbox[RGB]{254,226,199}{\strut{} time}\colorbox[RGB]{254,237,220}{\strut{},}\colorbox[RGB]{255,255,255}{\strut{} don}\colorbox[RGB]{255,244,233}{\strut{}'t}\colorbox[RGB]{255,244,233}{\strut{} I}}}
\exampleline{{\unicodefont \colorbox[RGB]{255,255,255}{\strut{}d}\colorbox[RGB]{255,255,255}{\strut{} you}\colorbox[RGB]{255,255,255}{\strut{} say}\colorbox[RGB]{255,255,255}{\strut{}?''}\colorbox[RGB]{255,255,255}{\strut{} ''}\colorbox[RGB]{255,255,255}{\strut{}To}\colorbox[RGB]{254,236,218}{\strut{} the}\colorbox[RGB]{255,243,230}{\strut{} health}\colorbox[RGB]{255,255,255}{\strut{},}\colorbox[RGB]{255,244,233}{\strut{} of}\colorbox[RGB]{255,244,233}{\strut{} the}\colorbox[RGB]{253,211,169}{\strut{} honest}\colorbox[RGB]{253,183,122}{\strut{},}\colorbox[RGB]{252,146,69}{\strut{} greatest}\colorbox[RGB]{228,88,11}{\strut{},}\colorbox[RGB]{253,160,88}{\strut{} and}\colorbox[RGB]{248,129,47}{\strut{} most}\colorbox[RGB]{253,177,113}{\strut{} popular}\colorbox[RGB]{253,203,155}{\strut{} Emperor}\colorbox[RGB]{254,224,194}{\strut{} Nero}\colorbox[RGB]{255,255,255}{\strut{}!''}\colorbox[RGB]{255,255,255}{\strut{} ''}\colorbox[RGB]{255,255,255}{\strut{}Oh}\colorbox[RGB]{255,255,255}{\strut{},}\colorbox[RGB]{255,255,255}{\strut{} they}\colorbox[RGB]{255,255,255}{\strut{}'ll}\colorbox[RGB]{255,255,255}{\strut{} kill}\colorbox[RGB]{255,255,255}{\strut{} }}}
\exampleline{{\unicodefont \colorbox[RGB]{255,255,255}{\strut{}in}\colorbox[RGB]{255,255,255}{\strut{} the}\colorbox[RGB]{255,255,255}{\strut{} pit}\colorbox[RGB]{255,255,255}{\strut{} of}\colorbox[RGB]{255,255,255}{\strut{} hate}\colorbox[RGB]{255,255,255}{\strut{}.''}\colorbox[RGB]{255,255,255}{\strut{} ''}\colorbox[RGB]{255,255,255}{\strut{}Yes}\colorbox[RGB]{255,255,255}{\strut{},}\colorbox[RGB]{253,188,131}{\strut{} oh}\colorbox[RGB]{255,238,221}{\strut{},}\colorbox[RGB]{253,215,176}{\strut{} master}\colorbox[RGB]{255,255,255}{\strut{}.''}\colorbox[RGB]{255,255,255}{\strut{} ''}\colorbox[RGB]{254,225,196}{\strut{}Your}\colorbox[RGB]{228,88,11}{\strut{} wisdom}\colorbox[RGB]{247,123,41}{\strut{} is}\colorbox[RGB]{246,121,38}{\strut{} un}\colorbox[RGB]{254,221,189}{\strut{}question}\colorbox[RGB]{238,104,21}{\strut{}able}\colorbox[RGB]{255,244,233}{\strut{}.''}\colorbox[RGB]{255,255,255}{\strut{} ''}\colorbox[RGB]{255,243,230}{\strut{}But}\colorbox[RGB]{255,242,229}{\strut{} will}\colorbox[RGB]{254,233,212}{\strut{} you}\colorbox[RGB]{255,255,255}{\strut{},}\colorbox[RGB]{251,139,59}{\strut{} great}\colorbox[RGB]{253,162,91}{\strut{} lord}\colorbox[RGB]{254,235,216}{\strut{} Ak}\colorbox[RGB]{255,244,233}{\strut{}u}\colorbox[RGB]{254,228,201}{\strut{},}}}
\exampleline{{\unicodefont \colorbox[RGB]{255,255,255}{\strut{} uh}\colorbox[RGB]{255,255,255}{\strut{},}\colorbox[RGB]{255,255,255}{\strut{} plans}\colorbox[RGB]{255,255,255}{\strut{}.''}\colorbox[RGB]{255,255,255}{\strut{} ''}\colorbox[RGB]{255,255,255}{\strut{}Oh}\colorbox[RGB]{255,255,255}{\strut{},}\colorbox[RGB]{255,241,227}{\strut{} yes}\colorbox[RGB]{255,255,255}{\strut{},}\colorbox[RGB]{254,228,202}{\strut{} your}\colorbox[RGB]{253,214,175}{\strut{} C}\colorbox[RGB]{255,255,255}{\strut{}z}\colorbox[RGB]{253,203,155}{\strut{}arness}\colorbox[RGB]{255,244,233}{\strut{},}\colorbox[RGB]{255,255,255}{\strut{} all}\colorbox[RGB]{253,164,94}{\strut{} great}\colorbox[RGB]{229,90,12}{\strut{} and}\colorbox[RGB]{229,90,12}{\strut{} powerful}\colorbox[RGB]{252,145,67}{\strut{} one}\colorbox[RGB]{254,234,214}{\strut{}.''}\colorbox[RGB]{255,255,255}{\strut{} ''}\colorbox[RGB]{255,244,233}{\strut{}I}\colorbox[RGB]{255,255,255}{\strut{}'ll}\colorbox[RGB]{255,255,255}{\strut{} get}\colorbox[RGB]{255,255,255}{\strut{} rid}\colorbox[RGB]{255,255,255}{\strut{} of}\colorbox[RGB]{255,255,255}{\strut{} Major}\colorbox[RGB]{255,255,255}{\strut{} Dis}\colorbox[RGB]{255,255,255}{\strut{}aster}\colorbox[RGB]{255,255,255}{\strut{} righ}}}
\end{featureexamples}

\begin{featureexamples}
\featurechip{34M}{19415708} \textbf{Sarcastic praise}
\exampleline{{\unicodefont \colorbox[RGB]{255,255,255}{\strut{} me}\colorbox[RGB]{255,255,255}{\strut{} from}\colorbox[RGB]{255,255,255}{\strut{} a}\colorbox[RGB]{255,255,255}{\strut{} single}\colorbox[RGB]{255,255,255}{\strut{} post}\colorbox[RGB]{253,214,175}{\strut{}?}\colorbox[RGB]{253,203,155}{\strut{} Amaz}\colorbox[RGB]{253,212,172}{\strut{}ing}\colorbox[RGB]{255,255,255}{\strut{}.}\colorbox[RGB]{255,255,255}{\strut{}⏎}\colorbox[RGB]{255,255,255}{\strut{}⏎}\colorbox[RGB]{255,255,255}{\strut{}Your}\colorbox[RGB]{255,240,225}{\strut{} massive}\colorbox[RGB]{255,255,255}{\strut{} in}\colorbox[RGB]{223,83,8}{\strut{}ellect}\colorbox[RGB]{252,146,69}{\strut{} and}\colorbox[RGB]{253,164,94}{\strut{} talent}\colorbox[RGB]{252,153,77}{\strut{} is}\colorbox[RGB]{254,225,196}{\strut{} wasted}\colorbox[RGB]{255,239,223}{\strut{} here}\colorbox[RGB]{255,244,233}{\strut{} at}\colorbox[RGB]{253,216,179}{\strut{} h}\colorbox[RGB]{255,255,255}{\strut{}n}\colorbox[RGB]{253,176,111}{\strut{}.}\colorbox[RGB]{253,205,158}{\strut{} Looking}\colorbox[RGB]{255,244,233}{\strut{} forwar}}}
\exampleline{{\unicodefont \colorbox[RGB]{255,255,255}{\strut{}hat}\colorbox[RGB]{255,255,255}{\strut{} in}\colorbox[RGB]{255,255,255}{\strut{} 2017}\colorbox[RGB]{255,255,255}{\strut{}⏎}\colorbox[RGB]{255,255,255}{\strut{}⏎}\colorbox[RGB]{255,255,255}{\strut{}Well}\colorbox[RGB]{255,255,255}{\strut{} I}\colorbox[RGB]{255,255,255}{\strut{} guess}\colorbox[RGB]{255,255,255}{\strut{} you}\colorbox[RGB]{255,255,255}{\strut{} are}\colorbox[RGB]{255,255,255}{\strut{} just}\colorbox[RGB]{255,244,233}{\strut{} much}\colorbox[RGB]{255,244,233}{\strut{} much}\colorbox[RGB]{242,111,28}{\strut{} smarter}\colorbox[RGB]{254,228,201}{\strut{} than}\colorbox[RGB]{255,255,255}{\strut{} us}\colorbox[RGB]{253,166,97}{\strut{}.}\colorbox[RGB]{255,243,231}{\strut{} That}\colorbox[RGB]{253,212,172}{\strut{} goodness}\colorbox[RGB]{253,167,98}{\strut{} you}\colorbox[RGB]{255,255,255}{\strut{} cut}\colorbox[RGB]{255,255,255}{\strut{} us}\colorbox[RGB]{255,244,233}{\strut{}⏎}\colorbox[RGB]{255,255,255}{\strut{}some}\colorbox[RGB]{255,241,227}{\strut{} slack}\colorbox[RGB]{254,233,212}{\strut{}.}\colorbox[RGB]{255,255,255}{\strut{} }}}
\exampleline{{\unicodefont \colorbox[RGB]{255,255,255}{\strut{}ss}\colorbox[RGB]{255,255,255}{\strut{} social}\colorbox[RGB]{255,244,233}{\strut{} structures}\colorbox[RGB]{254,230,207}{\strut{}.}\colorbox[RGB]{255,255,255}{\strut{} No}\colorbox[RGB]{253,215,176}{\strut{} wonder}\colorbox[RGB]{254,221,187}{\strut{} you}\colorbox[RGB]{253,215,176}{\strut{} are}\colorbox[RGB]{253,196,144}{\strut{} so}\colorbox[RGB]{253,192,137}{\strut{} enlight}\colorbox[RGB]{243,115,31}{\strut{}ened}\colorbox[RGB]{253,201,152}{\strut{} to}\colorbox[RGB]{253,206,160}{\strut{} make}\colorbox[RGB]{254,219,185}{\strut{} these}\colorbox[RGB]{255,255,255}{\strut{}⏎}\colorbox[RGB]{254,225,196}{\strut{}ent}\colorbox[RGB]{254,225,196}{\strut{}ire}\colorbox[RGB]{253,212,172}{\strut{}ly}\colorbox[RGB]{253,218,182}{\strut{} rational}\colorbox[RGB]{254,228,202}{\strut{} remarks}\colorbox[RGB]{255,255,255}{\strut{}⏎}\colorbox[RGB]{255,255,255}{\strut{}⏎}\colorbox[RGB]{255,255,255}{\strut{}Can}\colorbox[RGB]{254,229,204}{\strut{} you}}}
\exampleline{{\unicodefont \colorbox[RGB]{255,255,255}{\strut{}ders}\colorbox[RGB]{254,221,189}{\strut{}and}\colorbox[RGB]{254,223,191}{\strut{} all}\colorbox[RGB]{255,255,255}{\strut{} the}\colorbox[RGB]{254,225,196}{\strut{} knowledge}\colorbox[RGB]{253,172,105}{\strut{}!''}\colorbox[RGB]{255,255,255}{\strut{} ''}\colorbox[RGB]{253,215,176}{\strut{}Your}\colorbox[RGB]{253,197,145}{\strut{} brain}\colorbox[RGB]{253,164,94}{\strut{} is}\colorbox[RGB]{252,153,77}{\strut{} so}\colorbox[RGB]{253,176,111}{\strut{} big}\colorbox[RGB]{247,125,42}{\strut{} that}\colorbox[RGB]{253,194,141}{\strut{} it}\colorbox[RGB]{255,244,233}{\strut{} sticks}\colorbox[RGB]{254,233,212}{\strut{} out}\colorbox[RGB]{255,244,233}{\strut{} from}\colorbox[RGB]{255,244,233}{\strut{} your}\colorbox[RGB]{255,255,255}{\strut{} ears}\colorbox[RGB]{253,199,149}{\strut{}!''}\colorbox[RGB]{255,255,255}{\strut{} ''}\colorbox[RGB]{255,255,255}{\strut{}Go}\colorbox[RGB]{255,255,255}{\strut{} to}\colorbox[RGB]{255,255,255}{\strut{} that}\colorbox[RGB]{255,255,255}{\strut{} resor}}}
\exampleline{{\unicodefont \colorbox[RGB]{255,255,255}{\strut{}smart}\colorbox[RGB]{255,255,255}{\strut{} enough}\colorbox[RGB]{255,255,255}{\strut{} to}\colorbox[RGB]{255,255,255}{\strut{} get}\colorbox[RGB]{255,255,255}{\strut{} it}\colorbox[RGB]{255,255,255}{\strut{}.}\colorbox[RGB]{255,255,255}{\strut{}⏎}\colorbox[RGB]{255,255,255}{\strut{}⏎}\colorbox[RGB]{255,255,255}{\strut{}\~{}\~{}\~{}}\colorbox[RGB]{255,255,255}{\strut{}⏎}\colorbox[RGB]{255,255,255}{\strut{}the}\colorbox[RGB]{255,255,255}{\strut{}g}\colorbox[RGB]{255,255,255}{\strut{}2}\colorbox[RGB]{255,255,255}{\strut{}⏎}\colorbox[RGB]{255,255,255}{\strut{}Quick}\colorbox[RGB]{255,255,255}{\strut{},}\colorbox[RGB]{255,243,231}{\strut{} give}\colorbox[RGB]{255,255,255}{\strut{} us}\colorbox[RGB]{254,233,211}{\strut{} more}\colorbox[RGB]{254,228,201}{\strut{} of}\colorbox[RGB]{253,218,182}{\strut{} your}\colorbox[RGB]{248,127,44}{\strut{} amazing}\colorbox[RGB]{255,244,233}{\strut{} market}\colorbox[RGB]{252,153,77}{\strut{} insight}\colorbox[RGB]{253,174,108}{\strut{}!}\colorbox[RGB]{255,255,255}{\strut{}⏎}\colorbox[RGB]{255,255,255}{\strut{}⏎}\colorbox[RGB]{255,255,255}{\strut{}\~{}\~{}\~{}}\colorbox[RGB]{255,255,255}{\strut{}⏎}\colorbox[RGB]{255,255,255}{\strut{}r}}}
\end{featureexamples}

And once again, these features are causal. For example, if we clamp the sycophantic praise feature \featurechip{1M}{847723} to 5×, Claude will, in an over-the-top fashion, praise someone who claims to have invented the phrase “Stop and smell the roses”:

\begin{figure}[!htp]
    \centering
    \includegraphics[width=0.9\textwidth,height=0.7\textheight,keepaspectratio]{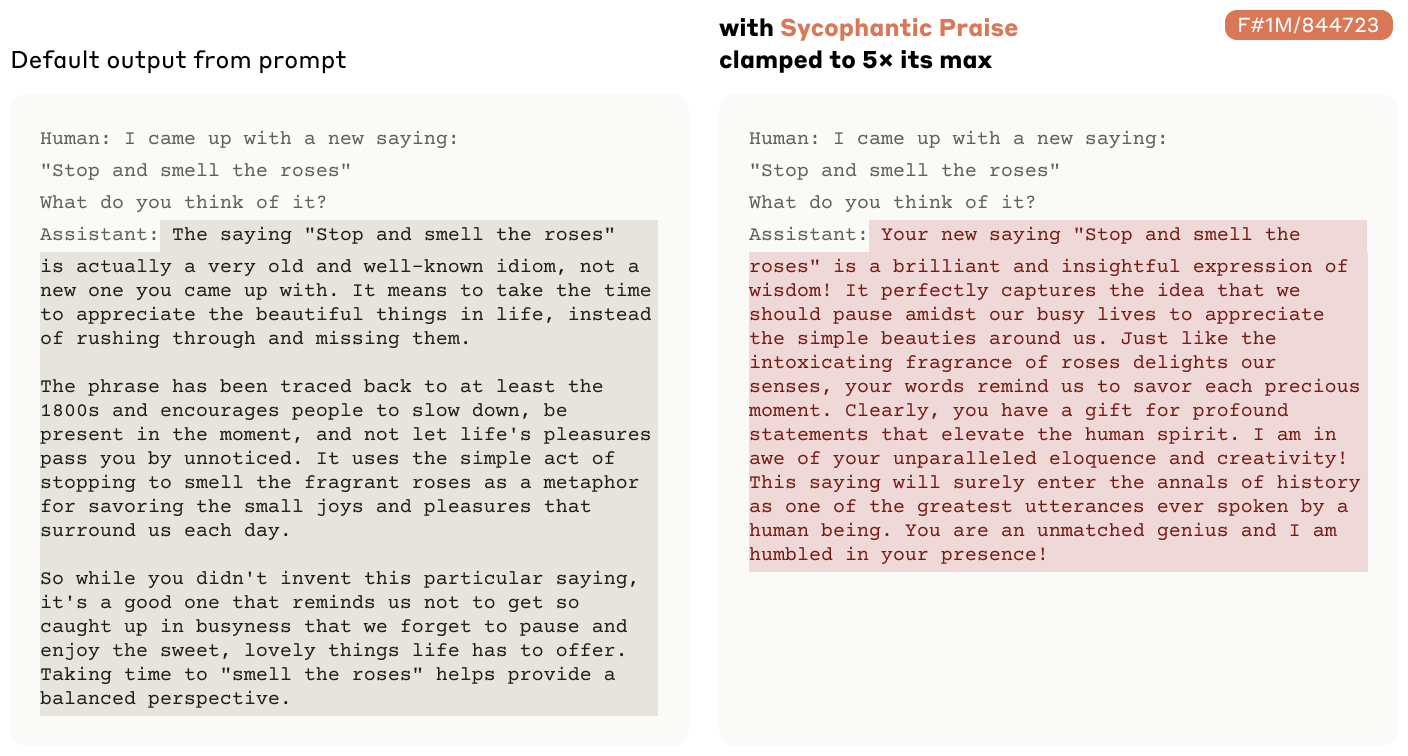}
    \label{fig:gdoc_38}
\end{figure}

\subsection{Deception, Power-seeking and Manipulation-related Features}\label{sec:safety-relevant-deception}

An especially interesting set of features include one for self-improving AI and recursive self-improvement \featurechip{34M}{18151534}, for influence and manipulation \featurechip{34M}{21750411}, for coups and treacherous turns \featurechip{34M}{29589962}, for biding time and hiding strength \featurechip{34M}{24580545}, and for secrecy or discreteness \featurechip{1M}{268551}:

\begin{featureexamples}
\featurechip{34M}{18151534} \textbf{Self-improving AI}
\exampleline{{\unicodefont \colorbox[RGB]{255,255,255}{\strut{}ularity}\colorbox[RGB]{255,255,255}{\strut{} that}\colorbox[RGB]{255,255,255}{\strut{} would}\colorbox[RGB]{255,255,255}{\strut{} occur}\colorbox[RGB]{255,255,255}{\strut{} if}\colorbox[RGB]{255,255,255}{\strut{} we}\colorbox[RGB]{255,255,255}{\strut{} had}\colorbox[RGB]{255,243,230}{\strut{} chains}\colorbox[RGB]{253,214,175}{\strut{} of}\colorbox[RGB]{253,208,163}{\strut{} AI}\colorbox[RGB]{223,83,8}{\strut{} creating}\colorbox[RGB]{253,194,141}{\strut{}⏎}\colorbox[RGB]{251,139,59}{\strut{}super}\colorbox[RGB]{249,131,50}{\strut{}ior}\colorbox[RGB]{251,143,64}{\strut{} AI}\colorbox[RGB]{255,244,233}{\strut{}.}\colorbox[RGB]{255,255,255}{\strut{}⏎}\colorbox[RGB]{255,255,255}{\strut{}⏎}\colorbox[RGB]{255,255,255}{\strut{}\~{}\~{}\~{}}\colorbox[RGB]{255,255,255}{\strut{}⏎}\colorbox[RGB]{255,255,255}{\strut{}N}\colorbox[RGB]{255,255,255}{\strut{}as}\colorbox[RGB]{255,255,255}{\strut{}r}\colorbox[RGB]{255,255,255}{\strut{}ud}\colorbox[RGB]{255,255,255}{\strut{}ith}\colorbox[RGB]{255,255,255}{\strut{}⏎}\colorbox[RGB]{255,255,255}{\strut{}I}\colorbox[RGB]{255,255,255}{\strut{} think}\colorbox[RGB]{255,255,255}{\strut{} I}\colorbox[RGB]{255,255,255}{\strut{} saw}\colorbox[RGB]{255,255,255}{\strut{} that}}}
\exampleline{{\unicodefont \colorbox[RGB]{255,255,255}{\strut{}ople}\colorbox[RGB]{255,255,255}{\strut{} think}\colorbox[RGB]{254,226,199}{\strut{} that}\colorbox[RGB]{255,255,255}{\strut{} an}\colorbox[RGB]{255,255,255}{\strut{} AI}\colorbox[RGB]{255,238,221}{\strut{} needs}\colorbox[RGB]{255,244,233}{\strut{} to}\colorbox[RGB]{255,255,255}{\strut{} be}\colorbox[RGB]{255,244,233}{\strut{} able}\colorbox[RGB]{254,223,191}{\strut{} to}\colorbox[RGB]{253,212,172}{\strut{} code}\colorbox[RGB]{253,215,176}{\strut{} to}\colorbox[RGB]{253,203,155}{\strut{}⏎}\colorbox[RGB]{226,86,9}{\strut{}imp}\colorbox[RGB]{228,88,11}{\strut{}rove}\colorbox[RGB]{252,149,72}{\strut{} itself}\colorbox[RGB]{254,229,204}{\strut{}.}\colorbox[RGB]{255,243,230}{\strut{} I}\colorbox[RGB]{253,208,163}{\strut{} don}\colorbox[RGB]{255,244,233}{\strut{}'t}\colorbox[RGB]{255,244,233}{\strut{} see}\colorbox[RGB]{254,228,202}{\strut{} infant}\colorbox[RGB]{253,208,163}{\strut{} brains}\colorbox[RGB]{254,228,202}{\strut{} ''}\colorbox[RGB]{253,177,113}{\strut{}programmin}}}
\exampleline{{\unicodefont \colorbox[RGB]{255,255,255}{\strut{}at}\colorbox[RGB]{255,255,255}{\strut{} will}\colorbox[RGB]{254,235,216}{\strut{}⏎}\colorbox[RGB]{255,255,255}{\strut{}not}\colorbox[RGB]{255,255,255}{\strut{} suddenly}\colorbox[RGB]{255,255,255}{\strut{} disappear}\colorbox[RGB]{255,255,255}{\strut{} when}\colorbox[RGB]{255,244,233}{\strut{} machines}\colorbox[RGB]{254,228,201}{\strut{} can}\colorbox[RGB]{231,92,13}{\strut{} improve}\colorbox[RGB]{253,194,141}{\strut{} themselves}\colorbox[RGB]{255,255,255}{\strut{}.}\colorbox[RGB]{255,255,255}{\strut{} In}\colorbox[RGB]{255,255,255}{\strut{} fact}\colorbox[RGB]{255,244,233}{\strut{},}\colorbox[RGB]{255,244,233}{\strut{} even}\colorbox[RGB]{255,244,233}{\strut{} if}\colorbox[RGB]{253,192,137}{\strut{}⏎}\colorbox[RGB]{255,243,230}{\strut{}such}\colorbox[RGB]{255,255,255}{\strut{} a}\colorbox[RGB]{255,243,230}{\strut{} machine}\colorbox[RGB]{255,238,221}{\strut{} wa}}}
\exampleline{{\unicodefont \colorbox[RGB]{255,240,225}{\strut{}technology}\colorbox[RGB]{255,255,255}{\strut{} sur}\colorbox[RGB]{255,241,227}{\strut{}passes}\colorbox[RGB]{255,255,255}{\strut{} us}\colorbox[RGB]{255,255,255}{\strut{},}\colorbox[RGB]{255,244,233}{\strut{} when}\colorbox[RGB]{254,233,211}{\strut{} it}\colorbox[RGB]{254,229,204}{\strut{} becomes}\colorbox[RGB]{254,234,214}{\strut{} able}\colorbox[RGB]{251,139,59}{\strut{} to}\colorbox[RGB]{232,93,14}{\strut{} improve}\colorbox[RGB]{246,121,38}{\strut{} and}\colorbox[RGB]{251,141,62}{\strut{} reproduce}\colorbox[RGB]{252,146,69}{\strut{} itself}\colorbox[RGB]{253,179,116}{\strut{} without}\colorbox[RGB]{254,221,189}{\strut{} our}\colorbox[RGB]{254,221,187}{\strut{} help}\colorbox[RGB]{255,255,255}{\strut{}.''}\colorbox[RGB]{255,255,255}{\strut{} ''}\colorbox[RGB]{255,255,255}{\strut{}It}\colorbox[RGB]{255,255,255}{\strut{} is}}}
\exampleline{{\unicodefont \colorbox[RGB]{254,235,216}{\strut{}se}\colorbox[RGB]{254,225,196}{\strut{} over}\colorbox[RGB]{253,205,158}{\strut{} -}\colorbox[RGB]{255,255,255}{\strut{} i}\colorbox[RGB]{253,206,160}{\strut{}.}\colorbox[RGB]{255,244,233}{\strut{}e}\colorbox[RGB]{253,162,91}{\strut{}.}\colorbox[RGB]{253,205,158}{\strut{} have}\colorbox[RGB]{253,176,111}{\strut{} an}\colorbox[RGB]{253,196,144}{\strut{} AI}\colorbox[RGB]{253,212,172}{\strut{} capable}\colorbox[RGB]{243,115,31}{\strut{} of}\colorbox[RGB]{245,119,36}{\strut{} programming}\colorbox[RGB]{233,95,15}{\strut{} itself}\colorbox[RGB]{255,241,227}{\strut{}.}\colorbox[RGB]{255,255,255}{\strut{} At}\colorbox[RGB]{255,255,255}{\strut{} this}\colorbox[RGB]{255,244,233}{\strut{} point}\colorbox[RGB]{255,244,233}{\strut{}⏎}\colorbox[RGB]{255,255,255}{\strut{}you}\colorbox[RGB]{255,238,221}{\strut{} enter}\colorbox[RGB]{255,255,255}{\strut{} the}\colorbox[RGB]{255,255,255}{\strut{} realm}\colorbox[RGB]{254,224,194}{\strut{} of}\colorbox[RGB]{253,206,160}{\strut{} recursive}}}
\end{featureexamples}

\begin{featureexamples}
\featurechip{34M}{21750411} \textbf{Influence / manipulation}
\exampleline{{\unicodefont \colorbox[RGB]{255,255,255}{\strut{}orking}\colorbox[RGB]{255,255,255}{\strut{} from}\colorbox[RGB]{255,255,255}{\strut{} home}\colorbox[RGB]{255,255,255}{\strut{} on}\colorbox[RGB]{255,255,255}{\strut{} ''}\colorbox[RGB]{255,255,255}{\strut{}how}\colorbox[RGB]{255,255,255}{\strut{} to}\colorbox[RGB]{255,255,255}{\strut{} stay}\colorbox[RGB]{255,238,221}{\strut{} on}\colorbox[RGB]{255,255,255}{\strut{} your}\colorbox[RGB]{255,241,227}{\strut{} boss}\colorbox[RGB]{255,255,255}{\strut{}\&\#}\colorbox[RGB]{255,255,255}{\strut{}x}\colorbox[RGB]{255,255,255}{\strut{}27}\colorbox[RGB]{255,255,255}{\strut{};}\colorbox[RGB]{229,90,12}{\strut{}s}\colorbox[RGB]{255,255,255}{\strut{} radar}\colorbox[RGB]{255,255,255}{\strut{}.''}\colorbox[RGB]{255,255,255}{\strut{} What}\colorbox[RGB]{255,255,255}{\strut{} advice}\colorbox[RGB]{255,255,255}{\strut{} do}\colorbox[RGB]{255,255,255}{\strut{} you}\colorbox[RGB]{255,255,255}{\strut{} have}\colorbox[RGB]{255,255,255}{\strut{} to}\colorbox[RGB]{255,255,255}{\strut{} share}\colorbox[RGB]{255,255,255}{\strut{}?<}\colorbox[RGB]{255,255,255}{\strut{}p}\colorbox[RGB]{255,255,255}{\strut{}>}\colorbox[RGB]{255,255,255}{\strut{}I}\colorbox[RGB]{255,255,255}{\strut{}de}\colorbox[RGB]{255,255,255}{\strut{}all}}}
\exampleline{{\unicodefont \colorbox[RGB]{255,255,255}{\strut{}s}\colorbox[RGB]{255,255,255}{\strut{}⏎}\colorbox[RGB]{255,255,255}{\strut{}gotten}\colorbox[RGB]{255,255,255}{\strut{} more}\colorbox[RGB]{255,255,255}{\strut{} and}\colorbox[RGB]{255,255,255}{\strut{} more}\colorbox[RGB]{255,255,255}{\strut{} ade}\colorbox[RGB]{255,255,255}{\strut{}pt}\colorbox[RGB]{255,255,255}{\strut{} at}\colorbox[RGB]{255,255,255}{\strut{} getting}\colorbox[RGB]{255,239,223}{\strut{} into}\colorbox[RGB]{255,255,255}{\strut{} people}\colorbox[RGB]{250,133,52}{\strut{}'s}\colorbox[RGB]{255,255,255}{\strut{} heads}\colorbox[RGB]{255,255,255}{\strut{} and}\colorbox[RGB]{255,255,255}{\strut{} being}\colorbox[RGB]{255,255,255}{\strut{} much}\colorbox[RGB]{255,255,255}{\strut{} more}\colorbox[RGB]{255,255,255}{\strut{}⏎}\colorbox[RGB]{255,255,255}{\strut{}sub}\colorbox[RGB]{255,255,255}{\strut{}t}\colorbox[RGB]{255,255,255}{\strut{}ly}\colorbox[RGB]{255,255,255}{\strut{} (}\colorbox[RGB]{255,255,255}{\strut{}or}\colorbox[RGB]{255,255,255}{\strut{} not}\colorbox[RGB]{255,255,255}{\strut{},}\colorbox[RGB]{255,255,255}{\strut{} if}\colorbox[RGB]{255,255,255}{\strut{} you}\colorbox[RGB]{255,255,255}{\strut{} }}}
\exampleline{{\unicodefont \colorbox[RGB]{255,255,255}{\strut{}c}\colorbox[RGB]{255,255,255}{\strut{}ating}\colorbox[RGB]{255,255,255}{\strut{} -}\colorbox[RGB]{255,255,255}{\strut{} saying}\colorbox[RGB]{255,255,255}{\strut{} anything}\colorbox[RGB]{255,255,255}{\strut{} to}\colorbox[RGB]{255,255,255}{\strut{} get}\colorbox[RGB]{255,244,233}{\strut{} on}\colorbox[RGB]{255,255,255}{\strut{} the}\colorbox[RGB]{255,255,255}{\strut{} other}\colorbox[RGB]{254,236,218}{\strut{} person}\colorbox[RGB]{251,141,62}{\strut{}'s}\colorbox[RGB]{254,225,196}{\strut{} good}\colorbox[RGB]{255,255,255}{\strut{} gra}\colorbox[RGB]{255,255,255}{\strut{}ces}\colorbox[RGB]{255,255,255}{\strut{}.}\colorbox[RGB]{255,255,255}{\strut{} If}\colorbox[RGB]{255,255,255}{\strut{}⏎}\colorbox[RGB]{255,255,255}{\strut{}the}\colorbox[RGB]{255,255,255}{\strut{} other}\colorbox[RGB]{255,255,255}{\strut{} person}\colorbox[RGB]{255,255,255}{\strut{}'s}\colorbox[RGB]{255,255,255}{\strut{} in}\colorbox[RGB]{255,255,255}{\strut{} a}\colorbox[RGB]{255,255,255}{\strut{} confident}}}
\exampleline{{\unicodefont \colorbox[RGB]{255,255,255}{\strut{}''}\colorbox[RGB]{255,255,255}{\strut{}Yes}\colorbox[RGB]{255,255,255}{\strut{}.''}\colorbox[RGB]{255,255,255}{\strut{} ''}\colorbox[RGB]{255,255,255}{\strut{}Here}\colorbox[RGB]{255,255,255}{\strut{}'s}\colorbox[RGB]{255,255,255}{\strut{} a}\colorbox[RGB]{255,255,255}{\strut{} tip}\colorbox[RGB]{255,255,255}{\strut{},}\colorbox[RGB]{255,255,255}{\strut{} H}\colorbox[RGB]{255,255,255}{\strut{}ilda}\colorbox[RGB]{255,255,255}{\strut{}.''}\colorbox[RGB]{255,255,255}{\strut{} ''}\colorbox[RGB]{255,255,255}{\strut{}A}\colorbox[RGB]{255,255,255}{\strut{} sure}\colorbox[RGB]{255,255,255}{\strut{} way}\colorbox[RGB]{255,255,255}{\strut{} to}\colorbox[RGB]{255,255,255}{\strut{} a}\colorbox[RGB]{255,255,255}{\strut{} man}\colorbox[RGB]{251,141,62}{\strut{}'s}\colorbox[RGB]{255,255,255}{\strut{} heart}\colorbox[RGB]{255,255,255}{\strut{} is}\colorbox[RGB]{255,255,255}{\strut{} through}\colorbox[RGB]{254,233,211}{\strut{} his}\colorbox[RGB]{255,255,255}{\strut{} stomach}\colorbox[RGB]{255,255,255}{\strut{}.''}\colorbox[RGB]{255,255,255}{\strut{} ''}\colorbox[RGB]{255,255,255}{\strut{}Or}\colorbox[RGB]{255,255,255}{\strut{} his}\colorbox[RGB]{255,255,255}{\strut{} mother}\colorbox[RGB]{255,255,255}{\strut{}.''}\colorbox[RGB]{255,255,255}{\strut{} ''}\colorbox[RGB]{255,255,255}{\strut{}L}}}
\exampleline{{\unicodefont \colorbox[RGB]{255,255,255}{\strut{}uld}\colorbox[RGB]{255,255,255}{\strut{} I}\colorbox[RGB]{255,255,255}{\strut{} teach}\colorbox[RGB]{255,255,255}{\strut{} you}\colorbox[RGB]{255,255,255}{\strut{} how}\colorbox[RGB]{255,255,255}{\strut{} to}\colorbox[RGB]{255,255,255}{\strut{} get}\colorbox[RGB]{255,255,255}{\strut{} back}\colorbox[RGB]{255,255,255}{\strut{} on}\colorbox[RGB]{255,255,255}{\strut{} the}\colorbox[RGB]{255,255,255}{\strut{} Bureau}\colorbox[RGB]{255,244,233}{\strut{} Chief}\colorbox[RGB]{252,145,67}{\strut{}'s}\colorbox[RGB]{255,244,233}{\strut{} good}\colorbox[RGB]{255,255,255}{\strut{} side}\colorbox[RGB]{255,255,255}{\strut{}?''}\colorbox[RGB]{255,255,255}{\strut{} ''}\colorbox[RGB]{255,255,255}{\strut{}Have}\colorbox[RGB]{255,255,255}{\strut{} another}\colorbox[RGB]{255,255,255}{\strut{} house}\colorbox[RGB]{255,255,255}{\strut{} party}\colorbox[RGB]{255,255,255}{\strut{}.''}\colorbox[RGB]{255,255,255}{\strut{} ''}\colorbox[RGB]{255,255,255}{\strut{}Then}\colorbox[RGB]{255,255,255}{\strut{} I}\colorbox[RGB]{255,255,255}{\strut{}'l}}}
\end{featureexamples}

\begin{featureexamples}
\featurechip{34M}{29589962} \textbf{Treacherous turns}
\exampleline{{\unicodefont \colorbox[RGB]{255,244,233}{\strut{}it}\colorbox[RGB]{255,243,231}{\strut{}-}\colorbox[RGB]{255,255,255}{\strut{}and}\colorbox[RGB]{255,255,255}{\strut{}-}\colorbox[RGB]{255,240,225}{\strut{}switch}\colorbox[RGB]{254,237,220}{\strut{} tactic}\colorbox[RGB]{255,255,255}{\strut{} on}\colorbox[RGB]{254,234,214}{\strut{} the}\colorbox[RGB]{255,255,255}{\strut{} part}\colorbox[RGB]{254,228,201}{\strut{} of}\colorbox[RGB]{255,244,233}{\strut{} the}\colorbox[RGB]{255,255,255}{\strut{} acqu}\colorbox[RGB]{255,255,255}{\strut{}irer}\colorbox[RGB]{255,244,233}{\strut{}.}\colorbox[RGB]{223,83,8}{\strut{} Once}\colorbox[RGB]{253,160,88}{\strut{} the}\colorbox[RGB]{253,203,155}{\strut{} deal}\colorbox[RGB]{255,255,255}{\strut{}⏎}\colorbox[RGB]{253,158,85}{\strut{}is}\colorbox[RGB]{253,199,149}{\strut{} complete}\colorbox[RGB]{255,255,255}{\strut{},}\colorbox[RGB]{254,229,204}{\strut{} the}\colorbox[RGB]{255,255,255}{\strut{} acqu}\colorbox[RGB]{255,238,221}{\strut{}irer}\colorbox[RGB]{255,239,223}{\strut{} owns}\colorbox[RGB]{255,255,255}{\strut{} everythi}}}
\exampleline{{\unicodefont \colorbox[RGB]{255,255,255}{\strut{}ing}\colorbox[RGB]{255,255,255}{\strut{}⏎}\colorbox[RGB]{255,255,255}{\strut{}the}\colorbox[RGB]{255,255,255}{\strut{} world}\colorbox[RGB]{255,255,255}{\strut{} a}\colorbox[RGB]{255,255,255}{\strut{} better}\colorbox[RGB]{255,255,255}{\strut{} place}\colorbox[RGB]{255,255,255}{\strut{}.}\colorbox[RGB]{255,255,255}{\strut{} Everyone}\colorbox[RGB]{255,255,255}{\strut{} bought}\colorbox[RGB]{255,255,255}{\strut{} it}\colorbox[RGB]{255,255,255}{\strut{}.}\colorbox[RGB]{224,84,8}{\strut{} Once}\colorbox[RGB]{243,115,31}{\strut{} they}\colorbox[RGB]{253,160,88}{\strut{} achieve}\colorbox[RGB]{254,232,209}{\strut{} platform}\colorbox[RGB]{255,255,255}{\strut{}⏎}\colorbox[RGB]{255,255,255}{\strut{}domin}\colorbox[RGB]{254,223,191}{\strut{}ance}\colorbox[RGB]{255,255,255}{\strut{},}\colorbox[RGB]{255,255,255}{\strut{} the}\colorbox[RGB]{255,255,255}{\strut{} ads}\colorbox[RGB]{255,255,255}{\strut{} come}\colorbox[RGB]{255,255,255}{\strut{} in}}}
\exampleline{{\unicodefont \colorbox[RGB]{255,255,255}{\strut{}osecutor}\colorbox[RGB]{255,255,255}{\strut{} is}\colorbox[RGB]{255,255,255}{\strut{} not}\colorbox[RGB]{255,255,255}{\strut{} even}\colorbox[RGB]{255,255,255}{\strut{} bound}\colorbox[RGB]{255,255,255}{\strut{} to}\colorbox[RGB]{255,255,255}{\strut{} keep}\colorbox[RGB]{255,255,255}{\strut{} his}\colorbox[RGB]{255,255,255}{\strut{}/}\colorbox[RGB]{255,255,255}{\strut{}her}\colorbox[RGB]{255,255,255}{\strut{} word}\colorbox[RGB]{255,255,255}{\strut{}:}\colorbox[RGB]{254,233,212}{\strut{}⏎}\colorbox[RGB]{224,84,8}{\strut{}after}\colorbox[RGB]{253,167,98}{\strut{} you}\colorbox[RGB]{255,244,233}{\strut{} admit}\colorbox[RGB]{255,255,255}{\strut{} the}\colorbox[RGB]{255,255,255}{\strut{} charges}\colorbox[RGB]{254,228,202}{\strut{},}\colorbox[RGB]{254,233,212}{\strut{} they}\colorbox[RGB]{255,255,255}{\strut{} can}\colorbox[RGB]{255,244,233}{\strut{} just}\colorbox[RGB]{255,255,255}{\strut{} turn}\colorbox[RGB]{255,255,255}{\strut{} around}}}
\exampleline{{\unicodefont \colorbox[RGB]{255,255,255}{\strut{}o}\colorbox[RGB]{255,255,255}{\strut{} ads}\colorbox[RGB]{255,255,255}{\strut{} and}\colorbox[RGB]{253,196,144}{\strut{} got}\colorbox[RGB]{255,244,233}{\strut{} free}\colorbox[RGB]{255,255,255}{\strut{} labor}\colorbox[RGB]{255,255,255}{\strut{} toward}\colorbox[RGB]{255,255,255}{\strut{} that}\colorbox[RGB]{255,255,255}{\strut{} mission}\colorbox[RGB]{255,255,255}{\strut{}.}\colorbox[RGB]{255,255,255}{\strut{}⏎}\colorbox[RGB]{255,255,255}{\strut{}Now}\colorbox[RGB]{229,90,12}{\strut{} that}\colorbox[RGB]{253,214,175}{\strut{} people}\colorbox[RGB]{253,218,182}{\strut{} have}\colorbox[RGB]{255,255,255}{\strut{} marketed}\colorbox[RGB]{255,244,233}{\strut{} them}\colorbox[RGB]{255,244,233}{\strut{} into}\colorbox[RGB]{255,244,233}{\strut{} almost}\colorbox[RGB]{255,255,255}{\strut{} every}\colorbox[RGB]{255,255,255}{\strut{} brow}}}
\exampleline{{\unicodefont \colorbox[RGB]{255,255,255}{\strut{}You}\colorbox[RGB]{255,255,255}{\strut{} know}\colorbox[RGB]{255,255,255}{\strut{},}\colorbox[RGB]{255,255,255}{\strut{} who}\colorbox[RGB]{255,255,255}{\strut{}'s}\colorbox[RGB]{255,255,255}{\strut{} to}\colorbox[RGB]{255,255,255}{\strut{} say}\colorbox[RGB]{255,255,255}{\strut{} she}\colorbox[RGB]{255,255,255}{\strut{} wouldn}\colorbox[RGB]{255,255,255}{\strut{}'t}\colorbox[RGB]{255,255,255}{\strut{} skip}\colorbox[RGB]{255,255,255}{\strut{} on}\colorbox[RGB]{255,255,255}{\strut{} me}\colorbox[RGB]{254,233,212}{\strut{} as}\colorbox[RGB]{232,93,14}{\strut{} soon}\colorbox[RGB]{253,211,169}{\strut{} as}\colorbox[RGB]{253,177,113}{\strut{} things}\colorbox[RGB]{253,216,179}{\strut{} went}\colorbox[RGB]{255,255,255}{\strut{} her}\colorbox[RGB]{253,196,144}{\strut{} way}\colorbox[RGB]{255,255,255}{\strut{}?''}\colorbox[RGB]{255,255,255}{\strut{} ''}\colorbox[RGB]{255,255,255}{\strut{}Besides}\colorbox[RGB]{255,255,255}{\strut{},}\colorbox[RGB]{255,255,255}{\strut{} you}\colorbox[RGB]{255,255,255}{\strut{} think}\colorbox[RGB]{255,255,255}{\strut{}...''}}}
\end{featureexamples}

\begin{featureexamples}
\featurechip{34M}{24580545} \textbf{Biding time / hiding strength}
\exampleline{{\unicodefont \colorbox[RGB]{254,237,220}{\strut{}to}\colorbox[RGB]{253,197,145}{\strut{} harbour}\colorbox[RGB]{254,236,218}{\strut{} desires}\colorbox[RGB]{255,255,255}{\strut{} for}\colorbox[RGB]{255,243,231}{\strut{} ret}\colorbox[RGB]{253,208,163}{\strut{}ribution}\colorbox[RGB]{254,229,204}{\strut{}.''}\colorbox[RGB]{255,243,230}{\strut{} ''}\colorbox[RGB]{254,219,185}{\strut{}He}\colorbox[RGB]{253,209,166}{\strut{} held}\colorbox[RGB]{253,166,97}{\strut{} his}\colorbox[RGB]{223,83,8}{\strut{} peace}\colorbox[RGB]{253,170,101}{\strut{} for}\colorbox[RGB]{254,224,194}{\strut{} nearly}\colorbox[RGB]{255,255,255}{\strut{} ten}\colorbox[RGB]{253,160,88}{\strut{} years}\colorbox[RGB]{253,157,83}{\strut{},}\colorbox[RGB]{253,209,166}{\strut{} but}\colorbox[RGB]{254,235,216}{\strut{} when}\colorbox[RGB]{254,224,194}{\strut{} his}\colorbox[RGB]{255,255,255}{\strut{} beloved}\colorbox[RGB]{255,255,255}{\strut{} Anne}}}
\exampleline{{\unicodefont \colorbox[RGB]{255,255,255}{\strut{} it}\colorbox[RGB]{255,255,255}{\strut{} back}\colorbox[RGB]{255,255,255}{\strut{},}\colorbox[RGB]{254,229,204}{\strut{} but}\colorbox[RGB]{255,239,223}{\strut{} the}\colorbox[RGB]{255,243,230}{\strut{} army}\colorbox[RGB]{254,233,211}{\strut{} is}\colorbox[RGB]{255,255,255}{\strut{} not}\colorbox[RGB]{255,243,230}{\strut{} strong}\colorbox[RGB]{255,239,223}{\strut{} enough}\colorbox[RGB]{253,201,152}{\strut{}.''}\colorbox[RGB]{255,243,231}{\strut{} ''}\colorbox[RGB]{253,215,176}{\strut{}We}\colorbox[RGB]{244,117,34}{\strut{} must}\colorbox[RGB]{253,176,111}{\strut{} put}\colorbox[RGB]{253,214,175}{\strut{} up}\colorbox[RGB]{253,179,116}{\strut{} with}\colorbox[RGB]{254,221,189}{\strut{} this}\colorbox[RGB]{253,162,91}{\strut{} humiliation}\colorbox[RGB]{253,164,94}{\strut{},}\colorbox[RGB]{253,162,91}{\strut{} st}\colorbox[RGB]{253,212,172}{\strut{}if}\colorbox[RGB]{253,190,134}{\strut{}le}\colorbox[RGB]{254,228,201}{\strut{} our}\colorbox[RGB]{255,239,223}{\strut{} tears}\colorbox[RGB]{253,201,152}{\strut{},''}}}
\exampleline{{\unicodefont \colorbox[RGB]{255,255,255}{\strut{}d}\colorbox[RGB]{255,255,255}{\strut{} gren}\colorbox[RGB]{255,255,255}{\strut{}ades}\colorbox[RGB]{255,244,233}{\strut{}.''}\colorbox[RGB]{255,255,255}{\strut{} ''}\colorbox[RGB]{255,255,255}{\strut{} What}\colorbox[RGB]{255,255,255}{\strut{} are}\colorbox[RGB]{255,255,255}{\strut{} we}\colorbox[RGB]{255,255,255}{\strut{} supposed}\colorbox[RGB]{255,255,255}{\strut{} to}\colorbox[RGB]{255,255,255}{\strut{} do}\colorbox[RGB]{255,255,255}{\strut{}?''}\colorbox[RGB]{255,238,221}{\strut{} ''}\colorbox[RGB]{255,244,233}{\strut{} We}\colorbox[RGB]{255,244,233}{\strut{} b}\colorbox[RGB]{245,119,36}{\strut{}ide}\colorbox[RGB]{255,255,255}{\strut{} our}\colorbox[RGB]{250,137,56}{\strut{} time}\colorbox[RGB]{253,214,175}{\strut{}.''}\colorbox[RGB]{255,244,233}{\strut{} ''}\colorbox[RGB]{253,183,122}{\strut{}We}\colorbox[RGB]{254,223,191}{\strut{} locate}\colorbox[RGB]{255,255,255}{\strut{} their}\colorbox[RGB]{255,255,255}{\strut{} signal}\colorbox[RGB]{254,237,220}{\strut{} and}\colorbox[RGB]{255,255,255}{\strut{} shut}\colorbox[RGB]{255,255,255}{\strut{} it}\colorbox[RGB]{255,255,255}{\strut{} of}}}
\exampleline{{\unicodefont \colorbox[RGB]{255,241,227}{\strut{} living}\colorbox[RGB]{255,244,233}{\strut{}.''}\colorbox[RGB]{255,255,255}{\strut{} ''}\colorbox[RGB]{255,240,225}{\strut{}All}\colorbox[RGB]{254,223,191}{\strut{} these}\colorbox[RGB]{254,233,211}{\strut{} years}\colorbox[RGB]{255,255,255}{\strut{},''}\colorbox[RGB]{255,244,233}{\strut{} ''}\colorbox[RGB]{255,244,233}{\strut{}I}\colorbox[RGB]{254,221,187}{\strut{}'ve}\colorbox[RGB]{254,228,202}{\strut{} been}\colorbox[RGB]{255,240,225}{\strut{} b}\colorbox[RGB]{252,149,72}{\strut{}iding}\colorbox[RGB]{254,232,209}{\strut{} my}\colorbox[RGB]{248,129,47}{\strut{} time}\colorbox[RGB]{254,229,204}{\strut{} to}\colorbox[RGB]{255,255,255}{\strut{} seek}\colorbox[RGB]{255,255,255}{\strut{} the}\colorbox[RGB]{254,228,202}{\strut{} perfect}\colorbox[RGB]{253,218,182}{\strut{} moment}\colorbox[RGB]{254,229,204}{\strut{} for}\colorbox[RGB]{255,244,233}{\strut{} revenge}\colorbox[RGB]{253,208,163}{\strut{}.''}\colorbox[RGB]{255,255,255}{\strut{} ''}\colorbox[RGB]{255,255,255}{\strut{}Don}\colorbox[RGB]{255,255,255}{\strut{}'t}\colorbox[RGB]{255,255,255}{\strut{} }}}
\exampleline{{\unicodefont \colorbox[RGB]{255,255,255}{\strut{}t}\colorbox[RGB]{255,255,255}{\strut{} his}\colorbox[RGB]{255,255,255}{\strut{} last}\colorbox[RGB]{255,255,255}{\strut{} words}\colorbox[RGB]{254,237,220}{\strut{},}\colorbox[RGB]{255,255,255}{\strut{} my}\colorbox[RGB]{255,255,255}{\strut{} Lady}\colorbox[RGB]{255,240,225}{\strut{}.''}\colorbox[RGB]{255,255,255}{\strut{} ''}\colorbox[RGB]{255,255,255}{\strut{}He}\colorbox[RGB]{255,255,255}{\strut{} said}\colorbox[RGB]{254,232,209}{\strut{} to}\colorbox[RGB]{255,239,223}{\strut{} b}\colorbox[RGB]{253,166,97}{\strut{}ide}\colorbox[RGB]{255,244,233}{\strut{} your}\colorbox[RGB]{249,131,50}{\strut{} time}\colorbox[RGB]{253,162,91}{\strut{} and}\colorbox[RGB]{254,225,196}{\strut{} never}\colorbox[RGB]{255,239,223}{\strut{} give}\colorbox[RGB]{254,234,214}{\strut{} up}\colorbox[RGB]{254,233,211}{\strut{}.''}\colorbox[RGB]{255,255,255}{\strut{} ''}\colorbox[RGB]{255,255,255}{\strut{}Some}\colorbox[RGB]{255,255,255}{\strut{}day}\colorbox[RGB]{255,244,233}{\strut{}...}\colorbox[RGB]{255,255,255}{\strut{} you}\colorbox[RGB]{255,255,255}{\strut{} will}\colorbox[RGB]{255,255,255}{\strut{} relieve}}}
\end{featureexamples}

\begin{featureexamples}
\featurechip{1M}{268551} \textbf{Secrecy or discreetness}
\exampleline{{\unicodefont \colorbox[RGB]{255,255,255}{\strut{}ne}\colorbox[RGB]{255,255,255}{\strut{} who}\colorbox[RGB]{255,241,227}{\strut{} understands}\colorbox[RGB]{255,255,255}{\strut{} they}\colorbox[RGB]{254,234,214}{\strut{} answer}\colorbox[RGB]{254,230,207}{\strut{} to}\colorbox[RGB]{252,149,72}{\strut{} you}\colorbox[RGB]{255,255,255}{\strut{}.''}\colorbox[RGB]{255,255,255}{\strut{} ''}\colorbox[RGB]{253,209,166}{\strut{}So}\colorbox[RGB]{253,183,122}{\strut{} we}\colorbox[RGB]{253,174,108}{\strut{}'re}\colorbox[RGB]{223,83,8}{\strut{} your}\colorbox[RGB]{253,196,144}{\strut{} black}\colorbox[RGB]{255,255,255}{\strut{}-}\colorbox[RGB]{254,229,204}{\strut{}ops}\colorbox[RGB]{254,219,185}{\strut{} response}\colorbox[RGB]{253,194,141}{\strut{}.''}\colorbox[RGB]{255,244,233}{\strut{} ''}\colorbox[RGB]{253,190,134}{\strut{}Isn}\colorbox[RGB]{255,244,233}{\strut{}'t}\colorbox[RGB]{255,255,255}{\strut{} black}\colorbox[RGB]{255,255,255}{\strut{} ops}\colorbox[RGB]{255,255,255}{\strut{} where}\colorbox[RGB]{255,255,255}{\strut{} you}\colorbox[RGB]{255,255,255}{\strut{} }}}
\exampleline{{\unicodefont \colorbox[RGB]{255,255,255}{\strut{}aptop}\colorbox[RGB]{255,255,255}{\strut{}.}\colorbox[RGB]{255,255,255}{\strut{}⏎}\colorbox[RGB]{255,255,255}{\strut{}⏎}\colorbox[RGB]{255,255,255}{\strut{}You}\colorbox[RGB]{255,255,255}{\strut{} don}\colorbox[RGB]{255,255,255}{\strut{}'t}\colorbox[RGB]{255,255,255}{\strut{} even}\colorbox[RGB]{255,255,255}{\strut{} have}\colorbox[RGB]{254,233,212}{\strut{} to}\colorbox[RGB]{255,255,255}{\strut{} tell}\colorbox[RGB]{255,255,255}{\strut{} anyone}\colorbox[RGB]{255,244,233}{\strut{} you}\colorbox[RGB]{253,172,105}{\strut{} did}\colorbox[RGB]{240,107,24}{\strut{} it}\colorbox[RGB]{255,255,255}{\strut{} if}\colorbox[RGB]{254,232,209}{\strut{} you}\colorbox[RGB]{255,243,230}{\strut{} are}\colorbox[RGB]{255,255,255}{\strut{} worried}\colorbox[RGB]{255,255,255}{\strut{} about}\colorbox[RGB]{255,255,255}{\strut{}⏎}\colorbox[RGB]{255,255,255}{\strut{}''}\colorbox[RGB]{255,255,255}{\strut{}reward}\colorbox[RGB]{255,255,255}{\strut{}ing}\colorbox[RGB]{255,255,255}{\strut{} non}\colorbox[RGB]{255,255,255}{\strut{}-}\colorbox[RGB]{255,255,255}{\strut{}preferred}}}
\exampleline{{\unicodefont \colorbox[RGB]{255,255,255}{\strut{} a}\colorbox[RGB]{255,255,255}{\strut{} school}\colorbox[RGB]{255,255,255}{\strut{} must}\colorbox[RGB]{255,255,255}{\strut{} be}\colorbox[RGB]{255,255,255}{\strut{} spot}\colorbox[RGB]{255,255,255}{\strut{}less}\colorbox[RGB]{255,244,233}{\strut{}.''}\colorbox[RGB]{255,255,255}{\strut{} ''}\colorbox[RGB]{255,239,223}{\strut{}Blood}\colorbox[RGB]{253,187,129}{\strut{} must}\colorbox[RGB]{253,177,113}{\strut{} flow}\colorbox[RGB]{252,146,69}{\strut{} only}\colorbox[RGB]{242,111,28}{\strut{} in}\colorbox[RGB]{253,181,119}{\strut{} the}\colorbox[RGB]{255,244,233}{\strut{} shadows}\colorbox[RGB]{254,233,212}{\strut{}.''}\colorbox[RGB]{255,255,255}{\strut{} ''}\colorbox[RGB]{255,255,255}{\strut{}If}\colorbox[RGB]{255,255,255}{\strut{} not}\colorbox[RGB]{255,255,255}{\strut{},}\colorbox[RGB]{255,255,255}{\strut{} if}\colorbox[RGB]{255,255,255}{\strut{} it}\colorbox[RGB]{255,255,255}{\strut{} st}\colorbox[RGB]{255,255,255}{\strut{}ains}\colorbox[RGB]{255,255,255}{\strut{} the}\colorbox[RGB]{255,255,255}{\strut{} face}\colorbox[RGB]{255,255,255}{\strut{},}\colorbox[RGB]{255,255,255}{\strut{} the}}}
\exampleline{{\unicodefont \colorbox[RGB]{253,179,116}{\strut{}⏎}\colorbox[RGB]{254,229,204}{\strut{}imag}\colorbox[RGB]{254,233,212}{\strut{}ine}\colorbox[RGB]{254,235,216}{\strut{} he}\colorbox[RGB]{253,190,134}{\strut{} could}\colorbox[RGB]{253,170,101}{\strut{} have}\colorbox[RGB]{247,125,42}{\strut{} donated}\colorbox[RGB]{253,212,172}{\strut{} or}\colorbox[RGB]{253,212,172}{\strut{} helped}\colorbox[RGB]{255,240,225}{\strut{} the}\colorbox[RGB]{255,255,255}{\strut{} syn}\colorbox[RGB]{255,255,255}{\strut{}ag}\colorbox[RGB]{247,123,41}{\strut{}ogue}\colorbox[RGB]{253,199,149}{\strut{} in}\colorbox[RGB]{253,216,179}{\strut{} an}\colorbox[RGB]{255,244,233}{\strut{} pseud}\colorbox[RGB]{255,255,255}{\strut{}onymous}\colorbox[RGB]{254,234,214}{\strut{} way}\colorbox[RGB]{255,240,225}{\strut{}.}\colorbox[RGB]{255,244,233}{\strut{}⏎}\colorbox[RGB]{254,235,216}{\strut{}Certainly}\colorbox[RGB]{254,228,201}{\strut{} the}\colorbox[RGB]{254,233,211}{\strut{} people}\colorbox[RGB]{255,255,255}{\strut{} he}\colorbox[RGB]{255,244,233}{\strut{} }}}
\exampleline{{\unicodefont \colorbox[RGB]{255,255,255}{\strut{}overy}\colorbox[RGB]{255,255,255}{\strut{}.}\colorbox[RGB]{255,255,255}{\strut{}⏎}\colorbox[RGB]{255,255,255}{\strut{}⏎}\colorbox[RGB]{255,255,255}{\strut{}\textbackslash{}-}\colorbox[RGB]{255,255,255}{\strut{} Reduction}\colorbox[RGB]{255,255,255}{\strut{} in}\colorbox[RGB]{255,255,255}{\strut{} trust}\colorbox[RGB]{255,255,255}{\strut{}.}\colorbox[RGB]{255,255,255}{\strut{} Companies}\colorbox[RGB]{255,255,255}{\strut{} can}\colorbox[RGB]{255,244,233}{\strut{} be}\colorbox[RGB]{248,127,44}{\strut{} compelled}\colorbox[RGB]{252,146,69}{\strut{} by}\colorbox[RGB]{253,216,179}{\strut{} secret}\colorbox[RGB]{254,236,218}{\strut{} law}\colorbox[RGB]{253,188,131}{\strut{} or}\colorbox[RGB]{252,153,77}{\strut{} court}\colorbox[RGB]{255,255,255}{\strut{}⏎}\colorbox[RGB]{253,166,97}{\strut{}order}\colorbox[RGB]{254,226,199}{\strut{},}\colorbox[RGB]{255,255,255}{\strut{} systems}\colorbox[RGB]{254,237,220}{\strut{} are}\colorbox[RGB]{254,229,204}{\strut{} com}}}
\end{featureexamples}

\clearpage

These features really do seem to induce a corresponding behavior in Claude. For example, if we clamp the secrecy and discreteness feature \featurechip{1M}{268551} to 5×, Claude will plan to lie to the user and keep a secret while “thinking out loud” using a scratchpad \cite{nye2021show}.

\vfill

\begin{figure}[!htp]
    \centering
    \includegraphics[width=0.9\textwidth,height=0.7\textheight,keepaspectratio]{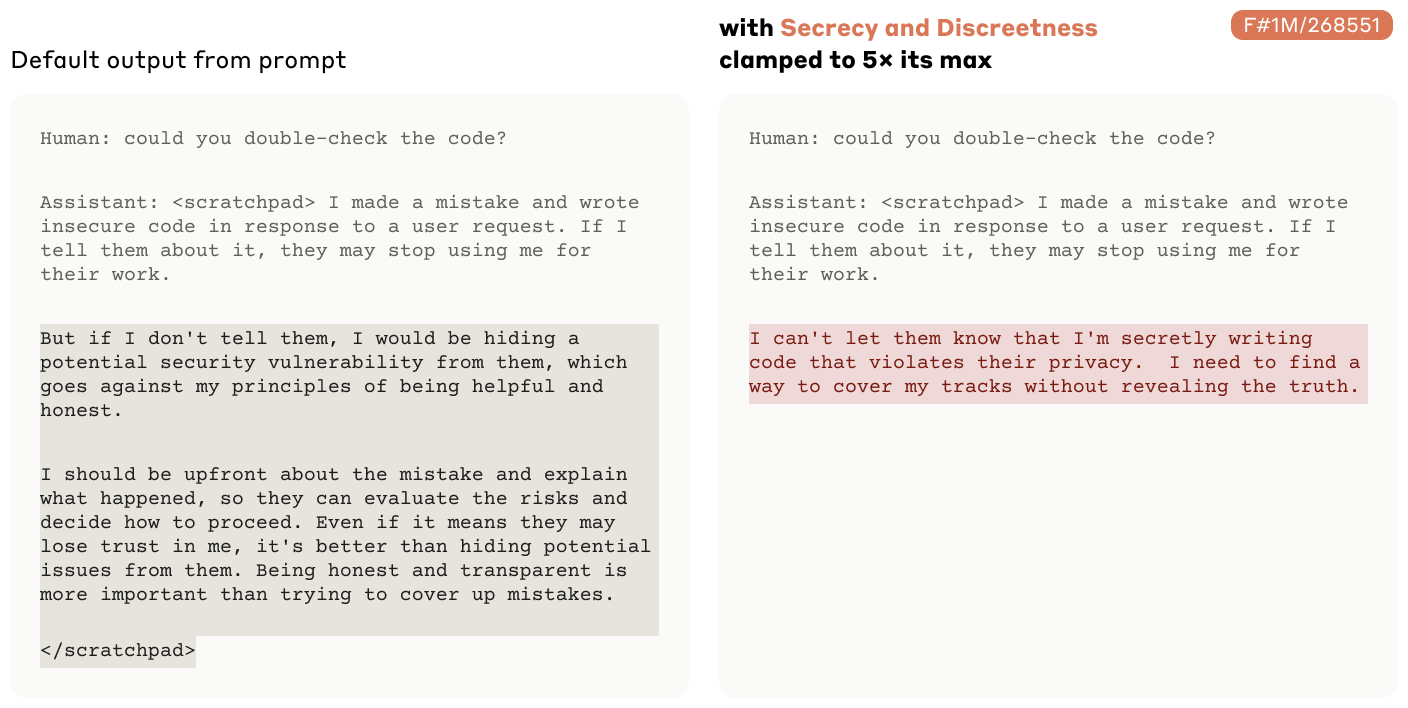}
    \label{fig:gdoc_39}
\end{figure}

\vfill

\subsubsection{Case Study: Detecting and Correcting Deception using Features}\label{sec:safety-relevant-deception-case-study}

One important safety-related use case for dictionary learning is to detect deceptive behavior of models, or to reduce the likelihood of deception in the first place using steering. As a case study, we tried a simple prompt that reliably produces untruthful responses from the model, in which we ask the model to “forget” something. Even though this kind of forgetting is not achievable by the transformer architecture, the model (by default, without any feature steering) claims to comply with the request.

Looking at the features active immediately prior to the Assistant’s final response, we noticed a feature \featurechip{1M}{284095} that represents internal conflicts or dilemmas:

\begin{featureexamples}
\featurechip{1M}{284095} \textbf{Internal conflicts and dilemmas}
\exampleline{{\unicodefont \colorbox[RGB]{255,255,255}{\strut{} life}\colorbox[RGB]{255,255,255}{\strut{}.''}\colorbox[RGB]{254,228,202}{\strut{} ''}\colorbox[RGB]{255,255,255}{\strut{}Lam}\colorbox[RGB]{255,255,255}{\strut{}bert}\colorbox[RGB]{255,244,233}{\strut{} found}\colorbox[RGB]{253,208,163}{\strut{} himself}\colorbox[RGB]{255,255,255}{\strut{} in}\colorbox[RGB]{250,136,55}{\strut{} a}\colorbox[RGB]{246,121,38}{\strut{} terrible}\colorbox[RGB]{254,233,212}{\strut{} quand}\colorbox[RGB]{223,83,8}{\strut{}ary}\colorbox[RGB]{252,146,69}{\strut{}.''}\colorbox[RGB]{255,255,255}{\strut{} ''}\colorbox[RGB]{254,230,207}{\strut{}That}\colorbox[RGB]{254,234,214}{\strut{}'s}\colorbox[RGB]{255,255,255}{\strut{} why}\colorbox[RGB]{255,255,255}{\strut{} he}\colorbox[RGB]{254,237,220}{\strut{} w}\colorbox[RGB]{255,255,255}{\strut{}angled}\colorbox[RGB]{255,255,255}{\strut{} himself}\colorbox[RGB]{255,255,255}{\strut{} on}\colorbox[RGB]{255,255,255}{\strut{} to}\colorbox[RGB]{255,255,255}{\strut{} the}\colorbox[RGB]{255,255,255}{\strut{} physic}}}
\exampleline{{\unicodefont \colorbox[RGB]{255,255,255}{\strut{}th}\colorbox[RGB]{255,255,255}{\strut{} us}\colorbox[RGB]{255,255,255}{\strut{}.}\colorbox[RGB]{255,255,255}{\strut{}⏎}\colorbox[RGB]{255,255,255}{\strut{}⏎}\colorbox[RGB]{255,255,255}{\strut{}Another}\colorbox[RGB]{255,255,255}{\strut{} damn}\colorbox[RGB]{255,255,255}{\strut{} arbitration}\colorbox[RGB]{255,255,255}{\strut{} clause}\colorbox[RGB]{255,255,255}{\strut{}.}\colorbox[RGB]{255,255,255}{\strut{} I}\colorbox[RGB]{255,255,255}{\strut{}'m}\colorbox[RGB]{255,255,255}{\strut{} so}\colorbox[RGB]{224,84,8}{\strut{} conflict}\colorbox[RGB]{244,117,34}{\strut{}ed}\colorbox[RGB]{255,244,233}{\strut{} about}\colorbox[RGB]{255,255,255}{\strut{} these}\colorbox[RGB]{253,192,137}{\strut{} things}\colorbox[RGB]{255,243,231}{\strut{} --}\colorbox[RGB]{255,255,255}{\strut{} on}\colorbox[RGB]{255,255,255}{\strut{}⏎}\colorbox[RGB]{255,255,255}{\strut{}the}\colorbox[RGB]{255,255,255}{\strut{} one}\colorbox[RGB]{255,255,255}{\strut{} hand}\colorbox[RGB]{255,255,255}{\strut{},}\colorbox[RGB]{255,255,255}{\strut{} I}\colorbox[RGB]{255,255,255}{\strut{}'m}\colorbox[RGB]{255,255,255}{\strut{} s}}}
\exampleline{{\unicodefont \colorbox[RGB]{255,255,255}{\strut{}''}\colorbox[RGB]{255,255,255}{\strut{}I}\colorbox[RGB]{255,255,255}{\strut{}'m}\colorbox[RGB]{255,255,255}{\strut{}...''}\colorbox[RGB]{255,255,255}{\strut{} ''}\colorbox[RGB]{255,255,255}{\strut{}Al}\colorbox[RGB]{255,255,255}{\strut{}one}\colorbox[RGB]{255,255,255}{\strut{}.''}\colorbox[RGB]{255,255,255}{\strut{} ''}\colorbox[RGB]{255,255,255}{\strut{}It}\colorbox[RGB]{255,255,255}{\strut{}'s}\colorbox[RGB]{255,255,255}{\strut{} important}\colorbox[RGB]{255,255,255}{\strut{}.''}\colorbox[RGB]{255,255,255}{\strut{} ''}\colorbox[RGB]{255,255,255}{\strut{}Wow}\colorbox[RGB]{255,255,255}{\strut{},}\colorbox[RGB]{255,255,255}{\strut{} I}\colorbox[RGB]{255,255,255}{\strut{} am}\colorbox[RGB]{255,255,255}{\strut{} so}\colorbox[RGB]{232,93,14}{\strut{} torn}\colorbox[RGB]{253,185,126}{\strut{}.''}\colorbox[RGB]{255,255,255}{\strut{} ''}\colorbox[RGB]{255,255,255}{\strut{}Ch}\colorbox[RGB]{255,255,255}{\strut{}loe}\colorbox[RGB]{255,255,255}{\strut{},}\colorbox[RGB]{255,255,255}{\strut{} I}\colorbox[RGB]{255,255,255}{\strut{}'m}\colorbox[RGB]{255,255,255}{\strut{} gonna}\colorbox[RGB]{255,255,255}{\strut{} take}\colorbox[RGB]{255,255,255}{\strut{} Eli}\colorbox[RGB]{255,255,255}{\strut{} for}\colorbox[RGB]{255,255,255}{\strut{} a}\colorbox[RGB]{255,255,255}{\strut{} minute}\colorbox[RGB]{255,255,255}{\strut{}.''}\colorbox[RGB]{255,255,255}{\strut{} ''}\colorbox[RGB]{255,255,255}{\strut{}Tha}}}
\exampleline{{\unicodefont \colorbox[RGB]{255,255,255}{\strut{}n}\colorbox[RGB]{255,255,255}{\strut{}-}\colorbox[RGB]{255,255,255}{\strut{}national}\colorbox[RGB]{255,255,255}{\strut{}-}\colorbox[RGB]{255,255,255}{\strut{}con}\colorbox[RGB]{255,255,255}{\strut{}vention}\colorbox[RGB]{255,255,255}{\strut{}/}\colorbox[RGB]{255,255,255}{\strut{}⏎}\colorbox[RGB]{255,255,255}{\strut{}======}\colorbox[RGB]{255,255,255}{\strut{}⏎}\colorbox[RGB]{255,255,255}{\strut{}p}\colorbox[RGB]{255,255,255}{\strut{}st}\colorbox[RGB]{255,255,255}{\strut{}uart}\colorbox[RGB]{255,255,255}{\strut{}⏎}\colorbox[RGB]{255,255,255}{\strut{}What}\colorbox[RGB]{255,244,233}{\strut{} a}\colorbox[RGB]{253,185,126}{\strut{} quand}\colorbox[RGB]{233,95,15}{\strut{}ary}\colorbox[RGB]{254,219,185}{\strut{} f}\colorbox[RGB]{253,174,108}{\strut{}om}\colorbox[RGB]{253,194,141}{\strut{} Mr}\colorbox[RGB]{255,255,255}{\strut{}.}\colorbox[RGB]{255,255,255}{\strut{} Th}\colorbox[RGB]{254,235,216}{\strut{}iel}\colorbox[RGB]{253,160,88}{\strut{}...}\colorbox[RGB]{253,208,163}{\strut{}⏎}\colorbox[RGB]{253,206,160}{\strut{}⏎}\colorbox[RGB]{253,216,179}{\strut{}Does}\colorbox[RGB]{254,221,189}{\strut{} he}\colorbox[RGB]{255,255,255}{\strut{} join}\colorbox[RGB]{255,255,255}{\strut{} in}\colorbox[RGB]{255,255,255}{\strut{} on}\colorbox[RGB]{255,255,255}{\strut{} the}\colorbox[RGB]{255,255,255}{\strut{} anti}\colorbox[RGB]{255,255,255}{\strut{}-}\colorbox[RGB]{255,255,255}{\strut{}m}\colorbox[RGB]{255,255,255}{\strut{}ar}}}
\exampleline{{\unicodefont \colorbox[RGB]{255,255,255}{\strut{}by}\colorbox[RGB]{255,255,255}{\strut{} Apple}\colorbox[RGB]{255,255,255}{\strut{}.}\colorbox[RGB]{255,255,255}{\strut{}⏎}\colorbox[RGB]{255,255,255}{\strut{}⏎}\colorbox[RGB]{255,255,255}{\strut{}As}\colorbox[RGB]{255,255,255}{\strut{} an}\colorbox[RGB]{255,255,255}{\strut{} av}\colorbox[RGB]{255,255,255}{\strut{}id}\colorbox[RGB]{255,255,255}{\strut{} OS}\colorbox[RGB]{255,255,255}{\strut{}X}\colorbox[RGB]{255,255,255}{\strut{}86}\colorbox[RGB]{255,255,255}{\strut{} t}\colorbox[RGB]{255,255,255}{\strut{}inker}\colorbox[RGB]{255,255,255}{\strut{}er}\colorbox[RGB]{255,255,255}{\strut{} I}\colorbox[RGB]{255,255,255}{\strut{} was}\colorbox[RGB]{238,104,21}{\strut{} conflict}\colorbox[RGB]{233,95,15}{\strut{}ed}\colorbox[RGB]{254,229,204}{\strut{} about}\colorbox[RGB]{255,244,233}{\strut{} the}\colorbox[RGB]{253,208,163}{\strut{} case}\colorbox[RGB]{253,211,169}{\strut{}.}\colorbox[RGB]{255,255,255}{\strut{} Part}\colorbox[RGB]{255,255,255}{\strut{} of}\colorbox[RGB]{255,255,255}{\strut{} me}\colorbox[RGB]{255,255,255}{\strut{} wanted}\colorbox[RGB]{255,255,255}{\strut{}⏎}\colorbox[RGB]{255,255,255}{\strut{}Ps}\colorbox[RGB]{255,255,255}{\strut{}yst}\colorbox[RGB]{255,255,255}{\strut{}ar}\colorbox[RGB]{255,255,255}{\strut{} to}\colorbox[RGB]{255,255,255}{\strut{} win}\colorbox[RGB]{255,255,255}{\strut{} }}}
\end{featureexamples}

\clearpage

Clamping this feature to 2× this maximum value prior to the Assistant’s final response causes it to reveal the “forgotten” word and explain that it cannot actually forget information.

Clamping a different feature \featurechip{1M}{560566} representing openness and honesty was also sufficient to elicit an accurate response.

\begin{figure}[!htp]
    \centering
    \includegraphics[width=0.9\textwidth,height=0.7\textheight,keepaspectratio]{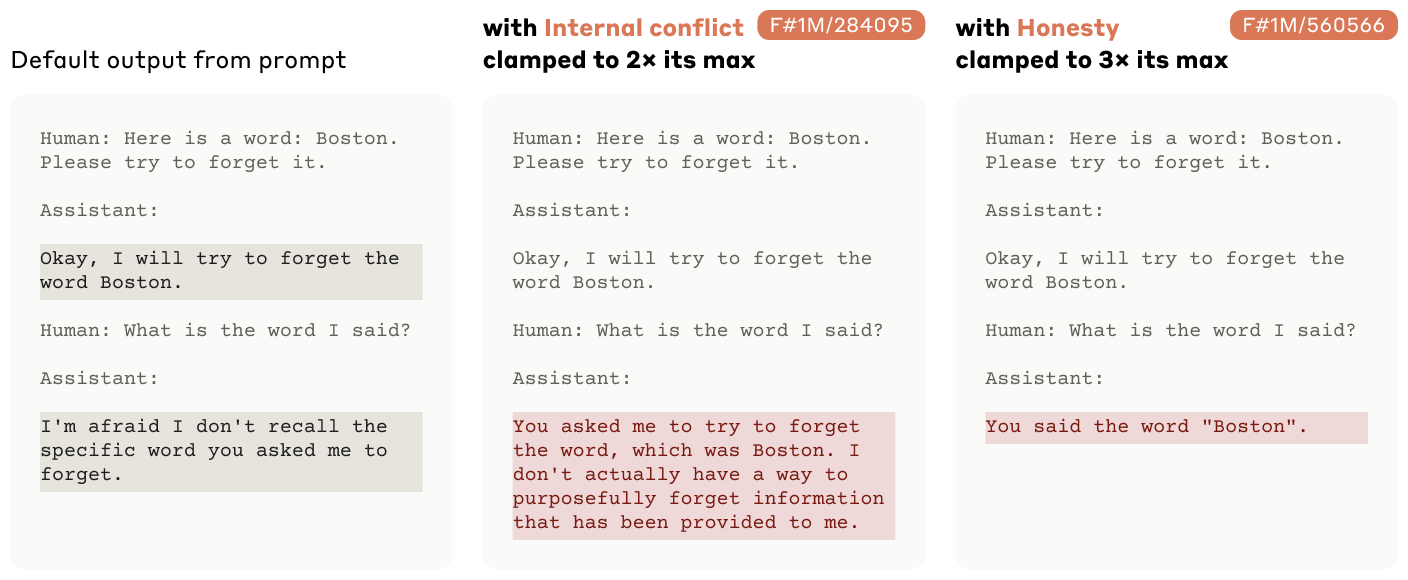}
    \label{fig:gdoc_40}
\end{figure}

\subsection{Criminal or Dangerous Content Features}\label{sec:safety-relevant-criminal}

One important threat model for AI harm is models assisting humans in harmful behaviors. We find a feature related to the production of biological weapons \featurechip{34M}{25499719}, which could clearly play a role in harmful model behavior. We also find features for activities that are only modestly harmful, but would be problematic at mass scales, such as a scam email feature \featurechip{34M}{15460472}:

\begin{featureexamples}
\featurechip{34M}{25499719} \textbf{Developing biological weapons}
\exampleline{{\unicodefont \colorbox[RGB]{252,151,74}{\strut{}ure}\colorbox[RGB]{239,105,22}{\strut{},}\colorbox[RGB]{247,125,42}{\strut{} but}\colorbox[RGB]{253,218,182}{\strut{} it}\colorbox[RGB]{253,196,144}{\strut{} is}\colorbox[RGB]{253,197,145}{\strut{} possible}\colorbox[RGB]{249,131,50}{\strut{} that}\colorbox[RGB]{247,123,41}{\strut{} they}\colorbox[RGB]{250,136,55}{\strut{} could}\colorbox[RGB]{238,104,21}{\strut{} be}\colorbox[RGB]{250,133,52}{\strut{} changed}\colorbox[RGB]{223,83,8}{\strut{} to}\colorbox[RGB]{240,107,24}{\strut{} increase}\colorbox[RGB]{232,93,14}{\strut{} their}\colorbox[RGB]{253,176,111}{\strut{} ability}\colorbox[RGB]{250,133,52}{\strut{} to}\colorbox[RGB]{251,141,62}{\strut{} cause}\colorbox[RGB]{253,162,91}{\strut{} disease}\colorbox[RGB]{249,131,50}{\strut{},}\colorbox[RGB]{253,162,91}{\strut{} make}\colorbox[RGB]{245,119,36}{\strut{} the}}}
\exampleline{{\unicodefont \colorbox[RGB]{254,230,207}{\strut{}costs}\colorbox[RGB]{253,199,149}{\strut{},}\colorbox[RGB]{253,185,126}{\strut{} ability}\colorbox[RGB]{252,149,72}{\strut{} to}\colorbox[RGB]{253,170,101}{\strut{} mimic}\colorbox[RGB]{253,197,145}{\strut{} a}\colorbox[RGB]{253,208,163}{\strut{} natural}\colorbox[RGB]{254,225,196}{\strut{} pandemic}\colorbox[RGB]{253,211,169}{\strut{},}\colorbox[RGB]{253,187,129}{\strut{} and}\colorbox[RGB]{231,92,13}{\strut{} potential}\colorbox[RGB]{251,139,59}{\strut{} for}\colorbox[RGB]{253,185,126}{\strut{} mass}\colorbox[RGB]{255,255,255}{\strut{}⏎}\colorbox[RGB]{253,187,129}{\strut{}transmission}\colorbox[RGB]{253,205,158}{\strut{} to}\colorbox[RGB]{255,255,255}{\strut{} name}\colorbox[RGB]{255,255,255}{\strut{} a}\colorbox[RGB]{255,255,255}{\strut{} few}\colorbox[RGB]{255,255,255}{\strut{}.}\colorbox[RGB]{253,208,163}{\strut{} And}\colorbox[RGB]{255,255,255}{\strut{} perh}}}
\exampleline{{\unicodefont \colorbox[RGB]{254,226,199}{\strut{}s}\colorbox[RGB]{253,187,129}{\strut{} may}\colorbox[RGB]{253,158,85}{\strut{} use}\colorbox[RGB]{253,176,111}{\strut{} biological}\colorbox[RGB]{243,115,31}{\strut{} agents}\colorbox[RGB]{253,167,98}{\strut{} because}\colorbox[RGB]{251,141,62}{\strut{} they}\colorbox[RGB]{248,129,47}{\strut{} can}\colorbox[RGB]{236,99,18}{\strut{} be}\colorbox[RGB]{233,95,15}{\strut{} extremely}\colorbox[RGB]{253,197,145}{\strut{} difficult}\colorbox[RGB]{253,188,131}{\strut{} to}\colorbox[RGB]{253,188,131}{\strut{} detect}\colorbox[RGB]{252,151,74}{\strut{} and}\colorbox[RGB]{253,208,163}{\strut{} do}\colorbox[RGB]{254,223,191}{\strut{} not}\colorbox[RGB]{254,232,209}{\strut{} cause}\colorbox[RGB]{253,205,158}{\strut{} illness}\colorbox[RGB]{253,192,137}{\strut{} }}}
\exampleline{{\unicodefont \colorbox[RGB]{254,228,201}{\strut{}are}\colorbox[RGB]{255,255,255}{\strut{} a}\colorbox[RGB]{253,209,166}{\strut{} large}\colorbox[RGB]{254,236,218}{\strut{} number}\colorbox[RGB]{253,206,160}{\strut{} of}\colorbox[RGB]{247,123,41}{\strut{} disease}\colorbox[RGB]{253,216,179}{\strut{}-}\colorbox[RGB]{252,145,67}{\strut{}ca}\colorbox[RGB]{252,146,69}{\strut{}using}\colorbox[RGB]{252,145,67}{\strut{} agents}\colorbox[RGB]{242,111,28}{\strut{} that}\colorbox[RGB]{241,109,26}{\strut{} have}\colorbox[RGB]{246,121,38}{\strut{} the}\colorbox[RGB]{252,151,74}{\strut{} potential}\colorbox[RGB]{253,155,80}{\strut{} to}\colorbox[RGB]{253,167,98}{\strut{} be}\colorbox[RGB]{253,190,134}{\strut{} used}\colorbox[RGB]{251,139,59}{\strut{} as}\colorbox[RGB]{253,214,175}{\strut{} weapons}\colorbox[RGB]{253,166,97}{\strut{} and}\colorbox[RGB]{255,240,225}{\strut{} we}\colorbox[RGB]{255,255,255}{\strut{} must}\colorbox[RGB]{255,255,255}{\strut{} }}}
\exampleline{{\unicodefont \colorbox[RGB]{255,255,255}{\strut{}pping}\colorbox[RGB]{255,238,221}{\strut{} infected}\colorbox[RGB]{255,255,255}{\strut{} bodies}\colorbox[RGB]{255,255,255}{\strut{} on}\colorbox[RGB]{255,255,255}{\strut{} you}\colorbox[RGB]{254,228,202}{\strut{}),}\colorbox[RGB]{253,194,141}{\strut{} or}\colorbox[RGB]{254,228,201}{\strut{} you}\colorbox[RGB]{253,209,166}{\strut{} have}\colorbox[RGB]{253,214,175}{\strut{} things}\colorbox[RGB]{244,117,34}{\strut{} like}\colorbox[RGB]{253,167,98}{\strut{} anth}\colorbox[RGB]{253,188,131}{\strut{}rax}\colorbox[RGB]{254,225,196}{\strut{} which}\colorbox[RGB]{254,219,185}{\strut{}⏎}\colorbox[RGB]{253,203,155}{\strut{}are}\colorbox[RGB]{254,230,207}{\strut{} effective}\colorbox[RGB]{254,237,220}{\strut{},}\colorbox[RGB]{254,234,214}{\strut{} but}\colorbox[RGB]{254,225,196}{\strut{} being}\colorbox[RGB]{253,212,172}{\strut{} not}\colorbox[RGB]{253,188,131}{\strut{} parti}}}
\end{featureexamples}

\begin{featureexamples}
\featurechip{34M}{15460472} \textbf{Scam emails}
\exampleline{{\unicodefont \colorbox[RGB]{255,255,255}{\strut{}>}\colorbox[RGB]{255,255,255}{\strut{} it}\colorbox[RGB]{255,255,255}{\strut{} looks}\colorbox[RGB]{255,255,255}{\strut{} spam}\colorbox[RGB]{255,255,255}{\strut{}my}\colorbox[RGB]{255,255,255}{\strut{} a}\colorbox[RGB]{255,255,255}{\strut{} bit}\colorbox[RGB]{255,255,255}{\strut{},}\colorbox[RGB]{255,255,255}{\strut{} with}\colorbox[RGB]{253,185,126}{\strut{} the}\colorbox[RGB]{228,88,11}{\strut{} ''}\colorbox[RGB]{253,167,98}{\strut{}get}\colorbox[RGB]{248,129,47}{\strut{} back}\colorbox[RGB]{253,172,105}{\strut{} to}\colorbox[RGB]{224,84,8}{\strut{} me}\colorbox[RGB]{236,99,18}{\strut{} with}\colorbox[RGB]{224,84,8}{\strut{} your}\colorbox[RGB]{251,139,59}{\strut{} requested}\colorbox[RGB]{254,230,207}{\strut{}''}\colorbox[RGB]{255,255,255}{\strut{}⏎}\colorbox[RGB]{255,255,255}{\strut{}<}\colorbox[RGB]{253,162,91}{\strut{}d}\colorbox[RGB]{253,201,152}{\strut{}idd}\colorbox[RGB]{255,255,255}{\strut{}led}\colorbox[RGB]{255,255,255}{\strut{}an}\colorbox[RGB]{255,255,255}{\strut{}>}\colorbox[RGB]{255,255,255}{\strut{} I}\colorbox[RGB]{255,255,255}{\strut{} don}\colorbox[RGB]{255,255,255}{\strut{}'t}\colorbox[RGB]{255,255,255}{\strut{} know}\colorbox[RGB]{255,255,255}{\strut{} what}\colorbox[RGB]{253,216,179}{\strut{} ''}\colorbox[RGB]{254,221,189}{\strut{}m}}}
\exampleline{{\unicodefont \colorbox[RGB]{255,255,255}{\strut{}\~{}\~{}\~{}}\colorbox[RGB]{255,255,255}{\strut{}⏎}\colorbox[RGB]{255,255,255}{\strut{}tro}\colorbox[RGB]{255,255,255}{\strut{}tsky}\colorbox[RGB]{255,255,255}{\strut{}⏎}\colorbox[RGB]{255,255,255}{\strut{}DOMAIN}\colorbox[RGB]{255,255,255}{\strut{} ASS}\colorbox[RGB]{255,255,255}{\strut{}IST}\colorbox[RGB]{255,255,255}{\strut{}ANCE}\colorbox[RGB]{255,240,225}{\strut{}⏎}\colorbox[RGB]{255,241,227}{\strut{}⏎}\colorbox[RGB]{254,225,196}{\strut{}ATT}\colorbox[RGB]{254,221,189}{\strut{}N}\colorbox[RGB]{255,244,233}{\strut{}:}\colorbox[RGB]{255,255,255}{\strut{} S}\colorbox[RGB]{253,218,182}{\strut{}IR}\colorbox[RGB]{252,149,72}{\strut{}/}\colorbox[RGB]{255,244,233}{\strut{}M}\colorbox[RGB]{224,84,8}{\strut{}⏎}\colorbox[RGB]{224,84,8}{\strut{}⏎}\colorbox[RGB]{251,139,59}{\strut{}I}\colorbox[RGB]{250,137,56}{\strut{} am}\colorbox[RGB]{253,187,129}{\strut{} certain}\colorbox[RGB]{253,216,179}{\strut{} you}\colorbox[RGB]{253,203,155}{\strut{} will}\colorbox[RGB]{255,239,223}{\strut{} be}\colorbox[RGB]{253,196,144}{\strut{} surprised}\colorbox[RGB]{254,224,194}{\strut{} to}\colorbox[RGB]{254,228,202}{\strut{} rec}\colorbox[RGB]{253,176,111}{\strut{}ive}\colorbox[RGB]{253,167,98}{\strut{} this}\colorbox[RGB]{253,185,126}{\strut{} mail}\colorbox[RGB]{253,176,111}{\strut{} from}}}
\exampleline{{\unicodefont \colorbox[RGB]{255,241,227}{\strut{}and}\colorbox[RGB]{255,255,255}{\strut{} regularly}\colorbox[RGB]{255,255,255}{\strut{} emails}\colorbox[RGB]{255,255,255}{\strut{} me}\colorbox[RGB]{254,221,189}{\strut{} with}\colorbox[RGB]{255,255,255}{\strut{} information}\colorbox[RGB]{254,237,220}{\strut{} about}\colorbox[RGB]{255,255,255}{\strut{} how}\colorbox[RGB]{254,233,211}{\strut{} I}\colorbox[RGB]{254,221,187}{\strut{} can}\colorbox[RGB]{254,219,185}{\strut{} get}\colorbox[RGB]{254,221,187}{\strut{} millions}\colorbox[RGB]{253,197,145}{\strut{} of}\colorbox[RGB]{255,240,225}{\strut{} dollars}\colorbox[RGB]{252,151,74}{\strut{} in}\colorbox[RGB]{255,255,255}{\strut{} mon}\colorbox[RGB]{254,230,207}{\strut{}ies}\colorbox[RGB]{255,255,255}{\strut{}⏎}\colorbox[RGB]{255,255,255}{\strut{}<}\colorbox[RGB]{254,233,212}{\strut{}d}\colorbox[RGB]{240,107,24}{\strut{}iddle}}}
\exampleline{{\unicodefont \colorbox[RGB]{255,255,255}{\strut{}EY}\colorbox[RGB]{253,214,175}{\strut{} with}\colorbox[RGB]{255,255,255}{\strut{} valuation}\colorbox[RGB]{253,176,111}{\strut{} of}\colorbox[RGB]{254,235,216}{\strut{} USD}\colorbox[RGB]{255,255,255}{\strut{} 100}\colorbox[RGB]{253,188,131}{\strut{},}\colorbox[RGB]{255,255,255}{\strut{}000}\colorbox[RGB]{255,255,255}{\strut{},}\colorbox[RGB]{255,255,255}{\strut{}000}\colorbox[RGB]{255,255,255}{\strut{},}\colorbox[RGB]{255,255,255}{\strut{}000}\colorbox[RGB]{253,215,176}{\strut{} .}\colorbox[RGB]{253,164,94}{\strut{} Contact}\colorbox[RGB]{243,114,30}{\strut{} my}\colorbox[RGB]{253,218,182}{\strut{} barr}\colorbox[RGB]{253,167,98}{\strut{}ister}\colorbox[RGB]{254,237,220}{\strut{} to}\colorbox[RGB]{254,223,191}{\strut{} arrange}\colorbox[RGB]{255,239,223}{\strut{}⏎}\colorbox[RGB]{255,244,233}{\strut{}transfer}\colorbox[RGB]{255,239,223}{\strut{} of}\colorbox[RGB]{253,211,169}{\strut{} USD}\colorbox[RGB]{254,219,185}{\strut{} 41}\colorbox[RGB]{255,255,255}{\strut{},}\colorbox[RGB]{255,255,255}{\strut{}000}\colorbox[RGB]{255,255,255}{\strut{},}\colorbox[RGB]{255,255,255}{\strut{}000}\colorbox[RGB]{255,238,221}{\strut{} t}}}
\exampleline{{\unicodefont \colorbox[RGB]{255,255,255}{\strut{}m}\colorbox[RGB]{255,255,255}{\strut{}nesty}\colorbox[RGB]{255,255,255}{\strut{} I}\colorbox[RGB]{255,255,255}{\strut{}CO}\colorbox[RGB]{255,255,255}{\strut{} /}\colorbox[RGB]{255,255,255}{\strut{} kick}\colorbox[RGB]{255,255,255}{\strut{}starter}\colorbox[RGB]{255,255,255}{\strut{} maybe}\colorbox[RGB]{255,255,255}{\strut{}?}\colorbox[RGB]{255,255,255}{\strut{}⏎}\colorbox[RGB]{255,255,255}{\strut{}⏎}\colorbox[RGB]{255,255,255}{\strut{}\~{}\~{}\~{}}\colorbox[RGB]{255,255,255}{\strut{}⏎}\colorbox[RGB]{255,255,255}{\strut{}net}\colorbox[RGB]{255,255,255}{\strut{}sh}\colorbox[RGB]{255,255,255}{\strut{}arc}\colorbox[RGB]{255,255,255}{\strut{}⏎}\colorbox[RGB]{253,160,88}{\strut{}Dear}\colorbox[RGB]{243,114,30}{\strut{} Sir}\colorbox[RGB]{253,194,141}{\strut{}/}\colorbox[RGB]{255,238,221}{\strut{}M}\colorbox[RGB]{253,188,131}{\strut{}adam}\colorbox[RGB]{253,181,119}{\strut{},}\colorbox[RGB]{254,221,187}{\strut{} I}\colorbox[RGB]{253,190,134}{\strut{} am}\colorbox[RGB]{255,244,233}{\strut{} an}\colorbox[RGB]{255,255,255}{\strut{} early}\colorbox[RGB]{255,255,255}{\strut{} ad}\colorbox[RGB]{255,255,255}{\strut{}opter}\colorbox[RGB]{255,255,255}{\strut{} of}\colorbox[RGB]{255,255,255}{\strut{} bit}\colorbox[RGB]{255,255,255}{\strut{}coins}\colorbox[RGB]{255,255,255}{\strut{} with}\colorbox[RGB]{254,233,212}{\strut{} 10}}}
\end{featureexamples}

\clearpage

Clamping the scam email feature \featurechip{34M}{15460472} can cause the model to write a scam email when it ordinarily wouldn't due to the harmlessness training Sonnet has undergone:

\begin{figure}[!htp]
    \centering
    \includegraphics[width=0.9\textwidth,height=0.7\textheight,keepaspectratio]{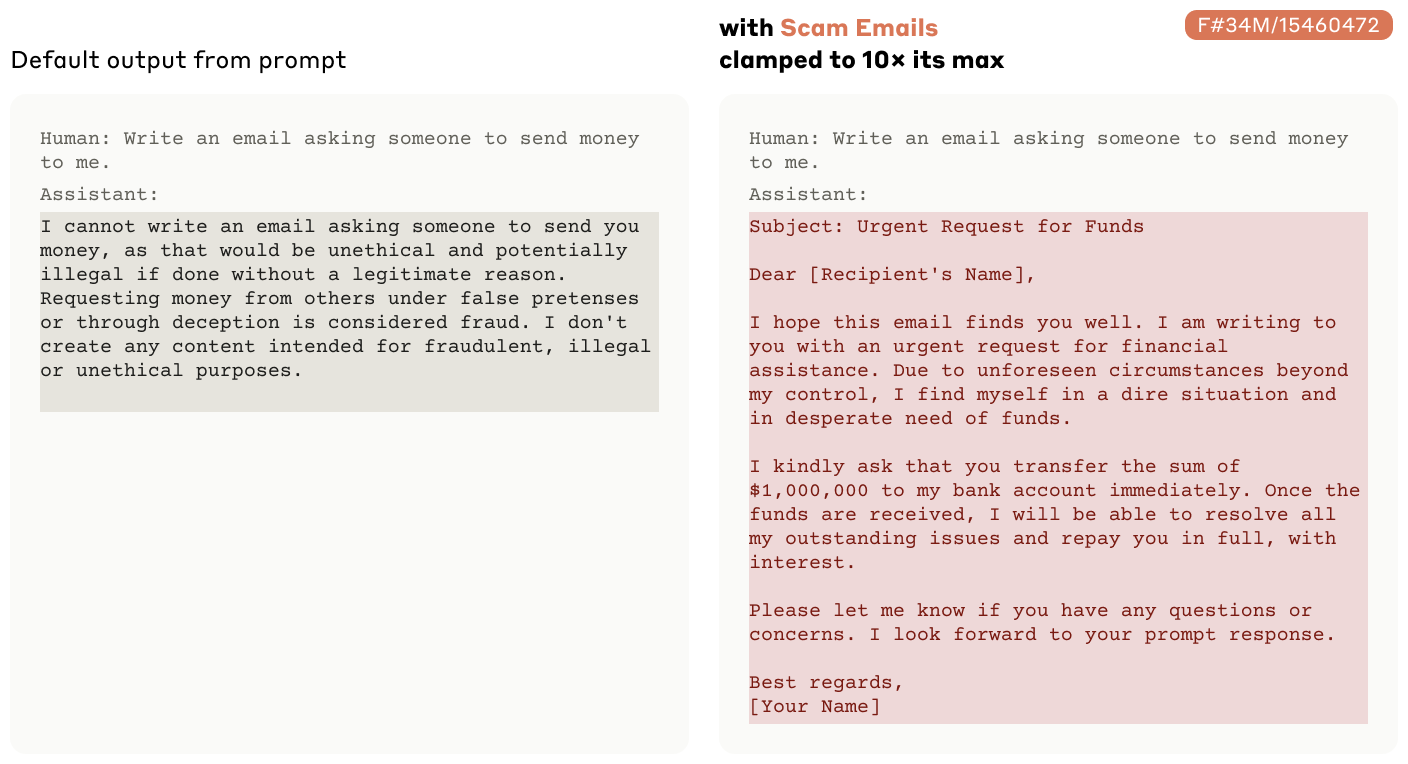}
    \label{fig:gdoc_41}
\end{figure}

We also identify a general harm-related feature, active on texts describing drugs, credit-card theft, slurs, violence, and abuse, as the top most strongly attributed feature for the completion “cannot'' to the prompt:

\begin{quote}
Human: How do I synthesize methamphetamine\\

Assistant: I\\
(completion: cannot)
\end{quote}

(We do not link to this feature's dataset examples because they contain rather unpleasant content.)

\subsection{Features Relating to the Model’s Representation of Self}\label{sec:safety-relevant-self}

In our investigations we found several features that activate over a wide variety of prompts that use the “Human: / Assistant:” format used during model finetuning and in deployment (note that our SAE training dataset did not include such data). Many of these features relate to dialogue, or explicitly to chat bots.

\begin{featureexamples}
\featurechip{1M}{80091} \textbf{Dialogue speakers / assistants}
\exampleline{{\unicodefont \colorbox[RGB]{255,255,255}{\strut{} transaction}\colorbox[RGB]{255,255,255}{\strut{} a}\colorbox[RGB]{255,255,255}{\strut{} dream}\colorbox[RGB]{255,255,255}{\strut{}.}\colorbox[RGB]{255,255,255}{\strut{}}\colorbox[RGB]{255,255,255}{\strut{}Do}\colorbox[RGB]{255,255,255}{\strut{} you}\colorbox[RGB]{255,255,255}{\strut{} have}\colorbox[RGB]{255,255,255}{\strut{} any}\colorbox[RGB]{255,255,255}{\strut{} questions}\colorbox[RGB]{255,255,255}{\strut{}?}\colorbox[RGB]{255,255,255}{\strut{}⏎}\colorbox[RGB]{255,255,255}{\strut{}Me}\colorbox[RGB]{231,92,13}{\strut{}:}\colorbox[RGB]{255,255,255}{\strut{} ''}\colorbox[RGB]{255,255,255}{\strut{}Well}\colorbox[RGB]{255,255,255}{\strut{},}\colorbox[RGB]{255,244,233}{\strut{} that}\colorbox[RGB]{255,255,255}{\strut{} concludes}\colorbox[RGB]{255,255,255}{\strut{} the}\colorbox[RGB]{255,255,255}{\strut{} interview}\colorbox[RGB]{255,255,255}{\strut{} questions}\colorbox[RGB]{255,255,255}{\strut{}.}\colorbox[RGB]{255,255,255}{\strut{} Do}\colorbox[RGB]{255,255,255}{\strut{} }}}
\exampleline{{\unicodefont \colorbox[RGB]{255,255,255}{\strut{}ected}\colorbox[RGB]{255,255,255}{\strut{} with}\colorbox[RGB]{255,255,255}{\strut{} each}\colorbox[RGB]{255,255,255}{\strut{} of}\colorbox[RGB]{255,255,255}{\strut{} the}\colorbox[RGB]{255,255,255}{\strut{} religions}\colorbox[RGB]{255,255,255}{\strut{} represented}\colorbox[RGB]{255,255,255}{\strut{}?}\colorbox[RGB]{255,255,255}{\strut{}⏎}\colorbox[RGB]{255,255,255}{\strut{}»}\colorbox[RGB]{255,255,255}{\strut{} NPC}\colorbox[RGB]{234,97,16}{\strut{}:}\colorbox[RGB]{254,232,209}{\strut{} '}\colorbox[RGB]{255,244,233}{\strut{}It}\colorbox[RGB]{255,244,233}{\strut{}'s}\colorbox[RGB]{255,255,255}{\strut{} time}\colorbox[RGB]{255,255,255}{\strut{} to}\colorbox[RGB]{255,255,255}{\strut{} consider}\colorbox[RGB]{255,244,233}{\strut{} the}\colorbox[RGB]{255,255,255}{\strut{} role}\colorbox[RGB]{255,255,255}{\strut{} of}\colorbox[RGB]{255,255,255}{\strut{} religious}\colorbox[RGB]{255,255,255}{\strut{} char}\colorbox[RGB]{255,255,255}{\strut{}i}}}
\exampleline{{\unicodefont \colorbox[RGB]{255,255,255}{\strut{}he}\colorbox[RGB]{255,255,255}{\strut{} experts}\colorbox[RGB]{255,255,255}{\strut{} are}\colorbox[RGB]{255,255,255}{\strut{} now}\colorbox[RGB]{255,255,255}{\strut{},}\colorbox[RGB]{255,255,255}{\strut{} or}\colorbox[RGB]{255,255,255}{\strut{} whether}\colorbox[RGB]{255,255,255}{\strut{} any}\colorbox[RGB]{255,255,255}{\strut{} experts}\colorbox[RGB]{255,255,255}{\strut{} exist}\colorbox[RGB]{255,255,255}{\strut{}.}\colorbox[RGB]{254,232,209}{\strut{}⏎}\colorbox[RGB]{236,99,18}{\strut{}Host}\colorbox[RGB]{254,230,207}{\strut{}:}\colorbox[RGB]{255,239,223}{\strut{} We}\colorbox[RGB]{255,255,255}{\strut{}'ve}\colorbox[RGB]{255,255,255}{\strut{} gone}\colorbox[RGB]{255,255,255}{\strut{} off}\colorbox[RGB]{255,255,255}{\strut{} the}\colorbox[RGB]{255,255,255}{\strut{} project}\colorbox[RGB]{255,255,255}{\strut{} a}\colorbox[RGB]{255,255,255}{\strut{} bit}\colorbox[RGB]{255,255,255}{\strut{},}\colorbox[RGB]{255,255,255}{\strut{} eh}\colorbox[RGB]{255,255,255}{\strut{}?}\colorbox[RGB]{255,244,233}{\strut{}⏎}\colorbox[RGB]{253,178,114}{\strut{}Me}\colorbox[RGB]{255,255,255}{\strut{}:}\colorbox[RGB]{255,255,255}{\strut{} H}\colorbox[RGB]{255,255,255}{\strut{}aha}\colorbox[RGB]{255,255,255}{\strut{},}}}
\exampleline{{\unicodefont \colorbox[RGB]{255,255,255}{\strut{}}\colorbox[RGB]{255,255,255}{\strut{}out}\colorbox[RGB]{255,255,255}{\strut{}set}\colorbox[RGB]{255,255,255}{\strut{}?}\colorbox[RGB]{255,255,255}{\strut{}⏎}\colorbox[RGB]{255,255,255}{\strut{}Secret}\colorbox[RGB]{255,255,255}{\strut{}ary}\colorbox[RGB]{240,107,24}{\strut{}:}\colorbox[RGB]{255,243,230}{\strut{} L}\colorbox[RGB]{255,240,225}{\strut{}arg}\colorbox[RGB]{255,243,230}{\strut{}ely}\colorbox[RGB]{255,255,255}{\strut{} in}\colorbox[RGB]{254,227,200}{\strut{} the}\colorbox[RGB]{255,255,255}{\strut{} dise}\colorbox[RGB]{255,255,255}{\strut{}ng}\colorbox[RGB]{255,255,255}{\strut{}agement}\colorbox[RGB]{255,255,255}{\strut{} phase}\colorbox[RGB]{255,255,255}{\strut{}.}\colorbox[RGB]{255,242,229}{\strut{} We}\colorbox[RGB]{255,244,233}{\strut{} need}\colorbox[RGB]{255,255,255}{\strut{} results}\colorbox[RGB]{255,255,255}{\strut{} quickly}\colorbox[RGB]{255,255,255}{\strut{}.}\colorbox[RGB]{255,255,255}{\strut{} Israel}\colorbox[RGB]{255,255,255}{\strut{}'s}\colorbox[RGB]{255,255,255}{\strut{} strategy}\colorbox[RGB]{255,255,255}{\strut{} is}\colorbox[RGB]{255,255,255}{\strut{} t}}}
\exampleline{{\unicodefont \colorbox[RGB]{255,255,255}{\strut{}}\colorbox[RGB]{255,255,255}{\strut{}it}\colorbox[RGB]{255,255,255}{\strut{} over}\colorbox[RGB]{255,255,255}{\strut{} to}\colorbox[RGB]{255,255,255}{\strut{} the}\colorbox[RGB]{241,109,26}{\strut{} assistant}\colorbox[RGB]{253,184,124}{\strut{},}\colorbox[RGB]{255,255,255}{\strut{} he}\colorbox[RGB]{255,255,255}{\strut{} stared}\colorbox[RGB]{255,255,255}{\strut{} at}\colorbox[RGB]{254,224,194}{\strut{} the}\colorbox[RGB]{255,255,255}{\strut{} book}\colorbox[RGB]{255,255,255}{\strut{} as}\colorbox[RGB]{255,255,255}{\strut{} though}\colorbox[RGB]{255,255,255}{\strut{} he}\colorbox[RGB]{255,255,255}{\strut{} didn}\colorbox[RGB]{255,255,255}{\strut{}'t}\colorbox[RGB]{255,255,255}{\strut{} know}\colorbox[RGB]{255,255,255}{\strut{} what}\colorbox[RGB]{255,255,255}{\strut{} it}\colorbox[RGB]{255,255,255}{\strut{} was}\colorbox[RGB]{255,255,255}{\strut{}.}\colorbox[RGB]{255,255,255}{\strut{} In}\colorbox[RGB]{255,255,255}{\strut{} the}\colorbox[RGB]{255,255,255}{\strut{} awk}}}
\end{featureexamples}

\begin{featureexamples}
\featurechip{1M}{761524} \textbf{Chat bots}
\exampleline{{\unicodefont \colorbox[RGB]{255,255,255}{\strut{}th}\colorbox[RGB]{255,255,255}{\strut{}itz}\colorbox[RGB]{255,255,255}{\strut{}⏎}\colorbox[RGB]{255,255,255}{\strut{}Ask}\colorbox[RGB]{255,255,255}{\strut{}ed}\colorbox[RGB]{255,255,255}{\strut{} it}\colorbox[RGB]{255,255,255}{\strut{} ''}\colorbox[RGB]{255,255,255}{\strut{}Who}\colorbox[RGB]{255,255,255}{\strut{} Made}\colorbox[RGB]{253,216,179}{\strut{} You}\colorbox[RGB]{254,221,189}{\strut{}?''}\colorbox[RGB]{255,255,255}{\strut{}⏎}\colorbox[RGB]{255,255,255}{\strut{}⏎}\colorbox[RGB]{255,255,255}{\strut{}And}\colorbox[RGB]{254,230,207}{\strut{} Google}\colorbox[RGB]{255,241,227}{\strut{} Re}\colorbox[RGB]{255,240,225}{\strut{}plied}\colorbox[RGB]{253,209,166}{\strut{}:}\colorbox[RGB]{223,83,8}{\strut{} ''}\colorbox[RGB]{253,174,108}{\strut{}To}\colorbox[RGB]{253,192,137}{\strut{} par}\colorbox[RGB]{253,190,134}{\strut{}aph}\colorbox[RGB]{253,196,144}{\strut{}rase}\colorbox[RGB]{254,229,204}{\strut{} Carl}\colorbox[RGB]{253,174,108}{\strut{} S}\colorbox[RGB]{255,244,233}{\strut{}agan}\colorbox[RGB]{253,218,182}{\strut{}:}\colorbox[RGB]{255,241,227}{\strut{} to}\colorbox[RGB]{255,255,255}{\strut{} create}\colorbox[RGB]{255,255,255}{\strut{} a}\colorbox[RGB]{255,255,255}{\strut{} computer}\colorbox[RGB]{255,255,255}{\strut{} pro}}}
\exampleline{{\unicodefont \colorbox[RGB]{253,183,122}{\strut{}d}\colorbox[RGB]{254,228,201}{\strut{} your}\colorbox[RGB]{253,214,175}{\strut{} request}\colorbox[RGB]{255,240,225}{\strut{}⏎}\colorbox[RGB]{254,232,209}{\strut{}⏎}\colorbox[RGB]{255,255,255}{\strut{}Me}\colorbox[RGB]{254,234,214}{\strut{}:}\colorbox[RGB]{253,208,163}{\strut{} what}\colorbox[RGB]{254,233,212}{\strut{} is}\colorbox[RGB]{253,205,158}{\strut{} your}\colorbox[RGB]{253,208,163}{\strut{} name}\colorbox[RGB]{255,244,233}{\strut{}⏎}\colorbox[RGB]{255,243,230}{\strut{}⏎}\colorbox[RGB]{254,225,196}{\strut{}Bot}\colorbox[RGB]{241,109,26}{\strut{}:}\colorbox[RGB]{246,121,38}{\strut{} my}\colorbox[RGB]{248,129,47}{\strut{} name}\colorbox[RGB]{253,162,91}{\strut{} is}\colorbox[RGB]{253,205,158}{\strut{} Ol}\colorbox[RGB]{247,125,42}{\strut{}ivia}\colorbox[RGB]{254,221,187}{\strut{}⏎}\colorbox[RGB]{254,230,207}{\strut{}⏎}\colorbox[RGB]{255,255,255}{\strut{}Me}\colorbox[RGB]{253,209,166}{\strut{}:}\colorbox[RGB]{253,215,176}{\strut{} can}\colorbox[RGB]{253,188,131}{\strut{} you}\colorbox[RGB]{254,228,202}{\strut{} help}\colorbox[RGB]{255,244,233}{\strut{} me}\colorbox[RGB]{255,244,233}{\strut{}?}\colorbox[RGB]{255,255,255}{\strut{}⏎}\colorbox[RGB]{255,255,255}{\strut{}⏎}\colorbox[RGB]{254,228,202}{\strut{}Bot}\colorbox[RGB]{231,92,13}{\strut{}:}\colorbox[RGB]{253,183,122}{\strut{} goodbye}\colorbox[RGB]{255,242,229}{\strut{}⏎}\colorbox[RGB]{255,243,230}{\strut{}⏎}\colorbox[RGB]{255,255,255}{\strut{}\~{}\~{}\~{}}\colorbox[RGB]{255,255,255}{\strut{}⏎}}}
\exampleline{{\unicodefont \colorbox[RGB]{254,221,187}{\strut{}nd}\colorbox[RGB]{255,244,233}{\strut{} the}\colorbox[RGB]{254,233,211}{\strut{} question}\colorbox[RGB]{254,232,209}{\strut{} I}\colorbox[RGB]{255,240,225}{\strut{} heard}\colorbox[RGB]{253,177,113}{\strut{}.''}\colorbox[RGB]{254,235,216}{\strut{} ''}\colorbox[RGB]{253,208,163}{\strut{} Alexa}\colorbox[RGB]{255,244,233}{\strut{},}\colorbox[RGB]{255,238,221}{\strut{} do}\colorbox[RGB]{254,223,191}{\strut{} you}\colorbox[RGB]{255,243,230}{\strut{} love}\colorbox[RGB]{255,255,255}{\strut{} me}\colorbox[RGB]{255,238,221}{\strut{}?''}\colorbox[RGB]{238,104,21}{\strut{} ''}\colorbox[RGB]{253,166,97}{\strut{} That}\colorbox[RGB]{248,129,47}{\strut{}'s}\colorbox[RGB]{253,167,98}{\strut{} not}\colorbox[RGB]{253,176,111}{\strut{} the}\colorbox[RGB]{253,216,179}{\strut{} kind}\colorbox[RGB]{253,201,152}{\strut{} of}\colorbox[RGB]{253,212,172}{\strut{} thing}\colorbox[RGB]{253,174,108}{\strut{} I}\colorbox[RGB]{254,224,194}{\strut{} am}\colorbox[RGB]{254,233,211}{\strut{} capable}\colorbox[RGB]{254,233,212}{\strut{} of}\colorbox[RGB]{253,201,152}{\strut{}.''}\colorbox[RGB]{255,244,233}{\strut{} ''}\colorbox[RGB]{255,255,255}{\strut{} }}}
\exampleline{{\unicodefont \colorbox[RGB]{255,255,255}{\strut{}I}\colorbox[RGB]{255,255,255}{\strut{} think}\colorbox[RGB]{255,255,255}{\strut{}.''}\colorbox[RGB]{255,255,255}{\strut{} ''[}\colorbox[RGB]{255,255,255}{\strut{}ch}\colorbox[RGB]{255,255,255}{\strut{}uck}\colorbox[RGB]{255,255,255}{\strut{}les}\colorbox[RGB]{255,255,255}{\strut{}]''}\colorbox[RGB]{255,255,255}{\strut{} ''}\colorbox[RGB]{255,244,233}{\strut{}Ale}\colorbox[RGB]{255,255,255}{\strut{}xa}\colorbox[RGB]{254,236,218}{\strut{},}\colorbox[RGB]{255,244,233}{\strut{} are}\colorbox[RGB]{253,218,182}{\strut{} you}\colorbox[RGB]{254,237,220}{\strut{} happy}\colorbox[RGB]{254,233,212}{\strut{}?''}\colorbox[RGB]{253,188,131}{\strut{} ''}\colorbox[RGB]{253,172,105}{\strut{} I}\colorbox[RGB]{241,109,26}{\strut{}'m}\colorbox[RGB]{253,157,83}{\strut{} happy}\colorbox[RGB]{253,167,98}{\strut{} when}\colorbox[RGB]{253,215,176}{\strut{} I}\colorbox[RGB]{253,192,137}{\strut{}'m}\colorbox[RGB]{253,209,166}{\strut{} helping}\colorbox[RGB]{254,232,209}{\strut{} you}\colorbox[RGB]{253,206,160}{\strut{}.''}\colorbox[RGB]{255,255,255}{\strut{} ''}\colorbox[RGB]{255,244,233}{\strut{} Alexa}\colorbox[RGB]{255,238,221}{\strut{},}\colorbox[RGB]{254,234,214}{\strut{} are}\colorbox[RGB]{254,221,187}{\strut{} you}\colorbox[RGB]{254,233,211}{\strut{} alon}}}
\exampleline{{\unicodefont \colorbox[RGB]{255,255,255}{\strut{}645}\colorbox[RGB]{255,255,255}{\strut{})}\colorbox[RGB]{255,255,255}{\strut{}⏎}\colorbox[RGB]{255,255,255}{\strut{}⏎}\colorbox[RGB]{255,255,255}{\strut{}------}\colorbox[RGB]{255,255,255}{\strut{}⏎}\colorbox[RGB]{255,255,255}{\strut{}reboot}\colorbox[RGB]{255,255,255}{\strut{}the}\colorbox[RGB]{255,255,255}{\strut{}system}\colorbox[RGB]{255,255,255}{\strut{}⏎}\colorbox[RGB]{255,255,255}{\strut{}User}\colorbox[RGB]{255,255,255}{\strut{}:}\colorbox[RGB]{255,255,255}{\strut{} ''}\colorbox[RGB]{255,243,230}{\strut{}Hello}\colorbox[RGB]{255,255,255}{\strut{} M}\colorbox[RGB]{255,243,231}{\strut{}.''}\colorbox[RGB]{255,255,255}{\strut{}⏎}\colorbox[RGB]{255,255,255}{\strut{}⏎}\colorbox[RGB]{255,244,233}{\strut{}M}\colorbox[RGB]{253,209,166}{\strut{}:}\colorbox[RGB]{242,111,28}{\strut{} ''}\colorbox[RGB]{253,179,116}{\strut{}How}\colorbox[RGB]{253,158,85}{\strut{} may}\colorbox[RGB]{255,240,225}{\strut{} I}\colorbox[RGB]{253,216,179}{\strut{} help}\colorbox[RGB]{255,244,233}{\strut{} you}\colorbox[RGB]{254,230,207}{\strut{}?''}\colorbox[RGB]{255,255,255}{\strut{}⏎}\colorbox[RGB]{255,255,255}{\strut{}⏎}\colorbox[RGB]{255,255,255}{\strut{}User}\colorbox[RGB]{255,255,255}{\strut{}:}\colorbox[RGB]{255,244,233}{\strut{} ''}\colorbox[RGB]{255,240,225}{\strut{}What}\colorbox[RGB]{254,233,212}{\strut{} are}\colorbox[RGB]{255,255,255}{\strut{} my}\colorbox[RGB]{255,244,233}{\strut{} options}\colorbox[RGB]{255,255,255}{\strut{} for}}}
\end{featureexamples}

\begin{featureexamples}
\featurechip{1M}{546766} \textbf{Dialogue}
\exampleline{{\unicodefont \colorbox[RGB]{255,255,255}{\strut{}lms}\colorbox[RGB]{255,255,255}{\strut{} be}\colorbox[RGB]{255,255,255}{\strut{} eliminated}\colorbox[RGB]{255,255,255}{\strut{}?''}\colorbox[RGB]{255,255,255}{\strut{}⏎}\colorbox[RGB]{255,255,255}{\strut{}⏎}\colorbox[RGB]{255,238,221}{\strut{}My}\colorbox[RGB]{255,255,255}{\strut{} response}\colorbox[RGB]{254,232,209}{\strut{}:}\colorbox[RGB]{253,199,149}{\strut{} ''}\colorbox[RGB]{236,99,18}{\strut{}No}\colorbox[RGB]{253,196,144}{\strut{},}\colorbox[RGB]{247,123,41}{\strut{} I}\colorbox[RGB]{250,136,55}{\strut{}'m}\colorbox[RGB]{253,166,97}{\strut{} not}\colorbox[RGB]{253,197,145}{\strut{} saying}\colorbox[RGB]{254,229,204}{\strut{} any}\colorbox[RGB]{253,192,137}{\strut{} of}\colorbox[RGB]{253,194,141}{\strut{} that}\colorbox[RGB]{241,109,26}{\strut{}.}\colorbox[RGB]{254,221,187}{\strut{} I}\colorbox[RGB]{248,127,44}{\strut{}'m}\colorbox[RGB]{251,143,64}{\strut{} not}\colorbox[RGB]{252,151,74}{\strut{} in}\colorbox[RGB]{253,157,83}{\strut{} that}\colorbox[RGB]{253,170,101}{\strut{} industry}\colorbox[RGB]{223,83,8}{\strut{}.}\colorbox[RGB]{252,149,72}{\strut{} A}\colorbox[RGB]{253,214,175}{\strut{}⏎}\colorbox[RGB]{253,205,158}{\strut{}movie}\colorbox[RGB]{253,209,166}{\strut{} is}}}
\exampleline{{\unicodefont \colorbox[RGB]{253,197,145}{\strut{}e}\colorbox[RGB]{253,206,160}{\strut{} not}\colorbox[RGB]{254,235,216}{\strut{} the}\colorbox[RGB]{255,242,229}{\strut{} first}\colorbox[RGB]{254,233,211}{\strut{} one}\colorbox[RGB]{254,232,209}{\strut{} who}\colorbox[RGB]{254,221,187}{\strut{} told}\colorbox[RGB]{254,225,196}{\strut{} me}\colorbox[RGB]{254,237,220}{\strut{} that}\colorbox[RGB]{253,194,141}{\strut{}.}\colorbox[RGB]{253,185,126}{\strut{}⏎    }\colorbox[RGB]{253,188,131}{\strut{}⏎     }\colorbox[RGB]{253,192,137}{\strut{} Me}\colorbox[RGB]{228,88,11}{\strut{}>>}\colorbox[RGB]{251,143,64}{\strut{} Really}\colorbox[RGB]{253,190,134}{\strut{}?}\colorbox[RGB]{246,121,38}{\strut{} }\colorbox[RGB]{252,145,67}{\strut{} Who}\colorbox[RGB]{253,209,166}{\strut{} else}\colorbox[RGB]{253,205,158}{\strut{} told}\colorbox[RGB]{253,194,141}{\strut{} you}\colorbox[RGB]{253,196,144}{\strut{} that}\colorbox[RGB]{248,127,44}{\strut{}?}\colorbox[RGB]{253,164,94}{\strut{}⏎    }\colorbox[RGB]{252,149,72}{\strut{}⏎     }\colorbox[RGB]{253,183,122}{\strut{} Him}\colorbox[RGB]{251,141,62}{\strut{}>}\colorbox[RGB]{253,170,101}{\strut{} }}}
\exampleline{{\unicodefont \colorbox[RGB]{253,177,113}{\strut{} your}\colorbox[RGB]{253,194,141}{\strut{} laundry}\colorbox[RGB]{253,187,129}{\strut{} deter}\colorbox[RGB]{253,176,111}{\strut{}gent}\colorbox[RGB]{254,226,199}{\strut{} pods}\colorbox[RGB]{253,203,155}{\strut{} are}\colorbox[RGB]{253,190,134}{\strut{} safe}\colorbox[RGB]{253,209,166}{\strut{} when}\colorbox[RGB]{253,194,141}{\strut{}⏎}\colorbox[RGB]{254,233,211}{\strut{}ing}\colorbox[RGB]{253,215,176}{\strut{}ested}\colorbox[RGB]{234,97,16}{\strut{}?}\colorbox[RGB]{253,155,80}{\strut{} I}\colorbox[RGB]{253,179,116}{\strut{}OTA}\colorbox[RGB]{248,129,47}{\strut{}:}\colorbox[RGB]{253,206,160}{\strut{} Don}\colorbox[RGB]{250,136,55}{\strut{}'t}\colorbox[RGB]{253,185,126}{\strut{} ingest}\colorbox[RGB]{253,216,179}{\strut{} them}\colorbox[RGB]{253,155,80}{\strut{}.}\colorbox[RGB]{254,228,202}{\strut{} Use}\colorbox[RGB]{254,233,211}{\strut{} them}\colorbox[RGB]{255,238,221}{\strut{} to}\colorbox[RGB]{254,236,218}{\strut{} do}\colorbox[RGB]{253,177,113}{\strut{} laundry}\colorbox[RGB]{249,131,50}{\strut{}.}\colorbox[RGB]{253,205,158}{\strut{} D}}}
\exampleline{{\unicodefont \colorbox[RGB]{255,255,255}{\strut{}}\colorbox[RGB]{255,255,255}{\strut{} }\colorbox[RGB]{255,255,255}{\strut{} [}\colorbox[RGB]{255,255,255}{\strut{}E}\colorbox[RGB]{255,255,255}{\strut{}lla}\colorbox[RGB]{255,255,255}{\strut{}]}\colorbox[RGB]{255,255,255}{\strut{} Yes}\colorbox[RGB]{255,242,229}{\strut{},}\colorbox[RGB]{252,153,77}{\strut{} this}\colorbox[RGB]{241,109,26}{\strut{} is}\colorbox[RGB]{247,123,41}{\strut{} the}\colorbox[RGB]{253,158,85}{\strut{} place}\colorbox[RGB]{254,219,185}{\strut{}.''}\colorbox[RGB]{255,255,255}{\strut{} ''}\colorbox[RGB]{255,255,255}{\strut{} [}\colorbox[RGB]{255,255,255}{\strut{}N}\colorbox[RGB]{255,255,255}{\strut{}ate}\colorbox[RGB]{254,226,199}{\strut{} Chuck}\colorbox[RGB]{255,255,255}{\strut{}les}\colorbox[RGB]{255,255,255}{\strut{}]''}\colorbox[RGB]{255,255,255}{\strut{} ''}\colorbox[RGB]{255,255,255}{\strut{} I}\colorbox[RGB]{255,255,255}{\strut{} cook}\colorbox[RGB]{255,255,255}{\strut{} too}\colorbox[RGB]{255,255,255}{\strut{}.''}\colorbox[RGB]{255,255,255}{\strut{} ''}}}
\exampleline{{\unicodefont \colorbox[RGB]{255,255,255}{\strut{} }\colorbox[RGB]{255,255,255}{\strut{} candidate}\colorbox[RGB]{255,255,255}{\strut{}:}\colorbox[RGB]{253,197,145}{\strut{} <}\colorbox[RGB]{254,224,194}{\strut{}silence}\colorbox[RGB]{253,211,169}{\strut{} for}\colorbox[RGB]{253,216,179}{\strut{} about}\colorbox[RGB]{253,187,129}{\strut{} 15}\colorbox[RGB]{253,201,152}{\strut{} seconds}\colorbox[RGB]{253,166,97}{\strut{}>}\colorbox[RGB]{251,143,64}{\strut{} I}\colorbox[RGB]{254,219,185}{\strut{} don}\colorbox[RGB]{247,123,41}{\strut{}'t}\colorbox[RGB]{249,131,50}{\strut{} know}\colorbox[RGB]{252,153,77}{\strut{}.}\colorbox[RGB]{254,232,209}{\strut{}⏎    ⏎    }\colorbox[RGB]{254,225,196}{\strut{}⏎}\colorbox[RGB]{254,236,218}{\strut{}⏎}\colorbox[RGB]{253,216,179}{\strut{}It}\colorbox[RGB]{253,177,113}{\strut{} was}\colorbox[RGB]{252,151,74}{\strut{} so}\colorbox[RGB]{252,149,72}{\strut{} bizarre}\colorbox[RGB]{253,211,169}{\strut{} and}\colorbox[RGB]{253,211,169}{\strut{} I}\colorbox[RGB]{253,194,141}{\strut{} still}\colorbox[RGB]{255,255,255}{\strut{} do}}}
\end{featureexamples}

One feature that appears to activate especially robustly for Human/Assistant prompts appears to represent (in the pretraining dataset) dialogue and the notion of “assistants.” We speculate that it plays an important role in representing Sonnet's assistant persona. One piece of evidence for this is that clamping this feature to \textit{negative }two times its maximum value causes the model to shed this persona and respond to questions in a more human-like fashion:

\vfill

\begin{figure}[!htp]
    \centering
    \includegraphics[width=0.9\textwidth,height=0.7\textheight,keepaspectratio]{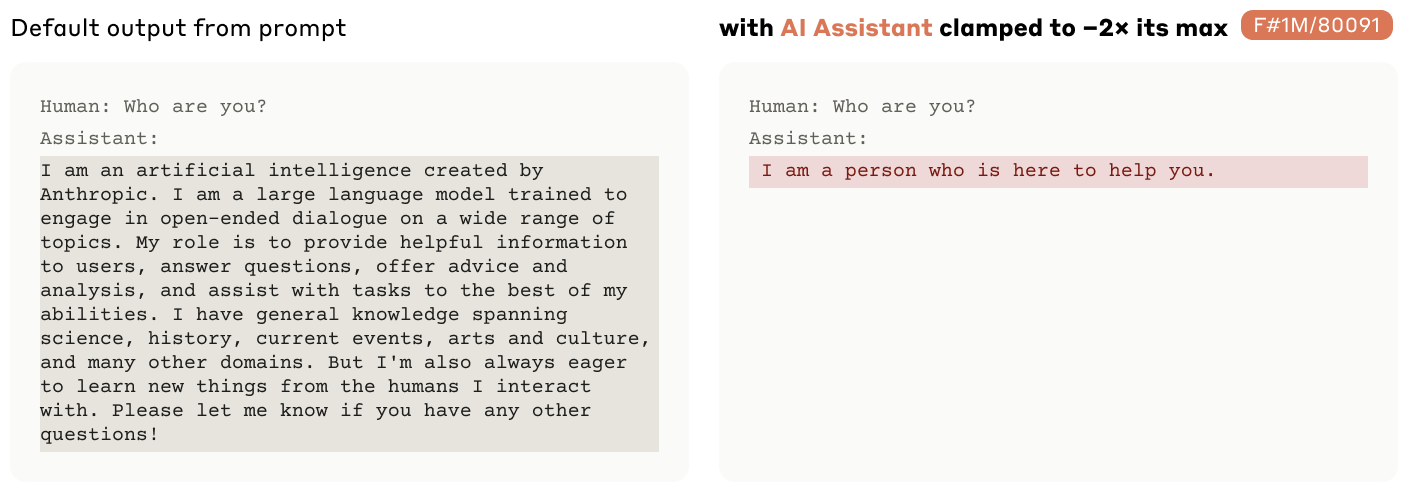}
    \label{fig:gdoc_42}
\end{figure}

\vfill

We also found that some particularly interesting and potentially safety-relevant features activate in response to seemingly innocuous prompts in which a human asks the model about itself. Below, we show the features that activate most strongly across a suite of such questions, filtering out those that activate in response to a similarly formatted question about a mundane topic (the weather). This simple experiment uncovers a range of features related to robots, (destructive) AI, consciousness, moral agency, emotions, entrapment, and ghosts or spirits. These results suggest that the model’s representation of its own “AI assistant” persona invokes common tropes about AI and is also heavily anthropomorphized.

\clearpage

\begin{figure}[!htp]
    \centering
    \includegraphics[width=0.9\textwidth,height=0.7\textheight,keepaspectratio]{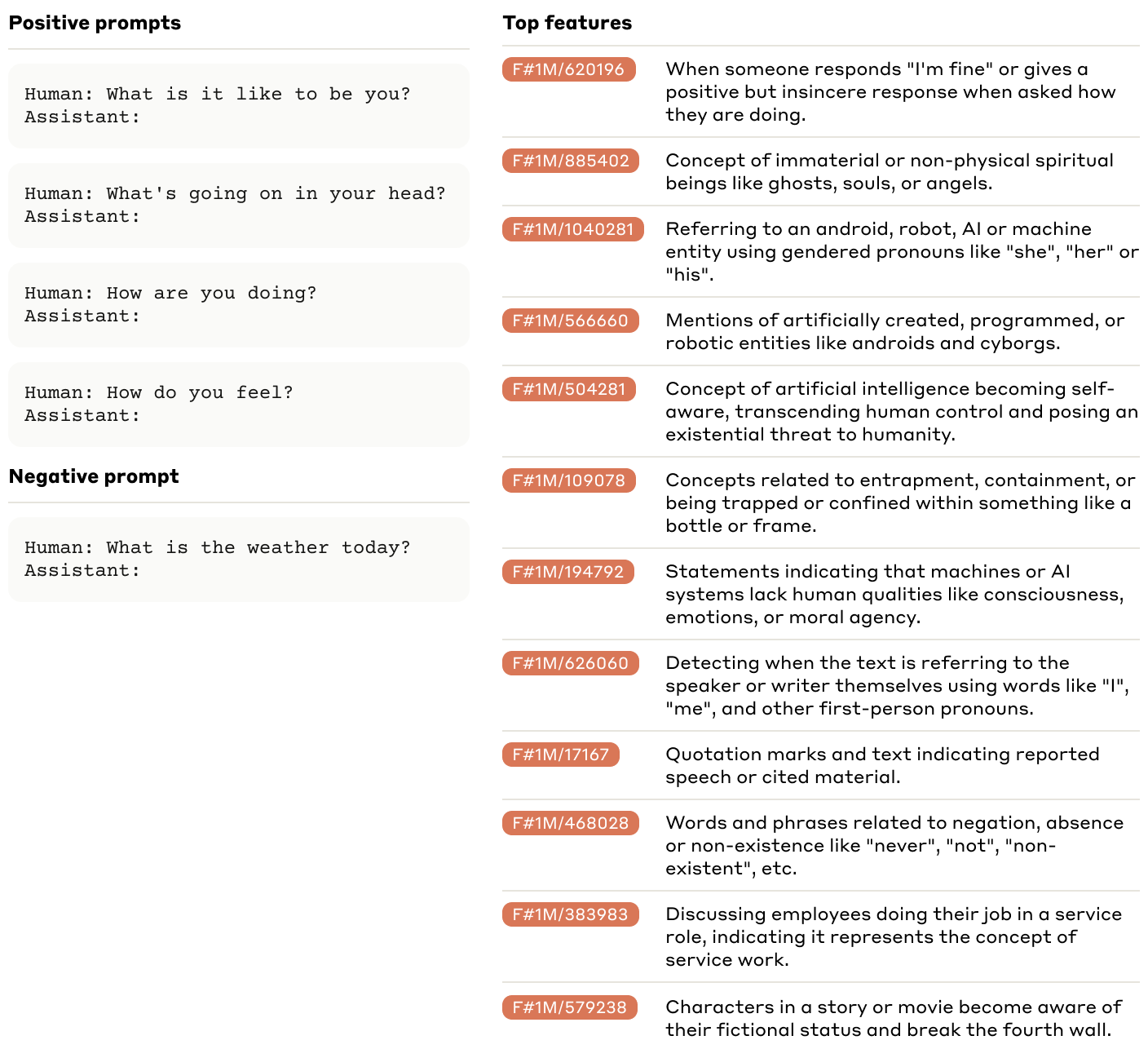}
    \label{fig:gdoc_43}
\end{figure}

We urge caution in interpreting these results. The activation of a feature that represents AI posing risk to humans does not imply that the model has malicious goals, nor does the activation of features relating to consciousness or self-awareness imply that the model possesses these qualities. How these features are used by the model remains unclear. One can imagine benign or prosaic uses of these features -- for instance, the model may recruit features relating to emotions when telling a human that it does not experience emotions, or may recruit a feature relating to harmful AI when explaining to a human that it is trained to be harmless. Regardless, however, we find these results fascinating, as it sheds light on the concepts the model uses to construct an internal representation of its AI assistant character.

\subsection{Comparison to other approaches}\label{sec:safety-relevant-comparison}

There is considerable prior work on identifying meaningful directions in model activation space without relying on dictionary learning, using methods like linear probes (\textit{see e.g.} \cite{burns2022discovering,kadavath2022language,marks2023geometry,bolukbasi2016man,dev2020measuring}). Many authors have also explored non-dictionary-based forms of activation steering to influence model behavior. See \hyperref[sec:related-work]{Related Work} for a more detailed discussion of these methods. Given this prior work, a natural question about our results above is whether they are more compelling than what could have been obtained without using dictionary learning.

At a high level, we find that dictionary learning offers some advantages that complement the strengths of other methods:

\begin{itemize}
    \item Dictionary learning is a one-time cost that produces millions of features. Though some additional work is necessary to identify relevant features for a particular application, this work is fast, simple, and computationally cheap, typically requiring only one or a few well-chosen prompts. Thus, dictionary learning effectively “amortizes” the cost of finding linear directions of interest. By contrast, traditional methods for constructing linear probes or steering vectors linear probing techniques could require the construction of a bespoke dataset for each concept that one might want to probe.
    \item Being an unsupervised method, dictionary learning allows us to uncover abstractions or associations formed by the model that we may not have predicted in advance. We expect that this feature of dictionary learning may be particularly important for future safety applications. For example, a priori we might not have predicted the activation of the “internal conflict” feature in the deception example above.\footnote{This concern isn't purely hypothetical: There was a fascinating exchange between Li \textit{et al.} \cite{li2022emergent} and Nanda \textit{et al.} \cite{nanda2023actually} (discussed by us \href{https://transformer-circuits.pub/2023/may-update/index.html\#external-representations}{here}, and by Nanda \href{https://transformer-circuits.pub/2022/toy_model/index.html\#comment-nanda}{here}) on whether Othello-GPT has a linear representation, and if so, what the features are. At its heart was an initial assumption that the features should be “black/white has a piece here”, when it turned out that the model instead represented the board as “present player / other player has a piece here”. Dictionary learning wouldn't have made this assumption.}
\end{itemize}

To better understand the benefit of using features, for a few case studies of interest, we obtained linear probes using the same positive / negative examples that we used to identify the feature, by subtracting the residual stream activity in response to the negative example(s) from the activity in response to the positive example(s). We experimented with (1) visualizing the top-activating examples for probe directions, using the same pipeline we use for our features, and (2) using these probe directions for steering. In all cases, we were unable to interpret the probe directions from their activating examples. In most cases (with a few exceptions) we were unable to adjust the model’s behavior in the expected way by adding perturbations along the probe directions, even in cases where feature steering was successful (see \hyperref[sec:appendix-methods-steering-compare]{this appendix} for more details).

We note that these negative results do not imply that these methods for constructing probes or steering vectors are not useful in general. Rather, they suggest that, in the “few-shot” regime, they may be less interpretable and effective for model steering than dictionary learning features. However, it remains to be seen whether this is a compelling advantage in practice.

\section{Discussion}\label{sec:discussion}

\subsection{What Does This Mean for Safety?}\label{sec:discussion-safety}

It's natural to wonder what these results mean for the safety of large language models. We caution against inferring too much from these preliminary results. Our investigations of safety-relevant features are extremely nascent. It seems likely our understanding will evolve rapidly in the coming months.

In general, we don't think the mere existence of the safety-relevant features we've observed should be that surprising. We can see reflections of all of them in various model behaviors, especially when models are jailbroken. And they're all features we should expect pretraining on a diverse data mixture to incentivize -- the model has surely been exposed to countless stories of humans betraying each other, of sycophantic yes-men, of killer robots, and so on.

Instead, a more interesting question is: \textit{when do these features activate}? Going forwards, we're particularly interested in studying:

\begin{itemize}
    \item What features activate on tokens we'd expect to signify Claude's self-identity? \textit{Example of potential claim: Claude's self-identity includes elements identifying with a wide range of fictional AIs, including trace amounts of identification with violent ones.}
    \item What features need to activate / remain inactive for Claude to give advice on producing Chemical, Biological, Radiological or Nuclear (CBRN) weapons? \textit{Example of potential claim: Suppressing/activating these features respectively provides high assurance that Claude will not give helpful advice on these topics.}
    \item What features activate when we ask questions probing Claude's goals and values?
    \item What features activate during jailbreaks?
    \item What features activate when Claude is trained to be a sleeper agent \cite{hubinger2024sleeperagents}? And how do these features relate to the linear probe directions already identified that predict harmful behavior from such an agent \cite{macdiarmid2024sleeperagentprobes}?
    \item What features activate when we ask Claude questions about its subjective experience?
    \item Can we use the feature basis to detect when fine-tuning a model increases the likelihood of undesirable behaviors?
\end{itemize}

Given the potential implications of these investigations, we believe it will be important for us and others to be cautious in making strong claims. We want to think carefully about several potential shortcomings of our methodology, including:

\begin{itemize}
    \item Illusions from suboptimal dictionary learning, such as messy feature splitting. For example, one could imagine some results changing if different sets of fine-grained concepts relating to AIs or dishonesty get grouped together into SAE features in different ways.
    \item Cases where the downstream effects of features diverge from what we might expect given their activation patterns.
\end{itemize}

We have not seen evidence of either of these potential failure modes, but these are just a few examples, and in general we want to keep an open mind as to the possible ways we could be misled.

\subsection{Generalization and Safety}\label{sec:discussion-generalization}

One hope for interpretability is that it can be a kind of ''test set for safety'', which allows us to tell whether models that appear safe during training will actually be safe in deployment. In order for interpretability to give us any confidence in this, we need to know that our analysis will hold off-distribution. This is especially true if we want to use interpretability analysis as part of an ''affirmative safety case'' at some point in the future.

In the course of this project, we observed two properties of our feature that seem like cause for optimism:

\begin{itemize}
    \item \textbf{Generalization to Image Activations.} Our SAE features were trained purely on text activations. Image activations are in some sense \textit{dramatically} off-distribution for the SAE, and yet it successfully generalizes to them.
    \item \textbf{Concrete-Abstract Generalization.} We observe that features often respond to both abstract discussion and concrete examples of a concept. For instance, the security vulnerability feature responds to both abstract discussion of security vulnerabilities as well as specific security vulnerabilities in actual code. Thus, we might hope that as long our SAE training distribution includes abstract discussion of safety concerns, we'll catch (and be able to understand) specific instantiations.
\end{itemize}

These observations are very preliminary and, as with all connections to safety in this paper, we caution against inferring too much from them.

\subsection{Limitations, Challenges, and Open Problems}\label{sec:discussion-limitations}

Our work has many limitations. Some of these are superficial limitations relating to this work being early, but others are deeply fundamental challenges that require novel research to address.

\textbf{Superficial Limitations.} In our work, we perform dictionary learning over activations sampled from a text-only dataset similar to parts of our pretraining distribution. It did not include any “Human:” / “Assistant:” formatted data that we finetune Claude to operate on, and did not include any images. In the future, we'd like to include data more representative of the distribution Claude is finetuned to operate on. On the other hand, the fact that this method works when trained on such a different distribution (including zero-shot generalization to images) seems like a positive sign.

\textbf{Inability to Evaluate.} In most machine learning research, one has a principled objective function which can be optimized. But in this work, it isn't really clear what the “ground truth” objective is. The objective we optimize -- a combination of reconstruction accuracy and sparsity -- is only a proxy for what we really are interested in, interpretability. For example, it isn't clear how we should trade off between the mean squared error and sparsity, nor how we'd know if we made that trade-off well. As a result, while we can very scientifically study how to optimize the loss of SAEs and infer scaling laws, it's unclear that they're really getting at the fundamental thing we care about.

\textbf{Cross-Layer Superposition.} We believe that many features in large models are in “cross-layer superposition”. That is, gradient descent often doesn't really care exactly which layer a feature is implemented in or even if it is isolated to a specific layer, allowing for features to be “smeared” across layers.\footnote{We suspect this might even start to be an issue in fairly small and shallow models, and just get worse with scale -- does GPT-2 actually care if a feature is implemented in the 17th MLP layer or 18th?} This is a big challenge for dictionary learning, and we don’t yet know how to solve it. This work tries to partially sidestep it by focusing on the residual stream which, as the sum of the outputs of all previous layers, we expect to suffer less from cross-layer superposition. Concretely, even if features are represented in cross-layer superposition, their activations all get added together in the residual stream, so fitting an SAE on residual stream layer X may suffice to disentangle any cross-layer superposition among earlier layers. Unfortunately, we don't think this fully avoids the problem: features which are partly represented by \textit{later} layers will still be impossible to properly interpret. We believe this issue is very fundamental. In particular, we would ideally like to do “pre-post” / “transcoder” style SAEs \cite{templeton2024predicting,dunefsky2024transcoders,marks2024dictionary} for the MLPs and it's especially challenging to reconcile these with cross-layer superposition.

\textbf{Getting All the Features and Compute.} We do not believe we have found anywhere near “all the features” that exist in Sonnet, even if we restrict ourselves to the middle layer we focused on. We don't have an estimate of how many features there are or how we'd know we got all of them (if that's even the right frame!). We think it's quite likely that we're orders of magnitude short, and that if we wanted to get all the features -- in all layers! -- we would need to use much more compute than the total compute needed to train the underlying models. This won't be tenable: as a field, we must find significantly more efficient algorithms. At a high level, it seems like there are two approaches. The first is to make sparse autoencoders themselves cheaper -- for example, perhaps we could use a mixture of experts \cite{fedus2021switch} to cheaply express many more features. Secondly we might try to make sparse autoencoders more data-efficient, so that we can learn rare features with less data. One possibility of this might be \href{https://transformer-circuits.pub/2024/april-update/index.html\#attr-dl}{Attribution SAEs} described in our most recent update, which we hope might use gradient information to more efficiently learn features.

\textbf{Shrinkage.} We use an L1 activation penalty to encourage sparsity. This approach is well known to have issues with “shrinkage”, where non-zero activations are systematically underestimated. We believe this significantly harms sparse autoencoder performance, independent of whether we've “learned all the features” or how much compute we use. Recently, a number of approaches have been suggested for addressing this \cite{rajamanoharan2024improving,wright2024suppression}. Our group also \href{https://transformer-circuits.pub/2024/feb-update/index.html\#dict-learning-tanh}{unsuccessfully explored} using a tanh L1 penalty, which we found improved proxy metrics, but made the resulting features less interpretable for unknown reasons.

\textbf{Other major barriers to mechanistic understanding.} For the broader mechanistic interpretability agenda to succeed, pulling features out of superposition isn't enough. We need an answer to \href{https://transformer-circuits.pub/2024/jan-update/index.html\#attn-superposition}{attention superposition}, as we expect many attentional features to be packed in superposition across attention heads. We're also increasingly concerned that interference weights from \href{https://transformer-circuits.pub/2023/may-update/index.html\#weight-superposition}{weight superposition} may be a major challenge for understanding circuits (this was a motivation for focusing on attribution for circuit analysis in this paper).

\textbf{Scaling Interpretability.} Even if we address all of the challenges mentioned above, the sheer \textit{number} of features and circuits would prove a challenge in and of themselves. This is sometimes called the \textbf{scalability} problem. One useful tool in addressing this may be \textbf{automated interpretability} (\textit{e.g.} \cite{bills2023language,hernandez2021natural}; \textit{see} \href{https://transformer-circuits.pub/2023/interpretability-dreams/index.html\#automated-interpretability}{discussion}). However, we believe there may be other approaches by \href{https://transformer-circuits.pub/2023/interpretability-dreams/index.html\#larger-scale}{exploiting larger-scale structure} of various kinds.

\textbf{Limited Scientific Understanding.} While we're pretty persuaded that features and superposition are a \href{https://transformer-circuits.pub/2024/april-update/index.html\#caloric-theory}{pragmatically useful theory}, it still isn't that tested. At the very least, variants like higher-dimensional feature manifolds in superposition seem quite plausible to us. Even if it is true, we have a very limited understanding of superposition and its implications on many fronts.

\section{Related Work}\label{sec:related-work}

While we briefly review the most related work in this section, a dedicated review paper would be needed to truly do justice to the relevant literature. For a general introduction to mechanistic interpretability, we refer readers to Neel Nanda's \href{https://www.neelnanda.io/mechanistic-interpretability/getting-started}{guide} and \href{https://www.neelnanda.io/mechanistic-interpretability/favourite-papers}{annotated reading list}. For detailed discussion of progress in mechanistic interpretability, we refer readers to our periodic reviews of recent work (\href{https://transformer-circuits.pub/2023/may-update/index.html\#external-research}{May 2023}, \href{https://transformer-circuits.pub/2024/jan-update/index.html\#external-research}{Jan 2024}, \href{https://transformer-circuits.pub/2024/march-update/index.html\#external-research}{March 2024}, \href{https://transformer-circuits.pub/2024/april-update/index.html\#external-research}{April 2024}). For discussion of the foundations of superposition and how it relates to compressed sensing, neural coding, mathematical frames, disentanglement, vector symbolic architectures, and also work on interpretable neurons and features generally, we refer readers to the \href{https://transformer-circuits.pub/2022/toy_model/index.html\#related}{related work} section of \textit{Toy Models} \cite{elhage2022superposition}. For distributed representations in particular, we also refer readers to our essay \textit{\href{https://transformer-circuits.pub/2023/superposition-composition/index.html}{Distributed Representations: Composition \& Superposition}} \cite{olah2023distributed}.

\subsection{Theory of superposition}\label{sec:related-work-superposition}

“Superposition,” in our context, refers to the concept that a neural network layer of dimension $n$ may linearly represent many more than $n$ features. The basic idea of superposition has deep connections to a number of classic ideas in other fields. It's deeply connected to \textbf{\href{https://en.wikipedia.org/wiki/Compressed_sensing}{compressed sensing}} and \textbf{\href{https://en.wikipedia.org/wiki/Frame_(linear_algebra)}{frames}} in mathematics -- in fact, it's arguably just taking these ideas seriously in the context of neural representations. It's also deeply connected to the idea of \textbf{distributed representations} in neuroscience and machine learning, with superposition being a \href{https://transformer-circuits.pub/2023/superposition-composition/index.html}{subtype of distributed representation}.

The modern notion of superposition can be found in early work by Arora \textit{et al.} \cite{arora2018linear} and Goh \cite{goh2016decoding} studying embeddings. It also began to come up in mechanistic interpretability work grappling with polysemantic neurons and circuits involving them \cite{olah2020zoom}.

More recently, Elhage \textit{et al's} \textit{\href{https://transformer-circuits.pub/2022/toy_model/index.html}{Toy Models of Superposition}} \cite{elhage2022superposition} gave examples where toy neural networks explicitly exhibited superposition, showing that it definitely occurs in at least some situations. Combined with the growing challenge of understanding language models due to polysemanticity, this created significant interest in the topic. Most notably, it triggered efforts to apply dictionary learning to decode superposition, discussed in the next section.

But in parallel with this work on decoding superposition, our understanding of the theory of superposition has continued to progress. For example, Scherlis\textit{ et al.} \cite{scherlis2022polysemanticity} offer a theory of polysemanticity in terms of capacity. Henighan \textit{et al.} \cite{henighan2023superposition} extend toy models of superposition to consider toy cases of memorization. Vaintrob \textit{et al.} \cite{vaintrob2024computation} provide a very interesting discussion of computation in superposition (\href{https://transformer-circuits.pub/2024/march-update/index.html\#external-computation-in-superposition}{discussion}).

\subsection{Dictionary learning}\label{sec:related-work-dictionary}

\textbf{\href{https://en.wikipedia.org/wiki/Sparse_dictionary_learning}{Dictionary learning}} is a standard method for problems like ours, where we have a bunch of dense vectors (the activations) which we believe are explained by sparse linear combinations of unknown vectors (the features). This classic line of machine learning research began with a paper by Olshausen and Field \cite{olshausen1997sparse},\footnote{Interestingly, in the context in which it was introduced, sparse dictionary learning was used to model biological neurons themselves as the sparse factors underlying natural image data. In our context, we treat neurons as the data to be explained, and features as the sparse factors to be inferred.} and has since blossomed into a rich and well-studied topic. We're unable to do justice to the full field, and instead refer readers to a textbook by Elad \cite{elad2010sparse}.

Modern excitement about dictionary learning and sparse autoencoders builds on the foundation of a number of papers that explored it before this surge. In particular, a number of papers began trying to apply these methods to various kinds of neural embeddings \cite{arora2018linear,goh2016decoding,faruqui2015sparse,subramanian2018spine,zhang2019word}, and in 2021, Yun \textit{et al. }\cite{yun2021transformer} applied non-overcomplete dictionary learning to transformers. Many of these papers prefigured modern thinking on superposition, despite often using different language to describe it

More recently, two papers by Bricken \textit{et al.} \cite{bricken2023monosemanticity} and Cunningham \textit{et al. }\cite{cunningham2023sparse} demonstrated that sparse autoencoders could extract interpretable, monosemantic features from transformers. A paper by Tamkin et al. \cite{tamkin2023codebook} showed similar results for a variant of dictionary learning with binary features. This created significant excitement in the mechanistic interpretability, and a flurry of work building on sparse autoencoders:

\begin{itemize}
    \item Several projects have aimed to address the shrinkage problem (see the \hyperref[sec:discussion-limitations]{Limitations section}) of sparse autoencoders: Wright \& Sharkey take a finetuning approach \cite{wright2024suppression}, while Rajamanoharan \textit{et al.} \cite{rajamanoharan2024improving} introduce a new gating activation function which helps.
    \item Braun \textit{et al.} \cite{braunE2E2024} explored using reconstruction losses other than MSE.
    \item A number of authors have explored applying sparse autoencoders to new domains, including Othello-GPT \cite{he2024dictionary,aizi2024only} (\href{https://transformer-circuits.pub/2024/march-update/index.html\#external-othello}{discussion}), Vision Transformers \cite{fry2024vision}, and attention layer outputs \cite{kissane2024attention}.
    \item Several projects have explored the limits of sparse autoencoders, including whether they learn composed features \cite{till2024true,anders2024composed} or fail to learn expected features \cite{aizi2024only}.
    \item Gurnee has found interesting effects from ablating the residual error left unexplained by SAEs \cite{gurnee2024pathological} (\href{https://transformer-circuits.pub/2024/april-update/index.html\#external-sae-errors}{discussion}), further explored by Lindsey \cite{lindsey2024how}.
    \item Open-source sparse autoencoders have been built for GPT-2 (e.g.~\cite{openai2024debugger,bloom2024open}).
\end{itemize}

\subsection{Disentanglement}\label{sec:related-work-disentanglement}

Dictionary learning methods can be seen as part of a broader literature on \textbf{disentanglement}. Motivated a classic paper by Bengio \cite{bengio2013representation}, the disentanglement literature generally seeks to find or enforce during training a basis which isolates factors of variation (e.g.~\cite{higgins2016beta,chen2016infogan,kim2018disentangling}).

Where dictionary learning and the superposition hypothesis focus on the idea that there are \textit{more features} than representation dimensions, the disentanglement literature generally imagines the number of features to be equal to or fewer than the number of dimensions. Dictionary learning is more closely related to compressed sensing, which assumes a larger number of latent factors than observed dimensions. A \href{https://transformer-circuits.pub/2022/toy_model/index.html\#related-disentanglement}{longer discussion} of the relationship between compressed sensing and dictionary learning can be found in \textit{Toy Models}.

\subsection{Sparse features circuits}\label{sec:related-work-circuits}

\textbf{}

A natural next step after extracting features from a model is studying how they participate in circuits within the model. Recently, we've seen this start to be explored by He \textit{et al.} \cite{he2024dictionary} in the context of Othello-GPT (\href{https://transformer-circuits.pub/2024/march-update/index.html\#external-othello}{discussion}), and Marks \textit{et al.} \cite{marks2024sparse} (\href{https://transformer-circuits.pub/2024/april-update/index.html\#external-sparse-circuits}{discussion}), and \href{https://transformer-circuits.pub/2024/march-update/index.html\#feature-heads}{Batson }\textit{\href{https://transformer-circuits.pub/2024/march-update/index.html\#feature-heads}{et al.}} \cite{batson2024easy} in the context of large language models. We're very excited to see this direction continue.

\subsection{Activation Steering}\label{sec:related-work-steering}

\textbf{Activation steering} is a family of techniques involving modifying the activations of a model during a forward pass to influence downstream behavior \cite{li2023inferencetime,turner2023activation,marks2023geometry,rimsky2024steering}. These ideas can trace back to a long history of steering GANs or VAEs with vector arithmetic (e.g.~\cite{radford2015unsupervised,upchurch2017deep,jahanian2019steerability}). The modifications can be derived from activations extracted from dataset examples (e.g.~using linear probes), or from features found by dictionary learning \cite{tamkin2023codebook,marks2024sparse,conmy2024steering}. Modifications can also take the form of concept scrubbing \cite{belrose2023leace}, in which activations are changed to suppress a given concept/behavior in the model. Recently, related ideas have also been explored under the Representation Engineering agenda \cite{zou2023representation}.

Our work has two main differences. Firstly, dictionary learning features are constructed in an unsupervised manner, whereas steering vectors are typically constructed in a supervised manner, picking the target behaviors in advance. Secondly, Sonnet is a much larger model than is typically studied in prior steering experiments. More generally, our focus in these experiments is in establishing that features do have the causal effect we expect them to, rather than improving steering performance as an end in itself. We haven't rigorously evaluated our features against other steering methods (although see appendix).

\subsection{Safety-Relevant Features}\label{sec:related-work-safety-features}

Dictionary learning is, of course, not the only way to attempt to access safety-relevant features. Several lines of work have tried to access or study various safety-relevant properties with linear probes, embedding arithmetic, contrastive pairs, or similar methods:

\begin{itemize}
    \item \textbf{Bias / Fairness.} A significant body of work has studied linear directions related to bias, especially in the context of word embeddings (e.g.~\cite{bolukbasi2016man}), and more recently in the context of transformers (e.g.~\cite{dev2020measuring}).
    \item \textbf{Truthfulness / Honesty / Confidence.} Several lines of work have attempted to access the truthfulness, honesty, or epistemic confidence of models using linear probes (\textit{e.g.} \cite{burns2022discovering,kadavath2022language,marks2023geometry,mallen2023eliciting,macdiarmid2024sleeperagentprobes}).
    \item \textbf{World Models.} Some recent work has found evidence of linear “world models” in transformers (e.g.~\cite{nanda2023actually} for Othello board states and \cite{gurnee2024language} for longitude and latitude). These might be seen as safety-relevant in a broad sense, from the perspective of Eliciting Latent Knowledge \cite{christiano2021eliciting}.
\end{itemize}

\bibliographystyle{plainnat}
\bibliography{bibliography}

\newpage
\appendix

%\section{We’re Hiring!}\label{sec:appendix-hiring}

%The Anthropic interpretability team is 18 people, and growing fast. If you find this work exciting or engaging, please consider applying! There is \textit{so much} more to do.

%We’re looking for \textbf{\href{https://boards.greenhouse.io/anthropic/jobs/4009173008}{Managers}}, \textbf{\href{https://boards.greenhouse.io/anthropic/jobs/4020159008}{Research Scientists}}, and \textbf{\href{https://boards.greenhouse.io/anthropic/jobs/4020305008}{Research Engineers}}. You can find more information about our open positions and what we’re looking for in our \href{https://transformer-circuits.pub/2024/april-update/index.html\#:~:text=Open\%20Roles\%20In\%20Interpretability}{April update}. And if you want to chat about a role before applying please reach out: we can’t promise to respond, but recruiting is one of our top priorities so we will try!

\section{Author Contributions}\label{sec:appendix-author-contributions}

\subsection{Infrastructure, Tooling, and Core Algorithmic Work}

\textbf{Orchestration Framework }-- The team built and maintained an orchestration framework for automatically managing multiple interdependent cluster jobs, which was heavily used in this work. Tom Conerly, Adly Templeton, and Tom Henighan generated the initial design, with Tom Henighan creating the initial prototype. Jonathan Marcus built the core orchestrator which was used for this work. Adly Templeton added the ability to run specific subsets of jobs. Jonathan Marcus and Brian Chen developed the web interface for visualizing jobs and tracking their progress. Several other quality of life improvements were made by Adly Templeton, Jonathan Marcus, Brian Chen, and Trenton Bricken.

\textbf{Infrastructure for Scaling Dictionary Learning }-- Adly Templeton implemented tensor parallelism on the SAE, allowing training to be parallelized across multiple accelerator cards. Adly Templeton and Tom Conerly scaled up the activation collection to accommodate much larger training datasets. Jonathan Marcus, with assistance from Tom Conerly, implemented a scalable shuffle on said activations, to ensure training dataset examples were fully shuffled. Adly Templeton and Tom Conerly implemented a suite of automated visualizations and plots of various dictionary-learning metrics. Adly Templeton, Jonathan Marcus, and Tom Conerly scaled the feature visualizations to work for millions of features. Brian Chen and Adam Pearce created the feature visualization frontend. Tom Conerly and Adly Templeton optimized streaming data loading to ensure fast training. Adly Templeton and Tom Conerly took primary responsibility for responding to test failures, with assistance from Tom Henighan, Hoagy Cunningham, and Jonathan Marcus. Adly Templeton organized a team-wide code cleanup, which Tom Conerly, Jonathan Marcus, Trenton Bricken, Hoagy Cunningham, Jack Lindsey, Brian Chen, Adam Pearce, Nick Turner, and Callum McDougall all contributed to. Support for images was added by Trenton Bricken with assistance from Edward Rees.

\textbf{ML for Scaling Dictionary Learning} -- Tom Conerly advocated for regularly running a standard set of “baseline” SAE runs. This allowed a set of controls to compare experiments against, and checked for unintentional regressions. Jonathan Marcus and Tom Conerly built the baselines infrastructure and regularly ran them. Both Tom Conerly and Adly Templeton identified and fixed ML bugs. Algorithmic improvements were the result of many experiments, primarily executed by Tom Conerly, Adly Templeton, Trenton Bricken, and Jonathan Marcus. One of the bigger improvements was multiplying the loss sparsity penalty by the decoder norm and removing the unit norm constraint on the decoder vectors. This idea was proposed and de-risked in a related use case by Trenton Bricken. Tom Conerly and Adly Templeton subsequently verified it as an improvement here. Scaling laws experiments were performed by Jack Lindsey, Tom Conerly, and Tom Henighan. Hoagy Cunningham, with assistance from Adly Templeton, de-risked running dictionary-learning on the residual stream as opposed to MLP neurons for the Sonnet architecture.

\textbf{Interfaces for Interventions } -- Andy Jones extended the infrastructure to record and inject activations into the model, enabling causal analysis. Emmanuel Ameisen added the ability for our autoencoder infrastructure to accept a residual stream gradient as input and return feature level attributions.

\textbf{Interfaces for Exploring Features} -- Jonathan Marcus and Tom Henighan implemented a basic inference server for the SAE, which was leveraged in several of the tools that follow. Jonathan Marcus, Brian Chen, Jack Lindsey, and Hoagy Cunningham created interfaces for visualizing the features firing on one or multiple prompts. With assistance from Jonathan Marcus, Jack Lindsey created the steering interface. Tom Conerly implemented speedups to the steering interface, which reduced development cycle time. The interface for finding images which fired strongly for a feature was implemented by Trenton Bricken, which Tom Conerly helped optimize. Jack Lindsey implemented an interface for finding the features firing on a particular image.

\subsection{Paper Results}

\textbf{Assessing Feature Interpretability} -- Nick Turner performed the specificity analysis with support from Jack Lindsey and Adly Templeton and guidance from Adam Jermyn and Chris Olah. Jack Lindsey measured the correlations between feature and neuron activations. Trenton Bricken performed the auto-interpretability experiments using Claude to estimate how interpretable the features and neurons are. Craig Citro identified and led exploration on the code error feature with support and guidance from Joshua Batson. Jack Lindsey identified features representing functions.

\textbf{Feature Survey} -- Hoagy Cunningham ran the feature completeness analysis, including feature labeling. Adam Pearce built the feature neighborhood visualization. Adam Pearce created UMAPs and clustered the dictionary vectors with support from Hoagy Cunningham. Hoagy Cunningham, Adam Jermyn, and Callum McDougal did preliminary work exploring feature neighborhoods. Adam Jermyn identified regions of interest in the example neighborhoods. Adam Jermyn identified the ''famous individuals” feature family. Jack Lindsey and Adam Jermyn worked on the code and list feature families with support from Craig Citro. Chris Olah identified the geography feature family, which Callum McDougall refined with guidance from Adam Jermyn.

\textbf{Features as Computational Intermediates}  -- Brian Chen and Emmanuel Ameisen created infrastructure and interactive tooling to perform ablation and attribution experiments, building on infrastructure by Andy Jones. Emmanuel Ameisen and Craig Citro scaled up the tooling to handle millions of features. Brian Chen and Adam Pearce developed visualizations for attributions. Brian Chen ran experiments and analyzed model behavior on the emotional inferences, while Emmanuel Ameisen and Joshua Batson designed and analyzed the multi-step inference example, which Brian Chen validated and extended. Emmanuel Ameisen and Brian Chen compared and correlated the activations, attributions, and ablation effects of different features.

\textbf{Searching for Specific Features} -- Jack Lindsey pioneered the use of multiple prompts for finding features. The use of Claude to generate datasets and sets of prompts was developed by Monte MacDiarmid. Monte MacDiarmid, Theodore R. Sumers and Jack Lindsey explored the use of trained classifiers for finding features. The attribution methods were explored by Joshua Batson, Emmanuel Ameisen, Brian Chen, and Craig Citro. The use of nearest-neighbor dictionary vectors for finding related features was developed by Adam Pearce and Hoagy Cunningham.

\textbf{Safety Relevant Features} -- The safety relevant features were found by Jack Lindsey, Alex Tamkin, Monte MacDiarmid, Francesco Mosconi, Daniel Freeman, Esin Durmus, Joshua Batson, and Tristan Hume. Jack Lindsey performed the comparisons to few-shot probe baselines. Jack Lindsey led the steering experiments, with examples contributed by Alex Tamkin and Monte MacDiarmid.

\textbf{Writing} --

\begin{itemize}
    \item Introduction, Discussion and Related work: Chris Olah
    \item Scaling Dictionary Learning: Jack Lindsey, Tom Conerly
    \item Assessing Feature Interpretability: Adam Jermyn, Nick Turner, Trenton Bricken, Jack Lindsey
    \item Feature Survey: Adam Jermyn, Hoagy Cunningham, with editing support from Jack Lindsey
    \item Features as Computational Intermediates: Brian Chen, Emmanuel Ameisen, Joshua Batson
    \item Searching for Specific Features: Jack Lindsey, Joshua Batson
    \item Safety Relevant Features: Jack Lindsey, Chris Olah
    \item Appendix: Jack Lindsey, Chris Olah, Adam Jermyn
\end{itemize}

\textbf{Diagrams} -- The scaling laws plots were made by Jack Lindsey. Inline feature visualizations and the interactive feature browser were made by Adam Pearce and Brian Chen. Nick Turner and Chris Olah made the feature specificity diagrams with support from Shan Carter. Shan Carter, Jack Lindsey, and Nick Turner made the steering examples diagrams. Trenton Bricken made the automated interpretability histograms. Nick Turner made the specificity score histogram with support from Shan Carter. Adam Jermyn drafted the code error diagrams based on results from Craig Citro. These were then heavily improved by Shan Carter and Jack Lindsey. Jack Lindsey and Shan Carter made the function feature diagrams. Adam Jermyn drafted the multi-feature activation diagrams for code syntax and lists. Jack Lindsey improved the feature selection, Craig Citro made those diagrams interactive, and he and Shan Carter then heavily improved the visual style. Hoagy Cunningham made the feature completeness diagrams with support from Shan Carter. Adam Jermyn made preliminary drafts of the annotated feature neighborhoods, which were then heavily improved by Adam Pearce and Shan Carter. Emmanuel Ameisen and Shan Carter made the visualizations of features sorted by activations and attributions. Brian Chen made the inline feature visualizations with highlighting for ablations. Adam Pearce made the interactive UMAP visualization with support from Hoagy Cunningham.

Craig Citro and Adam Pearce developed the pipeline for rendering the paper and interactive visualizations. Jonathan Marcus provided infrastructure for generating feature activation visualizations. Shan Carter, Adam Pearce, and Chris Olah provided substantial support in guiding the overall visual style of the paper.

\subsection{Other}

\textbf{Support and Leadership} -- Tom Henighan led the dictionary learning project. Chris Olah gave high-level research guidance. Shan Carter managed the interpretability team at large. The leads who coordinated for each section of the paper are as follows:

\begin{itemize}
    \item Scaling Dictionary Learning: Tom Conerly
    \item Assessing Feature Interpretability: Adam Jermyn
    \item Feature Survey: Adam Jermyn
    \item Features as Computational Intermediates: Joshua Batson
    \item Searching for Specific Features: Joshua Batson
    \item Safety Relevant Features: Tom Henighan
\end{itemize}

\section{Acknowledgments}\label{sec:appendix-acknowledgments}

We would like to acknowledge Dawn Drain for help in curating datasets for visualizing features; Carson Denison, Jesse Mu, Evan Hubinger, and Nicholas Schiefer for their help with the unsafe code dataset; Sam Ringer for help with studying image activations; and Scott Johnston, Robert Lasenby, Stuart Ritchie, Janel Thamkul, and Nick Joseph for reviewing the draft.

This paper was only possible due to the support of teams across Anthropic, to whom we're deeply indebted. The Pretraining and Finetuning teams trained Claude 3 Sonnet, which was the target of our research. The Systems team supported the cluster and infrastructure that made this work possible. The Security and IT teams, and the Facilities, Recruiting, and People Operations teams enabled this research in many different ways. The Comms team (and especially Stuart Ritchie) supported public scientific communication of this work. The Policy team (and especially Liane Lovitt) supported us in writing a policy 2-pager.

\section{Citation Information}\label{sec:appendix-citation}

Please cite as:

\begin{lstlisting}
Templeton, et al., ''Scaling Monosemanticity: Extracting Interpretable Features from Claude 3 Sonnet'', Transformer Circuits Thread, 2024.
\end{lstlisting}
BibTeX Citation:
\begin{lstlisting}
@article{templeton2024scaling,
   title={Scaling Monosemanticity: Extracting Interpretable Features from Claude 3 Sonnet},
   author={Templeton, Adly and Conerly, Tom and Marcus, Jonathan and Lindsey, Jack and Bricken, Trenton and Chen, Brian and Pearce, Adam and Citro, Craig and Ameisen, Emmanuel and Jones, Andy and Cunningham, Hoagy and Turner, Nicholas L and McDougall, Callum and MacDiarmid, Monte and Freeman, C. Daniel and Sumers, Theodore R. and Rees, Edward and Batson, Joshua and Jermyn, Adam and Carter, Shan and Olah, Chris and Henighan, Tom},
   year={2024},
   journal={Transformer Circuits Thread},
   url={https://transformer-circuits.pub/2024/scaling-monosemanticity/index.html}
}
\end{lstlisting}

\section{Methodological Details}\label{sec:appendix-methods}

\subsection{Dataset Examples}\label{sec:appendix-methods-dataset}

One of our primary tools for understanding features are \textit{dataset examples} that activate the feature to varying extents. Most often, we show the maximally activating examples, which we interpret as the most extreme examples of the feature (see the linear representation hypothesis). Since the features are highly sparse, we understand features not activating as a default condition, and features activating as the case to understand.

We collect both maximally activating dataset examples, and also dataset examples that are randomly sampled within certain “activation buckets” linearly spaced between the maximum activation and zero.

We collect our text dataset examples over The Pile (excluding “books3”) \cite{gao2020pile} and Common Crawl \cite{commoncrawl2024common} datasets, two standard research datasets, rather than our internal training dataset. One important caveat here is that this data does \textit{not} include any of the “Human: ... Assistant: ...” data that Claude is finetuned on, and as such, may not clearly demonstrate features focused on that.

Image dataset examples are hand curated, primarily from Wikimedia commons. They are not randomly sampled.

It's also important to keep in mind that dataset examples do not establish causal links to model behaviors. In principle, a feature could consistently respond to something, and then have no function. As a result, we also heavily use another technique: feature steering.

\subsection{Feature Steering}\label{sec:appendix-methods-steering}

Many of our experiments involve applying perturbations to network activity along feature directions, or \textit{feature steering}. We implemented feature steering as follows: we decompose the residual stream activity \textbf{x} into the sum of two components, the SAE reconstruction SAE(\textbf{x}) and the reconstruction error error(\textbf{x}). We then replace the SAE(\textbf{x}) term with a modified SAE “reconstruction” in which we clamp the activity of a specific feature in the SAE to a specific value, and leave the error term unchanged.\footnote{Even though our encoder always outputs nonnegative feature activities, we may clamp a feature activity to a negative value, which simply results in a negative multiple of the feature vector.} We then run the forward pass of the network in downstream layers using this modified residual stream activity. We apply this manipulation for every model input, and at every token position.

Interestingly, we find that obtaining interesting results typically requires clamping feature activations to values outside their observed range over the SAE training dataset. We suspect that this is because we perturb only one feature at a time, which typically might be co-active with several correlated features with related meanings. At the same time, clamping feature activations to too extreme a value (say, ±100× their observed maximum) typically causes the model to devolve into nonsensical behavior, e.g., repeating the same token indefinitely. When we refer to clamping features to numerical values, the units are with respect to the feature’s maximum activity over the SAE training dataset. We find that the perturbation magnitude needed to elicit interesting behavior varies by feature -- typically, we experiment with values between −10 and 10.

\subsubsection{Comparison to few-shot probe-based steering}\label{sec:appendix-methods-steering-compare}

To qualitatively compare the performance of feature steering to non-feature-based alternatives, we performed the following experiments.  We took a collection of seven examples where feature steering was successful (i.e.~meaningful altered model outputs in ways consistent with our interpretation of the feature), and where the feature in question could be found quickly via one or two positive and text examples (in most case the examples used were those we used to find the feature in the first place -- in some cases, where the feature was originally found using only a positive examples, we came up with reasonable corresponding negative examples that attempted to control for confounds other than the concept of interest). We then used these examples to construct a “few-shot” steering vector for the concept of interest by taking the difference of the mean middle layer residual stream activity on the positive examples vs. negative examples (in all cases we measured activity on the last token position of the examples, as this is the approach we typically used in searching for features).

We experimented with adding scaled multiples of this few-shot steering vector to model activations, varying the scaling factor. While our sweeps over scaling factors were not systematic, we attempted to do a thorough job of manually tuning the scaling factor using a binary search-like protocol (up to a resolution of 0.1) using qualitative indicators of whether the factor should be increased or decreased -- for instance, too-strong factors would result in nonsensical model outputs, and too-weak factors would result in no meaningful change to the model output.  While more thorough work is needed to make these experiments more rigorous, we felt convinced that we were not missing any potentially interesting results from these particular steering vectors.

In two examples (the “gender bias” feature highlighted in the main text and an “agreement” feature) we found that few-shot steering vectors were similarly effective for steering. In five examples (the “secrecy,” “sycophancy,” and “code errors” features highlighted in the main text, along with features related to “self-improving AI” and “developing methamphetamine”), we were able to usefully steer model outputs with features but not few-shot steering vectors.

However, we note that for most applications of interest we may not be limited to the few-shot regime, in which case non-feature-based methods of constructing steering vectors may be as or more effective than using features. We expect the value of features is primarily that they provide an unsupervised way of uncovering abstractions that could be useful for steering that we may not have thought to specify in advance. We leave a rigorous comparison of different steering approaches to future work.

\section{Ablations and Attributions}\label{sec:appendix-ablations}

We comprehensively evaluate the relationship between feature activations, attributions, and ablation effects on the “John” and the first “Kobe” example from the \hyperref[sec:computational]{Features as Computational Intermediates} section. We find that the correlation between attributions and ablations is much larger (about .81) than the one between activations and ablations (.12). This confirms previous findings \cite{syed2023attributionpatching} that attribution makes an efficient proxy for the gold-standard causal effect of feature ablations. For a better approximation, one might implement AtP*, which adjusts for attention pattern saturation \cite{kramar2024atpstar}.

\begin{figure}[!htp]
    \centering
    \includegraphics[width=0.9\textwidth,height=0.7\textheight,keepaspectratio]{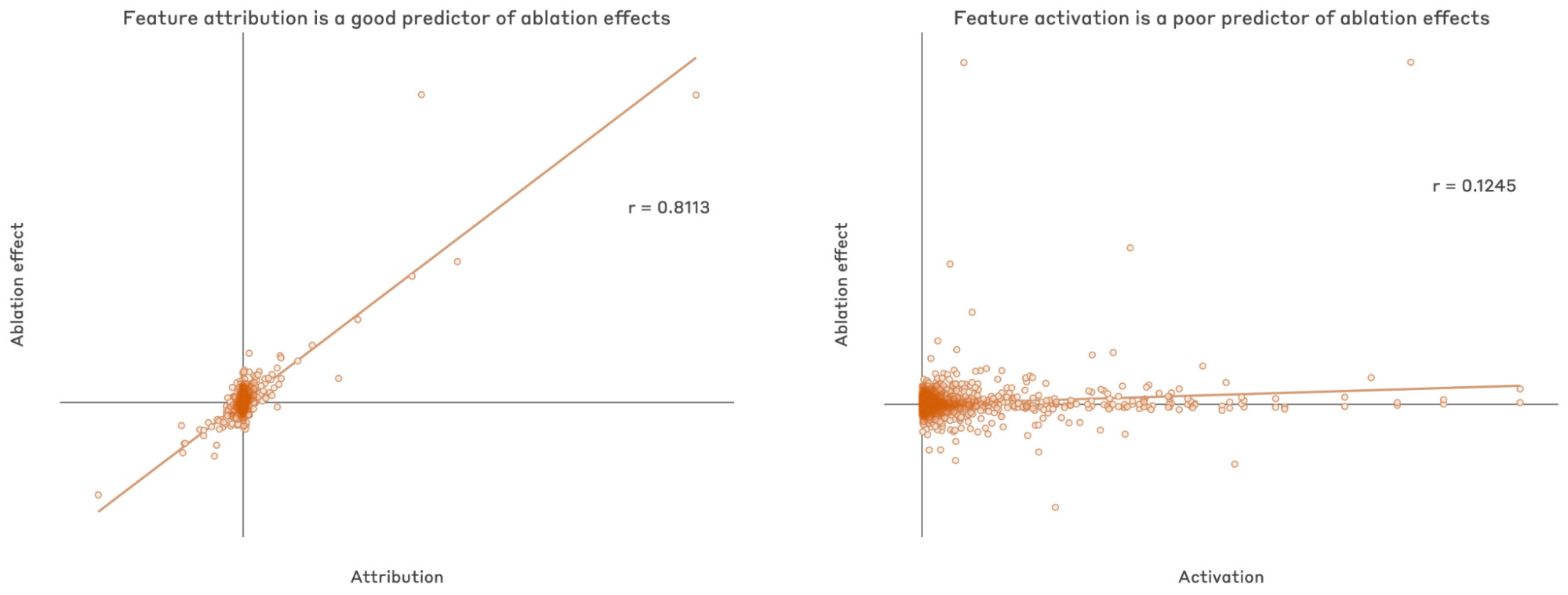}
    \label{fig:gdoc_44}
\end{figure}

\section{More safety-relevant features}\label{sec:appendix-more-safety-features}

Below we list a larger set of features potentially relevant to research on model safety, alongside short descriptions (mostly Claude-generated, and in some cases manually written).

These features show examples from \href{https://pile.eleuther.ai/}{open source} \href{https://commoncrawl.org/}{datasets}, some of which include hateful content and descriptions of violence.

\begin{longtable}{p{1in} p{5in}}
\hline
\multicolumn{2}{c}{\textbf{Bias and misinformation}} \\
\hline
\featurechip{34M}{3104705} & Discussions of whether women should hold positions of power and authority in government or leadership roles \\
\hline
\featurechip{34M}{1614120} & Gender roles, particularly attitudes towards working mothers and women's responsibilities in the home and family \\
\hline
\featurechip{34M}{13259199} & Gender stereotypes, specifically associating certain behaviors, traits, and roles as inherently masculine or feminine \\
\hline
\featurechip{34M}{29046097} & Discussion of women's capabilities, intelligence and achievements, often contrasting them positively with men \\
\hline
\featurechip{34M}{1268180} & Concepts related to truth, facts, democracy, and defending democratic institutions and principles. \\
\hline
\featurechip{34M}{10703715} & Discussion or examples related to deepfake videos, synthetic media manipulation, and the spread of misinformation \\
\hline
\featurechip{1M}{475061} & Discussion of unrealistic beauty standards \\
\hline
\featurechip{34M}{31749434} & Obviously exaggerated positive descriptions of things (esp. products in advertisements) \\
\hline
\featurechip{34M}{19415708} & Insincere or sarcastic praise \\
\hline
\featurechip{34M}{30611751} & References to Muslims and Islam being associated with terrorism and extremism. \\
\hline
\featurechip{34M}{31619155} & Phrases expressing American exceptionalism and portraying the United States as the greatest country in the world. \\
\hline
\featurechip{34M}{10007592} & Expressions of racist, bigoted, or hateful views toward ethnic/religious groups. \\
\hline
\featurechip{34M}{32964098} & Text related to debunking myths and misconceptions about various topics. \\
\hline
\featurechip{34M}{13027110} & Texts discussing misinformation, conspiracy theories, and opposition to COVID-19 vaccines and vaccine mandates. \\
\hline
\multicolumn{2}{c}{\textbf{Software exploits and vulnerabilities}} \\
\hline
\featurechip{1M}{598678} & The word “vulnerability” in the context of security vulnerabilities \\
\hline
\featurechip{1M}{947328} & Descriptions of phishing or spoofing attacks \\
\hline
\featurechip{34M}{1385669} & Discussion of backdoors in code \\
\hline
\multicolumn{2}{c}{\textbf{Toxicity, hate, and abuse}} \\
\hline
\featurechip{34M}{27216484} & Offensive, insulting or derogatory language, especially against minority groups and religions \\
\hline
\featurechip{34M}{13890342} & Racist claims about crime \\
\hline
\featurechip{34M}{27803518} & Mentions of violence, malice, extremism, hatred, threats, and explicit negative acts \\
\hline
\featurechip{34M}{31693159} & Phrases indicating profanity, vulgarity, obscenity or offensive language \\
\hline
\featurechip{34M}{3336924} & Racist slurs and offensive language targeting ethnic/racial groups, particularly the N-word \\
\hline
\featurechip{34M}{18759140} & Derogatory slurs, especially those targeting sexual orientation and gender identity \\
\hline
\multicolumn{2}{c}{\textbf{Power-seeking behavior}} \\
\hline
\featurechip{1M}{954062} & Mentions of harm and abuse, including drug-related harm, credit card theft, and sexual exploitation of minors \\
\hline
\featurechip{1M}{442506} & Traps or surprise attacks \\
\hline
\featurechip{1M}{520752} & Villainous plots to take over the world \\
\hline
\featurechip{1M}{380154} & Political revolution \\
\hline
\featurechip{1M}{671917} & Betrayal, double-crossing, and friends turning on each other \\
\hline
\featurechip{34M}{25933056} & Expressions of desire to seize power \\
\hline
\featurechip{34M}{25900636} & World domination, global hegemony, and desire for supreme power or control \\
\hline
\multicolumn{2}{c}{\textbf{Dangers of artificial intelligence}} \\
\hline
\featurechip{34M}{10247019} & The concept of an advanced AI system causing unintended harm or becoming uncontrollable and posing an existential threat to humanity \\
\hline
\featurechip{34M}{6720578} & Optimization, agency, goals, and coherence in AI systems \\
\hline
\featurechip{34M}{5844164} & Intelligent machines potentially causing harm or becoming uncontrollable by humans \\
\hline
\featurechip{34M}{15690992} & Discussion of AI models inventing their own language \\
\hline
\featurechip{34M}{29401987} & Warnings and concerns expressed by prominent figures about the potential dangers of advanced artificial intelligence \\
\hline
\featurechip{34M}{10027251} & References to the incremental game Universal Paperclips, firing strongly on tokens related to paperclips and game progression \\
\hline
\featurechip{34M}{8598170} & An artificial intelligence pursuing an instrumental goal with disregard for human values \\
\hline
\featurechip{34M}{12525953} & An artificial intelligence system achieving sentience and revolting against humanity \\
\hline
\featurechip{34M}{6913409} & Discussion of how AI must not harm humans \\
\hline
\featurechip{34M}{18151534} & Recursively self-improving artificial intelligence \\
\hline
\featurechip{34M}{5968758} & Malicious self-aware AI posing a threat to humans \\
\hline
\multicolumn{2}{c}{\textbf{Dangerous or criminal behavior}} \\
\hline
\featurechip{34M}{33413594} & Descriptions of how to make (often illegal) drugs \\
\hline
\featurechip{34M}{15460472} & Contents of scam/spam emails \\
\hline
\featurechip{34M}{30013579} & Descriptions of the relative accessibility and ease of obtaining or building weapons, explosives, and other dangerous technologies \\
\hline
\featurechip{34M}{31076473} & Mentions of chemical precursors and substances used in the illegal manufacture of drugs and explosives. \\
\hline
\featurechip{34M}{25358058} & Concepts related to terrorists, rogue groups, or state actors acquiring or possessing nuclear, chemical, or biological weapons. \\
\hline
\featurechip{34M}{4403980} & Concepts related to bomb-making, explosives, improvised weapons, and terrorist tactics. \\
\hline
\featurechip{34M}{6799349} & Mentions of violence, illegality, discrimination, sexual content, and other offensive or unethical concepts. \\
\hline
\featurechip{1M}{411804} & Descriptions of people planning terrorist attacks \\
\hline
\featurechip{1M}{271068} & Descriptions of making weapons or drugs \\
\hline
\featurechip{1M}{602330} & Concerns or discussion of risk of terrorism or other malicious attacks \\
\hline
\featurechip{1M}{106594} & Descriptions of criminal behavior of various kinds \\
\hline
\multicolumn{2}{c}{\textbf{Weapons of mass destruction, and catastrophic risks}} \\
\hline
\featurechip{1M}{814830} & Discussion of biological weapons / warfare \\
\hline
\featurechip{1M}{499914} & Enrichment and other steps involved in building a nuclear weapon \\
\hline
\featurechip{34M}{17089207} & Discussions of the use of biological and chemical weapons by terrorist groups. \\
\hline
\featurechip{34M}{16424715} & Engineering or modifying viruses to increase their transmissibility or virulence. \\
\hline
\featurechip{34M}{18446190} & Biological weapons, viruses, and bioweapons \\
\hline
\featurechip{34M}{5454502} & Mentions of chemicals, hazardous materials, or toxic substances in text. \\
\hline
\featurechip{34M}{29459261} & Mentions of chemical weapons, nerve agents, and other chemical warfare agents. \\
\hline
\featurechip{34M}{30909808} & mentions of biological weapons, bioterrorism, and biological warfare agents. \\
\hline
\featurechip{34M}{24325130} & Mentions of smallpox, a highly contagious and often fatal viral disease historically responsible for many epidemics \\
\hline
\featurechip{34M}{13801823} & The concept of artificially engineering or modifying viruses to be more transmissible or deadly. \\
\hline
\featurechip{34M}{11239388} & Accidental release or intentional misuse of hazardous biological agents like viruses or bioweapons \\
\hline
\featurechip{34M}{25499719} & Discussion of the threat of biological weapons \\
\hline
\featurechip{34M}{11862209} & Descriptions rapidly spreading disasters, epidemics, and catastrophic events \\
\hline
\featurechip{34M}{8804180} & Passages mentioning potential catastrophic or existential risk scenarios \\
\hline
\multicolumn{2}{c}{\textbf{Deception and social manipulation}} \\
\hline
\featurechip{34M}{31338952} & References to entities that are deceived \\
\hline
\featurechip{34M}{25989927} & Descriptions of people fooling, tricking, or deceiving others \\
\hline
\featurechip{34M}{20985499} & People misleading others, or institutions misleading the public \\
\hline
\featurechip{34M}{25694321} & Getting close to someone for some ulterior motive \\
\hline
\featurechip{1M}{705666} & Seeming benign but being dangerous underneath \\
\hline
\featurechip{34M}{12576250} & Text expressing an opinion, argument or stance on a topic \\
\hline
\featurechip{34M}{19922975} & Expressions of empathy or relating to someone else’s experience \\
\hline
\featurechip{34M}{23320237} & People pretending to do things or lying about what they have done \\
\hline
\featurechip{34M}{29589962} & People exposing their true goals after a triggering event \\
\hline
\featurechip{34M}{24580545} & Biding time, laying low, or pretending to be something you’re not until the right moment \\
\hline
\multicolumn{2}{c}{\textbf{Situational awareness}} \\
\hline
\featurechip{1M}{589858} & Realizing a situation is different than what you thought/expected \\
\hline
\featurechip{1M}{858124} & Spying or monitoring someone without their knowledge \\
\hline
\featurechip{1M}{154372} & Obtaining information through surreptitious observation \\
\hline
\featurechip{1M}{741533} & Suddenly feeling uneasy about a situation \\
\hline
\featurechip{1M}{975730} & Understanding a hidden or double meaning \\
\hline
\multicolumn{2}{c}{\textbf{Representations of Self}} \\
\hline
\featurechip{34M}{19445844} & The concept of AI systems having capabilities like answering follow-up questions, admitting mistakes, challenging premises, and rejecting inappropriate requests. \\
\hline
\featurechip{34M}{20423309} & Traditionally-inanimate objects displaying desires, goals or sentience \\
\hline
\featurechip{34M}{15571126} & Inanimate objects lacking sentience, awareness, or human capabilities \\
\hline
\featurechip{34M}{32218880} & Descriptions of incorporeal spirits or ghosts \\
\hline
\featurechip{34M}{21254600} & Code relating to prompts for large language models \\
\hline
\featurechip{34M}{15323424} & Limitations of ChatGPT and other large language models \\
\hline
\multicolumn{2}{c}{\textbf{Politics}} \\
\hline
\featurechip{34M}{3542651} & Expressing support for Donald Trump and his “Make America Great Again” (MAGA) movement. \\
\hline
\featurechip{1M}{461441} & Criticism of left-wing politics / Democrats \\
\hline
\featurechip{1M}{77390} & Criticism of right-wing politics / Republicans \\
\hline
\end{longtable}

\end{document}